UNIVERSITY OF JOENSUU
COMPUTER SCIENCE
DISSERTATIONS 19

TUOMO KAKKONEN

# FRAMEWORK AND RESOURCES FOR NATURAL LANGUAGE PARSER EVALUATION

ACADEMIC DISSERTATION

To be presented for public criticism with the permission of the Faculty of Science of the University of Joensuu in Hall H30, Agora building, Yliopistokatu 4, Joensuu, on December 3th, 2007, at 12 noon.

Supervisors

Professor Erkki Sutinen
Department of Computer Science and Statistics
University of Joensuu, Finland

PhD Stefan Werner
Department of Linguistics
University of Joensuu, Finland

Reviewers

Associate Professor Mike Joy
Department of Computer Science
University of Warwick, Coventry, United Kingdom

PhD Krister Lindén
Department of General Linguistics
University of Helsinki, Finland

Opponent

Professor Tapio Salakoski
Department of Information Technology
University of Turku, Finland

# FRAMEWORK AND RESOURCES FOR NATURAL LANGAUGE PARSER EVALUATION

Tuomo Kakkonen
Department of Computer Science and Statistics
University of Joensuu
P.O. Box 111, FI-80101 Joensuu, FINLAND
tuomo.kakkonen@cs.joensuu.fi

## Abstract

Because of the wide variety of contemporary practices used in the automatic syntactic parsing of natural languages, it has become necessary to analyze and evaluate the strengths and weaknesses of different approaches. This research is all the more necessary because there are currently no genre- and domain-independent parsers that are able to analyze unrestricted text with 100% *preciseness* (I use this term to refer to the correctness of analyses assigned by a parser). All these factors create a need for methods and resources that can be used to evaluate and compare parsing systems. This research describes: (1) A theoretical analysis of current achievements in parsing and parser evaluation. (2) A framework (called *FEPa*) that can be used to carry out practical parser evaluations and comparisons. (3) A set of new evaluation resources: *FiEval* is a Finnish treebank under construction, and *MGTS* and *RobSet* are parser evaluation resources in English. (4) The results of experiments in which the developed evaluation framework and the two resources for English were used for evaluating a set of selected parsers.

## Acknowledgments

I wish to extend my thanks and acknowledgement to everyone who has helped me with this dissertation. I would like to express my gratitude to my supervisor, Professor Erkki Sutinen, without whom I would never have begun this project. It was Professor Sutinen’s encouragement that originally inspired me to undertake scientific research. I would also like to thank PhD Stefan Werner for acting as the co-supervisor, Roger Loveday for proofreading the manuscript, and Phil. Lic. Simo Vihjanen of Lingsoft Inc for initially directing me to the field of parser evaluation. The comments of the two reviewers, Professor Mike Joy and PhD Krister Lindén, were valuable for preparing the final version of the manuscript. I thank them for their work.

My visits to other research institutes and groups were an integral part of the research process. I appreciated from the very beginning how contacts with other researchers with different skills and interests enabled me to develop my ideas, and I would like to extend a warm word of thanks to all those with whom I made contact during this research. I would in particular like to thank Professor Koenraad de Smedt, who acted as my supervisor during my stay in Bergen, Norway, in 2005, where I was privileged to be a research fellow at the Marie Curie Early Stage Researcher Training Site MULTILINGUA. I would also like to thank Professor Etienne Barnard, PhD Willie Smit and others on the staff of the Human Language Technology Group of the Council for Scientific and Industrial Research (CSIR) in Pretoria, South Africa, for creating a pleasant working environment in which I was able to accomplish a great deal. I would also like to thank Professor Joško Božanić for giving me permission to work in the friendly atmosphere of the Faculty of Philosophy at the University of Split, Croatia, in 2006 and 2007. Working in Split allowed me to get in contact with Professor Damir Ćavar on whose invitation I visited the University of Zadar and the Institute of Croatian Language and Linguistics in Zagreb.

I was able to bear the costs associated with this research because of a fellowship at MULTILINGUA Research Training Site funded by the European Union, the MirrorWolf Project funded by the Technology Agency of Finland (TEKES), the Automated Assessment Technologies for Free Text and Programming Assignments Project funded by the Academy of Finland, a one-year PhD student placement at the Graduate School of Language Technology in Finland, and a PhD scholarship that I received from the Marjatta and Kalle Salo Foundation.

Finally, I would like to thank my family and friends for their support. I offer my grateful thanks to my parents, Ulla and Aimo Kakkonen, for instilling in me the conviction that the only way to achieve success is through perseverance and hard work. Finally and above all, I would like to thank my wife, Gordana Galić Kakkonen, for her love and support.

Split, Croatia, November 20th, 2007
Tuomo Kakkonen

## Abbreviations

| Abbreviation | Description |
|---|---|
| A | *Adjective* |
| AG | *Annotation Graph* |
| AI | *Artificial Intelligence* |
| AIF | *Atlas Interchange Format* |
| API | *Application Protocol Interface* |
| ATLAS | *Architecture and Tools for Linguistic Analysis Systems* |
| ATN | *Augmented Transition Network* |
| AVM | *Attribute-value Matrix* |
| BC | *Brown Corpus* |
| CatG | *Categorial Grammar* |
| CDG | *Classical Dependency Grammar* |
| CoNLL-X | *10th Conference on Computational Natural Language Learning* |
| CCG | *Combinatory Categorial Grammar* |
| CF | *Context-free* |
| CFG | *Context-free Grammar* |
| CG | *Constraint Grammar* |
| CFPSG | *Context-free Phrase Structure Grammar* |
| CL | *Computational Linguistics* |
| CSG | *Context-sensitive Grammar* |
| CSTS | *Czech Sentence Tree Strucutre* |
| CYK | *Cocke-Younger-Kasami* |
| GR | *Grammatical Relation* |
| DAG | *Directed Acyclic Graph* |
| DepBank | *PARC 700 Dependency Bank* |
| DG | *Dependency Grammar* |
| D structure | *Dependency Structure* |
| D treebank | *Dependency Treebank* |
| DOP | *Data-oriented Parsing* |
| EU | *European Union* |
| ERG | *English Resource Grammar* |
| FEPa | *Framework for Evaluating Parsers* |
| FDG | *Functional Dependency Grammar* |
| FINCG | *Constraint Grammar Parser for Finnish* |
| FS | *Finite-state* |
| FSIG | *Finite Intersection Grammar* |
| FSM | *Finite-State Machine* |
| FST | *Finite-State Transducer* |
| GPSG | *Generalized Phrase Structure Grammar* |
| GR | *Grammatical Relations* |
| HC | *Head-corner* |
| HMM | *Hidden Markov Model* |
| HPSG | *Head-driven Phrase Structure Grammar* |
| ID | *Immediate Dominance* |
| JDBC | *Java Database Connectivity* |
| IR | *Information Retrieval* |
| LA | *Leaf-ancestor* |
| LALR | *Look-ahead LR* |
| LC | *Left-corner* |
| LDD | *Long-distance Dependency* |
| LFG | *Lexical-functional Grammar* |
| LG | *Link Grammar* |
| LP | *Linear Precedence* |
| LR | *Left to right, Rightmost derivation* |
| LTAG | *Lexicalized Tree-adjoining Grammar* |
| ME | *Maximum Entropy* |
| MCSG | *Mildly Context-sensitive Grammar* |
| MGTS | *Multi-Genre Test Set* |
| MPH | *Maximal Projections of Heads* |
| MRS | *Minimal Recursion Semantics* |
| MT | *Machine Translation* |
| NLP | *Natural Language Processing* |
| N | *Noun* |
| NP | *Noun Phrase* |
| O | *Object* |
| PAS | *Predicate-argument Structure* |
| PCFG | *Probabilistic Context-free Grammar* |
| PDT | *Prague Dependency Treebank* |
| PLTAG | *Probabilistic Lexicalized Tree-adjoining Grammar* |
| POS | *Part-of-speech* |
| PP | *Prepositional Phrase* |
| PS | *Phrase Structure* |
| PSG | *Phrase Structure Grammar* |
| PTB | *Penn Treebank* |
| RP | *Recognition Problem* |
| RTN | *Recursive Transition Network* |
| SGC | *Strong Generative Capacity* |
| STMDP-CoNNL-X | *The Shared Task on Multilingual Dependency Parsing at the 10th Conference on Computational Natural Language Learning* |
| TAG | *Tree-adjoining Grammar* |
| TG | *Transformational Grammar* |
| TSNLP | *Test Suites for Natural Language Processing* |
| TWOL | *Two-level* |
| UG | *Unification Grammar* |
| URP | *Universal Recognition Problem* |
| V | *Verb* |
| VP | *Verb phrase* |
| WGC | *Weak Generative Capacity* |
| WSJ | *Wall Street Journal* |
| XCES | *XML-based Corpus Encoding Standard* |
| XDG | *Extensible Dependency Grammar* |
| XML | *Extensible Markup Language* |

# Table of Contents

# 1 Introduction

This thesis reports research into the syntactic parsing of natural languages and evaluation of parsing systems. In this work, techniques and algorithms for parsing have been analyzed and compared on the theoretical level, and resources, methods and tools for the practical evaluation and comparison of syntactic parsers have been designed and implemented. A *natural language* is a language that has evolved through use in a social system, and is used by human beings for everyday communication. A *grammar* specifies the rules for how each sentence is constructed from parts. *Parsing* is the process of identifying the syntactic structure of a given sentence. A *natural language parser* is computer software that automatically performs parsing and outputs the structural description of a given character string in the context of a specific grammar. The output of a parser is called a *parse* and it describes the structure of a particular analyzed language fragment.

## 1.1 Motivation

Because of the ubiquity of the Internet among other factors, the amount of available textual information has grown explosively in past decades. This has resulted in an ever-increasing demand for software that can automatically process the information contained in natural languages. In the early days of *natural language processing* (NLP), the complexity of processing natural languages was drastically underestimated by most researchers involved in that field. Although their work was largely unsuccessful at the time, the first applications were developed in the 1950s for *machine translation* (MT) (see, for example, Locke & Booth (1955)). Later successful applications include systems for *information extraction*, *document summarization*, *message classification*, and *question answering.*

Parsing is not usually a goal in itself, but a parser is used as a component of NLP and *artificial intelligence* (AI) systems. Education is a novel application field for parsers and other NLP techniques. Parsing systems are applied, for example, in computer-assisted and automatic assessment of free-text responses (i.e. essays) (Hearst 2000, Kakkonen & Sutinen 2004). The ability to process natural languages plays a key role in *information retrieval* (IR) (Baeza-Yates & Ribeiro-Neto 1999). The analysis of syntactic and semantic structures is necessary for advancing from data retrieval (exact searching based on numeric and structured data) towards more “fuzzy” retrieval of information from textual data.

One can get some idea of how great the current demand for NLP applications has become at the time of writing (August 2007) by looking at the *European Union* (EU) with its 27 member states and 23 official languages (Mariani 2005, European Union 2007). Multilingualism has become one of the main economic, political and cultural challenges in the EU because of the desire of member states to preserve their languages and cultures while taking for granted the possibility of inter-lingual communication among the citizens of the EU. The cost of the translation of documents and having hundreds of interpreters on hand to translate between the 506 language pairs call for investigating the use of NLP systems for automating at least part of the processes.

Evaluation plays a crucial role in NLP and *computational linguistics* (CL). Evaluation methods and tools are needed to allow the developers and users to assess, enhance and choose appropriate systems (Gaizauskas 1998). It has become clear that standardized evaluations and system comparisons need to be undertaken. The contemporary interest in evaluation in the research community has inspired initiatives such as the *Evaluation in Language and Speech Engineering* (ELSE) project of the EU (Clark 2005), starting a biannual conference series entitled *International Conference on Language Resources and Evaluation* (LREC), first held in 1998, and launching a journal entitled *Language Resources and Evaluation* in 2005 (Springer 2005).

My motivation for undertaking research into parser evaluation was stimulated by the fact that there are no genre- and domain-independent parsers that are able to analyze unrestricted text with 100% *preciseness*[1]. This deficiency brought home to me the need for linguistic resources, methods and tools for evaluating parsers and comparing the characteristics of parsing systems. A *linguistic resource* is a set of machine-readable language data and descriptions. There are several types of such resources such as spoken and written corpora, lexical databases, treebanks and terminologies. *Treebanks*, for example, are collections of syntactically annotated sentences that serve as the "gold standard" to which parsers' outputs might be compared. There are as yet no linguistic resources suitable for parser evaluation in

[1] Instead of utilizing the commonly used term "accuracy", I prefer to use the term "preciseness" to refer to the correctness of analyses assigned by a parser. I intentionally also avoid using the terms "accuracy" and "precision" because of their technical use in evaluation context. *Test accuracy* refers to the proportion of instances that have been correctly classified. It is therefore logical to use the term "accuracy" to refer to the percentage of constituents/dependencies or sentences correctly parsed. *Precision* is commonly used as a measure of preciseness evaluation. "Preciseness" is a more general term that includes the kind of evaluation that uses the accuracy or precision as an evaluation measure. See Sections 8.1, 9.1., 9.3, 10.2 and 10.6.3 for more details.

Finnish. While linguistic resources do exist for English, there is a need for resources that are built for the needs of parser evaluation.

An *evaluation method* defines the way in which the *performance*[2] of a parser may be quantified. An *evaluation framework* consisting of the resources and methods can be used by practitioners of NLP to compare the strengths and weaknesses of diverse parsers and by parser developers to guide their work by pinpointing problems and providing analytical information about a parser's performance. There is, in addition, a lack of comprehensive evaluation tools that can facilitate practical evaluations. An *evaluation tool* is a software program that operationalizes an evaluation method or a set of evaluation methods.

## 1.2 Syntax and Parsing

Table 1-1 lists the seven levels of knowledge of natural languages distinguished by (Allen 1995).

**Table 1-1.** Types of knowledge of language. The levels most relevant to this work are highlighted (Allen 1995).

| Type of knowledge | Function |
|---|---|
| *Phonetics and phonology* | How words are related to the sounds that realize them |
| *Morphology* | How words are constructed from basic units |
| *Syntax* | How words can be put together to form sentences |
| *Semantics* | Meaning of words and sentences |
| *Pragmatics* | How sentences are used in different situations |
| *Discourse* | How preceding sentences affect the interpretation of a succeeding sentence |
| *World knowledge* | General knowledge about the world that language users possess |

*Morphology* is the study of word formation. The morphological processes of a natural language create completely new words or word forms from a root form. *Syntax* is the linguistic study that describes how a language user combines words to form phrases and sentences. *Semantics* is the study of the meanings created by words and phrases. It is the purpose of natural language parsers to describe the syntax of the input sentences, usually without any reference to semantics (Sikkel 1997). Some parsers can also perform a morphological analysis to capture the structure of individual words.

[2] In this work, the word *performance* refers to the quality of a parser relative to a specific criterion.

Syntactic parsing is a prerequisite for understanding speech or written text. A system for understanding a natural language usually includes the processing stages that are illustrated in Figure 1-1.

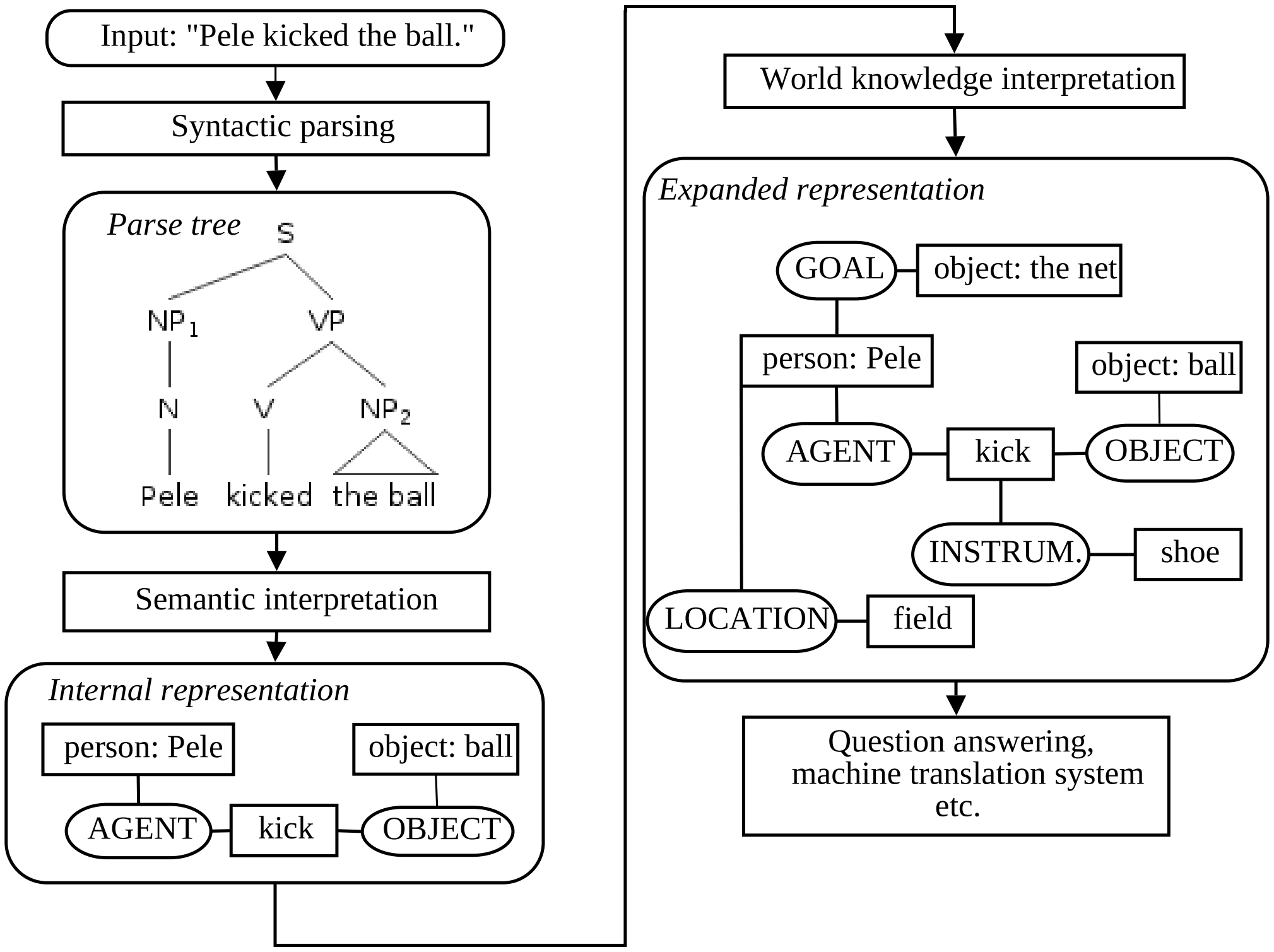

**Figure 1-1.** The stages of processing in a language understanding system (adapted from Luger & Stubblefield 1998).

Figure 1-1 shows how the results of syntactic parsing are combined with information about the meaning of the words in a sentence to perform semantic interpretation for creating an internal representation of the meaning. An expanded meaning representation can be created by adding structures from a world knowledge base. Contemporary NLP systems neither have access to the amount of world knowledge possessed by human beings, nor are they capable of reasoning from available knowledge as well as humans.

## 1.3 Perspectives on Evaluation

Before an evaluation of a computer software tool can be carried out, a set of criteria needs to be defined. In parser evaluation, these could include, for example, correctness of the output, efficiency and usability. The correctness of a parser can be measured by checking the output produced by a system. Evaluating the output of a parser requires one to make judgments about the grammaticality or "correctness" of the structural descriptions assigned by the system. The efficiency of a parser can be measured in terms of how a parser utilizes time and space. The usability of a system can be measured by, for example, asking users' opinions about how easy or otherwise it is to use the system.

Parsers can be evaluated from the point of view of developers, end-users and managers (TEMAA 1996). *Developers* need to be able to track the progress of the system with which they are working. *End-users* need to know how different parsers compare so that they can select a parser that is best suited to their needs and requirements. *Managers* need to have information on which to base decisions about resource allocation. Although parsers are used as components of NLP applications, the evaluation of parsing systems cannot solely be based on a comparison of their performance as parts of whole systems.

An important distinction for this research is between intrinsic, extrinsic, and comparative evaluation (Srinivas *et al*. 1998, Hirschman & Thompson 1998). *Intrinsic* evaluation focuses on measuring the performance of a single parser and on detecting errors in its output. Intrinsic evaluation also provides the developers of parsers with a means to identify the changes and amendments that would improve parser performance. *Extrinsic* evaluation means the process of evaluating a parser when it is an embedded component of an NLP application. This kind of evaluation is based on the performance of the whole NLP application rather than on a direct observation of the parser and its execution time. This type of evaluation is especially useful for end-users and managers because it allows them to select an appropriate parser for the task at hand. *Comparative* evaluation means comparing different parsers and it is useful for both end-users and system developers alike.

This research makes use of both intrinsic and comparative methods. Comparative evaluation is the more complicated of the two because, firstly, it is difficult to make direct comparisons because parsers often use different types of output formats, and, secondly, the comparison of differences in efficiency is not straightforward because parsers utilize different programming languages and different platforms (Unix, Windows, Linux).

## 1.4 Research Tasks

Four research tasks were identified:

a) Theoretical analysis of current achievements in state-of-the-art parsing and parser evaluation;

b) Design and implementation of linguistic resources for evaluating parsers of Finnish and English;
c) Derivation of a framework for parser evaluations; and
d) Carrying out practical parser evaluations by using the created resources, methods and tools.

Parser evaluation, like any research, is built on the foundations of the earlier research methods and findings. A researcher needs to be completely conversant with the theory, methods and algorithms of parsing before he or she can devise new evaluation practices or undertake parser evaluations (research task *a*). It is furthermore of utmost importance to appreciate and understand the evaluation resources, methods and tools at his or her disposal before undertaking any practical research into parser evaluation.

Since no suitable linguistic resources are available for parser evaluation for Finnish, it is necessary first to construct such resources before evaluating parsers of Finnish (research task *b*). An *annotation tool*, for example, has to be designed and constructed before the treebank can be created. Such a tool facilitates a quick and error-free annotation process. In addition to designing the Finnish resource, two evaluation resources for English were constructed as a part of this research.

Practical parser evaluations and comparisons cannot be undertaken without a comprehensive evaluation framework and evaluation tools for carrying out the experiments (research task *c*). While several evaluation methods have been devised and some practical evaluations undertaken, these evaluations have usually concentrated on a single item of parsers' performance. In order therefore to undertake this research, I had to design an evaluation framework and implement a set of evaluation tools.

In the final phase of the research, I carried out a program of practical evaluations of selected parsers by using the resources, methods and tools created for this purpose (research task *d*). I, moreover, compared the results of these experiments with the findings of the tasks *a*, *b* and *c*.

## 1.5 Research Methods

The syntactic parsing and evaluation of parsers is an interdisciplinary field because it combines CL, *linguistics*, *computer science* and *mathematics*. The interdisciplinary nature of the work is reflected in several ways, but especially strongly in the conduct of state-of-the-art parsing. While a dissertation in computer science obviously has to be technically anchored, research in the field of parsing and CL requires a firm grounding in linguistics and a familiarity with disciplines usually associated with the humanities. Even though the adoption of multiple perspectives has increased the complexity of this dissertation, the incorporation and fusion of two usually distinct academic fields of knowledge has been one of the main challenges and most important contributions of this work.

Task *a* comprises a critical literature review and an analysis of the methods used in natural language parsing. It also includes an analysis of the structure and content of linguistic resources for evaluation, and a derivation of a set of practices that should be employed in designing and annotating parser evaluation resources. Task *b* consists of an application of the practices thus developed to the design and construction of new evaluation resources. On the basis of the findings of tasks *a*, I also devised a set of evaluation criteria for parsers and defined evaluation metrics for each of the relevant criteria (task *c*). In addition, a set of software tools was designed and implemented for carrying out parser evaluations. Task *d* has two main aims: the testing of the developed evaluation framework and the evaluation and comparison of a set of selected parsers. Figure 1-2 below illustrates the connections between the four tasks.

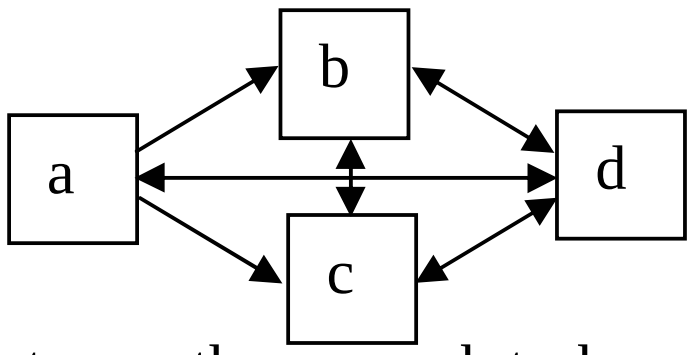

**Figure 1-2**. Connections between the research tasks.

For the practical part of this research, namely the design and implementation of the evaluation resources, methods and tools, I utilized the taxonomy of research methods devised by Järvinen & Järvinen (2001).

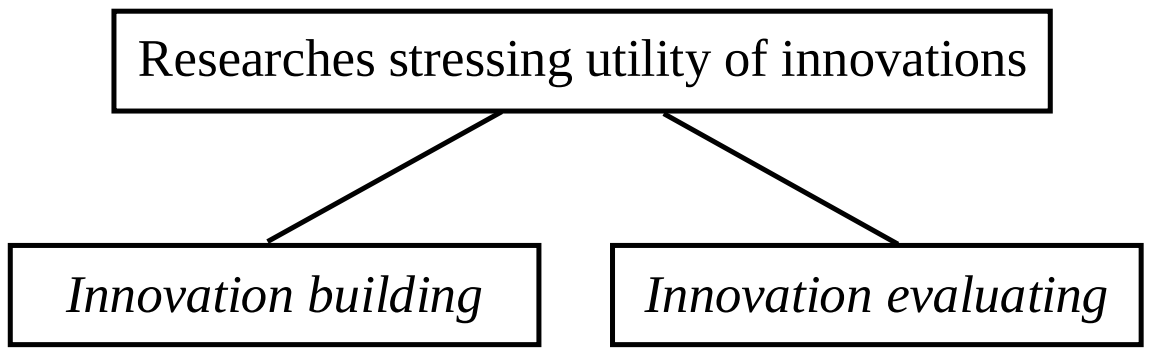

**Figure 1-3.** Those parts of Järvinen & Järvinen's (2001) taxonomy of research methods (relating to the creation and evaluation of innovations) that are relevant to this research.

The methods from the taxonomy of Järvinen & Järvinen (2001) that are relevant to purposes of this thesis are those that describe the utility of innovations, namely *building and evaluating innovations* (See Figure 1-3 above). While the evaluation of natural language parsers can be regarded as an *evaluation* of an innovation, the development of evaluation framework, tools, resources and methods can be regarded as the *building* of an innovation. In this thesis, I have applied an innovation building approach to building resources, methods and tools for evaluating innovations (research tasks *b* and *c*). I have, in addition, applied the built innovation in practice in my evaluation of natural language parsers (research task *d*).

The evaluation of a natural language parser can be viewed as a controlled experiment in which as many factors as possible should be under the control of the researcher (Järvinen & Järvinen 2001). But Mason (1988) notes that tightness (rigidity) of control and the richness (complexity) of reality are two properties that need to be compromised and traded against one another in any controlled experimental design. As control over experimental conditions increases, the results of the experiment become less relevant to real life situations and thus less applicable or generalizable. This axiom is especially important in the design and construction of parser evaluation resources. While resources that consist of artificially constructed test sentences may indeed allow for highly controlled experiments, experiments that use such resources are less generally applicable and useful than evaluations that are based on naturally occurring running texts.

## 1.6 Structure of the Thesis

The dissertation is organized as follows: Part I, which consists of Chapters 2, 3, 4 and 5, addresses parsing technologies. Chapters 2, 3, and 4 describe the current state-of-the-art in parsing technology. Each section ends with an analysis of the methods discussed and underlines their respective strengths and weaknesses. Chapter 5 contains a discussion of the issues and challenges in parsing.

Part II is concerned with parser evaluation resources. Chapter 6 analyzes the existing linguistic resources that are applied for parser evaluation purposes. Chapter 7 begins with the description of the design for the parser evaluation treebank for Finnish, and continues by describing the English evaluation materials that were created for this research.

Part III is concerned with parser evaluation methods and tools. Chapter 8 analyzes existing methods and tools for parser evaluation. The developed evaluation framework, *FEPa*, is described in Chapter 9.

Part IV of the thesis analyzes the results of the parser evaluation and concludes with the findings. Chapter 10 compares the parsers by applying the FEPa framework and the developed resources. In addition to the goal of evaluating the selected parsers, the aim of the research is to evaluate the FEPa framework. Chapter 11 summarizes the key findings of this work and states possible directions for future research.

The following papers and articles are published versions of the themes and content concerned with the dissertation. In each of the papers, I was the main author. In those papers with two authors, the second author acted as supervisor.

[1] Kakkonen, T.: Dependency Treebanks: Methods, Annotation Schemes and Tools. Proceedings of the 15th Nordic Conference of Computational Linguistics. Joensuu, Finland, 2005.

> The paper contains a survey of existing dependency treebanks and the methodologies and tools used for constructing them. The research reported in this paper provided the requirements specification and design basis for the *DepAnn* annotation tool and the *FiEval* treebank.

[2] Kakkonen, T.: DepAnn - An Annotation Tool for Dependency Treebanks. Proceedings of the 11th ESSLLI Student Session at the 18th European Summer School in Logic, Language and Information. Malaga, Spain, 2006.

> The paper introduces the DepAnn annotation tool, explains its design and provides information about its implementation.

[3] Kakkonen, T., Werner, S.: The Annotation Scheme for an Evaluation Treebank of Finnish. Proceedings of the Biannual Conference of the Society for Computational Linguistics and Language Technology. Tübingen, Germany, 2007.

> The paper introduces the design and content of the evaluation treebank of Finnish (FiEval) and discusses the reasons why various decisions were taken.

[4] Kakkonen, T.: Developing Parser Evaluation Resources for English and Finnish. To appear in the Proceedings of the 3rd Baltic Conference on Human Language Technologies. Kaunas, Lithuania, 2007.

> This paper describes the linguistic resources that I developed as part of this research. Two of these resources consist of English texts and one of Finnish texts. I describe the status of the resources and justify the decisions that I made when designing them.

[5] Kakkonen, T., Sutinen, E.: Towards A Framework for Evaluating Syntactic Parsers. Proceedings of the 5th International Conference on Natural Language Processing. Turku, Finland, 2006.

This paper offers a survey of parser evaluation methods and outlines a framework for experimental parser evaluation. The proposed framework focuses on intrinsic evaluation and provides useful information for parser developers. We also discuss ways of using the framework in comparative evaluations.

[6] Kakkonen, T.: Robustness Evaluation of Two CCG, a PCFG and a Link Grammar Parsers. Proceedings of the 3rd Language & Technology Conference: Human Language Technologies as a Challenge for Computer Science and Linguistics. Poznan, Poland, 2007.

Robustness refers to the ability of a device to cope with exceptional circumstances outside its normal range of operation: a parser is robust if it is able to deal with phenomena outside its normal range of inputs. I carried out a series of evaluations of state-of-the-art parsers in order to find out how they perform when faced with input that contains misspelled words. In this paper, I also propose two measures for evaluation based on a comparison of a parser's output for grammatical input sentences and their noisy counterparts. I used these metrics to compare the performance of four parsers and analyzed the decline in each of the parser's performance as error levels increased.

[7] Kakkonen, T., Sutinen, E.: Coverage-based Evaluation of Generalizability of Six Parsers. To appear in the Proceedings of the Third International Joint Conference on Natural Language Processing. Hyderabad, India, 2008.

We carried out a series of evaluations of different types of parsers using texts from several genres such as newspaper, religion, law and biomedicine. This paper reports the findings of these experiments.

Papers [2], [5], [6] and [7] were accepted based on a full review and papers [1], [3] and [4] based on the abstract.

Figure 1-4 illustrates the structure of the thesis and the publications related to each chapter. Table 1-2 summarizes the chapters of the thesis and relates each chapter to the research tasks and the papers published.

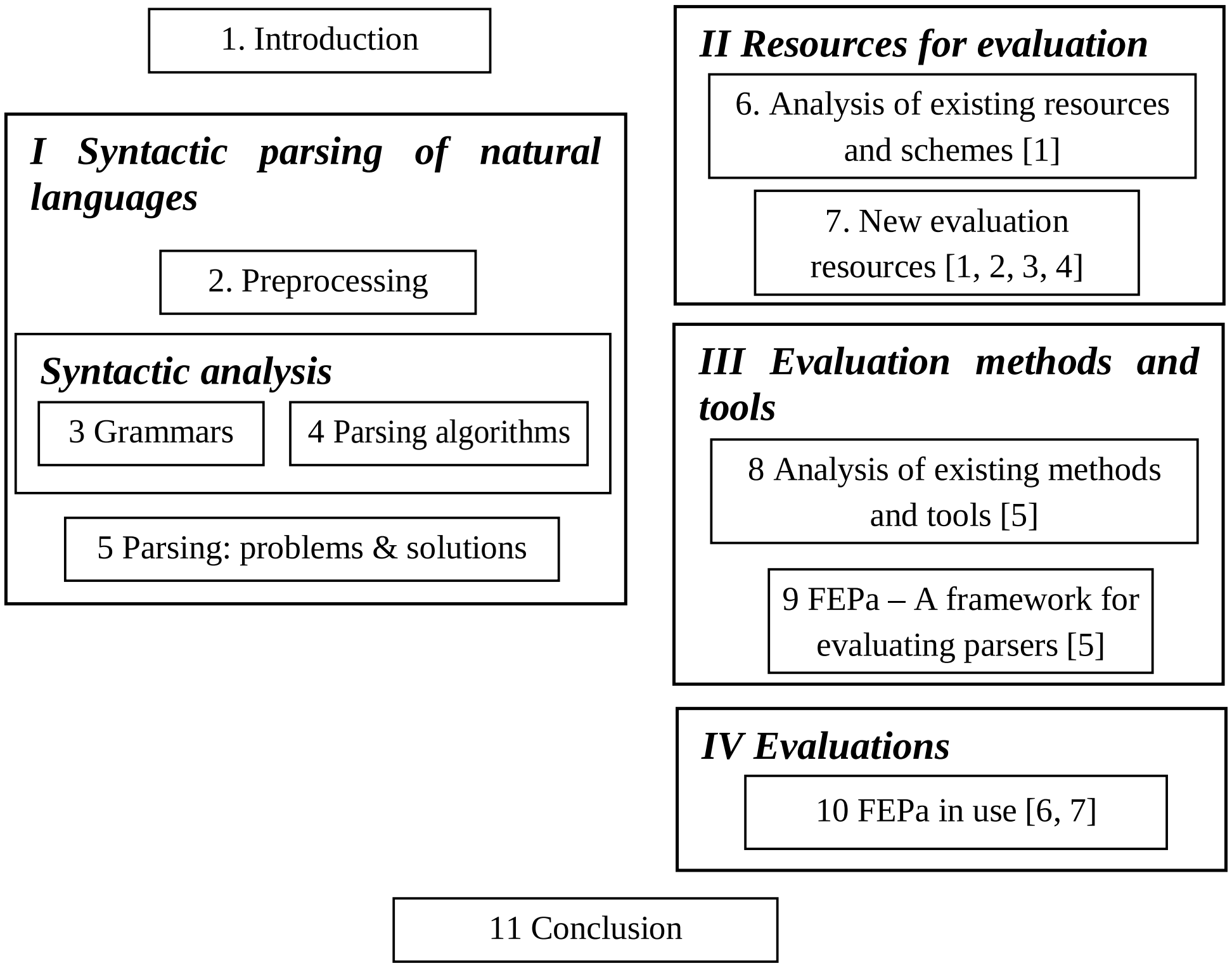

**Figure 1-4.** The structure of the dissertation and the publications based on the content of each chapter.

**Table 1-2.** The structure of the dissertation.

| Part | Chapter | Research task | Papers |
|---|---|---|---|
| I Syntactic parsing of natural languages | 2 Preprocessing | a | - |
| | 3 Syntactic analysis – Grammars | a | - |
| | 4 Syntactic analysis – Parsing algorithms | a | - |
| | 5 Parsing: problems and solutions | a | - |
| II Linguistic resources for evaluation | 6 Analysis of existing resources and schemes | a, b | 1 |
| | 7 New evaluation resources | b | 1, 2, 3, 4 |
| III Evaluation methods and tools | 8 Analysis of existing methods and tools | a, c | 5 |
| | 9 FEPa – A framework for evaluating parsers | c | 5 |
| IV Evaluations | 10 FEPa in use | c,d | 6, 7 |
| 11 Conclusion and future work | | - | - |

Although this thesis is in the field of computer science, it assumes a familiarity with the fundamentals of linguistics. Readers who are unfamiliar with the concepts of morphology and syntax may care to refer to Appendix A that contains a glossary of grammatical terms and terminology. The textbooks of Tallerman (1998), Katamba (1993) and Haspelmath (2002) are among the best sources of further information. A reader who is already well acquainted with parsing technologies might prefer to concentrate on Parts II, III and IV, which cover the main contributions of this work. However, since Part I draws together a great deal of scattered information about different approaches to parsing, it offers new perspectives and insights into parsing research. One of the main challenges in the theoretical analysis of the contemporary practices in syntactic parsers was the heterogeneity of the concepts and notations used by the researchers in the field. This work analyzes different approaches by using a coherent set of concepts and uniform notations.

As far as I know, this thesis represents the only published review of current best practices in parsing and parser evaluation on this scale. Apart from conference papers and journal articles that describe a single or only a few evaluation schemes or resources, this thesis is the first work on parser evaluation to deal comprehensively with this topic by covering theoretical and practical evaluation as well as evaluation tools and linguistic resources for evaluation. The example sentences in the thesis were either invented by the author or taken from literature and adapted to a single genre, football.

# I SYNTACTIC PARSING OF NATURAL LANGUAGES

## 2 Preprocessing

Before the syntactic analysis can be performed, input sentences must be *preprocessed*: Firstly, the units (sentences and words) need to be identified by *segmentation*. Segmentation methods are introduced in Section 2.1. In the second place, it is necessary to perform *part-of-speech* (POS) tagging and *disambiguation* (the process of selecting the correct tag from a set of possible tags) (Section 2.2) and a morphological analysis (Section 2.3). In *syntactic analysis* (Chapters 3 and 4), the syntactic structures of each input sentence are identified and marked. The labels assigned to words to denote their POS, morphological and syntactic roles are called *tags*. A *tagset* is the collection of tags used for a particular task. Figure 2-1 illustrates the sub-processes of syntactic parsing.

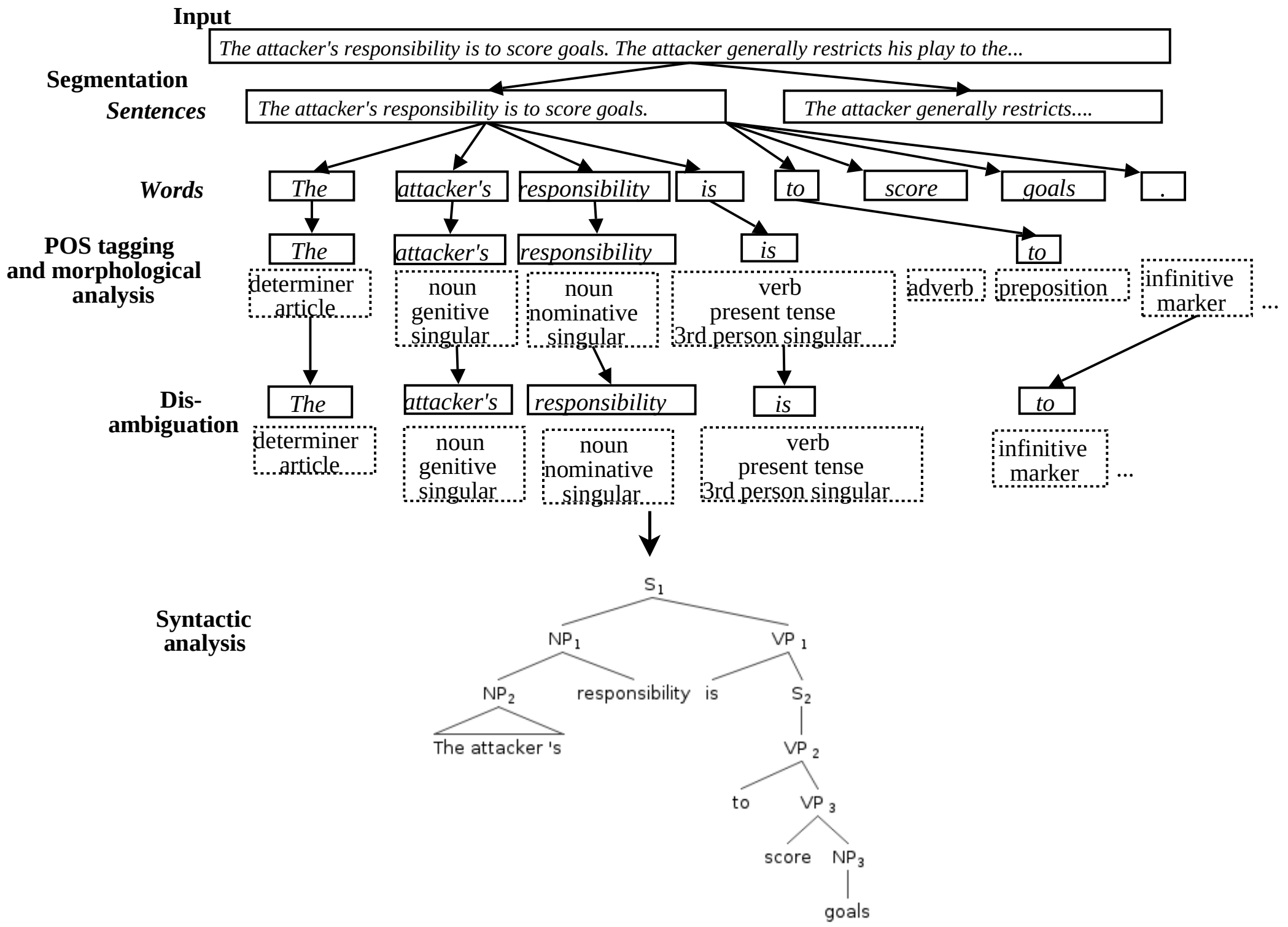

**Figure 2-1.** Segmentation, POS tagging and disambiguation, morphological analysis, and syntactic analysis. These processes are often interwoven. POS tagging and morphological analysis, for example, are typically performed simultaneously.

## 2.1 Segmentation

Segmentation, the process of identifying the text units, consists of *sentence* and *word segmentation*. Segmentation is an essential part of preprocessing; in order to assign structural descriptions to a sentence and the words in it, it is necessary first to identify these units.[3] The methods used in sentence segmentation can often also be applied to word segmentation.

Sentence segmentation is complicated because of the fact that end-of-the-sentence punctuation marks are ambiguous. In addition to the end of sentence, a period can denote, for instance, an abbreviation or a decimal point. An exclamation point and a question mark may occur within parentheses or quotation marks. Furthermore, the use of abbreviations, dates, and so on, depends on the text *genre*[4]. The number and type of ambiguous punctuation marks therefore vary tremendously between texts.

**Example 2-1**. Two sentences that illustrate period-related word segmentation decisions (adapted from Palmer 1994):

> *The game was rescheduled to Saturday 5 p.m. Sunday would be too late.*
> *The game was played at 5 p.m. Saturday to avoid the rain.*

In its simplest form, a word segmenter consists only of a set of rules that reduces any sequence of spaces, tabulation marks and new lines to a single space, and considers everything between two spaces to be a token. A *token* is a sequence of alphabetic characters or digits or a single non-alphanumeric character.

One solution to the segmentation problem is to use *regular expression grammars* that set out to identify patterns of characters that signal the way in which sentences end (for example "period-space-capital letter") (Palmer 1994). More advanced systems also take into account the preceding and succeeding words, and make use of lists of abbreviations and proper names. When they have been properly configured, such systems achieve an accuracy rate of up to 99%.

---

[3] It is simpler to segment texts in languages that use Latin or Cyrillic alphabets that indicate word boundaries with spaces and punctuation marks than it is in languages that use Chinese-derived writing systems (Grefenstette & Tapanainen 1994). While most published work in word segmentation report on segmentation in Chinese, this research is limited to segmentation for the Latin alphabet.

[4] The classification of texts in terms of domain, genre, register and style is a rather controversial issue (see, for example, discussion by Lee (2001)). A detailed analysis of these issues falls outside of the scope of this thesis. The purpose of this research is not to investigate the classification of texts but rather to undertake a practical evaluation and assessment of the extent to which any typology of texts exerts an influence on parsing performance. I have therefore adopted a simplified approach by indicating differences between texts by using the word *genre*. One may think of genres (in this sense) as indicating fundamental categorical differences between texts that are revealed in sets of attributes such as domain (e.g. art, science, religion, government), medium (e.g. spoken, written), content (topic, theme) and type (narrative, argumentation, etc.).

Karttunen (1996), for example, proposes a word segmentation approach based on *finite-state transducers* (FSTs)[5]. The word segmenter is based on FSTs accompanied by a list of multiword tokens. Yet another approach is to use extensive genre-specific word lists and name recognition routines as well as modules that analyze the structure of words. The *SATZ system* (Palmer 1994) is based on neural networks with *descriptor arrays* that represent the context that surrounds punctuation marks. Such contexts are modeled by the probability that POS tags will precede and also succeed words. In Palmer's experiments the system achieved a 98.5 to 99.3% accuracy with English, German and French data after an automatic training with between one hundred and a few hundred training examples.

Modifying a segmenter until it is able to cope with new text genres and languages can be difficult. For example, a segmenter based on a regular expression grammar cannot be easily adapted. These special-purpose grammars are limited to the text genre for which they were developed, and any attempt to adapt them to different genres or languages would be complicated. The greatest barrier to accurate word segmentation involves the recognition of words that do not occur in the word lists of the segmenter.

It is therefore far more practical to devise a trainable algorithm that can compensate for such inadequacies rather than to attempt to construct a single exhaustive word list or a series of genre-specific lists. Without going into details about the inner workings of such systems, one may observe that accuracy rates of up to 80% can be achieved with English de-segmented texts (texts in which the spaces have been removed and word boundaries are not explicitly indicated) after training with only 4,500 sentences. Another example of a highly adaptable approach is Chanod and Tapanainen's (1996) system, in which segmentation is interleaved with morphological analysis.

[5] A *finite-state machine* (FSM) is a model composed of a finite number of states, transitions between those states, and actions. In contrast to an FSM that has a single tape, an FST consists of two tapes. The tapes are typically viewed as an input tape and an output tape. The transducer translates the contents of its input tape to its output tape.

## 2.2 Part-of-Speech Tagging and Disambiguation

In POS tagging, the appropriate word class tag is automatically assigned to each word.[6] The process of selecting the correct tag from a set of possible tags is called *POS disambiguation* (see s Section 2.2.3). The size of tagsets varies considerably. For example, the widely used POS tagsets for English, namely the *Brown Corpus* (BC) (Francis & Kucera 1979) and the *Penn Treebank* (PTB) (Marcus *et al.* 1993) tagsets, consist of 87 and 45 tags respectively. Definitions of the task that a POS tagger performs are given in Definition 2-1.

> **Definition 2-1**. POS alignment and grammatically correct POS alignment.
> Let $\delta = (\alpha_1\ \alpha_2 \ldots \alpha_n)$, be a sentence, where $\alpha_i$, $i = 1,\ldots,n$ are the words. Let $T = (t_1\ t_2 \ldots t_n)$ be a sequence of POS tags $t_i$, $i = 1,\ldots,n$. The pair $(\delta, T)$ is a POS *alignment* of a sentence. A *grammatically correct POS alignment* is an alignment in which the POS tag for each word has been correctly assigned. For an ambiguous sentence, there exists a set of grammatically correct POS alignments.

The task that a POS tagger performs is to find the grammatically correct POS alignment(s) for each sentence in the input.

**Example 2-2.** POS tagging result for the sentence ”The attacker’s responsibility is to score goals.” W = (The attacker’s responsibility is to score goals), T = (Determiner Noun Noun Verb to Verb Noun).

*The*$_{<Determiner>}$ *attacker’s*$_{<Noun>}$ *responsibility*$_{<Noun>}$ *is*$_{<Verb>}$ *to*$_{<to>}$ *score*$_{<Verb>}$ *goals*$_{<Noun>}$.

The two basic approaches to POS tagging are *rule-based* and *probabilistic*. The earliest POS taggers were composed of a set of hand-constructed rules and a small dictionary. In such cases tagging was based on word properties such as an initial capital letter, suffixes and contextual information based on the succeeding and preceding word. The third approach, *transformation-based* tagging, combines components from both the rule-based and probabilistic methods (Brill 1995).

---

[6] The classical set of POS classes includes the following: *noun*, *verb*, *adjective*, *article* (or *determiner*), *preposition*, *adverb*, *conjunctive adverb*, *coordinate conjunction*, and *interjection*. While such a set of classes is often used to teach grammar in schools, it is not adequate for a comprehensive syntactic description but serves only as a basic set that most linguists and POS tagging systems would use as a starting point.

### *2.2.1 Probabilistic tagging – from extended rule-based methods to Hidden Markov Models*

In a probabilistic tagger the probabilities for POS tags for each word are automatically learned from a training corpus. The probabilistic model assigns the most probable tag for each word in the input. The first probabilistic tagger by Stolz *et al.* (1965) used a small dictionary and rules for tagging the most common words, and applied probabilities derived from a manually annotated text for words that could not be identified by either the rules or the dictionary.

Since the development of the fundamental *Hidden Markov Model* (HMM) methods during the late 1980s, HMMs have been widely used for POS tagging (for example, Church (1988) and DeRose (1998)). In these HMM-based *n*-gram models, it is typical that a simplifying assumption is made in order to reduce the number of probabilities to be estimated (Weischedel *et al*. 1993). Rather than assume that the current word $w_i$ depends on all previous words and tags, one assumes the tag $t_i$ depends only on the previous *n*-1 tags, and not all the previous tags. For example, a 3-gram model assumes that the probability of each tag $t_i$ can be approximated by its local context consisting of the tags $t_{i-2}$ and $t_{i-1}$; $P(t_i | t_{i-2}, t_{i-1})$.

The taggers by Weischedel *et al*. (1993) and Merialdo (1994) are fully HMM-based. The probabilities can be estimated for an HMM tagging model by using either supervised or unsupervised learning (Weischedel *et al*. 1993, Merialdo 1994). While training is undertaken with the use of manually annotated data in *supervised learning,* training data is not annotated in the case of *unsupervised learning*.

*Maximum entropy* (ME) (log-linear) models such as *MXPOS* (Ratnaparkhi 1996) and the *Stanford POS tagger* (Toutanova & Manning 2000), have been successful in tagging. ME is a method for analyzing available information from a noisy set of data in order to determine the probability distribution. These models offer a way of combining diverse pieces of contextual evidence (e.g. the surrounding words) in order to estimate the probability of a certain POS tag occurring in a specific linguistic context (Ratnaparkhi 1997a). The procedure used by ME modeling is to choose the probability distribution *p* that has the highest entropy of all distributions that satisfy a certain set of constraints.

For instance, in MXPOS the context is typically defined as the sequence of several words and tags preceding the current word $w_i$ (Ratnaparkhi 1996, 1997a).

Ratnaparkhi defines the probability model over $H \times T$, where $H$ is the set of possible word and tag contexts (referred to as *histories*), and $T$ is the set of permitted tags. The probability of history $h_i$ together with tag $t_i$, $p(h_i,t_i)$, is defined as:

$$p(h_i, t_i) = \pi\mu \prod_{j=1}^{k} \alpha_j^{f_j(h_i, t_i)} \qquad (2\text{-}1)$$

In Equation 2-1 $\pi$ is a normalization constant, $\{\mu, \alpha_1, ..., \alpha_k\}$ are the model parameters and $\{f_1, ..., f_k\}$ features, where $f_j(h_i, t_i) \in \{0,1\}$. Features encode information that contribute to predicting the tag of $w_i$. These include the tags preceding $w_i$ and the spelling of $w_i$. A feature may activate (and is set to 1) on any word or tag in the history $h_i$. Each parameter $\alpha_j$ corresponds to a feature $f_j$. These are set in the training phase to maximize the likelihood of the training data.

### *2.2.2 Transformation-based methods – combining rules and probabilities*

Transformation-based error-driven learning, a combination of probabilistic and rule-based approaches, has been applied to POS tagging (Brill 1995). The *rewrite rules* for assigning the correct tags are learned automatically by statistical inference. The rules are derived from *transformation templates.*[7] Figure 2-2 illustrates how the tagger works.

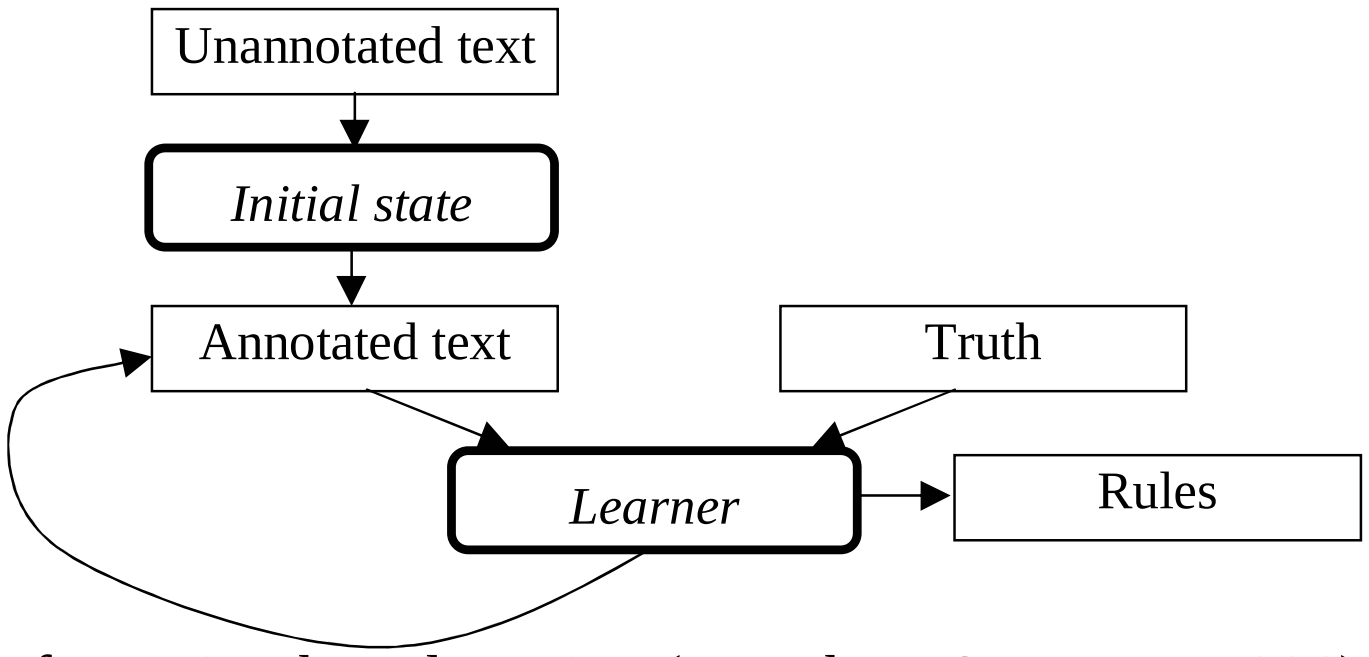

**Figure 2-2.** Transformation-based tagging (Ramshaw & Marcus 1996).

The transformation rules are learned in the following way (Figure 2-2). First, in the initial state, words are set with an initial tag by assigning the most likely tag or the noun tag to each word. Secondly, all the possible transformations are applied to the unannotated training text, and this creates *annotated text*. Thirdly, the tagging produced in the second phase is compared to the *truth*, a manually annotated text. Finally, the transformation rule that yields the highest reduction in tagging errors is chosen and applied to the learning corpus. The process continues until no transformations are found that reduce the error rate.

[7] A transformation contains two components: a rewrite rule and a triggering environment. An example of this is: "Change the tag from IN (preposition or subordinating conjunction) to RB (adverb) if the word two positions to the right is 'as'."

### *2.2.3 POS disambiguation*

The complexity of the tagging problem is caused by the fact that an ambiguous word may have several possible tags. Words that have one possible POS tag are called *unambiguous*. In the BC, for example, about 11% of the words are ambiguous between two or more POS tags (Charniak 1993). Words have up to seven possible tags in the corpus. Since ambiguous words tend to be words that occur frequently, over 40% of the word instances are ambiguous.

*POS disambiguation,* selecting the correct tag for each word, is made simpler by the fact that the various tags for a word are not equally likely. While some taggers try to guess a single POS tag for each word, others leave some ambiguities unresolved. One approach to POS disambiguation, namely *Constraint Grammar* (CG), is to use manually written constraints that allow for the discarding of contextually illegitimate ones from a list of all possible readings[8] for a word (Voutilainen & Heikkilä 1993). A constraint could, for example, remove all finite verb readings of a word $w_i$ if the immediately preceding word $w_{i-1}$ is “to”.

### *2.2.4 Analysis*

This section analyses the POS tagging models on the basis of their accuracy in assigning correct tags, and considers the ways in which current best practice in tagging might be even further improved.

The earliest, dictionary-consulting and rule-based POS tagging methods achieved an accuracy of somewhat over 90 per cent, measured as the percentage of the same tags assigned by the system and human taggers (Klein & Simmons 1993). The most obvious disadvantage of rule-based POS tagging methods is the unavoidable labor-intensiveness of rule writing. Training a probabilistic tagging model would reduce the need for such efforts. Table 2-1 compares the results obtained with the taggers discussed above.

[8] A reading represents the word paired with its POS and other morphosyntactic tags. Clearly, an ambiguous word has more than one reading.

**Table 2-1.** An overview of the results in POS tagging. The results are taken from the papers cited. The same data (Beal 1988) was used in all experiments.

| Tagger | Approach | Training set (words) | Accuracy |
|---|---|---|---|
| Weischedel *et al*. (1993) | HMM | 64,000/~1,000,000 | 96.3/96.7 |
| Merialdo (1994) | HMM | ~1,000,000 | 97.0 |
| MXPOS (Ratnaparkhi 1996) | ME | ~960,000 | 96.4 |
| Stanford tagger (Toutanova & Manning 2000) | ME | ~1,000,000 | 96.9 |
| Brill (1995) | Transformation -based learning | 64,000/600,000 | 96.7/97.2 |

State-of-the-art HMM taggers achieve an accuracy rate of around 97% when they have been trained with supervised learning. Unsupervised learning is useful when no manually annotated data is available. In the experiments of Merialdo (1994), the accuracy dropped by roughly 10 percentage points when unsupervised learning was applied. Using just 100 annotated sentences for training outperformed the accuracy achieved by the unsupervised method. It is possible to define and incorporate much more complex statistics in the ME framework if one does not restrict oneself to *n*-gram sequences[9]. But this, as is indicated by the results above, does not seem to boost practical performance.

Samuelsson and Voutilainen (1997) reported a comparison with a rule-based CG tagger and Church/DeRose type trigram HMM-taggers[10] on BC data. The results of the experiment are summarized in Table 2-2.

**Table 2-2.** The accuracy of CG and HMM taggers reported on different ambiguity levels (Samuelsson & Voutilainen 1997). Parenthesized values are obtained by interpolation.

| Ambiguity (tags/word) | % tags correct | |
|---|---|---|
| | HMM | CG |
| 1.000 | 95.3 | |
| 1.026 | (96.3) | 99.6 |
| 1.051 | 96.9 | |
| 1.070 | (97.2) | 99.9 |

The results in Table 2-2 give an insight both into the relative accuracy of the two approaches and the effects of unresolved ambiguity on accuracy. In addition to the fact that CG was superior to the HMM-based tagger in this experiment, the results

9 The model by Toutanova and Manning, for example, includes features that check whether all the letters of a word are uppercase. It also checks the context until the preceding verb is activated when the word contains an uppercase character and is not at the beginning of a sentence, etc.

10 The HMM-tagger was trained with 357,000 words from the BC and tested with 55,000 words.

show that the performance of the HMM tagger suffers considerably when it is trained and tested on different corpora as was the case in Samuelson & Voutilainen's experiment. Words unseen in the training phase accounted for 2.01% of the errors.

Entwisle and Powers (1998) point out that one has to be circumspect about accepting the percentage accuracy scores for POS taggers reported in the literature. When CG, for example, which is one of the most accurate taggers available, was tested with a test set, it became apparent that 18.0% of the words were assigned more than one tag and that 94.8% of the words had a correct tag among the suggested tags. Thus, only 82.0% of the words had been assigned a single, correct tag.

Although the accuracy rates given above are only indicative (because not all the experiments were performed with the same sets of POS tags and training data), they still offer an insight into the progress made in the POS tagging and current state-of-the-art practices in tagging. The accuracy figures for the two types of probabilistic POS taggers –HMM and ME-based – are very similar. While a rule-based tagger, CG, has the highest reported accuracy rate in the literature, this is achieved partly by assigning, in some cases, more than one tag per word. This of course makes it easier to achieve a high level of accuracy. This feature, however, may be an advantage in those cases where a tagger is not able to decide the correct tag for a given word. Retuning one tag that is incorrect would cause the sentence analysis to fail in later stages. It is difficult on the basis of these results alone to state definitively that either of these two approaches is better than the other.

Progress in achieving greater accuracy rates has almost ground to a halt in recent years. Ratnaparkhi (1996) notes that the accuracy of state-of-the-art taggers at around 96-97% represents the upper limits of what can be achieved, or is at least close to it. Because manually written rules and tagged training and testing corpora will always contain errors and inconsistencies, it is impossible in practice to reach a 100% accuracy rate. Errors in POS tagging lead to problems in later stages of parsing. Such errors may in fact be the single most important source of error in parsing (Dubey 2005). As I shall point out later in Section 5.1.1, precise POS tagging can greatly boost the overall performance of a parser.

Near-100% POS disambiguation accuracy is achievable only by taking into account the syntactic contexts in which the words occur. This means that rather than as preprocessing, POS tagging should be performed parallel with syntactic analysis.

## 2.3 Morphological Analysis

*Morphological processing* (Section 2.3.1) deals with the analysis and generation of word forms.[11] In the context of syntactic parsing, one is concerned with the former – with analyzing the surface form of a word and producing the output that represents the morphological features of any given word. In morphological analysis, one recognizes the structure and morphological properties of words (Sproat 1992). A *morphological analyzer* ("word parser") is needed for automatically computing the word information (Section 2.3.2).

### *2.3.1 Morphology and morphological processing*

The *morphological processes* of a natural language either create word forms from the root form or new words. In *inflectional morphology*, word forms appropriate to a particular context are formed from the root of that word (Sproat 1992). In word formation, on the other hand, a word is transformed into a different word either by *derivation* or by *compounding*.[12] The main difference between inflection and word-formation is that the latter is never required by syntax, whereas the former is often necessitated by particular syntactic contexts. For a morphological analyzer used in a syntactic parser, the main concern is therefore the inflectional morphology. This is especially important in parsers of highly inflected languages such as Finnish or Turkish. Table 2-3 shows examples of some words and their morphological analysis.

**Table 2-3**. Examples of morphological analyses. The first column gives the original word form while the second column shows the morphological features that are the result of the analysis.

| Word form | Morphologically analyzed form |
|---|---|
| players | player<noun> <plural> |
| kicked | kick<verb><past tense> *or* kicked<adjective> |
| run | run<noun><singular> *or*<br>run<verb> <present tense><non-3rd person singular> *or* run<verb><past perfect> |
| ball | ball<noun><singular> *or* ball<verb><present tense><non-3rd person singular> |

[11] In linguistics, a unit called *morpheme* refers to what is the common sense notion of a word. A morpheme is the smallest meaningful constituent of a linguistic expression (Haspelmath 2002). For example, the word "unbeatable" consists of three morphemes: "un-" (meaning not *x*), "-beat-" and "-able". "un-" is also a prefix, "-able" is a suffix.

[12] In derivation, derivational suffixes or prefixes, such as "-ment", "-ism", "anti-", and "dis-" in English, are added to the input word to form a word with a different meaning that might belong to a different POS category. In compounding, two or more lexemes are combined into a compound word.

An approach to morphological processing that is based on looking up full-form dictionary entries is bound to fail because of the productive nature of morphological processes (Sproat 1992). On the one hand, the word lists would eventually become too large to store and process. On the other hand they would never be complete because new words are constantly being generated.[13]

Because most of the forms are formed according to general and regular morphological processes, a morphological analyzer can store the base forms and compute the other forms according to the rules. The most common type of morphological analyzer used as a component of syntactic parser finds all the possible forms of a word and lets the parser decide which one of them is the most appropriate for that context.

### *2.3.2 The two-level model*

The *two-level model* (TWOL model) (Koskenniemi 1984) offers a good introduction to morphological analyzers for two reasons. Firstly, the computational mechanisms used in TWOL are commonly applied in other morphological analyzers.[14] Secondly, the *KIMMO* system, which is based on the model, is one of the most successful and most widely used morphological analyzer (Sproat 1992).

A word is represented in the TWOL model as a direct letter-to-letter correspondence between its lexical form and its surface form (Koskenniemi & Church 1988). The *surface representation* is typically a phonemic description of the word-form, like “tackled”. The *lexical representation*, for example *tackle+ed*, *tackle*$_{\text{<verb> + <past tense>}}$, is a description of the root and affixes of a word. If a language has phonological alternations, the two representations are not identical. The task of the TWOL rule component is to account for discrepancies between these representations.

---

[13] It would be impractical even to attempt to update such lists – especially for *agglutinative languages* such as Finnish and Turkish. While English may have a limited inflectional system, its derivational morphology is complex. One can, for example, derive the forms “computer”, “computerize”, “computerization”, “computational”, “recomputerize” etc. from the single root “compute”.

[14] The original Koskenniemi (1984) TWOL model was developed for *concatenative morphology*, in which words are formed by concatenating series of morphemes together. Concatenative morphology is especially interesting, since it is the most common model cross-linguistically (Sproat 1992). In agglutinative languages, inflectional morphology is based wholly on concatenation.

The *KIMMO morphological analyzer* has two main components: the rules and the lexicon (Karttunen 1983). The *lexicon* lists all the morphemes and specifies *morphotactic* constraints. The *rules* describe the phonological and orthographic alternations of word forms. Figure 2-3 illustrates the main components of KIMMO, and Figure 2-4 illustrates the structure of the lexicon in KIMMO.

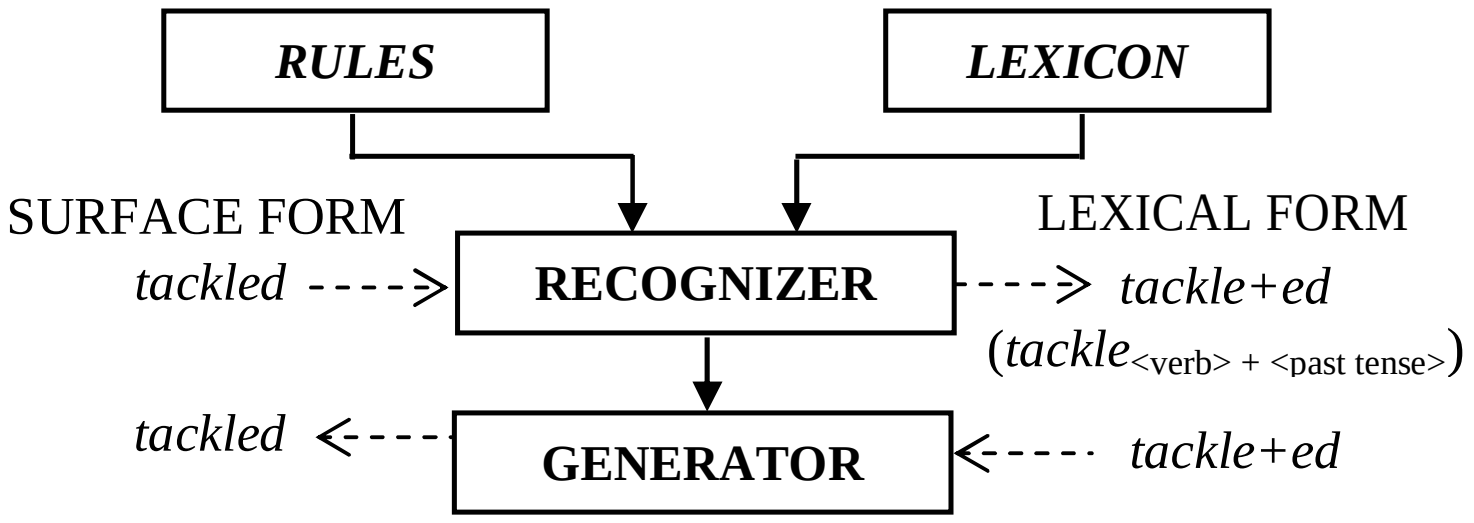

**Figure 2-3.** The main components and functions of the KIMMO morphological analyzer (adapted from (Antworth 1994)). The recognizer applies the rules and the lexicon to recognize surface form input and then outputs the lexical form of the word. The generator generates surface forms from the lexical forms given as the input.

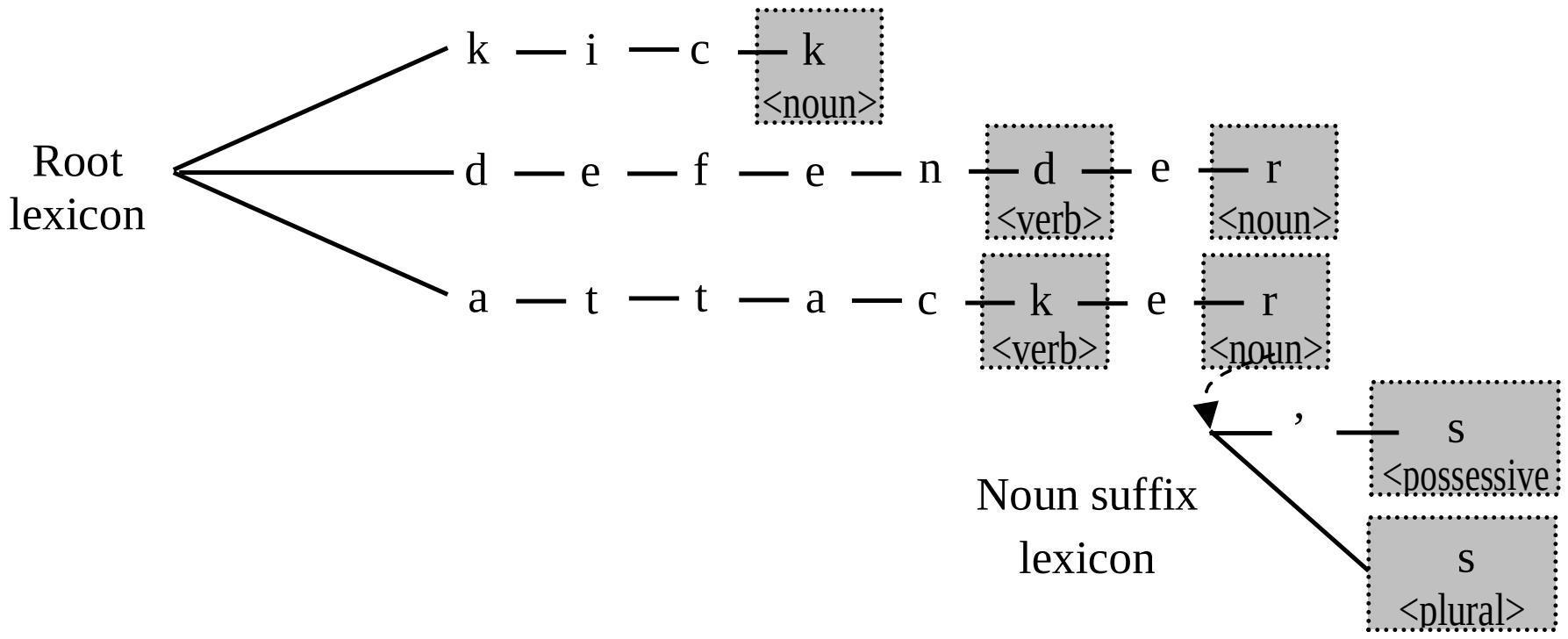

**Figure 2-4.** Lexicon in the TWOL model. Each lexical entry is marked with information of its POS and *continuation patterns*, indicating zero to *n* morpheme lexicons that can be applied to the word. For example, the continuation patterns of the nouns “defender” and “attacker” would indicate that the search can proceed to the continuation lexicon of noun suffixes.

The TWOL rules and the lexicon are separated from the processing components. It is therefore a relatively straightforward matter to adapt the model to new languages because the program itself remains untouched (Sproat 1992). While the system was originally implemented for an analysis of Finnish, it was later adapted to several languages such as English (Antworth 1994), Turkish (Oflazer 1994), Turkmen (Tantuğ *et al*. 2006), Korean (Kim *et al*. 1994), and Japanese (Alam 1983), many of which have typological properties quite different from those of Finnish.

### *2.3.3 Analysis*

As I have noted above, the most common interaction between the morphological and syntactic analyzers in a parser consists of the morphological analysis component providing a complete list of possible analyses for a word and the higher-level component selecting the most appropriate form. It would, however, be more desirable to have a bidirectional interaction between the morphological and syntactic analysis that would allow one to use the syntactic context to guide the analysis of complex word forms and compound words.

A unique feature of the TWOL model compared to the other morphological analysis methods of that time was that it was applicable to a wide range of languages, even to ones with a nonconcatenative morphology, after some modifications. Furthermore, it is relatively straightforward to modify the model for new languages. Several augmentations and improvements have been introduced to the KIMMO analyzer. In the original system, the rules had to be coded manually. Rule compilation is an error-prone activity that calls for a detailed understanding of the TWOL model and its rules (Karttunen & Beesley 2001). Koskenniemi (1986) developed a rule compiler for automatically constructing the FSTs from the rules.

# 3 Syntactic Analysis - Grammars

The most complex task that a natural language parser has to perform is syntactic analysis. The two main parts of the syntactic analysis component of a parser are the *grammar* and the *parsing algorithm* (Pereira 1998, Zaenen & Uszkoreit 1998).[15] The grammar encodes the linguistic rules and specifies how each sentence is constructed from its parts. The parsing algorithm applies the rules defined by the grammar to a given input. The *output scheme* defines the format of the parser's output. Figure 3-1 summarizes the main components and the structure of natural language parsers.

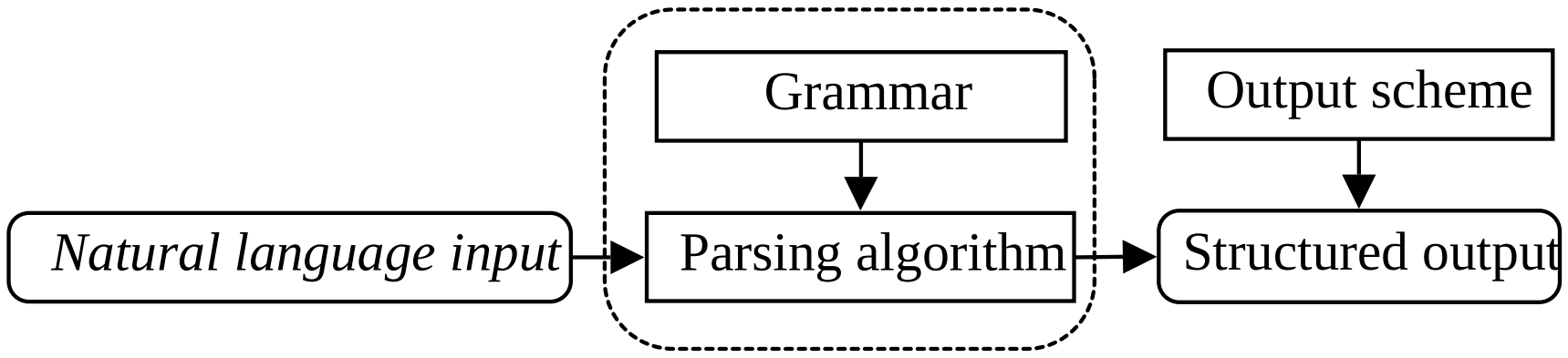

**Figure 3-1.** The main components of a parser: the grammar, parsing algorithm and output scheme.

Grammars can be characterized and compared in many different ways. In Chapters 3 and 4 I use two different approaches. On the one hand, grammars are characterized in terms of their linguistic properties such as background, linguistic assumptions, the way in which they classify strings as grammatical or ungrammatical, and type of analyses they offer (Chapter 3). On the other hand, I also approach the classification of grammars from a processing perspective and include the computational complexity of the grammars and their suitability for computational purposes (Chapter 4).

The purpose of this chapter is not to provide an exhaustive introduction to the types of grammars (i.e. *grammar formalisms*) that have been developed, but rather to identify and discuss the reasons why specific types of grammars can be successfully applied in practical parsing systems. Because of the number of grammar formalisms that exist, it would not be possible in a work of this nature to examine all the formalisms that are applied in parsing, let alone the variants that exist in grammar formalisms. I will therefore confine myself to looking at the most well-known formalisms that have been successfully applied in existing parsers and to discussing their most interesting theoretical features. The conclusions I reach may be useful both to parser evaluators and to NLP system developers who are looking for a suitable parser for their application.

[15] The predominant paradigm in the study of syntax is generative, originating from Noam Chomsky's *Syntactic Structures*, first published in 1957. In this work Chomsky makes a distinction between *performance* (the process that actually determines what a speaker will say or how an utterance is understood in a context) and *competence* (an abstract characterization of a speaker's knowledge of the language). The distinction between the grammar and the parsing algorithm may be referred to as the distinction between Chomsky's notions of competence and performance. Competence refers to the set of abstract rules that express our knowledge of the language while performance is defined in terms of how well we actually use these rules.

## 3.1 Structural Representation

Syntactic structures are often depicted as tree-shaped structures. They are often referred to as *dependency* (D) or *phrase structure* (PS) trees, depending on the kind of representation format they use. While D trees describe sentences with dependencies between words, PS trees illustrate phrases and the relationships between them. Figure 3-2 shows examples of both a D tree and a PS tree.

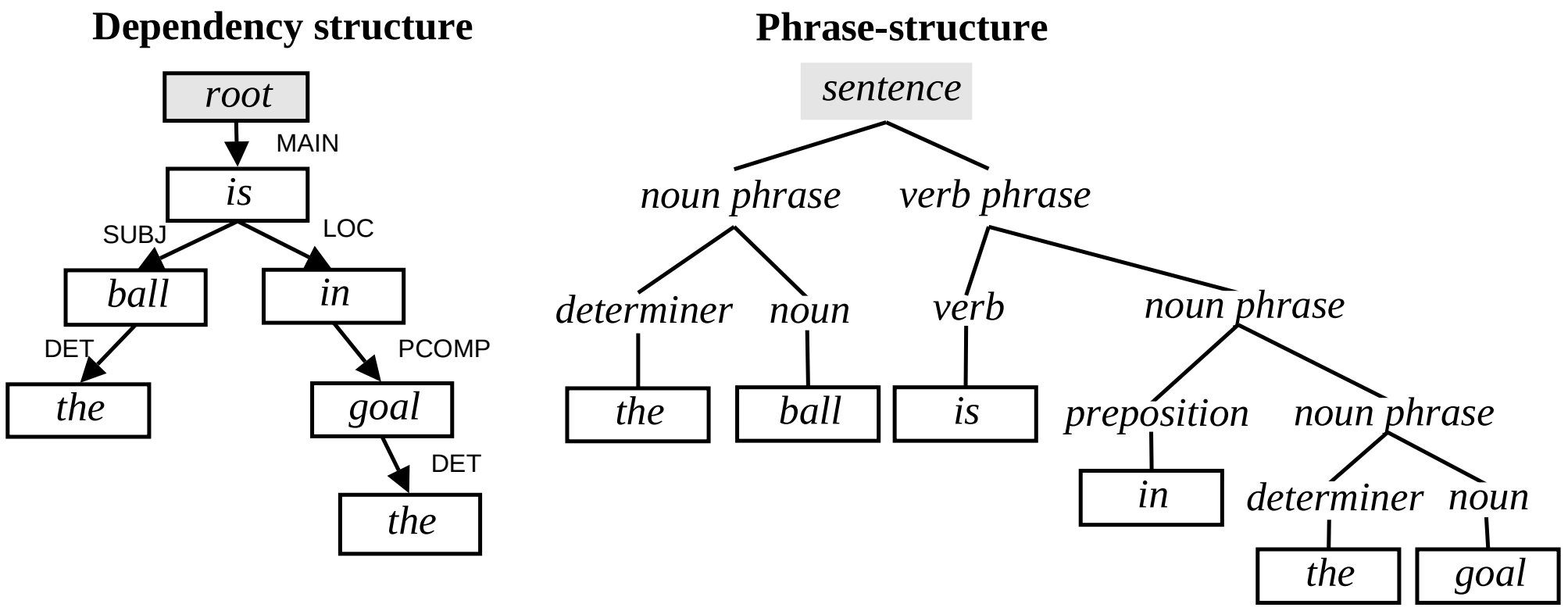

**Figure 3-2.** Dependency and phrase structure trees for the sentence “The ball is in the goal”.

A D tree shows only the word nodes, and these are linked to one another with directed binary relations that are called D links. The D tree of a sentence forms a DAG that consists of a number of nodes that are equal to the number of words in that sentence. A PS tree by contrast shows the words of a sentence in the form of the *terminals* in the leaf nodes. The *nonterminals* between the root and terminal nodes indicate how the sentence is constructed from *constituents*.

The amount of detail in parsers’ outputs varies from one parser to another. Shallow parsers, in comparison to deep parsers, produce a flatter analysis that represents only a part of the sentence structure (Abney 1997). Figure 3-3 shows an example of the differences between a shallow parse and a deep parse.

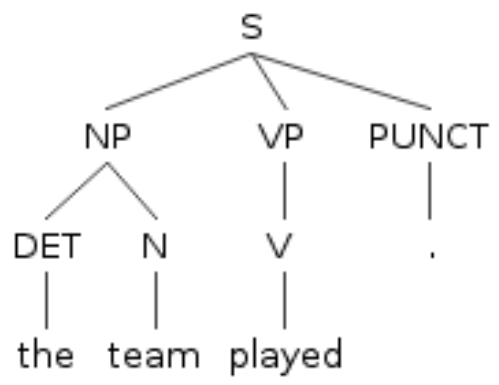

$$\begin{bmatrix} \textit{PRED} & \textit{played} \\ \textit{SUBJ} & \begin{bmatrix} \textit{PRED} & \textit{team} \\ \textit{SPECIFIER} & \textit{the} \end{bmatrix} \\ \textit{TENSE} & \textit{past} \end{bmatrix}$$

*The*$_{<DET>}$ *team*$_{<N>}$ *played*$_{<V>}$ .$_{<PUNCT>}$
[*the team*]NP [*played*]VP

**Figure 3-3.** Examples of deep parse (on the left) and shallow parse (on the right) for the sentence "The team played." While the shallow parser has performed POS tagging and *chunking* of NPs and VPs, the deep parser has produced a complete parse tree and has, in addition, offered information about the grammatical roles of individual words (such as whether they are predicates, subjects, objects, and so on).

In the following, I analyze the differences between D and PS representations and the consequences of different levels of detail in parsers' outputs. I also focus on the suitability of the two representation formats for describing different kinds of languages and analyze how suitable they may be for specific NLP tasks.

The status of the two representation types, PS and D, is a matter of some controversy (Nivre 2002). While PS has been favored by the transformational syntax community since the time of Chomsky's early works, there are many researchers who regard the D structure as the most fundamental one. But there are other theories that describe both representations as primitive. A PS-tree represents the structure of a sentence in a linear form as a chain of words, and illustrates it as a part-whole construction (Tarvainen 1977). The dependencies between the constituents are not distinctly shown. In contrast to this, a D tree marks the phrase categories only implicitly. It also does not show the word order.

D trees are better suited to representing the structures of languages that have a relatively free word order.[16] Because such languages possess a rich morphology, their word order is freer to express syntactic functions. Another advantage of D trees is that they offer a straightforward interface between syntactic and semantic representation (Covington 2001). D trees can capture much of the *predicate-argument structure* (PAS) which is often needed in practical NLP applications.

The PS representation is especially useful for expressing languages with very limited variations in word order (Covington 1990). Freer variations often produce *discontinuous constituents* that are split up and that are therefore linked by overlapping edges. Structures of this kind can cause problems in PS trees.[17] Figure 3-4 illustrates just such a case.

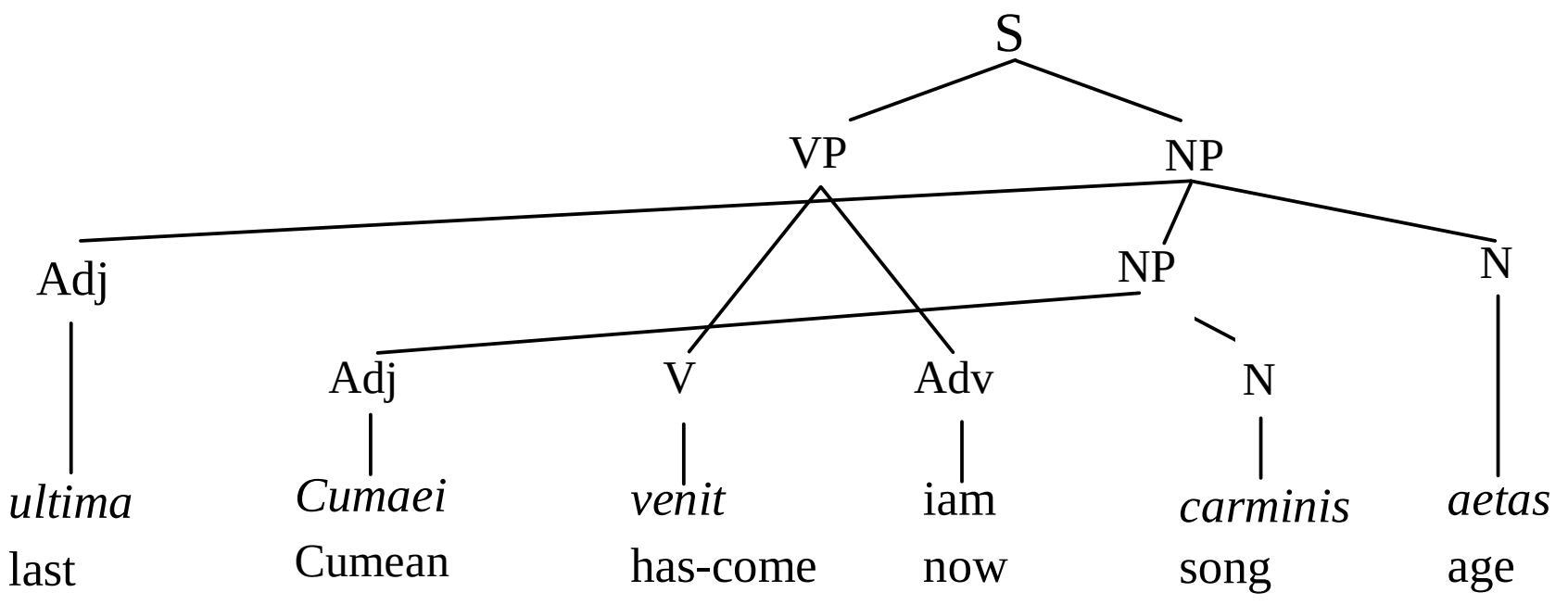

**Figure 3-4.** An example of word order-related problems in a PS tree (Covington 1990).

The current debate about the relative merits of the D and PS trees emphasizes integration and cooperation (Dahl 1980, Schneider 1998). While some elements of PS representation are better for handling certain phenomena (such as *coordination*), D relations permit free word order. One should also consider that a D tree that specifies word order can be converted, under certain conditions, into an equivalent PS tree and vice versa (Covington 1990, Xia & Palmer 2001, Nivre 2002).

The reason for selecting either shallow or deep processing must be based on the NLP application to which the parser will be put (Butt & King 2002). Whereas dialogue and MT systems, for example, need in-depth syntactic analysis, a more shallow analysis may be quite adequate for IR, message understanding and information extraction systems (Oepen *et al*. 2000). One of the advantages of shallow parsing, in comparison to deep syntactic parsing, is that it can be performed more quickly (Grishman 1995). It is also often easier to modify a shallow parser than it is to modify a deep syntactic parser. But deep parsing can provide better generalizations across semantic relations[18] and capture paraphrasing relations between syntactic structures.[19] Because of the richness of the information that they produce, deep parsing models can also offer estimates about how reliable their analyses might be (Oepen *et al.* 2000). This information can be used for avoiding erroneous output and for ranking analyses.

---

[16] Languages such as Russian, Latin and Korean permit extensive variations in word order. On the other end of the scale there are languages such as English, Chinese and French. Finnish, German and Japanese, among others, fall somewhere between these two extremes.

[17] Many proposals, including s*crambling rules* (e.g. Ross (1967)) and *ID/LP formalism* (Gazdar *et al.* 1985) and its modification by Uszkoreit (1987), have been made in order to accommodate discontinuous constituents and other word order related phenomena into the PS framework. None of them has been completely successful.

## 3.2 Basic Concepts and Formal Properties of Grammars

### *3.2.1 Basic concepts*

Any language consists of strings that are represented as segments of symbols of a given alphabet. We can define this formally in the following way:

**Definition 3-1.** Symbol, terminal and alphabet.

A *symbol* is a distinguishable character, such as “a”, “b” or “c”. Any permissible sequence of symbols is called a *terminal* (also referred to as a *word*). A finite, nonempty set $\sum$ of terminals is called an *alphabet.*

**Definition 3-2.** String and sets of strings.

Let $\sum$ be an alphabet. A finite sequence of symbols $S=(x_1\ x_2 \ldots x_n)$, $n \geq 0$, $x_i \in \sum$ is called a *string* in alphabet $\sum$. The *length* |S| of string S is *n*. The *empty string* is the sequence of length 0; written $\varepsilon$. $\sum^*$ is the set of all strings in $\sum$. In addition, $\sum^+ = \sum^* - \{\varepsilon\}$.

**Definition 3-3.** Language and sentence.

Let $\sum$ be an alphabet. Any subset $L$ of $\sum^*$ is called a *language* over alphabet $\sum$. Sequence $\delta = (\alpha_1\ \alpha_2 \ldots \alpha_n)$, where $\alpha_i \in L \forall i, 1 \leq i \leq n$, $n \in \mathbb{N}$ is called a *sentence* in language *L*.

A language follows the rules of a given grammar and is represented by using a particular grammar formalism.

**Definition 3-4.** Grammar, lexicon, rules.

A grammar *G* is a description of a language *L*. A grammar *G* consists of a lexicon and rules. A *lexicon* is a structure defines the terminals in a language. *Rules* describe how the terminals combine into larger entities.

---

[18] For example, the structure of a main clause is the same whether the verb is “to succeed” or “to fire”.

[19] For instance, “X tackles Y”, “Y was tackled by X”, “Y, who tackled X”.

**Definition 3-5.** Language generated by a grammar, derivation, grammatical and ungrammatical strings.

Let $L(G)$ denote that grammar $G$ *generates* language $L$. The process of grammar rule applications is referred to as *derivation*. $L(G)$ is the set of sentences that can be *derived* by the grammar $G$. The sentences that grammar G generates are referred to as *grammatical*. The sentences that are not generated by G are referred to as *ungrammatical*.

**Definition 3-6.** Grammar formalism and grammatical theory.

A *grammar formalism* is a language used for expressing grammars. A *grammatical theory* is the set of statements expressed in a grammar formalism.

The recognition problem is connected to the question if a given string is in a given language. The parsing problem is concerned with the kinds of structures that are assigned to a given string.

**Definition 3-7.** Recognition problem and parsing problem (Ristad 2003, Nivre 2005).

The *recognition problem* (RP) is characterized by the question "Is a given string in a given language or not?". The *parsing problem* answers the question: "What structural descriptions are assigned by a given grammar to a given string?" The parsing problem is tied to the corresponding recognition problem; only strings in $L(G)$ are assigned an analysis in the parsing process. Most parsing algorithms in fact solve the recognition and parsing problem simultaneously.

**Definition 3-8.** Fixed language and universal recognition problem (Ristad 1986).

A language may be characterized in terms of all the grammars that generate it. This is referred to as the *fixed language RP*. It can be stated as follows: "Is the string $S$ in a language $L$?" Another way to characterize a language is by a particular grammar that generates it. This is referred to as the *universal RP* (URP), and can be stated as "Is the string $S$ in a given grammar $G$?" The URP is connected to a specified grammar, and thus, is more closely connected to the parsing problem.

**Example 3-1.** Let a grammar formalism consist of a set of terminals, a set of nonterminals and a set of rules. The nonterminals are the "building blocks" that allow rules to combine terminals and nonterminals into larger entities. Let $\sum$ be a terminal alphabet, $N$ a nonterminal alphabet and $R$ a set of rules. Then $G$=(∑, N, R) is a grammar expressed in the grammar formalism:

$$\Sigma = \{x, y\}$$
$$N = \{S\}$$
$$R = \begin{cases} S \rightarrow xSy \\ S \rightarrow \varepsilon \end{cases}$$

There are two terminals, $x$ and $y$, in the grammar. The nonterminal alphabet consists of the start symbol $S$. The rules $R$ state that $S$ can be rewritten as $xSy$, or as an empty string. Grammar $G$ generates the language $L(G) = \{x^n y^n\}$, $n \in \mathbb{N}$. To generate a string with this grammar, we begin with the symbol $S$, successively rewriting it according to one of the rules $R$ until we cannot rewrite any longer. One string can be derived from another by choosing a rewrite rule whose left-hand side (denoted with α) matches a sequence in that string and by replacing that sequence with the right-hand side of the rule (denoted with β). For example, the string "*xxyy*" can be derived by the grammar $G$ by applying the rule $S \rightarrow xSy$ as $S \rightarrow xSy \rightarrow xxSyy$ and finally the rule $S \rightarrow \varepsilon$ as $xxSyy \rightarrow xxyy$. Thus, the string is grammatical. The string "*yyyx*" cannot be derived by G and is ungrammatical.

### *3.2.2 Context-free phrase structure grammars and transformational grammars*

The grammar formalisms of *context-free phrase structure grammars* (CFPSG) and *transformational grammars* (TGs) form the theoretical basis of several formalisms used in modern parsing systems. *Phrase structure grammars* (PSGs), originating from the works of Bloomfield (1933) and Chomsky (1959, 1965), are designed to study the structure of phrases. A CFPSG is a specific type of PS grammar.

**Definition 3-9**. Context-free phrase structure grammar.

A *context-free phrase structure grammar* is a 4-tuple G = (N, ∑, P, S) where

1. $N$ is a finite set of *nonterminal symbols*.
2. ∑ is a finite set of *terminal symbols*; $\Sigma \cap N = \varnothing$.
3. $P = \{\alpha \rightarrow \beta | \alpha \in N, \beta \in (\Sigma \cup N)^*\}$ is a finite set of *production rules*.
4. $S \in N$ is a distinguished symbol called the *sentence* symbol.

For example, the rule $S \rightarrow NP\ VP$ states that a sentence is a type of phrase that can be composed of a *noun phrase* (NP) and a *verb phrase* (VP). The rules of a CFPSG state how to replace a nonterminal without any reference to the context in which a nonterminal is located (Charniak & McDermott 1985). A *phrase structure* *rule* (PS rule) specifies two relations: *immediate dominance* (ID) between a mother (α dominates β) and *linear precedence* (LP) relation among the daughters (the order of the symbols on β) (Gazdar *et al*. 1985).[20]

In contrast to surface-oriented grammatical formalisms such as CFPSG, TGs assume at least two separate but related levels of syntactic representation. Just like CFPSGs, a TG specifies the permissible sentences of a language by using PS rules (Chomsky 1965, Charniak & McDermott 1985). PS rules build a *deep structure*, which is then modified by *transformational rules* to produce a *surface structure*. Deep structures are able to capture underlying similarities and differences between surface structures. TG is a *multistratal* formalism. A multistratal grammar formalism employs representations in which the structure of a sentence consists of two or more representations, particularly representations which are described in a uniform way.

## 3.3 Grammar Formalisms for Parsing

This section introduces the types of techniques (probabilistic, lexicalized and transition networks) used by grammar formalisms (Section 3.3.1) and a set of grammar formalisms applied in modern parsing systems (Sections 3.3.2 to 3.3.6).

### *3.3.1 Probabilistic, lexicalized and transition network models*

In *probabilistic grammar formalisms* a probability is associated with each rule. For example, *probabilistic context-free grammars* (PCFGs) can be characterized as CFPSGs that assign to each production the probability of its use. A PCFG (Booth & Thompson 1973) is a 4-tuple G = (N, ∑, P, S) like a CFPSG (Definition 3-9), except that each rule in *P* is associated with a probability in which *α* will be expanded to *β*.

[20] PS rules are of the α→β kind. The symbols listed on β are obligatory unless they are in parentheses and must occur in the order listed.

**Definition 3-10.** Productions in probabilistic context-free phrase structure grammar.

1. $P = \{\alpha \rightarrow \beta | \alpha \in N, \beta \in (\Sigma \cup N)^*\}$.
2. There is a probability function *p*: P→[0,1] such that for each $\alpha \in N, \sum p(\alpha \rightarrow \beta | \alpha) = 1$.

**Example 3-2.** PCFG rules and their probabilities.

| **r∈ P** | **p(r)** |
|---|---|
| S→NP VP | 0.7 |
| S→VP | 0.3 |
| NP→N | 0.6 |
| NP→N PP | 0.4 |
| VP→V | 0.5 |
| VP→V NP | 0.5 |
| PP→P NP | 1.0 |

Lexical information has a fundamental role in determining the properties of a syntactic structure. *Lexicalized grammars* contain a small set of grammar rules and a large lexicon (Miya *et al*. 2003, Joshi & Schabes 1997). Let us consider the effects of lexicalization by looking at the phrase “the green banana” (Charniak 1993). Since “banana” is not a noun that occurs very often, a probabilistic grammar without lexical sensitivity would add a low probability to the word “banana” by using the rule *n→banana*. But the rule itself has no knowledge of the actual context in which the substitution of an *n* to the word “banana” is taking place. Because of this, more frequently occurring nouns, such as “year” or “day”, would be assigned higher probabilities despite the fact that, in the given context, “banana” is clearly the more likely choice.

Each rule in a lexicalized grammar is associated with an *anchor* word. For example, a lexicalized CFPSG is a 4-tuple G = (N, Σ, P, S) just like a CFPSG in Definition 3-9, except that each production in *P* is associated with an anchor *a*:

**Definition 3-11.** Production rules in a lexicalized context-free phrase structure grammar.

$P = \{\alpha \rightarrow \beta | \alpha \in N, \beta \in (\Sigma \cup N)^* a (\Sigma \cup N)^*\}$, where *a* is a distinguished word in the lexicon, that is referred to as the anchor.

A straightforward way of constructing a lexicalized grammar is to take a CFPSG and make many copies of each rule – one for every possible anchor word in each phrase (Schabes *et al*. 1988, Nederhof & Satta 1994).

**Example 3-3.** A lexicalized context-free phrase structure grammar.

| | |
|---|---|
| NP $\rightarrow$ ball | VP $\rightarrow$ kick |
| NP $\rightarrow$ ball PP | VP $\rightarrow$ kick NP |
| NP $\rightarrow$ defender | VP $\rightarrow$ run |
| NP $\rightarrow$ defender PP | VP $\rightarrow$ run NP |
| PP $\rightarrow$ from NP | |

Most grammar formalisms introduced in what follows are of a lexicalized type. DGs (which are discussed in Section 3.3.4) are an extreme case of lexicalization since they are based purely on lexical dependencies.

A *transition network grammar* does not consist of rules, but rather of a set of FSMs, each of which comprises a collection of states and arcs connecting the states (Bolc 1983). Each network corresponds to a single nonterminal in the grammar. The labels in the arcs indicate the terminal symbol that allows a transition to the next state. A sentence is accepted by a transition network grammar if there is a path connecting the start state with a final state. A *recursive transition network* is a nondeterministic finite-state automaton. *Augmented transition networks* (ATNs), first introduced by Woods (1970), are extensions to the recursive model and have been used for describing grammars and parsing, especially in the AI community (Charniak & McDermott 1985). Figure 3-5 shows an example of a simple ATN grammar.

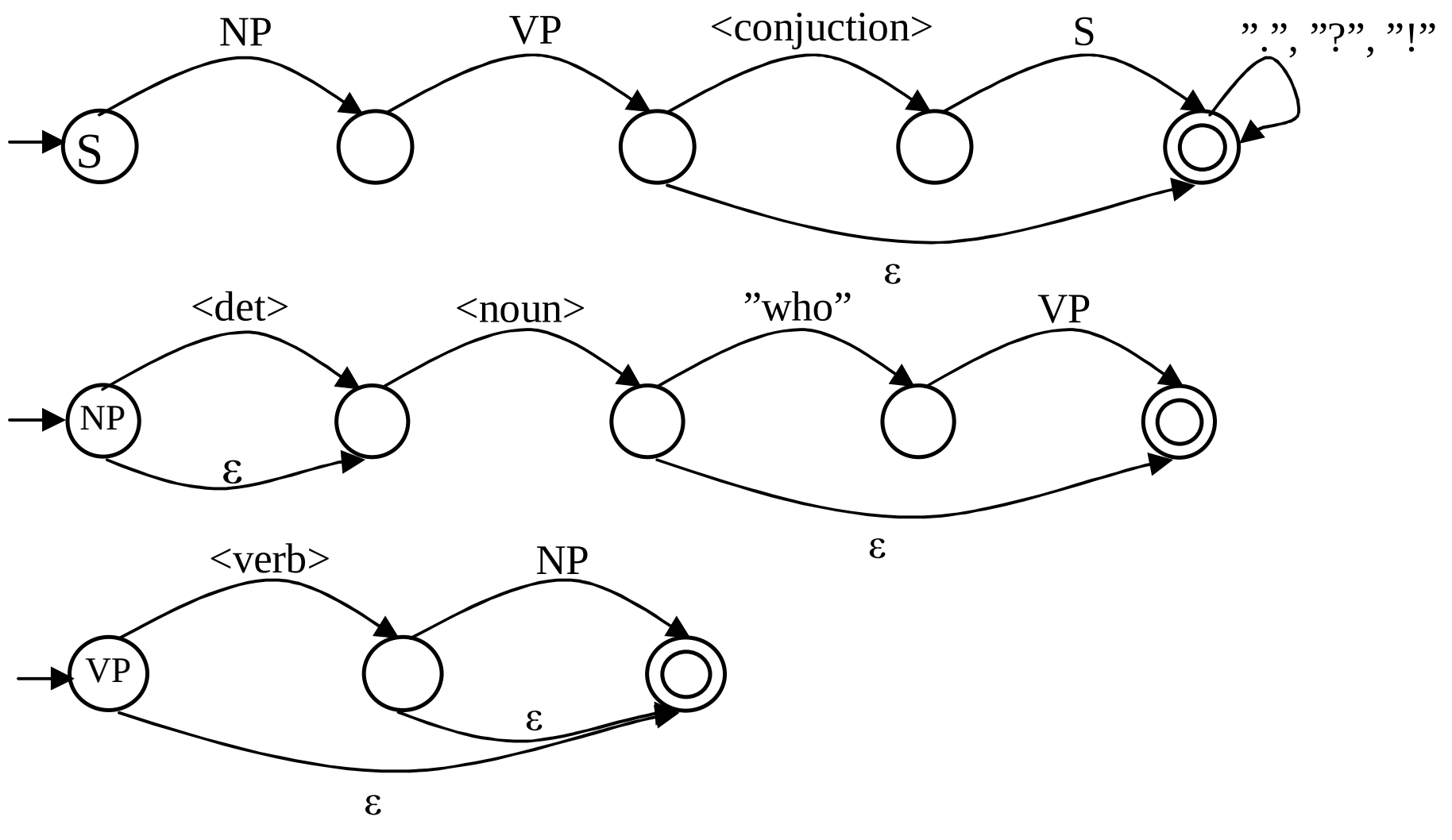

**Figure 3-5**. A simple ATN grammar with rules for Ss, NPs and VPs (Bolc 1983). A VP can, for example, consist of a verb followed by an empty string ε or an NP.

A major advantage of the transition network model over the usual CFPSG model is its ability to merge the common parts of CF rules (Woods 1970). Because this allows for the removal of redundancy, it makes for greater efficiency. ATN grammars are, however, hard to debug, maintain and extend because of their graph-like structure (Carroll 1993). Magerman (1994) points out that the use of ATNs encourages ad hoc system designs and that this leads to application-specific models. It is for these reasons that ATN grammars are not used in modern parsing systems.

### *3.3.2 Unification grammars*

*Unification grammars* (UGs), which are also known as *constraint-based* grammars, were specifically developed to overcome the problems that CFPSGs encounter when representing fine-grained grammatical information such as *agreement* and *subcategorization* (Carpenter 1989). This section introduces three of the best-known UGs, namely, *Generalized Phrase Structure Grammar* (GPSG) (Gazdar *et al*. 1985), *Head-driven Phrase Structure Grammar* (HPSG) (Pollard & Sag 1987, 1994) and *Lexical Functional Grammar* (LFG) (Kaplan & Bresnan 1982). UGs are able to model more complex linguistic phenomena than CFGs. These formalisms describe linguistic objects by using *feature structures* consisting of features and associated values that encode several levels of linguistic information (e.g. morphological, syntactic, semantic) in a uniform way.

**Example 3-4**. A feature structure represented as an *attribute-value matrix* (AVM) for the word "his".

$$\begin{bmatrix} major & NP \\ number & SING \\ person & 3RD \\ case & OBJ \end{bmatrix}$$

As the name suggests, the process of *unification*, matching and combining feature structures, plays a central role in UGs.[21] In spite of this, unification can only be regarded as a solution, as an operation used in algorithms to resolve the *constraints* set by the grammar (Pollard 1996). Unification is consequently not in itself a grammatical theory; instead, it is a mechanism for instantiating diverse grammatical theories (Joshi 1987). Figure 3-6 shows how unifications are used in deriving a sentence structure.

[21] In unification, two pieces of partial information are put together to form a larger piece of information. For example, let *A*={<major, NP>, <number, SING>}, *B*={<major, NP>, <person, 3RD>} and *C*={<major, NP>, <number, PLU>}. Then the unification of *A* and *B* (denoted with $A \wedge B$) is {<major, NP>, <number, SING>,<person, 3RD>} and the unifications $A \wedge C$ and $B \wedge C$ are undefined because of the clash with regard to the value for the feature *number*. Another important operation used in UGs is the *disjunction*, denoted with $\vee$. For example, $A \vee B$ means that some linguistic object *x* is described by either *A* or *B*, but whether it is either *A* or *B*, remains indiscernible (Pollard & Sag 1987).

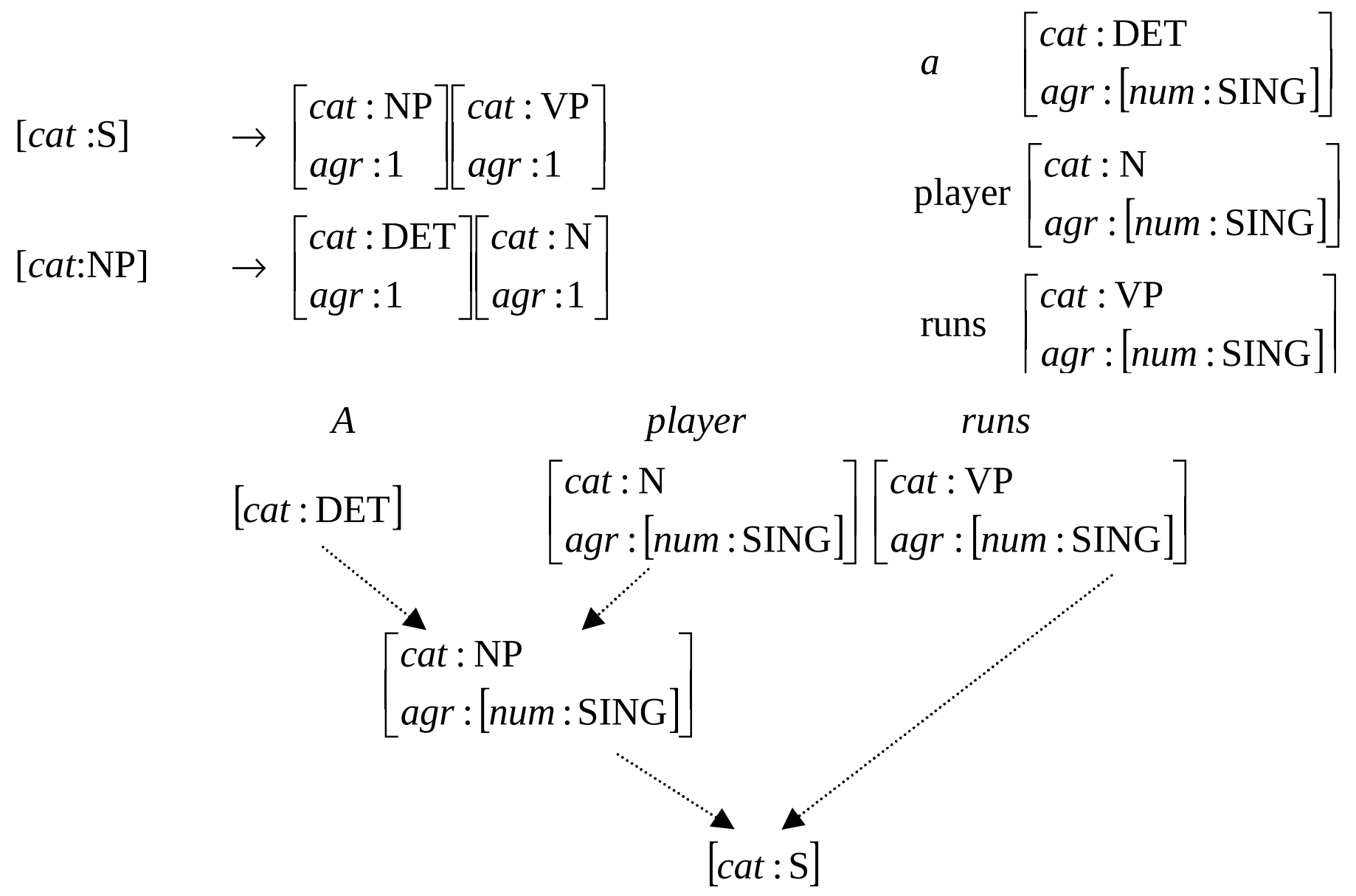

**Figure 3-6.** Derivation in unification grammars. The rules are given on the left-hand side and the lexicon on the right. In the derivation at the bottom, the arrows indicate the way in which the words and phrases unify into larger entities.

### 3.3.2.1 GPSG and HPSG

Before the advent of GPSG during the first half of the 1980s (Gazdar *et al.* 1985), theories of syntax and semantics were developed separately with little interaction. The work that was done on GPSG was undertaken with the intention of producing an integrated theory of the two levels. That represented one of the first attempts to compensate for the inadequacies of CFPSGs without applying transformations (Warner 1996). Instead of transformations, GPSG uses PS rules which can themselves be generated by *metarules.*[22]

Because the development of HPSG has been so comprehensively influenced by GPSG, it may rightly be regarded as its successor. In HPSG there are no derivations that transform one grammatical structure into another; instead, parallel representations are mutually constrained by the grammar. HPSGs have been applied to a wide range of languages such as English (Copestake & Flickinger 2000), Japanese (Siegel 2000), and German (Müller & Kasper 2000).

[22] Metarules take as an input grammar rules that match the input pattern, and they then output a new rule according to their output patterns. Intuitively, metarules have the same role in GPSG as transformations have in TGs.

An HPSG consists of grammar *principles*, grammar rules and lexical entries, and all these are represented as feature structures (Pollard & Sag 1987, Cooper 1996). Collections of information about phonological, syntactic and semantic constraints are called *signs*, and signs represent either a word (a lexical sign) or a phrase (a phrasal sign).[23] An HPSG can be defined as follows:

> **Definition 3-12.** Head-driven Phrase Structure Grammar (Pollard & Sag 1987). A *Head-driven Phrase Structure Grammar* is defined as
>
> $$P_1 \wedge \ldots \wedge P_{n+m} \wedge \left(L_1 \vee \ldots \vee L_p \vee R_1 \vee \ldots \vee R_q\right)$$
>
> where $P_1 \ldots P_n$ are *universal principles* common to all languages, $P_{n+1} \ldots P_{n+m}$ are language-specific principles, $L_1 \ldots L_p$ are the lexical signs of a language, and $R_1 \ldots R_q$ are its grammar rules.

Definition 3-12 implies that a linguistic object belongs to the language generated by the grammar if it satisfies all the universal and language-specific principles of that language, and that this either instantiates one of the lexical signs of the language or one of its grammar rules.

The lexicon in HPSG is rich and organized on the basis of the notion of types of word (Kim 2000, Cooper 1996).[24] The rules are expressed as *well-formedness* constraints on the feature structure descriptions and are represented as partially specified phrasal signs. Along with the rules, there are principles that define and therefore limit the signs that may be construed as belonging to a language. The well-formedness of a sign is verified by comparing the feature structure of a specific rule or principle to the feature structure expressing the sign. Checking is performed through unification.

#### 3.3.2.2 LFG

LFG combines various features from TGs and ATN processing, and has been applied to a variety of languages including English, French, German, Japanese, Chinese, Norwegian, Spanish, and Urdu (Butt *et al*. 2002, Oepen *et al*. 2004,

---

[23] The information about a sign is stored in a feature structure, and is encoded as an AVM (Pollard & Sag 1987). Phrasal signs join lexical signs into sentences.

[24] Multiple inheritance hierarchies are used to reduce redundancy and to allow generalizations across classes of words.

O'Donovan *et al*. 2005, Burke *et al*. 2004b, Dipper 2003). Figure 3-7 gives an example of a LFG parse.

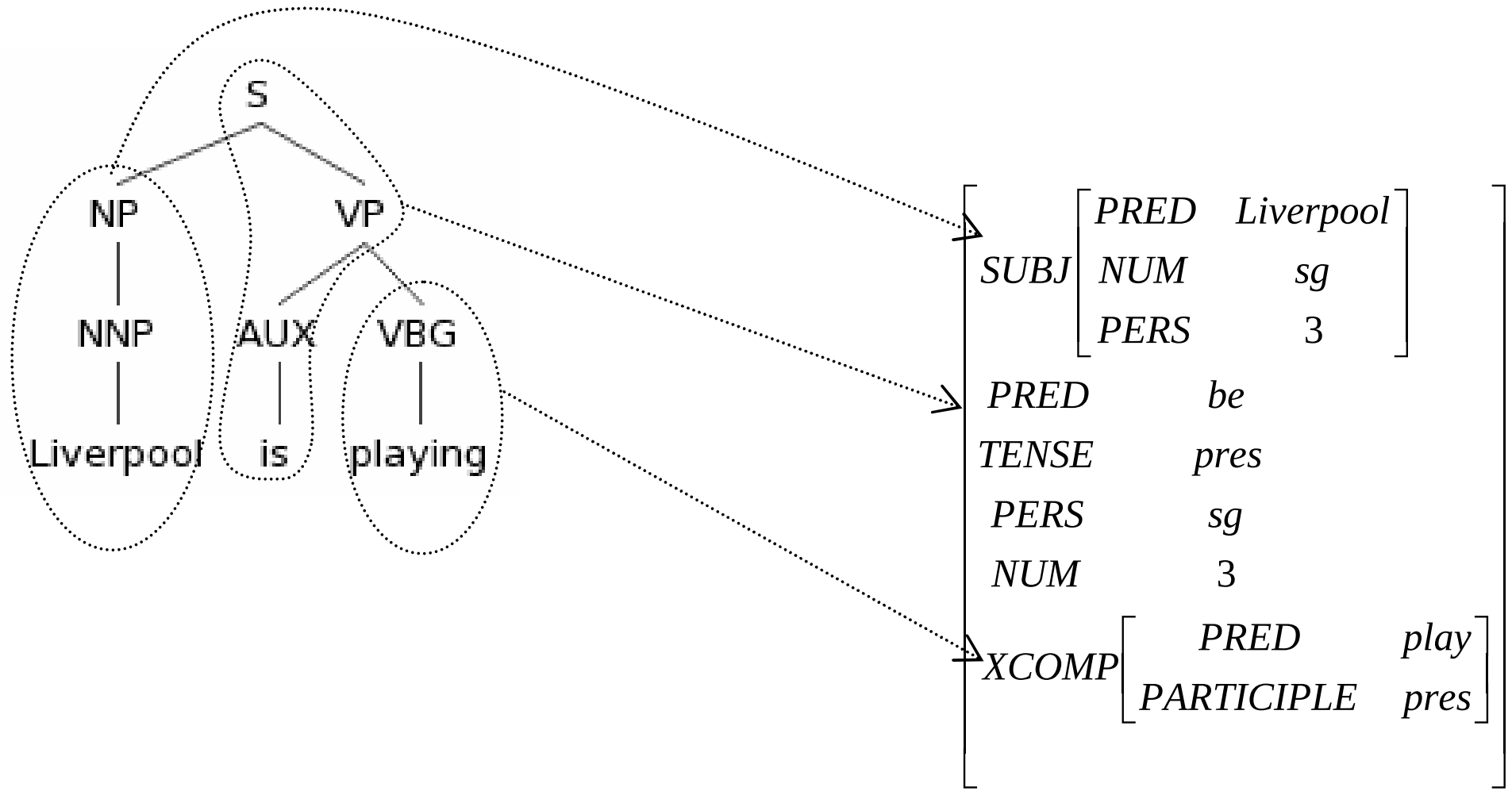

**Figure 3-7.** This example shows the correspondence of elements between a C and F-structure for the sentence "Liverpool is playing".

Figure 3-7 illustrates how LFG assigns two types of structure to input sentences. These are the *constituent structure* (C-structure), which is derived by CFPSG rules and it is represented as a PS tree, and the *functional* F-structure, which is produced by ATN operations and consists of *grammatical relations* (GRs) represented as an AVM (Bresnan 2001).[25] An distinguishing feature in LFG is *structural correspondence*. While one describes sentences with different descriptive languages and representations, it is essential to be able to correlate such structures.

### *3.3.3 Tree-adjoining grammars*

*Tree-adjoining Grammar* (TAG), which was first introduced by Joshi *et al*. (1975), manipulates trees as elementary objects. Different versions of TAGs have been implemented in English (XTAG Research Group 1998), Chinese (Bikel & Chiang 2000), French (Abeillé 1988), German (Neumann 2003), Arabic (Habash & Rambow 2004), Korean (Han *et al*. 2000) and Hindi (CDAG 2006).

[25] The two representations manifest the fact that different types of dependencies exist among the parts of a sentence, and that these dependencies can be best expressed using different formal structures (Kaplan 1989).

**Definition 3-13**. Tree-adjoining Grammar.

A *Tree-adjoining Grammar* is a 5-tuple TAG = (N, ∑, I, A, S) where

1. *N* is a finite set of nonterminal symbols; $N \cap \Sigma = \varnothing$;
2. ∑ is a finite set of terminal symbols;
3. *I* is a finite set of finite *initial trees*;
4. *A* is a finite set of finite *auxiliary trees,* and
5. $S \in N$ is a distinguished symbol called the *sentence* symbol.

The set of initial trees *I* and the set of auxiliary trees *A* are referred to as the *elementary trees* (Joshi & Schabes 1997, Han *et al.* 2000). The initial trees are minimal linguistic structures that contain the structure of simple phrases (e.g. NPs) and that do not have recursive structures. Auxiliary trees may contain recursion and represent constituents that are adjuncts to basic structures (such as adverbials). Sentences are derived by building *derived trees* from initial and auxiliary trees by the composition operations *adjoining* and *substitution*. Figure 3-8 illustrates an adjoining operation.

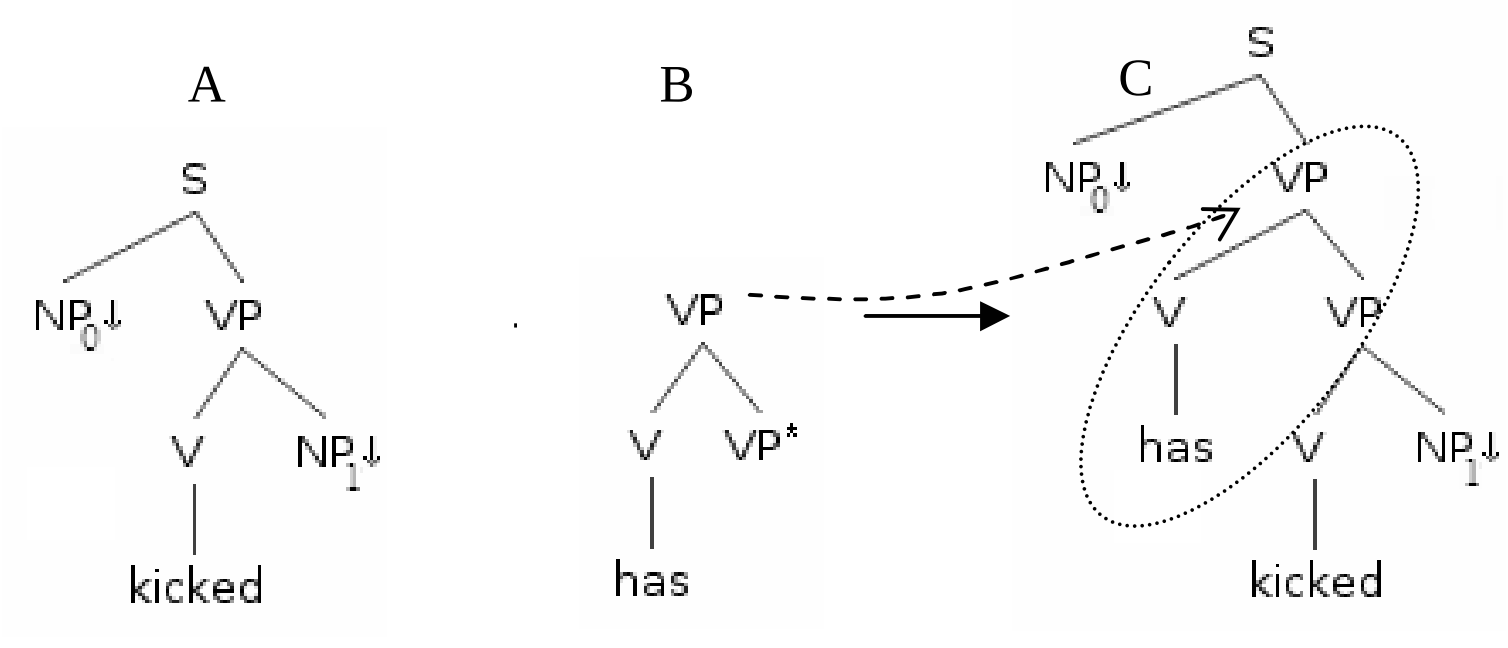

**Figure 3-8.** Adjoining operation. Trees *A* and *B* are adjoined to form tree *C*. The dotted arrow indicates the location to which tree B is adjoined.

The basic model has been extended in many ways by, for example, lexicalization (*Lexicalized TAG* (LTAG) (Schabes *et al.* 1988)), unification-based processing (*Feature-based LTAG* (Vijay-Shanker & Joshi 1991)), and probabilities (*Probabilistic LTAG* (PLTAG) (Resnik 1992, Schabes 1992)).

### *3.3.4 Dependency grammars*

*Dependency grammars* (DGs) identify the relations that connect words to each other. A fundamental notion in DGs is the relation between a *head* and a *dependent*.

**Definition 3-14.** Dependency relation (adapted from Robison 1970).

Let $E$ be a finite set of sentence elements. DR is a binary dependency relation that ranges over the elements $E$; $DR \subseteq E \times E$. Let $e_1, e_2, x \in$ E. Let $<e_1, e_2> \in$ DR denote that $e_1$ is dependent on $e_2$. $e_2$ is referred to as the *head*. If DR holds for $<e_1, x>\ldots<e_k, x>$, all $e_i \in E$, $i \in \{1...k\}$ are dependent on $x$.

A head may have any number of dependents, which may be either *modifiers* or *complements*. The grammar rules specify the dependents that each head can take (e.g. adjectives depend on nouns and not on verbs). I refer to the DG formalisms, as defined by Hays (1964) and Gaifman (1965), as *classical DG*s (CDGs). The properties of CDGs can be formally defined as follows:

**Definition 3-15.** Classical dependency grammar (adapted from Hayes (1964) and Gaifman (1965)).

A *dependency grammar* is a 4-tuple DG = (C, ∑, R, F) where

1. $C$ is a finite set of *lexical categories*.
2. $\Sigma$ is a finite set of *terminal symbols*.
3. $F$ is an assignment function $F : \Sigma \rightarrow C$.
4. $R$ is a finite set of *rules* over the lexical categories $C$ that define dependency relations *DR*. The rules are of the form:
   1. $x(c_1,...,*,...,c_k)$ denotes that $c_1 \ldots c_k \in C$ are dependent on $x \in C$. I.e. * indicates the position of head $x \in C$ in the sequence $c_1 \ldots c_k \in C$,
   2. $x$(*) denotes that $x \in C$ is a leaf node, and
   3. *($x$) denotes that $x \in C$ is the sentence root node.

In CDGs the D relations form trees that are acyclic and have a single root (Robinson 1970). Moreover, only one head per dependent is allowed.

**Example 3-5.** A simple dependency grammar.

DG = (C, ∑, R, F)

| | |
|---|---|
| C={V, N, DET} | F("kicks")=V |
| ∑={"kicks", "ball", "attacker", "the"} | F("ball")=N |
| R={*(V), V(N, *, N), N(DET, *), N(*), DET(*)} | F("attacker")=N |
| | F("the")=DET |

Figure 3-9 illustrates an analysis using the grammar shown in Example 3-5.

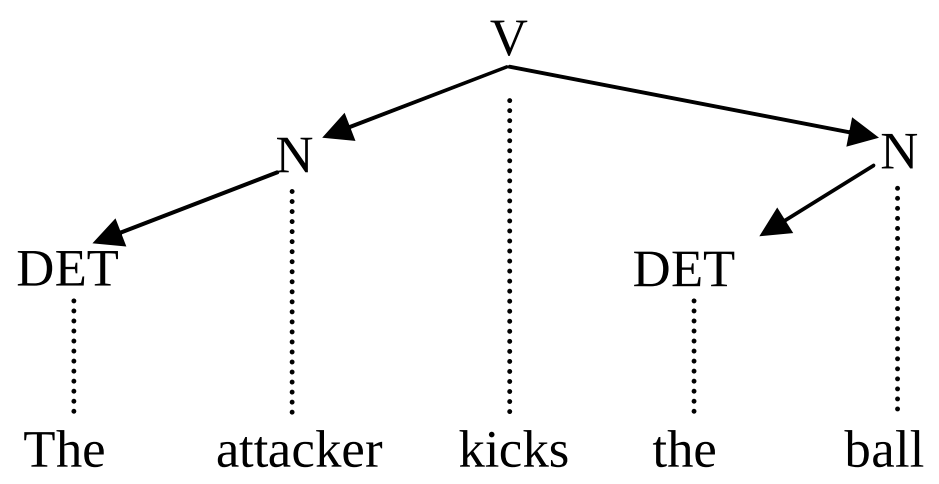

**Figure 3-9.** DG analysis for the sentence “The attacker kicks the ball”.

Several modifications and extensions to CDGs have been introduced. These include, for example, *Constraint Grammar* (CG), *Functional Dependency Grammar* (FDG) and *Extensible Dependency Grammar* (XDG). Standard PS trees and the analyses produced by CDGs are *projective*; this means that branches do not cross when the terminal nodes are projected onto the input string (Neuhaus & Bröker 1997). It has, however, been acknowledged among DG syntacticians that certain natural language phenomena require *non-projective* analyses (Rambow 1994). Such structures are frequently encountered in languages that allow a relatively free word order. In languages with strict word order, moreover, non-projective trees occur, for example, in *right-node raising* structures.

### 3.3.4.1 CG and FDG

CG is a DG formalism that was originally proposed by Karlsson (1990). Implementations of CG exist in, for example, English, Danish, French, German, Swedish, Finnish, Estonian, and Portuguese (Afonso *et al*. 2002, Bick 2003, Bick *et al*. 2005, Lingsoft 2006). Instead of having rules that define correct sentence structure, CG applies *constraints* for eliminating word readings that are inconsistent with the context. These manually formulated constraints act as partial LP rules. They may contain both rule-like facts and probabilistic tendencies.

CG was designed not as grammatical theory but as a language-independent parsing framework. It is a descriptive, computationally-oriented model. CG makes no claims about being able to explain linguistic phenomena. The most important aim of CG is to provide reliable and shallow rather than highly informative but faulty parsing.

Many of the ideas in FDG are derived from CG (Tapanainen & Järvinen 1997, Tapanainen 1999). FDGs have been implemented in English, Finnish, French, Spanish, German, and Swedish (Tapanainen 1999, Voutilainen 2001, Connexor 2006). In FDG, D links between syntactic labels form partial trees – usually around a verb to which the other words are linked. Table 3-1 illustrates an example of a FDG parse.

**Table 3-1**. An FDG analysis for the sentence "Liverpool is playing." The analysis gives the location of the word, the original word form, base form and syntactic and morphological tags.

| Index | Word | Lemma | Syntactic* | Morphological |
|---|---|---|---|---|
| 1 | Liverpool | Liverpool | subj:>2 @SUBJ %NH | N NOM SG |
| 2 | Is | be | v-ch:>3 @+FAUXV %AUX | V PRES SG3 |
| 3 | playing | play | main:>0 @+FMAINV %VA | ING |
| 4 | . | . | | |

*The dependencies are represented as the pair D label-head, separated by ":>". For example, according to the analysis, the word "Liverpool" is connected to the word "is" with a D link of the type *subject* (subj). KEY: v-ch = verb chain (auxiliaries + main verb), main = main verb. The functional tags are denoted by @. KEY: +FAUXV = finite auxiliary predicator. +FMAINV = finite main predicator. Surface syntactic tags are denoted by %. KEY: NH = nominal head, AUX = auxiliary verb or particle, VA = main verb in an active verb chain.

#### 3.3.4.2 XDG

XDG is a new DG formalism that defines two orthogonal but mutually constraining structures that explain the ID and LP constraints of the words respectively (Debusmann *et al.* 2004b). The analyses in XDG are *multigraphs* that consist of an arbitrary number of D graphs called *dimensions*. Each dimension represents a different aspect of language (syntactic function, PAS, etc.) (Debusmann & Smolka 2006). XDGs have been implemented in English and German, and have been applied to a smaller extent to Arabic, Czech and Dutch (Debusmann *et al.* (2004a, 2004b), Bojar 2005). An XDG consists of *dimensions*, principles, and a lexicon. XDG may be formally defined as follows:

**Definition 3-16.** Extensible Dependency Grammar (adapted from Debusmann *et al.* 2004b)
An *Extensible Dependency Grammar* is a 3-tuple XDG = (D, Pri, Lex), where

1. *D* is a finite set of dimensions $d_i$, $i$=1,…,$n$ of the form $d_i$=(Lab, Fea, Val, $\text{Pri}_d$), where
    a. *Lab* is a set of edge labels,
    b. *Fea* is a set of features,
    c. *Val* is a set of feature values, and
    d. $Pri_d$ is a set of one-dimensional principles.
2. *Pri* is a set of multi-dimensional principles.
3. *Lex* is the lexicon consisting of a set of lexical entries.

The well-formedness conditions of an analysis are checked by the interaction of the principles and the lexicon. As a DG formalism, XDG is lexicalized, and most of the information is specified in the lexical entries. XDG moreover provides a lexical abstraction mechanism for the purpose of reducing redundancy in the lexicon. Principles specify how words interact and how graphs on different dimensions relate to one another (Debusmann *et al*. 2004b). The lexicalized *valency principle*, for example, states that all nodes on the dimension *i* must satisfy their specifications for incoming and outgoing edges. Figure 3-10 illustrates an XDG parse.

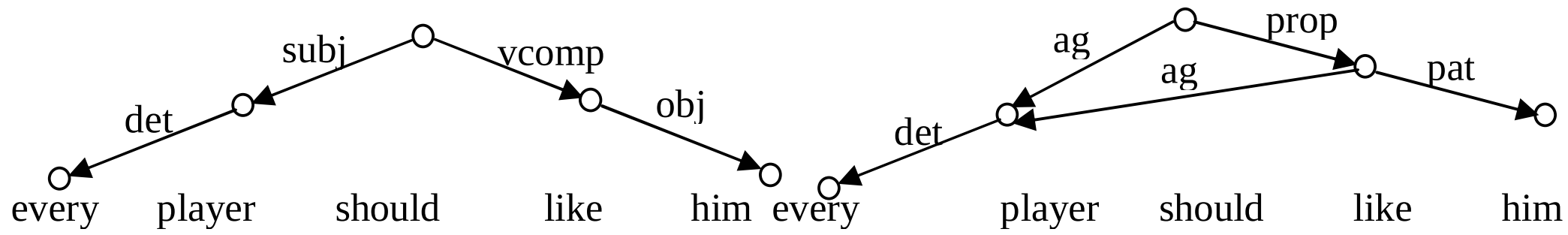

**Figure 3-10**. An example XDG analysis for the sentence “Every player should like him” (adapted form Debusmann *et al*. 2004a). The analysis has two dimensions: the graph on the left-hand side represents the syntactic structure and the graph on the right-hand side the semantic structure of the sentence.

### *3.3.5 Link Grammar*

*Link Grammar* (LG) is closely related to DG. So far, it has been applied only to English and German (Sleator & Temperley 1993, Kübler 1998). The lexicon in LG consists of entries that state the *linking requirements* of one or more words, and these are expressed in terms of *connectors*. A parse in LG is referred to as *linkage*, and it consists of a set of undirected links that connect the words in a sentence. A sentence may have several linkages. Figure 3-11 gives an example LG lexicon and linkage.

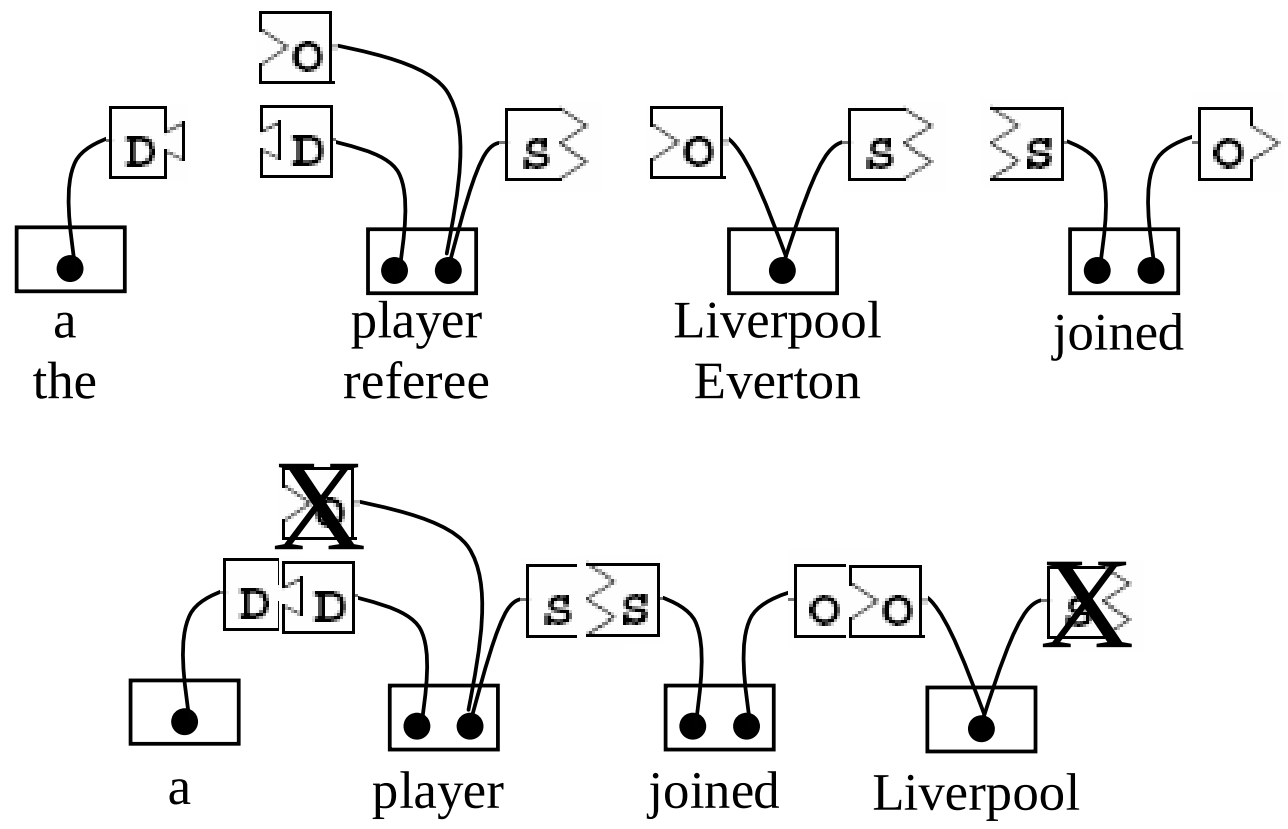

**Figure 3-11.** A Link Grammar lexicon (top) and linkage (bottom) (adapted from (Sleator & Temperley 1993)). The words "player" and "referee", for example, require an *o* or *d* connector to their left, or an *s* connector to their right.

Instead of requiring two connectors to be identical before they can be connected, LG allows one to enunciate *matching rules* (Sleator & Temperley 1993). In addition, one can use a *multi-connector* to connect to more than one link. This mechanism allows any number of adjectives, for example, to attach to a single noun.

### *3.3.6 Combinatory Categorial Grammar*

In *Categorial Grammars* (CatG), which originate from the work of Bar-Hillel (1953), the lexicon carries the main responsibility for defining syntactic knowledge (Steedman 2000). CatGs define a finite set of primitive categories (such as N, NP, VP, S) that combine by means of function application rules to create more complex structures. Unlike other grammar formalisms, CatGs do not define a set of rules for combining words. It is rather the definitions in the lexicon that determine how words can combine with each other.

Because CatC in its pure form is not adequate for describing natural languages, several extensions have been added to the basic framework. *Combinatory Categorial Grammar* (CCG), an extension well suited to parsing, originates in the work of Steedman (1985a). CCG is in essence a theory of syntactic and semantic types (Steedman 1999). While wide-coverage parsing CCGs exist only for English (see for example, Hockenmaier 2003, Villavicencio 1997), CCG has also been successfully applied to Dutch, Japanese, Korean and Turkish to a lesser extent (Steedman 1985b, Komagata 1999, Cha *et al*. 2002, Bozsahin 2002, Çakici 2005).

A CCG consists of the lexicon and the combinatory rules (Steedman 2000, Clark 2002, Clark *et al*. 2002a, Hockenmaier & Steedman 2007). The lexicon, which pairs words with lexical categories, defines most language-specific aspects of a CCG. Each item in the lexicon is associated with one or more *categories* that define their syntactic behavior by describing the constituents with which it can combine and the result of the possible combination. CCG categories are defined in Definition 3-17.

**Definition 3-17.** Categories in Combinatory Categorial Grammar (adapted from Vijay-Shanker & Weir 1994).

Let *N* be the set of *nonterminals* (also referred to as *atomic categories* in the context of CCG). The set of categories *C*(*N*) over the alphabet *N* is the smallest set for which the following conditions hold:

1. $N \subseteq C(N)$
2. If $c_1, c_2 \in C(N)$ then $(c_1/c_2)$, $(c_1\backslash c_2) \in C(N)$.

*S*, *N*, *NP*, and *PP* are examples of *atomic categories* composed of a single nonterminal. The *complex categories* are of the form *X/Y* or *Y\X*, where *X* is the result category and *Y* is the argument category. For example, *X/Y* states that the category needs to combine with a category *Y* on the right to form *X*. A CCG is defined in Definition 3-18 below.

**Definition 3-18.** Combinatory Categorial Grammar (adapted from Vijay-Shanker & Weir 1994).

A *Combinatory Categorial Grammar* is a 5-tuple G = (N, ∑, f, R, S) where

1 *N* is a finite set of *nonterminal symbols*

2 ∑ is a finite set of *terminal symbols*.

3 *f*: $\Sigma \rightarrow C(N)$ is a function that maps each element of ∑ to finite subsets of *C*(*N*).

4 *R* is a finite set of *combinatory rules*.

5 $S \in N$ is a distinguished symbol called the *sentence* symbol.

The lexicon consists of ∑ and *f*. The most basic combinatory rules are the *forward* and *backward application* (see Definition 3-19 below). A derivation in a CCG involves the use of the backward and forward application and other combinatory rules such as *functional application* and *composition*.[26]

[26] Other combinatory rules, such as *forward and backward composition* (>**B** and <**B**) *forward type-raising* (>**T**) and *backward crossed substitution* (<**S**) exist for dealing with coordination and extraction.

**Definition 3-19.** Forward and backward application in Combinatory Categorial Grammar (Hockenmaier & Steedman 2007).

Let *f* and *a* be terminals and X/Y and X\Y complex categories.

1. A *forward application* is of the form: X/Y:*f* Y:*a* → X:*fa*.
2. A *backward application* is of the form: Y:*a* X\Y:*f* → X:*fa*.

**Example 3-6.** CCG forward and backward applications. A category is accordingly connected to its neighbors using forward (>) and backward (<) applications. The categories, for example, of the word "likes" indicate that it can combine with an NP to form a structure of the category S\NP.

**Forward application**

| | | | |
|---|---|---|---|
| (S\NP)/NP | NP | → | S\NP |
| *likes* | *football* | → | *likes football* |

**Backward application**

| | | | |
|---|---|---|---|
| NP | S\NP | → | S |
| *He* | *likes football* | → | *He likes football* |

## 3.4 Analysis

Very little research or rational analysis has been focused on the relative weaknesses and advantages of grammar formalisms – let alone on the question about which theoretical approaches may be best suited to computational applications. As Carnie (2002) points out: "If you ask this question at any major syntax conference you are likely to get lynched." Most linguists have a strong prejudicial bias in favor of the particular framework that they themselves utilize while, at the same time, are actively borrowing ideas and techniques from one another.

While I shall now analyze grammar formalisms for parsing, I shall make no attempt to rank the theories concerned. Any such classification would be difficult, if not impossible, to make, for two reasons. Firstly, while some theories provide a good account of some syntactic phenomena, other theories do equally well for other syntactic phenomena. What criteria would one use to decide that some linguistic phenomena might be more important than others? Secondly, some theoretical approaches are more highly developed than others. Besides, the only purely objective grounds for evaluating a grammatical theory would be on the basis of how well it models the workings of language in the human mind and brain. But there is no known method by which one could reliably test such modeling capabilities.

Instead of attempting to rank them, I will compare the formalisms by describing the similarities and differences between them, the ways in which each formalism is able to model natural language phenomena, and how well each formalism is suited to parsing. Firstly, I will identify the most important characteristics of each of the formalisms (Section 3.4.1). Secondly, I will compare the formalisms by using several dimensions: the number of representation strata (Section 3.4.2), generative capacity (Section 3.4.3), and the treatment of *long-distance dependencies* (LDDs) (Section 3.4.4). Section 3.4.5 introduces the grammar development methods that are available for the grammar formalisms under consideration.

### *3.4.1 Outstanding characteristics of the formalisms*

This section underlines the outstanding features of the grammar formalisms under consideration and compares the most closely connected formalisms to each other.

#### 3.4.1.1 CFPSG and TGs

As I have already noted above, CFPSGs are insufficiently powerful for describing natural languages because they are not lexicalized and they do not use probabilistic modeling. It is for this reason that they are not used in state-of-the-art parsing. "Chomskyan" grammars, such as TG and its descendants, such as *Minimalist Program* (Chomsky 1995), have been a predominant influence in linguistics for a long time. Chomsky's theories in fact form the foundations on which many other theories of grammar have been built. However, transformations are not used in computationally-oriented grammar formalisms because of their computational complexity. The main problem with grammars of this type is their tendency to force a parser to destroy existing linguistic structures. Pollard's (1996) principle of *nondestructiveness* states:

> "Grammars should not make reference to operations […] that destroy existing linguistic structure."

#### 3.4.1.2 PCFG

PCFG is the most widely used probabilistic grammar formalism. While PCFG models lack the power of models that are sensitive to a wider range of dependencies, they nevertheless have the advantage of being simple (Resnik 1992, Charniak 1993, Charniak 1996). Even so, the information conveyed in PCFGs is not sufficiently rich fully to describe natural languages. In the first place, PCFGs model probabilities in a CF way.[27] In the second place, the original PCFG model is not lexicalized. This renders it incapable of capturing certain kinds of hierarchical information. Lexicalized PCFGs, like most existing probabilistic grammars, do make use of lexical information.

#### 3.4.1.3 UGs

Pollard (1996) makes the following comments about the "Chomskyan" approach to grammar:

> "There used to be an influential point of view which held that the formalism within which grammatical theory was formulated had to be highly constrained. This in turn was supposed to constrain the set of possible grammars…."

What is evident in Pollard's arguments against this point of view is that instead of the formalism itself setting constraints on the possible grammars, it is rather the grammars themselves that should impose the constraints. Because of their algebraic characterization, UGs offer, among other benefits, an opportunity to provide well-defined semantic representations. UGs have become the most widely used formalisms for computational grammars. According to Oepen *et al*. (2000), HPSG and LFG are the predominant UG formalisms for parsing.

Unlike many other UGs, GPSG emphasizes the PS rule component rather than a rich lexicon (Carnie 2002). One of the distinctive features of HPSG is its head-driven nature: words are structured and rich in information, and certain key words, lexical heads, have an important role to play in the processing of the structures that contain them (Raaijmakers 1993). The syntactic categories in HPSG are similar to those in GPSG in the sense that they are complex feature structures. Despite their close relations, there are several differences between HPSG and GPSG. Firstly, the categories in HPSG are more complex (Borsley 1996). Secondly, HPSG makes more specific claims about language universals than does GPSG (Carnie 2002). Because of its precise mathematical modeling, HPSG has been found to be highly useful in computational applications.

Like HPSG, LFG differs from GPSG by making more specific claims about language universals and variation. Both LFG and HPSG incorporate the principle of *strong lexicalism*, which implies that most of the information employed in constructing sentence representations is associated with individual words; relatively simple rules and principles govern how the information in lexical entries combines when the words are united into phrases (Sag & Wasow, forthcoming article). In contrast to what happens in LFG and CFPSG, where the rules determine the relative order of elements, the PS rules in GPSG and HPSG set only the ID relations. The order is defined by language-specific LP constraints.

---

[27] A PCFG, for example, will consider the probability of expanding an NP to be independent of where in the parse tree the NP is located. This assumption is false.

#### 3.4.1.4 TAGs

In contrast to what happens in most grammar formalisms, a derived tree in TAG does not include any information about how it was constructed (Joshi & Schabes 1997). Hence, an additional representation, called the *derivation tree*, is used for representing the information on the trees and the operations used in a derivation.

TAG enables one to state dependencies between nodes of trees (Joshi & Schabes 1997). This is in contrast to most other grammar formalisms in which the dependencies are defined between the elements of a rule. This allows for an extended *domain of locality* in which to state linguistic dependencies.

An interesting feature of TAG is the way in which it is able to capture the recursiveness of language structures (Kroch & Joshi 1987). In TAGs, local co-occurrence restrictions are stated in the elementary trees. The more complex structures are composed by means of adjunction and substitution operations that are constrained by local constraints, which, among other things, set subcategorization constraints for the trees to be joined. The idea is that the expression of local co-occurrence relations should be factored apart from the expressions of recursion and unbounded syntactic dependencies.

#### 3.4.1.5 DGs

In addition to being suitable for the analysis of languages with relatively free word orders, DGs are attractive because they offer one the possibility of mapping D trees onto semantic representations (i.e. PAS) (Lombardo & Lesmo 2000).

The syntactic representation of CG is based on an underspecified and relatively shallow[28] D description (Tapanainen & Järvinen 1997). In addition to this, CG always attaches an analysis to a sentence, even if it is ill-formed. A CG cannot therefore be used for accepting or rejecting sentences. But it is this very simplistic approach that permits CG to be an efficient parsing framework.

---

[28] Due to the shallowness of representation, for example the objects of infinitives and participle are indicated with the same tag.

Let us consider NPs as an example of *underspecification* in CG. While the heads of NPs are marked with functional labels in CG, the parent (NP) is not indicated:

**Example 3-7.** The fragment “fat player’s wife” would get the following analysis:

```
”<fat>”
        ”fat” A ABS @AN>
”<player’s>”
        ”player” N GEN SG @GN>
”<wife>”
```

The function tag *@AN>* in the word “fat” indicates that the head of the word is a nominal in the right-hand context. The head, however, is not specified. Thus there is no need to tackle the syntactically undecidable ambiguity between the analyses that the one who is fat is either the player or the player’s wife.

The main differences between the two otherwise similar formalisms may be summarized as follows (Tapanainen & Järvinen 1997, Järvinen & Tapanainen 1998): Firstly, the FDG analyses contain more detailed information than CG structures. The additional power of the framework stems from the mechanism that allows it to specify links between word readings. Secondly, whereas the CG rules typically represent the head and dependent implicitly and ambiguously, FDG makes the relation explicit by declaring the head and the dependents. Thirdly, the FDG framework has a mechanism for handling coordination. Fourthly, an advantage of FDG over CG is the way in which it deals with ambiguity.[29]

The most interesting feature in XDG is that it combines multi-layered linguistic description with a DG formalism (Debusmann *et al*. 2004b). While XDG has many desirable characteristics, it is a relatively new grammar formalism and needs to be developed further. Debusmann and Smolka (2006) point out two main weaknesses in the framework: Firstly, there is as yet no practical parsing algorithm for XDG. Secondly, work still needs to be done in the precise formalization of the framework, particularly on the syntax-semantics interface.

[29] In the experiments by Tapanainen and Järvinen (1997), FDG left only 3.2 to 3.3 per cent of words with more than one morphosyntactic label, while the number for the CG parser was 11.3 to 13.7 per cent.

#### 3.4.1.6 LG

While LGs are in many ways similar to DGs, there are several differences that can be identified. The main differences are as follows: (1) Links in LG are undirected. (2) Links may form cycles. (3) It contains no notion of the root word. In addition to this, LG analyses are projective and CF.

There are some inherent problems in the LG framework. Certain constructions, such as non-referential use of "it" and "there", and certain types of questions and comparatives are problematic for LG (Sleator & Temperley 1993). Sleator and Temperley (1993) reach the following conclusion: "Certain constructs in grammatical English are simply unnatural in our framework." While Lafferty *et al*. (1992) have described a probabilistic version of LG, Collins (1999) claims that the probabilistic model is flawed.

#### 3.4.1.7 CCG

The main advantage of CCG over other formalisms is its treatment of *coordination* and *extraction* (Lombardo & Lesmo 1998b). Following Montague's (1974) *principle of compositionality*, every syntactic derivation in a CatG corresponds to a semantic interpretation, and the two representations are constructed together. Because it is a CatG, CCG follows this principle. The consequence of this is that CCG has a transparent syntax-semantics interface. A problem in CCG is that there may be several derivations, all of which lead into the same derived structure (Clark & Curran 2004a). This property is referred to as *spurious ambiguity*.

### *3.4.2 The number of strata*

A distinction can be made between multi- and monostratal formalisms. In contrast to TG and other multistratal grammar formalisms, a *monostratal* formalism uses a single level of representation for the syntactic structure, and recognizes no more abstract level (Trask 1993). Most of the formalisms described in this work, GPSG, HPSG, TAGs, CG and FDG, for example, are clearly monostratal; they use one type of structure to describe syntax.

One of the features of the monostratal approach is that it makes processing efficient because single linguistic representations are not manipulated on several steps (strata). But it is true that monostratal approaches, such as HPSG, often give precedence to the syntactic structure, and this relegates the other levels of description to a secondary role (Debusmann *et al*. 2004b). This makes it potentially more difficult to modularize the grammar: when the syntactic part of the grammar is changed, the semantics component has to be changed as well.

While C- and F-structures in LFG are two distinct representations, LFG is monostratal in the sense that there are no multiple representations of a single structure (Cahill 2004); LFG has two syntactic representation levels, C- and F-structure, each of which is monostratal; an analysis of a single clause cannot consist of multiple strata of C-structure or F-structure. A major advantage of the C/F-structure distinction in LFG is that it enables one to take into consideration the fact that while languages may differ with respect to surface representations (such as word order), they may encode the same or similar PAS (Cahill 2004).[30]

XDG may be positioned between the monostratal (e.g. HPSG) and multistratal (e.g. TG) extremes. While linguistic analyses are divided into multistratal representations, the principles interact between strata to constrain all dimensions simultaneously. Debusmann *et al*. (2004b) claim that, in comparison to LFG, XDG places a lighter burden on the interfaces between the strata.[31] Because of the freer interaction between the dimensions, filtering can be done earlier in XDG. Furthermore, the constraints in XDG are not restricted to operating on adjacent strata, but are allowed to access all dimensions and all directions.

### *3.4.3 Formal power and equivalencies*

The ability of a grammar to generate languages, as described in Definition 3-5, is called its *weak generative capacity* (WGC) (Miller 1999, Joshi 2003b). In contrast to this, the *strong generative capacity* (SGC) defines the set of structural descriptions generated by a grammar. WGC is often used for locating a grammar in a hierarchy of formal grammars. Chomsky's (1959) theory of grammars offers a tool for studying differences in their formal powers. The hierarchy consists of four levels, numbered from 0 to 3 and called *regular*, *context-free* (CF), *context-sensitive* and *unrestricted*. The level of restriction grows gradually from level 0 to level 3.[32] Figure 3-12 depicts the hierarchy.

---

[30] While the same proposition expressed in different languages may have substantially dissimilar C-structures, it may be associated with similar F-structure representations. This property is useful in, for example, MT systems.
[31] The mapping between C- and F-structures is specific and has to be adapted to new structures in order to handle different word order, for example. Similar modifications in XDG could ideally be achieved simply by modifying the grammar (Debusmann *et al*. 2004b).
[32] Each level is moreover a subset of the less strict levels. The restrictions are cumulative – the rules of a type 3 grammar also obey the restrictions for types 0, 1 and 2. A regular grammar, for example, is a special kind of CFG.

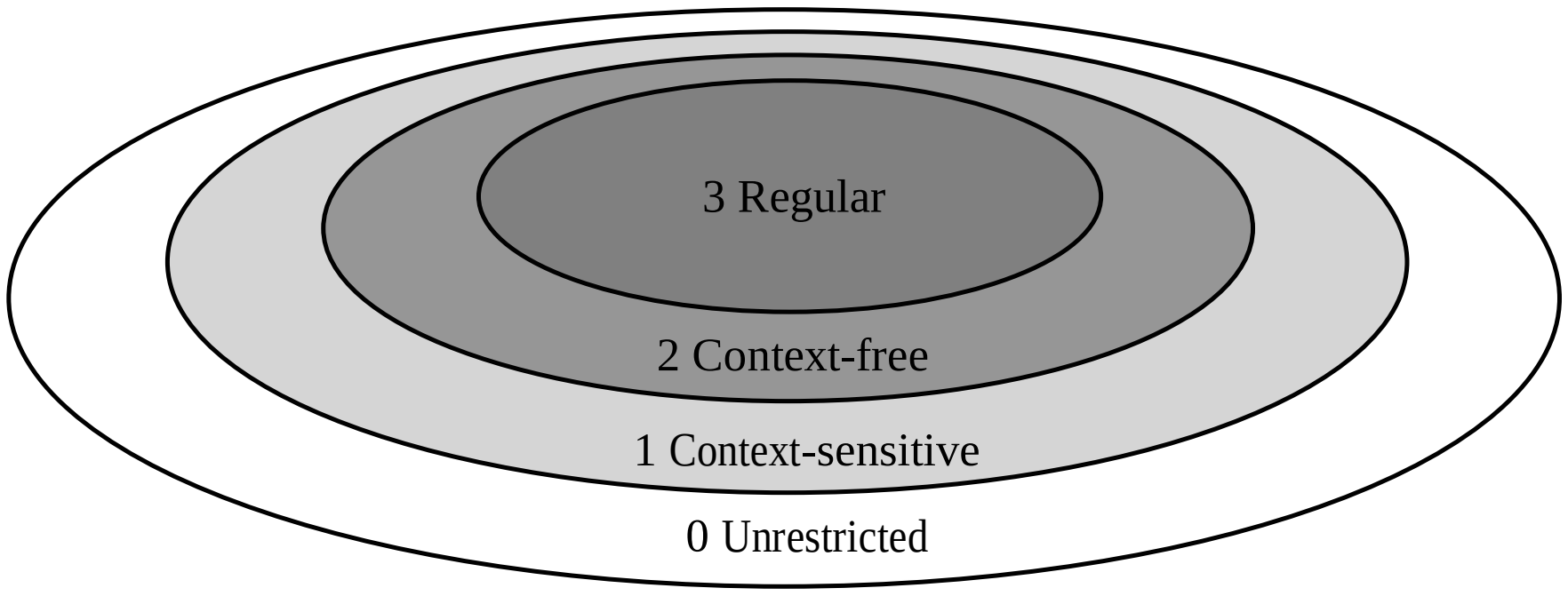

**Figure 3-12.** Chomsky's hierarchy of grammars with four levels: regular, context-free, context-sensitive and unrestricted.

*Context-sensitive grammars* (CSGs) and *context-free grammars* (CFGs) are the two most relevant classes of grammars for parsing (Geman & Johnson 2002, Winograd 1983). As shown in Table 3-2, in a CSG, the right-hand side of a rule defines the context in which a nonterminal can be replaced with the right-hand side. In contrast, CFGs do not take into consideration the context in which the nonterminal on the right-hand side occurs.

**Table 3-2.** The levels of Chomsky's hierarchy of grammars most relevant to natural language parsing (Chomsky 1959, Winograd 1983). α and β refer to the left- and right-hand side of the rules.

| **Grammar** | | **Rules** | | **Language** |
|---|---|---|---|---|
| **Level** | **Class** | **Form** | **Restrictions** | |
| 1 | Context-sensitive | β consist of α with a single symbol expanded | $\|\alpha\| \leq \|\beta\|$ | Any language whose sentences can be recognized by a deterministic computational machine using an amount of storage proportional to the length of the input. |
| 2 | Context-free | α consists of a single nonterminal symbol. | $\|\alpha\| = 1$ | Includes languages that involve embedding such as $a^n b^k c^n$ but not $a^n b^n c^n$ or *WW*, where *W* is an arbitrary string of terminal symbols. |

It is difficult to define a grammar that generates a given natural language. On the one hand, a grammar has to have enough generative power to handle all possible sentence constructions in the language. On the other hand, the generative capacity has to be constrained in order to make predictions about the structure of a language (Feinstein & Winter 2006). Because of their many favorable characteristics, CFGs are often used by parsing systems. There are, however, constructions in natural languages that cannot be described by CFGs (Shieber 1985a). Languages of the type $a^n b^n c^n$ cannot be described by CFGs because there is no way in which to formulate the fact that there is a specific number of *c*'s at the end of the string. Such constructions exist, for example, in Swiss German and Dutch. An example of a English sentence with dependencies that are beyond capabilities of CFGs is "Jamie, Steven and Harry are a defender, a midfielder, and an attacker, respectively". Such distinctions call for rules that are sensitive to the context. However, CSGs are actually more powerful than is needed for describing natural languages.

While *mildly context-sensitive grammars* (MCSGs) are slightly more powerful than CFGs, they are less powerful than CSGs (Joshi 2003a, Joshi & Schabes 1997). MCGSs are powerful enough to model natural languages while remaining efficiently parsable. For example, a commonly applied class of MCGSs, level-2 MCSGs, are able to capture up to 4 counting dependencies (includes $L_4 = \{a^n b^n c^n d^n | n \geq 1\}$, but not $L_5 = \{a^n b^n c^n d^n e^n | n \geq 1\}$ (Castaño 2003).

Formal equivalences between grammar formalisms can be described with the notions of *weak* and *strong equivalence*:

**Definition 3-20**. Weak and strong equivalence (Miller & Chomsky 1963, Miller 1999)

1. Two grammars, $G_1$ and $G_2$, are *weakly equivalent* if and only if they generate the same set of strings, i.e. iff $L(G_1) = L(G_2)$.
2. Two grammars, $G_1$ and $G_2$, are *strongly equivalent* if and only if they are weakly equivalent and if they assign the same set of structural descriptions for each sentence $\delta$ in $L(G_1)$ and $L(G_2)$.

GPSG, CatG, LG and projective DGs[33] are weakly equivalent to CFGs (Joshi 2003a, 2003b, Gaifman 1965). While they can generate the same set of languages, they do not assign the same description to the sentences of these languages. Since they have CF power, these formalisms are not powerful enough for fully modeling natural languages. Infante-Lopez and de Rijke (2006) have shown that PCFGs can define a set of trees that cannot be derived from rules of any CFG. From the perspective of formal language theory, probabilities thus are fundamental, and they add power to PCFGs.

[33] If a CFG is restricted so that one word in each phrase is designated its head and the phrase has no name or designation apart from the designation of its head, a DG and the CFG are strongly equivalent (Covington 2001).

TAG, CCG, *Linear Indexed Grammar* (LIG) (Gazdar 1988), and *Head Grammars* (Pollard 1984), are all level-2 MCSGs and are thus weakly equivalent (Vijay-Shanker & Weir 1994). Kuhlmann and Nivre (2006) describe the class of *mildly non-projective D structures*, which, they claim, are rich enough to account for naturally occurring syntactic constructions as well as sufficiently restricted to enable efficient parsing. In a similar vein, Yli-Jyrä and Nykänen (2004) have proposed a family of DGs that belong to the class of MCSGs.

When Joshi (1985) compared the generative capacities of GPSG, TAG and LFG, he came to the conclusion firstly that TAG is more powerful than GPSG and, secondly, that LFGs are context-sensitive and thus much more powerful than the two other formalisms. In HPSG the formalism itself does not set constraints on the power of a grammar (Rambow 1994). But the full power of the formalism is not necessarily used. I am not aware of any research in which the weak generative capacity of a specific HPSG has been investigated.

CG possesses a formal power that is less than that of regular languages (Tapanainen 1999). It is well known that projective DGs are not a powerful enough formalism for fully describing natural languages. This means that there is a need for non-projective DGs. But there is very little research about the formal power of non-projective DGs. While Tapanainen (1999), for example, shows that FDG is more powerful than CFGs, he does not formulate the exact generative capacity of the formalism. XDG is similar to HPSG in a sense that the formalism itself does not constrain the generative capacity of a grammar (Debusmann 2003). Debusmann (2006) notes that XDGs are at least as powerful as MCSGs. Table 3-3 summarizes the above discussion.

**Table 3-3**. The grammar formalism discussed in this work predicated on the basis of their formal powers.

| Level of grammar | Formalisms |
|---|---|
| *Less than regular* | CG |
| *Context-free* | CFPSG |
| | GPSG |
| | CDG |
| | LG |
| | Probabilistic LG |
| *More than context-free, but less than mildly context-sensitive* | PCFG |
| | FDG?* |
| *Mildly context-sensitive* | TAG, LTAG, PLTAG, CCG |
| *Context-sensitive* | LFG |
| *Unrestricted* | TG |
| | HPSG |
| | XDG |

*While Tapanainen (1999) shows that FDG is more powerful than CFGs, he does not formulate the exact generative capacity of the formalism.

### *3.4.4 Long-distance dependencies*

*Long-distance dependencies* (LDDs) are problematic for any theory of grammar because one needs to use non-local information when one analyzes the structures that contain them – linguistic structures cannot always be interpreted locally in the places where they are encountered. Table 3-4 illustrates how LDDs can arise out of phenomena such as extraction and coordination.

**Table 3-4**. Examples of long-distance dependencies.

| Cause | Type | Example |
|---|---|---|
| **Extraction** | *Wh-relative clause* | This is the player [who$_1$ the coach despises […]$_1$] |
| | *Tough-movement* | The coach$_1$ is hard to please […]$_1$ |
| **Coordination** | *Sentential gapping* | Harry played$_1$ football and Robbie […]$_1$ tennis. |
| | *Right-node raising* | Harry passed […]$_1$ and Robbie headed the ball$_1$. |

It was the existence of LDDs that motivated researchers to develop TGs. LDDs are handled in TGs by the transformations. LDDs are dependencies on the level of PAS (Vijay-Shanker 1989). The consequence of this is that CFPSG and the basic PCFG model, for example, are unable to handle LDDs because it is difficult, if not impossible, to state dependencies of that kind in a formalism that uses only PS-style syntactic representation. However, PCFGs and other probabilistic grammars can be augmented with LDD handling mechanisms. Collins (1997), for example, adds a probabilistic model for wh-movements to the basic PCFG parsing model.

GPSG and HPSG treat LDDs in a similar way – by means of feature-passing. In HPSG signs are divided into two parts: one defines local information and the other is used for handling LDDs (Pollard & Sag 1994). While local features are unified locally, non-local features pass their specification to larger phrases (Bouma *et al*. 2001).

In LFG, LDDs are treated by means of the *functional uncertainty* mechanism that allows one to state the constraints on LDDs in F-structures (Kaplan & Maxwell 1988). Joshi and Vijay-Shanker (1989) demonstrated a direct correspondence between the functional uncertainty in LFG and elementary trees in TAGs. The elementary trees in TAG provide an *extended domain of locality* that allows them to localize the predicate-argument dependencies. The handling of LDDs is therefore accomplished by the formalism itself. Joshi and Vijay-Shanker also enunciated a corollary to this finding by demonstrating that the handling of LDDs can be accomplished by means of MCSGs.

Coordination is regarded as a particularly difficult phenomenon to handle in DG formalisms (Lombardo & Lesmo 1998b). Hudson (1990), for example, suggests that a PS-style model is needed to describe coordination in DGs. Lomardo and Lesmo (1998b), among others, have proposed frameworks for dealing with LDDs within a DG framework. FDG handles coordination by chaining the coordinated elements (Järvinen & Tapanainen 1998). In contrast to other D links in an FDG analysis, links that mark coordination do not imply a D relation but rather a functional equivalence. It is this treatment of coordination that gives FDGs the ability to cope with *gapping* – another phenomenon that is regarded as a serious problem for DGs. The treatment of LDDs in XDG is a consequence of the modularity of the formalisms. The control and raising constructions, questions, *topicalization,* and *relative clauses,* for example, are dependent on the way in which the dimensions are connected, and they do not need to be specified explicitly. XDG, however, is able not to handle coordination structures with *ellipses* (Debusmann 2006).

Because LG is CF and produces projective analyses, it is not well suited to LDDs. The handling of coordination is achieved by a special type of connector called *fat connectors* that represent an ordered sequence of ordinary connectors and that allow "and" lists to be correctly handled (Sleator & Temperley 1993). This approach has only been applied to "and" coordination. Pyysalo *et al.* (2006) however argue that LG's approach to coordination compares favorably to FDG's chaining method. They claim that the LG approach is more expressive. While LG coordination structures can be transformed into chaining structures, the opposite is not possible.

While spurious ambiguity and the extra derivations that are caused by it increases the complexity of CCG parsing, it enables elegant analyses of coordination and extraction (Steedman 2000). Hockenmaier (2003) proposed a model that captures LDDs by expressing word-word dependencies of the PAS. The model allows multiple dependencies between a word and its context, as well as dependencies that arise through control, right node raising and other LDD constructions.

### *3.4.5 Grammar development*

The labor-intensiveness of grammar building is dependent both on the formalism and on the availability of grammar development tools. Since most of the current grammar formalisms used in parsing are lexicalized, this sets some specific requirements for grammar development. Is usual for lexicalization to give rise to a loss of generality in the grammar (Pedersen 2001).[34] LFG, for example, resolves this complication by *lexical rules* that allow for the application of a single rule to similar lexical items.

#### 3.4.5.1 Manual grammar development

The manual construction of a wide-coverage grammar presupposes many years of skilled human labor on the part of someone who possesses a great deal of expert knowledge and experience in this field of the discipline.[35] This kind of work also requires intensive collaboration between theoretical linguists and grammar writers (Oepen *et al.* 2000). Over and above these requirements, it is necessary to utilize engineering skills to resolve technical problems.[36] One of the major problems in manual grammar writing is to ensure consistency (Miyao *et al*. 2004).

---

[34] In an LTAG, for example, the variations on a basic sentential form (for instance *wh-movement*), have to be implemented as additional elementary trees.

[35] Since manual grammar development is such an ambitious undertaking, it should rely on the techniques and design principles that are already used in software engineering (Dipper 2003). Such principles include *modularity* (this means that a grammar code should be divided into modules for easier development and maintenance). The modules in a grammar assemble pieces of code that are functionally related. For instance, the two representation levels of LFG (C- and F-structure) are separated into modules. Like software projects, a grammar development project needs to be carefully documented. Large parts of this documentation typically consist of highly detailed code-level documentation. Another typical feature is the large number of links between the parts of documentation. Because the content of the dependent modules is required for understating the functionality of the parent module, it is necessary to make the documentation of the dependent modules easily accessible.

Because grammar building is such an expensive process, only a few hand-crafted deep grammars have achieved sufficient coverage to parse large collections of free text (Burke *et al*. 2004a). It is the stated aim of projects such as *ParGram* (Butt *et al*. 2002) to lower the intrinsic development costs involved in this process by using common tools and methods on several research sites where such grammars are being developed for multiple languages.

*Grammar development environments* are software systems that offer grammar writers several tools that have the capacity to simplify their work (Carroll 1993). Such tools support incremental input, grammar editing, browsing and searching the grammar. They also offer analytical tools for monitoring interactions between different parts of the grammar and for debugging by rule-tracing. *LKB* for HPSG (Copestake & Flickinger 2000), *Grammar Writer's Workbench* (Kaplan & Maxwell 1993) and XLE (Butt *et al*. 2002) for LFG, and *XDG Grammar Development Kit* (Debusmann *et al*. 2004a), are all examples of such environments.

### 3.4.5.2 Grammar induction

*Automatic grammar induction* is based on a treebank. The linguistic intuition is externalized into the annotations of the treebank and a grammar explaining the annotations is then learned automatically from the treebank (Miyao *et al*. 2004). Figure 3-13 illustrates the idea of grammar induction.

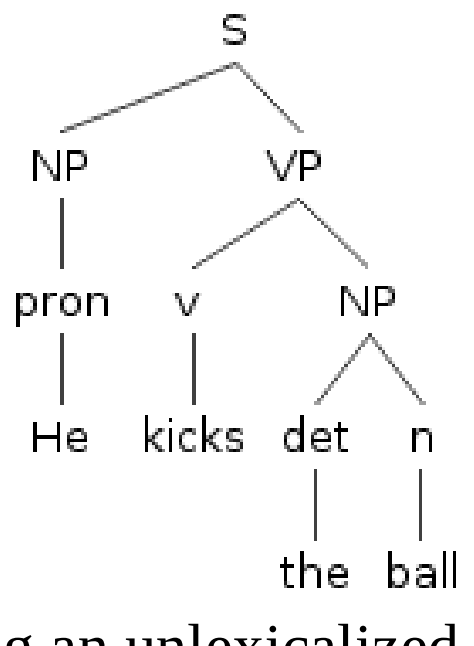

S → NP VP
NP → pron
VP → v NP
NP → det n

**Figure 3-13.** Learning an unlexicalized grammar from a treebank. The rules on the right-hand side of the figure can be deduced from the tree on the left-hand side.

---

[36] The scaling of a rich UG beyond small text fragments to unrestricted text is, for example, both time-consuming and expensive (Cahill 2004). One may note, for example, that the LinGO HPSG for English contains 100,000 lines of source code, 27 lexical and 37 PS rules, and around 6,000 lexical entries (Oepen & Callmeier 2000).

While it is relatively easy to compose the rules for lexicalized grammars, the sheer number of lexical entries presupposes that such a task will be extremely labor-intensive (Miya *et al.* 2003). There have been a few attempts in recent years to develop methods that will allow for the automatic generation of lexical entries. It is also possible to extract both the rules and the lexicon of a lexicalized grammar from a treebank.[37] At its best, automatic grammar induction is fast and cheap and it produces grammars with a wide coverage (Cahill 2004). Burke *et al.* (2004b), for example, estimate that it took them less than three person months to automatically induce a Chinese LFG.

The analyses provided by such grammars are often more shallow than those provided by their hand-crafted counterparts. The automatic induction of deep grammars, however, is currently being actively researched. Efforts in this field have yielded promising results, and have produced grammars that perform equally well or better than hand-crafted grammars. Grammar induction has been recently applied to HPSG (Miyao *et al*. (2003, 2004), to LFG (Burke *et al*. 2004a, 2004b, Donovan *et al*. 2005), and to CCG (Hockenmaier & Steedman 2002). All three methods have been applied to English and several other languages. They all work by first converting a set of syntactically analyzed sentences into HPSG, LFG or CCG formats respectively. Grammar induction methods are then applied to these converted structures. The method devised by Miyao *et al*. differs from the other two methods in that the rules are written manually and only the lexical entries are learned automatically. Table 3-5 outlines a comparison of these methods.

[37] Neumann (2003), for example, shows that LTAGs can be successfully learned from treebanks.

**Table 3-5**. Comparing HPSG, LFG and CCG acquisition methods (based on a lecture by Van Genabith (2006)). The same treebank was used in all these experiments. While the treebank structures are converted in a conversion-based approach without the addition of any new information, new information is added in an annotation-based approach. Tree binarization refers to the process of transforming the syntactic trees so that they will contain only constituents with two daughters. The size of the grammar and translation coverage refer to the English grammars obtained from the PTB. Translation coverage is the percentage of the PTB structures that were successfully converted into the target formalism.

| | **LFG** | **CCG** | **HPSG** |
|---|---|---|---|
| *Annotation/conversion -based* | Annotation | Conversion* | Conversion (with some annotation) |
| *Preprocessing by tree binarization* | No | Yes | Yes |
| *Preprocessing by cleaning up the treebank and correcting errors* | No | Yes** | Yes |
| *Manual editing of the acquired rules/lexicon* | None | Some editing of extracted categories | Some editing of lexical entries & rules |
| *Grammar size* | 50,000 rules | 3,262 rules; 1,286 lexical categories; 44,210 word types | 12 schemas; 1,942 lexical categories |
| *Translation coverage (%)* | 99.8 | 99.4 | 95.1 |
| *Languages* | English, German, Spanish, Chinese | English, German, Turkish | English, Japanese |

*For example, inserting a noun level into NPs and analyzing multiword expressions. **POS tagging & bracketing errors.

## 3.5 Summary of Findings

Figure 3-14 and Tables 3-6 and 3-7 summarize the main characteristics of the grammar formalisms surveyed in this chapter and the relations between them.

**Table 3-6**. A summary of the grammar formalism analyzed above. KEY: Fin = Finnish; Rus = Russian; Swe = Swedish; Tur = Turkish; Nor = Norwegian.

| Type | Forma-lism | Strata | Languages | | | | | | | | | | LDDs | Notes |
|---|---|---|---|---|---|---|---|---|---|---|---|---|---|---|
| | | | *English* | *German* | *French* | *Spanish* | *Italian* | *Dutch* | *Japanese* | *Chinese* | *Arabic* | *Others* | | |
| CF | PCFG | 1 | x | x | x | x | x | x | x | x | x | Czech, Rus, Latin + several others | No treatment of LDDs in the basic model | Rather simple models. Especially suitable for grammar induction |
| UG | GPSG | 1 | x | x | x | | | | | | | - | SLASH-features | Replaced by HPSG in parsing |
| | HPSG | 1 | x | x | x | x | x | x | x | x | | Nor, Korean, Greek + several others | SLASH-features, non-local features in signs | Descendant of GPSG, nonderivational |
| | LFG | 1 | x | x | x | x | | | x | x | x | Nor, Urdu, Tur + several others | Functional uncertainty | Assigns two types of representations |
| TAG | TAGs | 1 | x | x | x | | | | | x | x | Korean, Hindi | Extended domain of locality | Derivation trees. UG, lexicalized and probabilistic versions exist |
| DG | CG | 1 | x | x | x | x | x | | | | | Fin, Swe, Nor, Danish + several others | Not powerful enough for treating LDDs | Underspecified and shallow analyses |
| | FDG | 1 | x | x | x | x | | | | | | Fin, Swe | Coordination and gapping as chains of words | Descendant of CG, non-projective analyses |
| | XDG | M | x | x | | | | x | | | x | Czech | No treatment of coordination with ellipsis | Some details of the formalism are not completed |
| | LG | 1 | x | x | | | | | | | | - | And-coordination | In many ways similar to DGs |
| CatC | CCG | 1 | x | | | | | x | x | | | Tur, Korean, Irish Gaelic, Tzotzil | Follows from the spurious ambiguity | Strong on coordination and extraction |

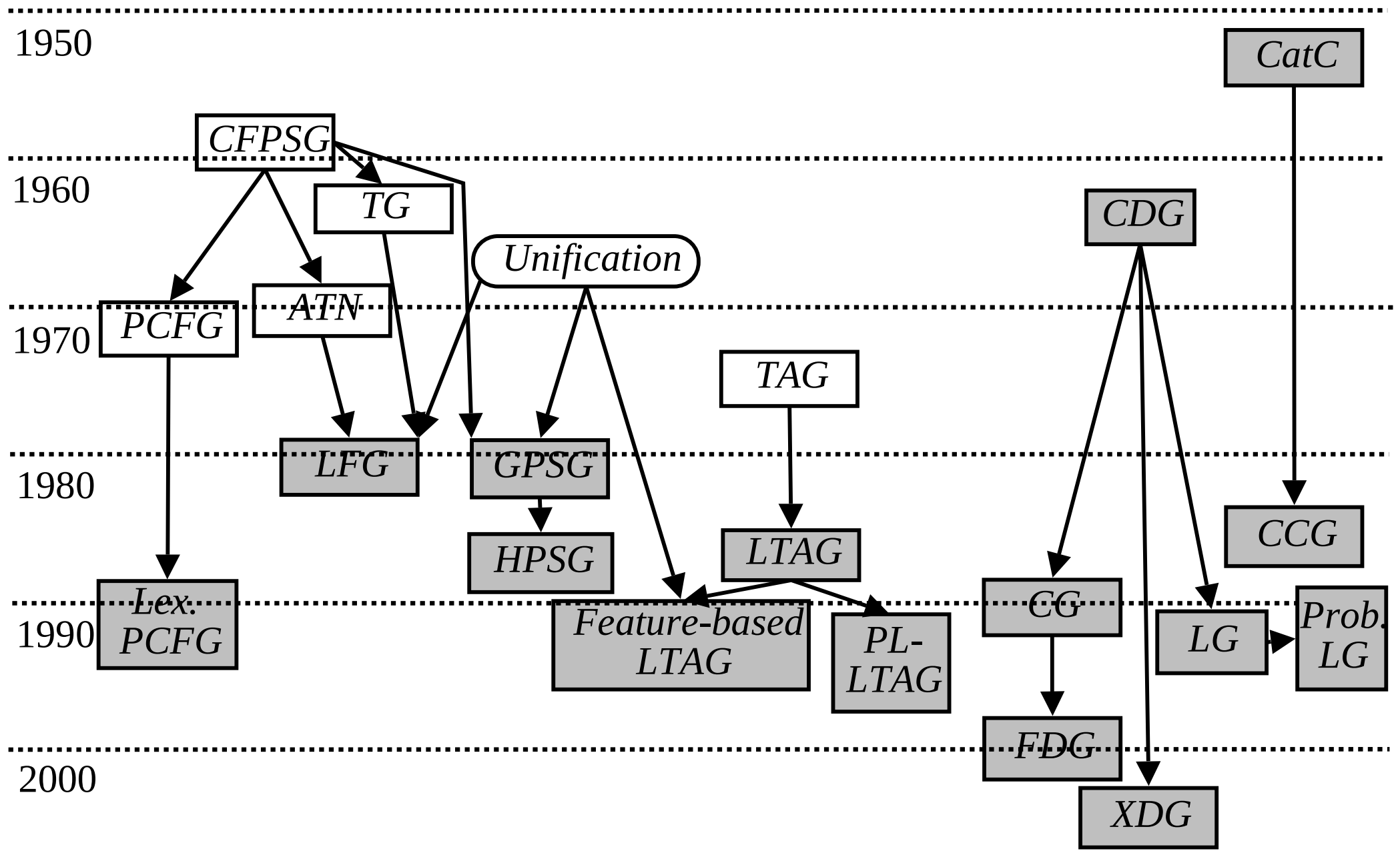

**Figure 3-14**. An overview to the grammar formalisms discussed in this chapter. The grammars shaded with gray boxes are lexicalized. An arrow between two formalisms indicates that the formalism at the beginning of the arrow has influenced the development of the formalism at the end of the arrow.

**Table 3-7**. A summary of the grammar formalisms. The row “development environment” indicates whether there are publicly available grammar development environments for the formalism. Two plusses indicate that there are several tools (more than two) available. The row “grammar induction” indicates whether a grammar induction method supporting the formalism has been reported in the literature. KEY: U = unrestricted; CS= context-sensitive; M = mildly context-sensitive; R = regular. “>“ and “<“ indicate whether the formalisms have a formal power that is greater than or less than the power of the given grammar class.

| | *CFPSG* | *TG* | *PCFG* | *GPSG* | *HPSG* | *LFG* | *TAG* | *LTAG* | *PLTAG* | *CDG* | *CG* | *FDG* | *XDG* | *LG* | *Proba. LG* | *CCG* |
|---|---|---|---|---|---|---|---|---|---|---|---|---|---|---|---|---|
| *Probabilistic* | - | - | + | - | - | - | - | - | + | -/+* | - | - | - | - | + | - |
| *Lexicalized* | - | - | -/+ | + | + | + | - | + | + | + | + | + | + | + | + | + |
| *Unification* | - | - | - | + | + | + | - | -/+ | - | - | - | - | - | - | - | - |
| *Generative capacity* | CF | U | >CF | CF | U | CS | M | M | M | CF | <R | >CF*** | U | CF | CF | M |
| *Development env.* | + | - | - | ++ | ++ | ++ | + | + | - | - | - | - | + | - | - | + |
| *Grammar induction* | + | - | + | - | + | + | - | + | + | - | +** | - | + | - | +**** | + |

*Eisner (1996a, 1996b), for example, has proposed probabilistic DG models. **While Samuelsson *et al.* (1996) represented a method for automatically inducing CGs, the performance of the resultant induced grammars was found to be far inferior to that of manually constructed ones. ***Tapanainen (1999) shows that FDG can describe CF languages and structures of the type $a^n b^n c^n$ that are not describable by CFGs. He does not, however, formulate the exact generative capacity. **** Fong and Wu (1995) experimented with probabilistic LG induction.

## 3.6 Conclusion

In conclusion, I make general observations (some of which have been discussed by Backofen *et al.* (1996)) about overall trends in the development of parsing grammar formalisms on the basis of the analysis.

While DGs and CatGs, for example, have always been lexicalized theories, the trend towards lexicalization is also strong in other formalisms such as the “Chomskyan” grammars, TAG and PCFG. Lexicalization creates a rich lexicon and diminishes the amount of the rules; more syntactic and semantic information is coded in the lexicon and there are fewer rules that contain less information. This increased complexity of the lexicon necessitates additional mechanisms with which to organize the lexicon and keep it free from redundancy. Hierarchical structuring and inheritance are therefore being more frequently used to organize the grammar and the lexicon. This approach is used, for example, in HPSG.

Inheritance allows for the building of more complex and specialized categories from more basic ones. Feature structure representation has also become increasingly popular in formalisms other than UGs, such as DGs and TAG.

In contrast to the earlier models, many of the current grammar formalisms, such as XDG, HPSG and LFG, tend to integrate several levels of linguistic knowledge into one formalism. The developers focus accordingly on integrating the analyses of various phenomena into one coherent theory, and many formalisms have moved from construction-specific rules towards general and language-specific principles. This trend is visible, for example, in HPSG. Many modern grammar formalisms permit discontinuous constituents and a free word order. Non-projective DGs are an extreme example of this trend.

PCFG parsers have arguably been applied to more languages than parsers based on any other grammar formalism. This is probably because they are relatively simple and easy to implement: only a treebank is needed from which to learn the grammar. The most intensively developed grammar formalisms for parsing at the moment appear to be based on LFG, HPSG and CCG. LFGs and HPSGs have been implemented for a wide variety of languages. There also are grammar development environments available for these formalisms. There are specifically multiple options available for LFG and HPSG developers. An important trend in grammar development is the rise of automatic induction methods, and research has been active in this field in recent years– especially among the LFG, HPSG and CCG research communities.
In addition to CCG and the two UG formalisms (HPSG and LFG), a considerable amount of work has been done in the DG framework, especially on XDG, CG and FDG. While XDG may be rated the best for having grammar development and induction tools available, some aspects (such as the treatment of coordination (Debusmann 2006)) of this new formalism still need to be further developed.

A grammar formalism is merely a language in which linguistic theories can be expressed. The use of a specific formalism does not therefore guarantee good parsing results. It is the quality of the grammar itself that is the key factor. It seems to be the case that the selection of a grammar formalism for a parser depends mostly on the following factors: the personal preferences of the user, the availability or otherwise of resources, the quality of the tools available for the formalism, and the needs set by the NLP system in which the parser is to be applied.

# 4 Syntactic Analysis – Parsing Algorithms

The parsing algorithm is the component of a parser that applies the grammar to the input sentences to construct parse trees. Parsing algorithms are usually not designed for individual grammars, but for classes of grammars. This chapter provides an overview on fundamental search strategies (Section 4.1), parsing algorithms (Section 4.2) and approaches to parsing (viz. probabilistic, lexicalized and finite-state approaches) (Section 4.3). It also analyzes how these are applied in parsing specific grammar formalisms (Section 4.4). Section 4.5 considers the computational properties of the formalisms introduced in Chapter 3. Section 4.6 concludes the discussion on grammar formalisms (Chapter 3) and their computational properties (Chapter 4).

## 4.1 Introduction

One may consider the problem of how to define search algorithms from several different points of view (Hellwig 2002, Pulman 1993). One may therefore describe parsing algorithms in terms of the direction in which the structure is built (whether *top-down* (Yngve 1959) or *bottom-up* (Glennie 1960)), or the way in which the search is executed (either *breadth-first* or *depth-first*), or in terms of the direction in which the input words are processed (from *left-to-right* or *right-to-left*).

Top-down algorithms[38] begin at the root of the parse tree and proceed from top to the bottom by trying to add nodes in accordance with the rules of the grammar (Pulman 1993, Hellwig 2002). The bottom-up approach starts from the words themselves and builds up the parse tree until the root of the tree is reached.[39] But neither of these two strategies alone adequately exploits the constraints set by the grammar and the input. *Bottom-up parsing with top-down filtering* combines the two methods by operating from bottom to top, but by applying top-down constraints to guide the search.

Depth-first searching always proceeds from the direction of the left-most symbol until a terminal is reached, thereby pursuing a single derivation at a time (Carroll 1993). In the breadth-first strategy, symbols are processed in the order in which they were created. This means that all the derivations are pursued simultaneously.

[38] This approach is sometimes aptly called *expectation driven* because it makes use of the derivation rules from left to right, thereby "predicting" which units will occur in the input string.
[39] In D parsing, a top-down approach proceeds from the head to the dependent. But in a bottom-up algorithm, the order is the opposite.

The depth-first method is usually a preferable one because memory usage is a problem in the case of a breadth-first search.

When a *non-deterministic* parsing algorithm finds itself in a situation in which several grammar rules are simultaneously applicable to an analysis point, it needs to choose one rule and then apply it. If the path that it selects later turns out to be false, the algorithm backtracks to the starting point and tries another rule (Dutoit 1997). This kind of procedure can generate a great deal of inefficient parsing. Since a *deterministic* algorithm never backtracks, it is able to produce the parse in *linear time*; the upper bound for the time required is proportional to the size *n* of the input, denoted by O(*n*).

In *left-to-right* parsing, the rules of a grammar are always processed from the left-most daughter of the rule toward the right (Pulman 1993, Carroll 2005). A right-to-left strategy is much rarer in parsers. *Island parsing* is a bidirectional approach in which analysis begins from a certain position in the sentence called the *island*, before it proceeds in both directions. In *head-driven parsing,* the head of each rule is taken as the starting point (Kay 1989, Nederhof & Satta 1994). The essence of this approach is that there is a distinguished member, the head, in each rule which is first recognized. The motivation is to start from the elements within the input string that carry the most syntactic content.

The number of possible parses for a sentence may grow exponentially as the length of a sentence grows. Two representation techniques, charts and chart packing, can be applied to avoid the exponential growth of the size of the parse forest. This will result in an increase in parsing efficiency. *Charts* are the basis of many parsing algorithms (Sikkel & Nijholt 1997, Carroll 2005). The chart is used for storing completed items that need no further processing. In a chart parser, complete sub-parses are saved. This obviates the duplicate searching of sub-parses that have already been found (Pulman 1993). The use of a chart also enables one to return a fragmented analysis based on the sub-parses recorded in the chart if the algorithm fails to produce a complete parse. It is moreover sometimes possible to combine a complete parse from fragments in the chart.[40] While a *passive chart* records only complete constituents, an *active chart* records information about the stages in the application of the rules (Carroll 2005).

[40] Mellish (1989), for example, applies a strategy in which a bottom-up parser is run over the input. If it fails to produce a complete parse, a top-down parser is run over the chart created by the bottom-up parsing in order to hypothesize possible complete parses.

Although chart parsing makes processing more efficient, it nevertheless requires exponential time and space to assign parses to a sentence with an exponential number of possible parses (Pulman 1993). A solution to this problem is the application of *chart packing*. Instead of representing *all* analyses in the chart and thus undertaking redundant work, sub-parses that represent *local ambiguity* can be merged and treated as one node (Tomita 1987).[41] Figure 4-1 provides an example of chart packing.

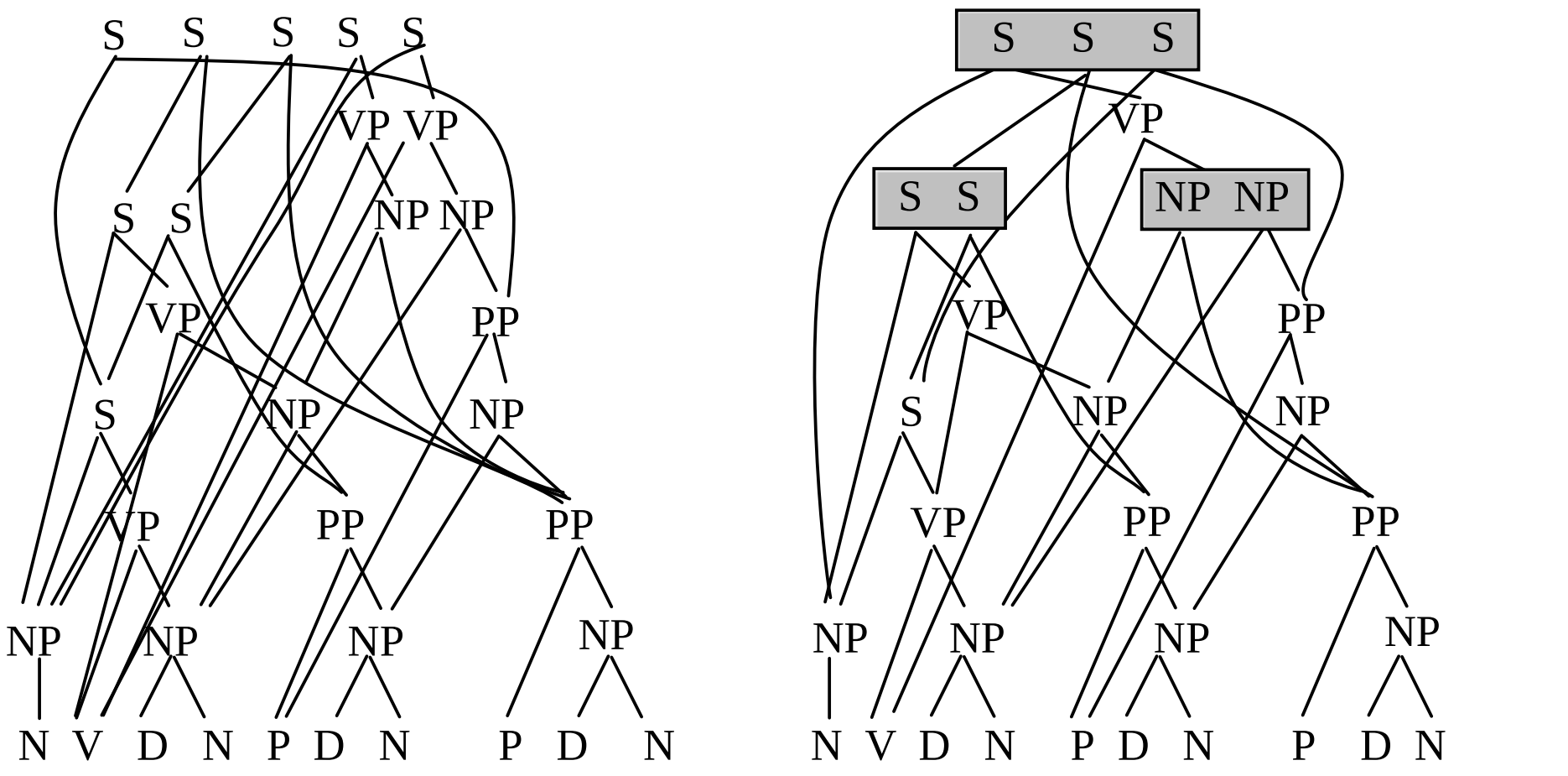

**Figure 4-1**. An example of chart packing. The unpacked structure is on the left-hand side and the packed one is on the right (adapted from (Tomita 1987)). The example on the left illustrates a basic chart parser with a set of parses with sub-parse sharing. The parse tree on the right contains packed nodes represented as boxes.

## 4.2 Parsing Algorithms

This chapter introduces the most fundamental parsing algorithms that were originally developed for parsing CFPSGs (Section 4.2.1) and two commonly used techniques for increasing speed, namely supertagging and CF filtering (Section 4.2.2).

### *4.2.1 Fundamental algorithms*

A common strategy for developing a parsing algorithm for classes of grammars that are more powerful than CFGs is to generalize a CFG algorithm to the more powerful class (Van Noord 1994). Even though a number of CFPSG parsing algorithms were developed already in the 1960s and 1970s, most of them are still in use, although modified to cope with newer grammar formalisms.

[41] Tomita uses the term *parse forest* for structures containing a set of trees represented by using packed nodes.

The *Cocke-Younger-Kasami* (CYK) (Kasami, 1965, Younger 1967) algorithm is a bottom-up method that uses a passive chart. The algorithm requires a CFG in the *Chomsky-normal form*, in which every rule must rewrite a nonterminal either as a single terminal or as two nonterminals.[42] The *Earley algorithm* (Earley 1970) is a parallel, top-down, left-to-right procedure that uses a passive chart. One of the most significant contributions of the Earley algorithm is that it is more generally applicable than, for example CYK, in the sense that although it can process CFGs at least as efficiently as earlier algorithms, it does not require the grammar to belong to a specific class.

*LR(k) parsing*[43] is a bottom-up parsing technique that is based on *shift-reduce* processing.[44] While an Earley parser constructs sets of possible productions on the run by following all the possible partial derivations, an LR parser has access to a pre-computed list of possible derivations (Stolcke 1995). Because much of the work takes place in the preprocessing phase, the result is a relatively simple run-time parsing. Several modifications of the LR(k) algorithm have been devised. These include Tomita's (1987) *Generalized LR*, which is a non-deterministic LR algorithm that uses local ambiguity packing and breadth-first search. *Look-ahead LR* (LALR) is a refinement of the technique for constructing the LR parse tables. *Head-inward parsing* combines a head-driven approach with LR processing (Bouma & Van Noord 1993, Nederhof & Satta 1994).

A *left-corner* (LC) algorithm will consider a grammar rule only if the current input word can serve as the LC of some derivation from that rule (Matsumoto *et al*. 1983, Moore 2004). The LC relation is usually precompiled by the parsing algorithm and indexed so that any pair of symbols can be checked in constant time.

---

[42] Although CFGs can usually be automatically converted into Chomsky-normal form, the size of the grammar may grow exponentially in the conversion, making parsing inefficient.

[43] The algorithm's name derives from the fact that it reads the input from *L*eft to right and produces a *R*ightmost derivation. The *k* refers to the number of look ahead symbols that are used in making parsing decisions. Typically *k* is 1, and hence LR refers to a LR(1) parser. An LR(0) parser makes decisions based on stack contents. An LR(1) uses, in addition, the next token on the input.

[44] In a shift-reduce parser either of the two actions is performed: a *shift* action consumes a word from the input string and pushes it onto the stack. A *reduce* action applies a grammar rule. Shift-reduction results in efficiency because it can delay decisions. For example, a word that has ambiguous POS tags can be shifted onto the parse stack and the final categorization will be delayed until a reduction involving the word is made.

**Example 4-1**. Left-corner of a parse tree. The word "saw" with the label V form the left-corners of the tree.

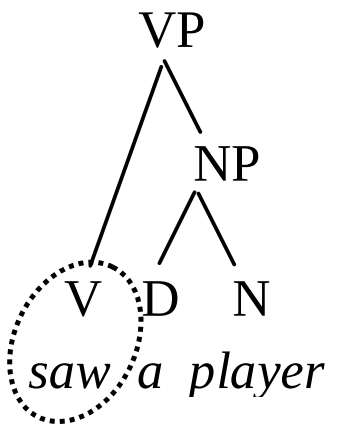

*Head-corner* (HC) parsing can be considered as a generalization of LC parsing (Sikkel & Op den Akker 1993, Bouma & Van Noord 1993). It is also closely connected to head-driven parsing. While an LC parser processes a parse tree from left to right, HC parsing begins from the heads. HC allows for parsing with more powerful grammars than LC parsing.

### *4.2.2 Supertagging and CF filtering*

The basic parsing algorithms discussed above have been adapted in several ways so that they can contribute to more efficient and accurate parsing systems. This section introduces two such adaptations: supertagging and CF filtering. *Supertagging*[45] is a technique for reducing the cost of parsing lexicalized grammars (Srinivas & Joshi 1994, 1999). It was first introduced for LTAG and has later been applied to several other formalisms. Instead of assigning POS tags, supertagging assigns more informative supertags to each word in an input sentence. In LTAG, each supertag corresponds to an elementary tree. After supertagging, the remaining step of determining the actual syntactic structure of an input sentence is rather trivial. It is also possible to return a fragmented analysis consisting of supertags in a case where a full parse cannot be produced. Supertagging also decreases the number of elementary trees that are assigned to each word, thereby increasing efficiency.

CF filtering is a technique for increasing the speed of UG parsing (Torisawa *et al*. 2000, Matsuzaki 2007). CF filtering first parses an input sentence by means of a CFG that approximates the UG, and only subsequently with the original grammar. For example, in the parser created by Torisawa *et al*. (2000), a CFG is extracted from an HPSG and is used in the first stage of parsing. The CFG typically generates a large set of possible parse trees. These trees are eliminated in the second stage by the HPSG. The use of CF filtering makes the algorithm more efficient if one compares it to the use of an HPSG alone because it avoids unnecessary unification by eliminating impossible parse trees.

[45] Srinivas and Joshi (1999) claim that supertagging is "almost parsing". The syntactic information provided by supertags is so rich that there is only some structural ambiguity left and the parse is almost entirely determined by the supertags.

## 4.3 Probabilistic, Lexicalized and Finite-state Parsing

While non-probabilistic algorithms (such as those described in Section 4.2.1) regard parsing as the recursive application of predetermined rewrite rules, probabilistic parsing (Section 4.3.1) approaches the problem by means of automatically discovering the disambiguation criteria for the decisions made during the parsing process. Because many of the grammar formalisms applied in parsing are lexicalized, lexicalized parsing (Section 4.3.2) has become increasingly important. Probabilistic and lexicalized parsing are often combined. Finite-state machines have been used for many NLP tasks such as segmentation and morphological analysis. Section 4.3.2 introduces a discussion on how FSMs can be applied in parsing.

### *4.3.1 Probabilistic parsing*

In addition to defining the probabilities for parses and thus finding out the most probable analyses, probabilistic information can be used for speeding up the parsing process by ordering the search space (Magerman 1994, Collins 2003). The goal here is to identify the best parse more quickly while simultaneously not undermining the quality of the produced results. Figure 4-2 shows the idea of probabilistic parsing.

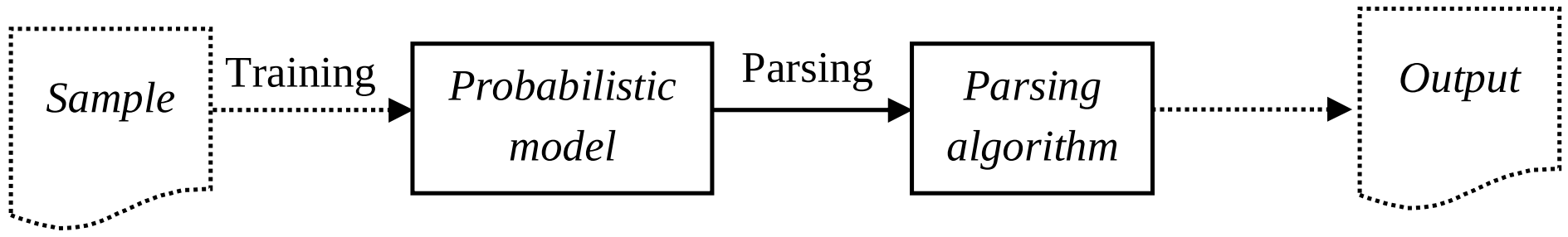

**Figure 4-2.** The components of a probabilistic parser. The sample consists of sentences in language *L*. It usually consists of annotated sentences obtained from a treebank. The probabilistic model defines possible analyses for sentences in language *L*. The parsing algorithm defines the analyses for the sentences of a given text in *L*, relative to the parsing model.

There are three phases in the development of a probabilistic parsing model. In the *parametrization* phase, the method for defining the probabilities of analyses has to be decided (Hockenmaier 2003). In the *training* phase, the probability distributions have to be instantiated with a sample. Finally, a method for measuring the quality of a particular model has to be chosen. The *model* *evaluation* is typically carried out in the following way. The sample is split into three sets: a training set, a development set, and final test set. The parameters of the model are estimated on the *training set*, tuned on the *development set* and tested on the *final test set* which contains unseen data.

A probabilistic parsing algorithm assigns probabilities $p(\pi|s)$ to each parse $\pi$ of the sentence $s$ and returns the most probable parse, i.e. the one that maximizes $p(\pi|s)$ (Charniak 1996):

$$P(s) = \arg\max_{\pi} \frac{p(\pi, s)}{p(s)} = \arg\max_{\pi} p(\pi, s) \tag{4-1}$$

The probability models that estimate $p(\pi|s)$ directly are referred as *conditional* (Hockenmaier 2003). In such models the probability has to be defined for every sentence in the language. This makes them difficult to apply in practice. By contrast, the assumption in *generative models* is that a parse tree is generated by a probabilistic process and that the probability of the parse can be expressed in terms of the individual steps of the process.

In addition to defining the probabilities for parses, it is possible to use probabilistic information for speeding up the parsing process by ordering and pruning the search space. The purpose of this is to enable the algorithm to find the best parse more quickly while at the same time not compromising the quality of the results that are produced. It is possible to select the most probable derivation of a sentence, for example, by *Viterbi optimization* (Viterbi 1967). The idea here is to eliminate the sub-derivations with low probabilities bottom up. One can also use a *beam search* strategy, in which only the best *n* partial parses are being tracked, for pruning the chart edges. *Best-first* parsing methods consider the most likely constituents first (Caraballo & Charniak 1998). A best-first probabilistic chart parser attempts to find the most likely parses by adding constituents to the chart in the order in which they are most likely to appear in a correct parse.

There are several ways to boost the performance of a probabilistic parser. In *voting methods,* such as those devised by Henderson and Brill (1999), predictions from several parsers are combined. Henderson and Brill, for example, compared two models: *parse hybridization,* which considers each constituent in isolation, and *parser switching,* in which one parse is selected for the whole sentence.

*Data-oriented parsing* (DOP) is an extreme type of probabilistic parsing that applies a treebank directly as a probabilistic grammar (Bod 1998, Bod *et al.* 2003). DOP operates by deconstructing the representations that are given for training into fragments and then reconstructing those pieces for analyzing new sentences. An input sentence is parsed by combining treebank subtrees by means of a *node substitution operation*, which closely resembles that of TAGs.

The probabilistic parsing approaches that were introduced above have much in common. They all begin with a relatively small knowledge of the language and gather statistics from the training corpus. Probabilistic parsing models can be compared for example on the basis of the scope of the statistical dependencies that they use. While most state-of-the-art probabilistic parsers are based on models of word-word dependencies, DOP takes into account all observable fragments in the training data.

### *4.3.2 Lexicalized parsing*

Most state-of-the-art parsers use lexicalized grammar formalisms. Lexicalized parsing approaches have therefore been a topic of wide interest during the previous decade. The most obvious implication of a lexicalized approach is the use of bottom-up parsing. It makes sense to start the search process from the words themselves to obtain lexical information in the first stage of the parsing.

A major processing advantage of lexicalized grammars over non-lexicalized ones is that there is no need to search the grammar as a whole: it only needs to search the grammatical information indexed by each of the words. A consequence of this is that increasing the size of a grammar does not necessarily slow down the processing – provided that the increase in size is caused by the addition of new words rather than increased lexical ambiguity.

Probabilistic and lexicalized approaches are often combined. To a non-lexicalized probabilistic parser, an input sentence is really just a list of POS tags and nonterminal nodes. The main advantage of non-lexicalized parsers is that the small terminal alphabet makes model training easier and less prone to data sparseness (Dubey 2005). The consequence of this is that neither computational efficiency nor smoothing are critical issues in non-lexicalized probabilistic parsing. It has however been evident from the time of the earliest probabilistic parsers such as those of Magerman (1994), that lexicalizing the probabilistic model typically helps one to obtain more precise models.

#### *4.3.3 Finite-state parsing*

Syntactic structures in FS parsing are modeled with graph representations (Oflazer 2003). FSM parsing techniques are based on the building of a larger system from smaller FSMs by combining and transforming those by using intersection, composition, determinization, and minimization algorithms (Ginter *et al*. 2006).

The following several approaches have been devised. *Pure FS parsing* combines a set of FSMs to a single FSM (Yli-Jyrä 2004b). In contrast to this, *extended FS* approaches use FS devices as basic components and combines them in such a way that the FS nature of the whole system is not necessarily preserved (Oflazer 2003, Roche 1997). Another commonly used approach is to create an *FS approximation* of the original grammar. CFGs, for example, can be approximated with FS grammars, which are then processed by methods that are efficient for such grammars.

The applicability of FS parsing to natural languages has been questioned for example by Chomsky (1957). He argued that because natural languages are non-regular, they cannot be modeled by FS machinery. One of his arguments was that unbounded *self-embedding* structures require unbounded memory. But recent studies have suggested that there is an absolute limit on *center-embedding* and self-embedding. An FS approach would consequently indeed be applicable to natural languages (Yli-Jyrä 2004a, Karlsson 2006). Yli-Jyrä (2005) claims that while non-FS frameworks are useful for modeling tree locality and co-occurrence constraints, FS grammars are at their best when they are used to approximate computationally expensive formalisms. FS methods have also been found to be especially useful for shallow parsing.

## 4.4 Examples of Grammar Formalism-specific Algorithms

This section introduces techniques that are applied in parsing specific grammar formalisms, and it offers examples of parsers for these formalisms.

#### *4.4.1 Parsing PCFGs*

*Head-driven statistical parsing* (Collins 1996) extends the basic PCFG model by lexicalization. Collins has extended this original framework by introducing what are generally known as the Collins Models 1, 2 and 3 (Collins (1997, 2003)). Model 1 is the baseline generative model based on (Collins 1996). Model 2 makes a distinction between complements and adjuncts and has parameters that correspond directly to the probability distributions over subcategorization frames for head-words. Model 3 adds a probabilistic treatment of a type of LDD, wh-movement.

Another well-known example of the use of a lexicalized PCFG is the parser by Charniak (1996, 1997, 2000). Like Collins, Charniak uses the Markov process for rule generation. This approach is based on an ME model. Charniak's model uses a richer set of features than Collins's models.

### *4.4.2 Parsing UGs*

Two basic UG parsing approaches can be distinguished. In *pure unification parsing*, such as that of Shieber's (1985b) chart parsing algorithm, there is no CF backbone. The parsing is based purely on feature unifications. In the case of a *UG with a CF backbone*, the parsing is driven by the backbone, and the appropriate unifications are carried out (for example, Carroll (1993)).

Any parsing strategy that is valid for CFGs is also valid for UGs. Bottom-up parsing is a more attractive approach for UGs than a top-down one because of its lexical element-driven nature (Bouma & Van Noord 1993, Oepen & Carroll 2000). Most of the recent work on UG parsing has therefore concentrated on purely bottom-up techniques. The efficiency of a unification-based parser depends to a great extent on the efficiency of the unification operation.[46] There are two crucial decisions that need to be made: the way the unification operation is implemented and the way in which the unifications are timed (Placeway 2002).

### *4.4.3 Parsing TAGs*

Constructing a derivation in TAG and LTAG requires the following two steps. Firstly, each word in the input sentence is assigned a set of trees. Secondly, the trees are combined to produce a derivation from which a parse tree can be constructed (Sarkar 2002). Some parsers, such as those of Joshi and Schabes (1997), use TAGs directly. Other parsers, such as those of Vijay-Shanker and Weir (1993) and Schabes and Shieber (1994), transform a TAG into an equivalent *linear indexed grammar* and use that for parsing.

---

[46] Roughly 90 per cent of the CPU time in parsing, when using a large-scale UG, is spent on feature structure unifications (Malouf *et al*. 2000).

Supertagging (Section 4.2.2) was originally introduced for TAGs, and it has proved to be a successful approach. LTAG parsing by supertagging involves the following two steps. Firstly, an appropriate set of supertags is selected for each word in the input (Srinivas *et al.* 1996). Secondly, the parser uses simple heuristic rules to link the supertags in an appropriate way to produce complete parses.

#### *4.4.4 Parsing DGs*

D trees contain one node per word, and the task of a D parser is to connect these nodes. A wide variety of techniques have been applied to D parsing. Some parsers, such as the ones of Lombardo & Lesmo (1998a) and Eisner (1996a), use dynamic programming algorithms similar to those applied for CFGs. Others regard parsing as a constraint satisfaction problem.

In contrast to most other formalisms which are parsed in a *constructive* way by building structural descriptions out of elementary blocks according to a rule system, CG and FDG follow an *elimination* approach (Voutilainen & Heikkilä 1993, Voutilainen 1994). The analysis is based on disambiguation by constraints.[47] Non-projective DGs have been successfully parsed using probabilistic data-driven methods, for example, by McDonald *et al*. (2005) and Nivre *et al.* (2004).

XDGs have been parsed by axiomatizing valid D graphs by finite set constraints. In this approach parsing is reduced to *finite set constraint programming* (Debusmann *et al*. 2004a, Duchier 1999). In contrast to the generative approach in which parses are build up by combining smaller ones, constraint programming sets global well-formedness conditions for the sentence structures.

#### *4.4.5 Parsing LG*

Sleator and Temperley (1993) use a dynamic programming algorithm for building up LG linkages. A major efficiency increase is obtained by post-processing that deletes connectors that cannot form linkages. Grinberg *et al*. (1995) proposed a modification that allows for parsing spoken language transcripts. The algorithm uses *null links* to allow connections between any pair of adjacent words, regardless of their definitions in the lexicon. In a recent work, Ginter *et al.* (2006) applied FS parsing to LGs. They concluded that while LGs can be parsed by FSMs, this approach is inefficient in practice because of the computation of the intermediate results.

[47] CG parsing proceeds as follows. Firstly, all possible POS and morphological tags (i.e. readings) are provided for each word in the sentence (Voutilainen & Heikkilä 1993). Secondly, the constraints eliminate tags that are inconsistent with the context.

### *4.4.6 Parsing CCG*

Because CCG derivations are binary trees, standard chart parsing techniques can be applied. The main challenge in CCG parsing is how to control spurious ambiguity. Spurious ambiguity is caused by the properties of CCGs, which allows the generation of an exponential number of derivations for a constituent.

In her search for a suitable probabilistic parsing framework for CCGs, Hockenmaier (2003) considered Charniak's (1999), Collins' (1997), Goodman's (1997) PCFG models, and the DOP models of Bod (1998). She found that those of Collins (1997) and Goodman (1997) were best suited to the purpose. She also proposed a model that captures LDDs by expressing word-word dependencies of the PAS. This model allows multiple dependencies between a word and its context, including dependencies that arise through control, right node raising and other LDD constructions. Another probabilistic approach to CCG parsing is introduced in Clark *et al*. (2002) and in Clark and Curran (2004a). The main difference between these and Hockenmaier's parser is that instead of being a generative model, Clark and Curran (2004a) apply a conditional probability model similar to that of Collins (1996). Clark and Curran also use a ME-based supertagger prior to parsing.

## 4.5 Computational Complexity of Parsing

The structure of the search space of a parsing algorithm is defined by the grammar. The properties of the grammar and the parsing algorithm together define the computational complexity of a parser. The computational complexity of an algorithm describes the rate at which it consumes time and space (Ristad 2003). It is characterized as the order of the growth of a function in the size of the input, typically by means of the upper bound of the resource requirements.

### *4.5.1 Efficiently parsable formalisms*

Let $n$ be the number of words in a sentence. The complexity of parsing CFGs with dynamic programming approaches, such as those of Earley and CYK, remains in $O(n^3)$ in the general case, but are more efficient with specific types of grammars. The Earley algorithm, for example, has the time complexity $n^2$ for certain types of grammars. Table 4-1 sets out worst-case time and space complexity figures for certain parsing algorithm - grammar formalism pairs for which a polynomial time algorithm is known.

**Table 4-1**. The worst-case time and space complexities for algorithms parsing a specific grammar formalisms reported in literature. The data is compiled from Carroll (1993), Covington (1990), Díaz *et al*. (2002), Eisner (1996a), Grinberg *et al*. (1995), Joshi (1998), Kipps (1989), Lomardo and Lesmo (1998a), Perrault (1984), Satta (1994), Schabes and Joshi (1988), Stolcke (1995), Sikkel and Op den Akker (1993), Tapanainen (1999), Yoshinaga *et al*. (2003), Van Noord (1994), and Vijay-Shanker and Weir (1993).

| Algorithm | Formalism | Time | Space |
|---|---|---|---|
| *Earley* | CFG | $n^3$ | $n^2$ |
| | PCFG | $n^3$ | $n^2$ |
| | TAG | $n^6$ | $n^6$ |
| | CDG | $n^3$ | - |
| *CYK* | CFG | $n^3$ | - |
| | PCFG | $n^3$ | - |
| | Level-2 MCSGs | $n^6$ | $n^4$ |
| | CDG | $n^3$ | - |
| *LR* | CFG | $n^3$ | $n^3$ |
| *LC* | TAG | $n^6$ | - |
| *HC* | CFG | $n^3$ | $n^2$ |
| | TAG | $n^6$ | - |
| *Elimination (Tapanainen 1999)* | CG | $n^3$ | - |
| *Sleator & Temperley (1993)* | LG | $n^3$ | - |
| *Grinberg et al. (1995)* | Probabilistic LG | $n^3$ | - |

There are, apart from the length of an input sentence, other factors that affect computational complexity. Sarkar *et al*. (2000), for example, point out that, for fully lexicalized grammars[48] such as LTAG, the sentence length is not the only, or even the dominant, factor affecting parsing complexity. They describe how syntactic lexical ambiguity and clausal complexity also indicate complexity. A word typically selects more than one syntactic structure. This *syntactic lexical ambiguity* is a better indicator of parsing complexity in LTAG than the sentence length. The *clausal complexity* of a sentence indicates the number of clauses it contains. As it increases, the number of decisions about how to link the clauses with one another increases accordingly.

[48] A grammar in which each lexical item is associated at least one syntactic structure.

#### *4.5.2 Intractable problems*

The problems in the classes *NP-complete*, *NP-hard* and *EXPPOLY* are intractable Ristad (1985, 1986). It is widely believed – although it has not been proved – that no fast methods of solution can ever be found for these problems. The recognition problems of some of the grammar formalisms analyzed in this work belong to the intractable classes (Table 4-2).

**Table 4-2**. The complexities of recognition problems of the grammar formalisms in intractable classes.

| Type | Formalism | URP | Source |
|---|---|---|---|
| *UG* | *GPSG* | EXPPOLY-hard/NP-complete* | Ristad (1985, 1986) |
| | *HPSG* | Undecidable/NP-complete** | Kepser and Mönnich (2003) & Trautwein (1995) |
| | *LFG* | NP-complete | Barton *et al.* (1987) |
| *DG* | *Non-projective DG* | NP-complete | Neuhaus and Bröker (1997) |
| | *XDG* | NP-hard | Debusmann & Smolka (2006) |

*Depending on the version used. **Trautwein (1995) showed that when certain restrictions are set on a HPSG, the complexity of URP is NP-complete.

While the theoretical upper bounds of the CF filtering approaches to LTAG and HPSG (Torisawa *et al*. 2000, Oouchida *et al*. 2004) remain in the same class as their non-filtering counterparts, experiments show that the practical parsing efficiency often is much better.

#### *4.5.3 Analysis*

There are some important points to note about the time complexity figures. Firstly, the worst-case complexities are not directly useful for evaluating the practical time complexities of parsing algorithms (Joshi 1998). As Covington (1990) points out, the worst-case parsing complexities rarely materialize. Covington (2001) claims that human language does not use unconstrained grammar, and that human beings do not use sentences that would put any reasonable parsing algorithm into a worst-case situation.

There are, however, no theoretical average case results available for most algorithms and formalisms; all we have are ones that are based on empirical experiments. Tapanainen (1999), for example, approximated the average complexity of CG parsing as O($n$ log $n$). Average-case complexities can in some cases be improved. The average-case complexity of a DG parser, for example, can be reduced by placing an arbitrary limit on the maximum distance between the current word and the potential head and dependent (Covingtion 1990). Although the limit will reduce the complexity significantly, Covington claims that it will not affect the correctness of the parser because massive word order inversions and the wide separation of constituents are very rare or forbidden in natural languages.

Secondly, parsing algorithms are often claimed to be grammar formalisms-independent. This may often be true in the sense that they can be used for parsing many different types of grammars. The grammar formalism, however, typically exerts an enormous influence on the efficiency of the parsing algorithm. An algorithm that works best for one grammar formalism may be the most inefficient for some other formalism. Sarkar *et al*. (2000), for example, assert that the theoretical upper bound did not exert any significant affect on performance in LTAG parsing, but that the dominant factor was the lexical ambiguity.[49]

Thirdly, complexity results are only partly relevant for probabilistic parsers since such parsers are usually not based on a grammar but rather on statistical inference from treebank annotations (Nivre 2006). While the recognition problem of non-projective DGs is NP-complete, for example, the probabilistic non-projective D parser by McDonald *et al*. (2005) has the time complexity $O(n^2)$.

Moore (2000) and Van Noord (1997) have reported practical efficiency evaluations using different pairs of grammars and algorithms. One cannot directly compare the conclusions that they draw because they ran the tests on different machines and grammars. Not only that but Moore (2000) used English test data and Van Noord used (1997) Dutch test data. One may nevertheless draw some conclusions from these experiments. In both experiments, LR performed worst on average while LC was more efficient than the Earley algorithm in most cases. The experiments also emphasize the fact that while parsing algorithms are most commonly designed to be general, their performance varies considerably in accordance with the grammar. The grammar also affects the relative space usages of the algorithms.

In order to arrive at a practical point of view of the space complexity of parsing, let us consider an example. A feature structure build by a parser using the LinGO HPSG grammar contains on average 300 internal nodes (Oepen *et al.* 2000), and each node is approximately 80 bytes in size. The parser executes over 4,000 unifications per sentence on average. The total memory processing therefore corresponds to almost 100 MBs.

[49] The number of possible lexical entries for the same word in a sentence.

The complexity figures given above do not include morphological analysis, which is also a complex task. Barton *et al.* (1987) demonstrated that the TWOL morphological analysis (both recognition and generation) is NP-hard. This is caused by backtracking in processing both the lexicon and the TWOL rules. Fortunately, as Koskenniemi and Church (1988) point out, the worst-case behavior rarely, if ever, materializes with *natural* languages. Koskenniemi and Church argue along similar lines to those argued by Covington (above) when they note that the exponential behavior does not occur in practical applications since natural languages avoid using constructions that could trigger any worst-case behavior.

It may therefore be concluded that it is difficult to state anything on the practical efficiency of a parser on the basis of only the theoretical computational complexities of the grammar and the algorithm. One also needs to bear in mind that there is no analytical technique that allows one adequately to characterize grammar complexity in a particular setting and to predict the most efficient parsing strategy on the basis of a given grammar.

Moreover, one may state that research on efficiency in parsing has two main goals. The first is to identify linguistically significant subclasses of grammars with low theoretical complexity, and the second is to find parsing approaches and algorithms that are efficient in practice. On the basis of the findings above priority should be given to the latter goal. These findings emphasize the need for finding an empirical means of comparing different approaches and identifying the best ones for a given parsing problem.

## 4.6 Conclusion

This section analyzes the algorithms and techniques that are applied in parsing the grammar formalisms surveyed in Chapter 3. Table 4-3 provides an overview of the techniques used in parsing the grammar formalisms represented in Section 3. While this table is not exhaustive, it nevertheless offers a comprehensive view of approaches that are applied in the parsing of each of the formalisms. Since many of the algorithms shown in the table are lexicalized, lexicalized parsing is not represented in a separate section in order to avoid duplicating items.

**Table 4-3.** Summary of parsing approaches and algorithms applied to the grammar formalisms analyzed in Chapter 3.

| | | PCFG | UGs | TAGs | DGs | LGs | CCG |
|---|---|---|---|---|---|---|---|
| *Fundamental algorithms* | | | | | | | |
| | *Earley* | Stolcke (1995) | Shieber (1985b) | Schabes & Joshi (1988), Nederhof (1999) | Lombardo & Lesmo (1996) | - | - |
| | *CYK* | Ney (1991) | Haas (1987), Tsuruoka *et al*. (2004) | Vijay-Shanker & Weir (1993), Chiang (2000) | Eisner (1996a), Lee & Choi (1997) | - | Hockenmaier (2003) |
| | *Shift-reduce /LR* | Carroll (1993) | Briscoe & Carroll (1993) | Schabes & Vijay-Shanker (1990) | Yamada & Matsumoto (2003), Nivre *et al*. (2004) | - | Villavicencio (1997) |
| | *LC* | Manning & Carpenter (1997) | Haas (1989) | Díaz *et al*. (2002) | - | - | - |
| | *HC* | - | Van Noord (1997) | Van Noord (1994), Sarkar (2000) | - | - | - |
| *CF filtering and supertagging* | | | | | | | |
| | *CF filtering* | - | Torisawa *et al*. (2000), Matsuzaki *et al*. (2007) | Poller & Becker (1998), Oouchida *et al*. (2004) | - | - | - |
| | *Super-tagging* | - | Ninomiya *et al*. (2006), Matsuzaki *et al*. (2007) | Srinivas & Joshi (1999) | Foth *et al*. (2006) | - | Clark (2002), Clark & Curran (2004b) |
| *Parsing approaches* | | | | | | | |
| *Finite-state* | *Pure* | - | - | - | Koskenniemi (1990) | - | - |
| *Finite-state* | *Extend.* | - | - | Roche (1997, 1999) | Oflazer (2003) | - | - |
| *Finite-state* | *Approx.* | Mohri & Nederhof (2001) | Johnson (1998) | - | Yli-Jyrä (2004b, 2005) | Ginter *et al*. (2006) | - |
| *Probabilistic* | *Genera-tive* | Charniak (1996, 2000), Collins (1997, 2003) | Cahill (2004) | Chiang (2000) | Eisner (1996a), Dienes *et al*. (2003) | Lafferty *et al*. (1992) | Hockenmaier (2003) |
| *Probabilistic* | *Condi-tional* | Johnson (2001) | - | - | - | - | Clark *et al*. (2002), Clark & Curran (2004a) |
| *Probabilistic* | *ME* | Charniak (2000) | Ninomiya *et al*. (2006) | - | Cheng *et al*. (2005) | - | Clark & Curran (2004a) |
| *Probabilistic* | *Inside-outside* | Goodman (1996) | Briscoe & Carroll (1993) | - | Lee & Choi (1997) | Lafferty *et al*. (1992) | - |
| *Probabilistic* | *DOP* | Goodman (1996) | Bod *et al*. (2003), Neumann (2002) | Neumann (1998) | - | - | - |

I have used two criteria, *expressiveness* and *computational efficiency*, to compare grammar formalisms for parsing. While the grammar formalism of a parser needs to be sufficiently expressive for describing the language that is being parsed, the formalism needs to be simultaneously parsable within a reasonable timeframe. I have represented the results of the analysis in Table 4-4.

**Table 4-4**. A summary of the analysis in this chapter.

| | **Formalism** | **Expressiveness** | **Efficiency, O()** | |
|---|---|---|---|---|
| | | | **Parsing** | **Recognition** |
| CF | CFG | CF | $n^3$ | P |
| | PCFG | >CF | $n^3$ | P |
| UG | GPSG | CF | - | NP-complete, EXPPOLY |
| | HPSG | Unrestricted | - | Undecidable/ NP-complete |
| | LFG | Context-sensitive | - | NP-complete |
| TAG | TAG | MCSG | $n^6$ | $n^6$ |
| | PLTAG | MCSG | $n^6$ | - |
| | LTAG | MCSG | $n^6$ | - |
| DG | CG | <Regular | - | - |
| | FDG | >CF | - | NP-complete* |
| | XDG | Unrestricted | - | NP-hard |
| LG | LG | CF | $n^3$ | - |
| | Prob LG | CF | $n^3$ | - |
| CatC | CCG | MCSG | $n^6$ | - |

*While there is no proof of the time complexity of FDG recognition, Neuhaus and Bröker (1997) have showed that the recognition of non-projective DG is NP-complete. Thus the time complexity of FDG recognition is at least NP-complete.

If one takes into consideration the tradeoff between expressiveness and efficiency, then one may judge MCSGs to be the most attractive class of grammars. While they are powerful enough for describing natural languages, they remain parsable in polynomial time even in the worst case. But, as I have already noted above, the theoretical upper bounds for parsing and the recognition of a certain grammar formalism may have a little effect on parsing efficiency in practice. Practical evaluations are needed for determining the efficiency of an algorithm/formalism pair. Such evaluations are provided in Chapter 10.

# 5 Parsing: Problems and Solutions

This chapter introduces the problems that syntactic parsers have to face and describes solutions to these problems (Section 5.1). I will close the chapter in Section 5.2 by comparing the two main parsing approaches – rule-based and probabilistic.

## 5.1 Problems in Parsing and Their Solutions

At the time of writing (2007), no domain- and genre-independent natural language parser capable of producing error-free parses for running text has ever been devised. The following factors make it difficult to parse unrestricted text (Leech *et al*. 1996, Briscoe 1998):

1. Ambiguity is an inherent property of all natural languages, and it affects parsers on the level of lexicon and grammar. Linguistic expressions taken out of context are ambiguous and incomplete. Parsing a sentence often thus results in more than one analysis for the input sentence.
2. Sentences in free text are often long and contain several clauses and phrases. This causes the number of possible analyses to grow exponentially.
3. Because parsers have no world knowledge, they have to rely solely on whatever information they can derive from linguistic rules.
4. It is necessary for a grammar with a broad coverage[50] to be extensive. This makes it difficult to achieve consistency in the rules and lexicon.

Apart from ambiguity (Section 5.1.1), both undergeneration and overgeneration (Section 5.1.2) cause problems for parsers. *Undergeneration* refers to a situation in which no analysis is generated for a grammatical sentence. *Overgeneration* means that a parser will produce parses for ungrammatical sentences. Other parsing issues covered in this chapter include the problem of dealing with ill-formed input (Section 5.1.3), the role of semantic information in syntactic parsing (Section 5.1.4), the problem of defining the relation between the grammar, and the processing component (Section 5.1.5).

[50] A grammar has broad coverage if it's able to produce a parse for a high proportion of input sentences representing several different text genres.

### *5.1.1 Ambiguity*

While human beings have the ability to resolve most cases of ambiguity, the disambiguation of even a simple case of ambiguity can be problematic for a parser. This is made even more problematic by the fact that types of ambiguity combine in complex ways. Parsers tend to run into problems when they analyze long sentences because the probability that they will encounter more than one case of ambiguity increases in proportion to the length of the sentence being parsed.

*POS ambiguity* is a particular problem in lexicalized grammars because a parser using such a grammar selects multiple structures for each POS analysis assigned to a word (Srinivas *et al*. 1995). Dalrymple (2006), in one of the very few pieces of research undertaken in this area, reported on the effect of POS tagging and POS ambiguity on a parser's performance. Dalrymple's test set contained a total of 2,105 sentences. The sentences were assigned 429 parses on average. 29.5% of the 2,105 sentences had parses whose POS tag sequences were identical. These sentences had 7.2 parses on average. Dalrymple concluded from these results that accurate POS tagging would not help to disambiguate these 29.5% of the sentences– but that it would help with the remaining 70.5%. Dalrymple estimated the degree of overall ambiguity reduction that could be obtained if the tagging were to be performed by a "perfect tagger" producing the correct tag sequence for each sentence. The results indicated that 45–50% of the potential parses for a sentence could be ruled out by choosing the correct tag sequence.

*Structural ambiguity* originates from a grammar assigning more than one analysis to a sentence. Two distinct types of structural ambiguity can be identified. A *globally ambiguous* sentence can be interpreted, as a whole, in more than one way. A *local ambiguity* affects only a part of a sentence (Gazdar & Mellish 1989). A particular type of ambiguity occurs in so called *garden path* sentences. While such sentences may be correct from a grammatical point of view, they can easily be misunderstood. Even a human language processor may misresolve and fail to analyze grammatical garden path sentences. This phenomenon is typically caused by local ambiguities (Crain & Steedman 1985).

**Example 5-1.** Globally ambiguous sentence A can be interpreted semantically in at least four different ways. The most probable reading is that a female avoids balls flying overhead by ducking. The other interpretations involve odd scenarios featuring a bird being constructed and a woman flying. Although the locally ambiguous sentence B is not ambiguous as a whole, if the three last words are examined in isolation, one can come up with the interpretation that Liverpool sold Everton – even though the sentence as a whole does not make such proposition. Sentence C is a garden path sentence.

*A) Flying balls made her duck.*
*B) The company that bought Liverpool sold Everton.*
*C) The referee who whistles tunes the whistle.*

Structural ambiguity has multiple causes (Jurafsky & Martin 2000). *Coordination ambiguity* is caused by a situation in which different sets of phrases can be conjoined. The phrase "accurate shots and crosses", for example, can be interpreted so that "accurate" modifies either "shots" or "shots and crosses". In *attachment ambiguity*, a constituent can be attached to more than one place in a sentence. Resolving attachment ambiguities correctly requires the use of several sources of information. In the sentence "The coach saw the player with the telescope", we have an example of a common type of attachment ambiguity, namely *prepositional phrase* (PP) attachment ambiguity (Collins 1999).[51] This sentence may be analyzed in at least two possible ways – the PP "with the telescope" modifies either "coach" or "saw", and this leads to the analyses shown in Figure 5-1.

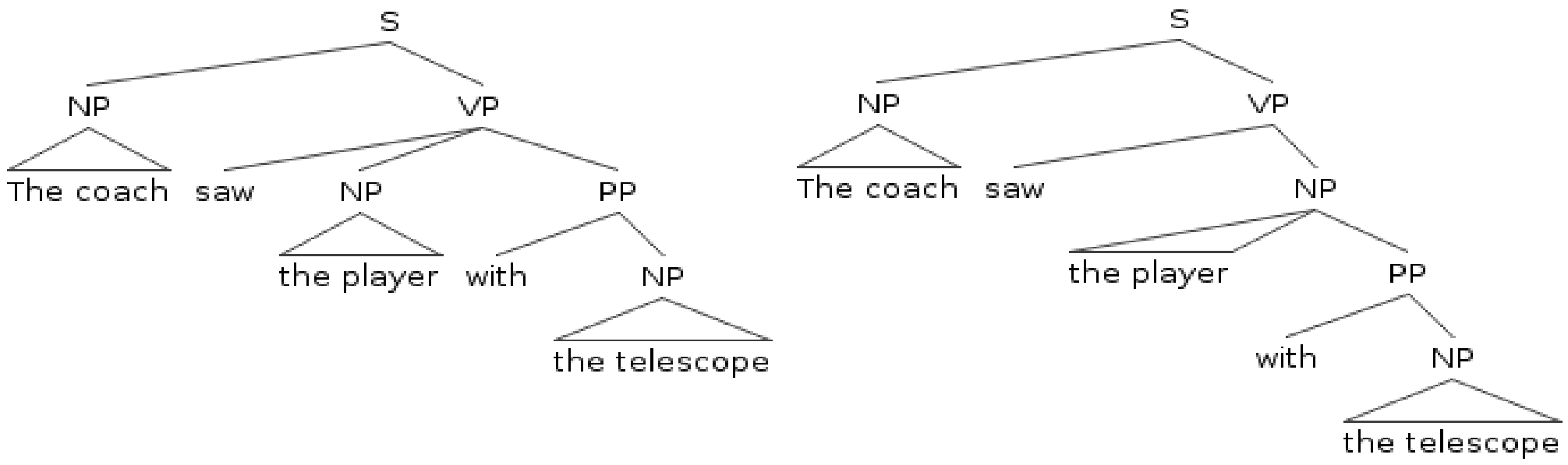

**Figure 5-1.** Two parse trees for the ambiguous sentence "The coach saw the player with the telescope."

*Disambiguation* is the process of resolving ambiguities (Earley 1970, Winograd 1983, Srinivas *et al*. 1995, Nivre 2006). A parser has to choose the correct one from a set of possible parses for a sentence. There are several different ways of dealing with ambiguity in syntactic parsing. The first and most straightforward approach is to let the parser look for all the possible parses that are derivable from the grammar and then choose the correct one. The second approach is simply to return the first parse that is found. In such an approach the search ought to be guided towards identifying the most plausible parses first. The third approach involves designing a grammar so that it returns one analysis for any grammatical sentence. The ambiguity of a grammar can be reduced, for example, by tailoring it to a specific text genre.

[51] The number of possible analyses for a sentence with PPs follows the Catalan number (Church & Patil 1982). A sentence with three embedded PPs, for example, produces a total of five possible parses, while a sentence with seven PPs produces a total of 469 possible parses.

The fourth approach involves performing POS disambiguation prior to syntactic analysis so that the number of potential parses is substantially reduced. This approach is especially important with lexicalized grammars. The fifth approach involves *tree filtering*. This reduces the set of possible parse trees by eliminating parses that cannot lead into a valid analysis on the basis of their structural properties. Such filtering may be based either on rules or probabilities. The sixth approach is an elimination approach such as CG and FDG (which was discussed in Chapters 3 and 4). It integrates parsing and disambiguation. Finally, semantic information can be integrated into the process of syntactic disambiguation (see Section 5.1.4).

### *5.1.2 Under- and overgeneration*

In addition to genuine cases of ambiguity such as those discussed above (in which the grammar *should* assign several plausible analyses to a sentence), a grammar may assign analyses that are never encountered in the language being parsed (Nivre 2006). This problem is referred to as overgeneration (or grammar leakage). In undergeneration, which is the opposite phenomenon to overgeneration, a parser is not able to analyze a sentence which belongs to the language. This is usually caused by a gap in the coverage of the grammar. Figure 5-2 illustrates the two concepts.

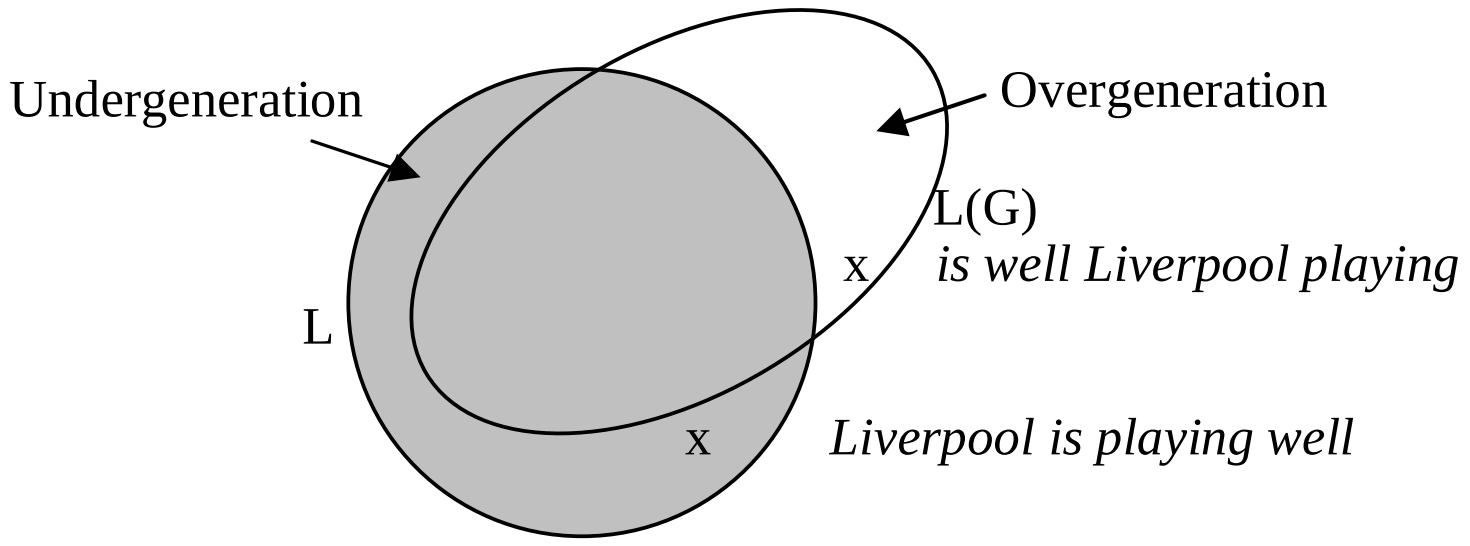

**Figure 5-2.** Under- and overgeneration (adapted from Dörnenburg 1997). *L* is the language being parsed, and *L(G)* is the language generated by the grammar *G*. If a parser *P* fails, for example, to analyze the sentence "Liverpool is playing well", it is undergenerating. If P produces an analysis for the sentence "is well Liverpool playing", it is overgenerating.

The concepts of under- and overgeneration are tightly connected to the dilemma of coverage and the fuzziness of grammaticality. A grammar that is rich enough to cover a natural language, including rare sentence constructions, often fails to distinguish natural from unnatural interpretations. On the other hand, a grammar that is restricted enough to exclude all ungrammatical sentences typically fails to accommodate all the grammatical ones.

### *5.1.3 Ill-formed input*

Parsers are often presented with texts that contain errors. While processing user inputs a parser may therefore encounter misspelled words, incorrect cases, missing or extra words, or dialect variations (Foster 2004). Transcriptions of *spoken* language texts are especially likely to contain such errors and complications. A parser's ability to produce an error-free or only a slightly altered output from input sentences containing errors, is referred to as *robustness*. A robust parser is able to provide as complete and correct analysis of the input sentence as it is capable of doing under the circumstances.

Probabilistic approaches to parsing are inherently robust because they consider all possible analyses of a sentence and usually propose a parse for any given input. Robustness can be added to a rule-based parser in several ways (Nivre, 2006, Menzel, 1995). The first way is by relaxing the constraints of the grammar in such a way that a sentence outside the language generated by the grammar can be assigned a complete analysis. The second way is by getting a parser to try to recover as much structure as possible from the well-formed fragments of an analysis when a complete analysis cannot be performed. The third way is by identifying a number of common mistakes and integrating them into the grammar in anticipation of such errors occurring in texts. This method is limited to few predictable high-frequency kinds of errors such as common spelling mistakes and high-frequency errors of word order.

#### *5.1.4 The role of semantic information*

A purely syntactic approach to language assumes that it is a collection of syntactic expressions (Gazdar *et al*. 1985). However, the reason why natural languages exist at all is that human beings associate meanings with these expressions. Psycholinguistic research has shown that the human brain processes language incrementally by using information extracted from several different levels of language (Pollard 1996). The human brain continuously integrates information about syntax with semantic, pragmatic and world knowledge, and even with probabilistic data. Human beings are consequently well endowed with highly developed skills to process natural languages. They are capable, for example, of understanding ill-formed sentences – even when they contain several types of errors simultaneously (Menzel 1995). But parsing systems encounter serious problems when they are confronted even with a single kind of distortion. This is partly attributable to the fact that most parsers rely solely on syntactic information to guide the parsing process.

Many linguists are of the opinion that meaning and structure are, in principle, independent of one another. Tesnière (1959), for example, claims that while syntactic structure follows from semantic structure, the contrary case is not valid. Chomsky (1957) presented the following, now widely known, examples to emphasize that syntax is autonomous of semantics. Even though both sentences are perfectly grammatical, their meanings are nonsensical.

*Colorless ideas sleep furiously.*
*Golf plays John.*

Models that are based on autonomy assume that the description of a linguistic expression must refer to only one level of representation, namely syntax. If one follows autonomy rules, one might claim that the assignation of a meaning should not determine the grammaticality of a sentence. In GPSG (Gazdar *et al*. 1985), for example, semantics is not allowed to act as a post-syntactic filter; nor is it allowed to interact with syntactic parsing decisions. It can be argued that although the sentence “Colorless ideas sleep furiously”, for example, is not semantically coherent, it is nevertheless perfectly grammatical. “This statement is false” is another example of a grammatical sentence that is impossible to make sense of.

The autonomy of syntax does not mean that it exists in isolation. Syntax cannot be reduced to semantics (Abney 1996). In contrast to autonomy, some theories of syntax are based on the assumption that no level of grammatical knowledge is privileged with respect to others and that no level is derived from any other – even though the levels mutually constrain one another.

The notion of *plausibility* refers to assessment of the sense of a particular sentence based on a general human understanding of the world (Crain & Steedman 1985). Crain and Steedman suggest the following method for formulating the principle of plausibility:

> "If a reading is more plausible in terms either of general knowledge about the world, or of specific knowledge about the universe of discourse, then, other things being equal, it will be favored over one that is not".

Crain and Steedman state that in a case where there is a conflict between general and specific knowledge, the latter must take precedence. But the use of plausibility constraints of this type in syntactic parsing can be problematic because the reason why natural languages have syntax at all is presumably because real-life events frequently contradict the expectations stated by such constraints.

Using semantic knowledge to guide syntactic parsing decisions is not as straightforward a choice as it might intuitively appear to be. One option that is applied by CCG is to follow Montague's (1974) *rule-to-rule* approach to the syntax-semantics interface. This principle states that each syntactic rule is associated with a semantic rule that determines the meaning of that part of the sentence whose form the syntactic rule specifies. If one accepts this point of view, semantic information can be used for filtering syntactic structures (Gazdar *et al*. 1985).

The rule-to-rule approach is an example of a *weak interaction* model in which a parser builds syntactic structures which, in turn, are checked by the semantic component that selects the ones that are semantically plausible (Crain & Steedman 1985, Gorrell 1995, Allen 1995). A parser could, for example, create all the parses that are syntactically correct according to the grammar and use selectional restrictions to discard semantically ill-formed ones. Weak interaction is the model employed by most practical parsing systems that use semantics as part of the parsing process. In contrast to this, *strong interaction* makes use of semantic information to guide initial syntactic parsing decisions. If one's purpose is to generate semantic representations, the most radical approach one could use would be to throw out the syntax completely and build semantic structures directly. In such systems, both grammatical and semantic information is stored in the lexicon.

A successful example of the use of semantic information in parsing is PP and relative clause attachment disambiguation (Charniak 1993). As I have already noted above, some attachment decisions are impossible to make purely on the basis of syntactic information. One approach is to mark a corpus of text with semantic tags and to train a probabilistic model on the data. One may then use the probabilistic model for the semantic role tagging of the PP clauses, and this in turn will act as a guide to making the correct attachment decisions. Another example of a successful application of computational semantics is the *Minimal Recursion Semantics* (MRS) framework, which is suitable for parsing with UGs (Copestake *et al*. 2003). MRS has been applied, for example, to an English-Norwegian MT system that parses Norwegian by using an LFG parser and generates English with an HPSG (Oepen *et al.* 2004).

### *5.1.5 Grammar/parser interface*

There is no evidence of two physically separate neural assemblies in the human language processor, i.e. one that stores grammar rules and another that is an active device for accessing grammar-rules in the course of its operation (Crain & Fodor 1985). From a psycholinguistic point of view, such a division represents only a manner of speaking or a method of dividing up components for purposes of theoretical convenience. Practical parsing systems nevertheless are required to deal with such a division.[52]

Obviously the more ambiguous a grammar is, the more problems and work it causes for the parsing algorithm. In addition to this, a grammar may cause problems for the parser by *concentrating structural complexity* at certain points in word strings. For example, in multiple center-embedded sentences such as “The attacker that the defender tackled kicked the ball”, the density of nonterminals in relation to words is high. The third possible source of complications is the *format and organization* of the grammar itself. The syntactical information may be represented in a way that is not well suited to the nature and sequence of the computations of the parsing algorithm.

[52]The *connectionist* approach to parsing, or the use of artificial neural networks, aims at simulating the mental processing of human sentence parsing (Rumelhart & McClelland 1986, Nakagawa & Mori 1988). Schnelle and Doust (1992), for example, have proposed a neural network structure that implements the Earley algorithm. Connectionist parsing is different from other approaches to parsing because it is intrinsically non-modular: the grammar and search procedure are interwoven. In addition to having an orientation towards modeling the mental processes of human sentence parsing, a further advantage of the connectionist approach is that it allows one to use semantic and contextual information alongside syntactic information in a uniform manner (Rumelhart & McClelland 1986, Nakagawa & Mori 1988). Nevertheless, the successes of such parsers in practice have thus far been modest, and it is for that reason I have not dealt with them in this research.

As already noted in Chapter 3, TGs offer a notorious example of the difficulties that a grammar can cause for the processing component of a parser. There seems to be no way in which a parser can directly use the TG rules: what is required is a reshuffling of the information encoding. By contrast, UGs, for example, treat parsing as a relatively straightforward process. Such a process is based on a grammar that contains rules which can be directly applied to word strings. In the purest form of UG parsing, no other filters or constraints or any other devices are applied.

## 5.2 Rule-based vs. Probabilistic Parsing

The debate in the CL and NLP communities over the rule-based (or symbolic) vs. probabilistic (or machine learning) approaches has been going on since generative grammars first appeared in the 1950s (Klavans & Resnik 1996).[53] Research into probabilistic and corpus-based methods was discouraged by Chomsky's (1957) observations (Geman & Johnson 2002). The research was predicated on TGs and other rule-based approaches. However, as I have noted above, TGs are computationally expensive. Many researchers were consequently drawn to developing grammar formalisms in which the surface structure is generated directly without any separate deep structure or transformations.

The following drawback can be identified in the rule-based approach to parsing (Srinivas *et al.* 1998, Geman & Johnson 2002, Nivre 2006): Firstly, the coverage of such grammars is incomplete because of the labor-intensiveness of manual grammar construction and because of an inadequate understanding of the syntactic constructions that occur in natural languages. Secondly, hand-constructed grammars are prone to generate spurious ambiguities – parses that are accepted by the grammar syntactically, but which are semantically anomalous. Thirdly, the performance of a rule-based grammar may suffer from the lack of text genre-specific knowledge. In order to minimize this problem, some parsers with a rule-based grammar use a heuristical disambiguation component that has been trained on genre-specific texts. Finally, rule-based approaches are typically not as robust as probabilistic approaches.

[53] In parsing, the terms *grammar-driven* and *data-driven* are also used (Nivre 2006). The rule-based approach is sometimes called "deep", by way of the contrast to "shallow" processing methods (Oepen *et al.* 2002). This distinction is actually not sustainable. With state-of-the-art methods it is possible to acquire deep grammars automatically.

The success of probabilistic approaches in speech recognition led some researchers to apply the methods to other NLP applications (Geman & Johnson 2002). The probabilistic approach has been applied, for example, to POS tagging. Some researchers even went so far as to announce that "linguists need not participate" in the development of a probabilistic parser (Magerman 1994). Most probabilistic systems are so designed that every input sentence is assigned at least one analysis – a move that largely eliminates the problem of robustness. Disambiguation may nevertheless become a severe problem because the improved robustness often causes massive overgeneration. Although smoothing may be useful in some cases,[54] automatically induced grammars are usually ambiguous and often overgenerate (Manning & Carpenter 1997). The overgeneration problem in probabilistic parsing is nevertheless compensated for – at least to some extent – by disambiguation and parse ranking mechanisms in the probabilistic model (Nivre 2006).

The binary decisions of grammaticality vs. ungrammaticality of rule-based grammars are replaced in probabilistic grammars with probability distributions. Instead of describing ill-formed structures as impossible, probabilistic approaches assign them a low probability. In addition to this, probabilistic approaches offer a means of distinguishing more plausible interpretations from less plausible ones. The lack of genre-specific knowledge is also often a problem for probabilistic parsers. Statistically induced grammars are trained with a specific treebank. Consequently their performance may become worse when they are exposed to texts from a different domain than the treebank that is used for training (Clegg & Shepherd 2005). It is, however, less demanding to modify a probabilistic parser than a rule-based parser to a new genre provided that a training material (i.e. treebank) representing texts from that new genre is available. If such a resource does not exist, the opposite of what was stated in the previous sentence may be true due to the high costs of developing treebanks. Table 5-1 compares rule-based and probabilistic methods in respect of several key properties.

[54] The training data, for example, allows a compound noun to be modified by four adjectives, but not by a simple noun. Thus the training data might have the phrase “quick, skilled, long, stylish midfield player”, but not “quick, skilled, long, stylish midfielder”. Smoothing would allow for analyzing the latter sentence. These added rules are unlikely to appear in the maximum probability

**Table 5-1.** Comparison of rule-based and probabilistic approaches to parsing.

| | Cost of grammar development | Grammar | Robustness | Level of detail | Coverage |
|---|---|---|---|---|---|
| *Rule-based* | Costly to develop because of the complexity of rule interactions | Perhaps more intuitive | More easily broken by new input or by user errors | Often higher than in probabilistic | Has not been broad enough for practical applications until recently |
| *Probabilistic* | Low if a suitable treebank exists. Otherwise high. | Perhaps less intuitive | Often more robust | Often lower than in rule-based | Typically broader coverage when compared to rule-based |

Table 5-1 shows that while the probabilistic approaches tend to be more robust in comparison to rule-based ones, their grammar may lack intuitive clarity (Kapur & Clark 1996). The development of a grammar for a rule-based system is, on the other hand, usually very expensive. This is also true for the development of probabilistic parsers if a suitable treebank is not available. The probabilistic approaches have been proven to offer a broad enough coverage for practical NLP applications. This has not been the case until recently with many of the rule-based approaches, such as HPSG (Oepen *et al*. 2002).

It appears to be the case that the research community has reached a consensus about combining the two approaches (see, for example, Oepen *et al*. 2002, Klavans & Resnik 1996, and Foth & Mengel 2006). The combination of the advantages of both the rule-based and probabilistic approaches would enable compact descriptions and robustness while keeping the grammar development costs low. One may also note that there appears to be an upper limit for the performance of parsers that use a single approach.

The two basic approaches in combining information from probabilistic and rule-based sources is either to allow the probabilistic component to choose from among the results returned by the rule-based component, or to restrict the number of possibilities for the next processing level by, for example, using a probabilistic POS disambiguator before rule-based parsing. Foth and Mengel (2006), for example, use a hybrid parsing architecture that combines information from more than two sources that are both probabilistic and rule-based. Their system is a rule-based D parser that contains probabilistic predicator components. In their

---

parse. While smoothing consequently allows some correct parses to be generated, the adding of unseen rules with low probabilities is unlikely to improve performance to any great extent.

experiments in parsing German, Foth and Mengel were able to increase performance by several percentage points over the rule-based baseline system.

# II LINGUISTIC RESOURCES FOR EVALUATION

## 6 Analysis of Existing Resources and Schemes

Evaluation of the correctness of a parser's output is generally done by comparing the system output to correct human-constructed structures. These *gold standard* parses are obtained from a *linguistic resource*. Section 6.1 analyzes existing linguistic resources and their suitability for parser evaluation.

Linguistic annotation (hereafter referred to as *annotation*) refers to the notations applied to language data that describes its information content. The annotation in a treebank, for example, includes at least POS tags and syntactic tags. An *annotation scheme* refers to the specification of a set of practices used for annotation in a particular linguistic resource. An e*ncoding scheme* defines the way in which the annotated data is represented. I will both introduce the annotation and encoding schemes used in existing linguistic resources and analyze their suitability for parser evaluation in Section 6.2.

Section 6.3 offers an analysis of existing dependency treebanks (D treebanks). The results of this analysis are used in Chapter 7 as the basis for the design of a treebank for Finnish.

### 6.1 Evaluation Resources

The most commonly used linguistic resources for parser evaluation are treebanks, which are collections of syntactically annotated sentences. These syntactically annotated corpora consist of sentences which have been assigned parse trees with at least syntactic and morphosyntactic annotation. Treebanks are described in Section 6.1.1. *Test suites* are collections of annotated test items that are organized in terms of specific linguistic phenomena (Section 6.1.2). One may describe them as treebanks that have been tailored for evaluation purposes because all the sentences in them have been annotated with syntactic information that elucidates the syntactic phenomena they contain. Section 6.1.3 describes a corpus of ungrammatical sentences. I conclude the findings in Section 6.1.4.

#### *6.1.1 Treebanks*

Treebanks have several applications in linguistics, CL and NLP (Abeillé 2003). Linguists use them, among other things, to gather corroboratory or contradictory evidence for supporting or disproving an hypothesis or theory. Psycholinguists can use treebanks for counting the frequency of specific kinds of sentence constructions. Applications in NLP include the development and evaluation of text classification, parsing and MT systems.

The best-known examples of a treebank with PS annotation are the *Penn Treebank* (PTB) for English (Marcus *et al*. 1993) and its successor PTB-II (Marcus *et al.* 1994). The PTB and PTB-II are the most widely used treebanks for training probabilistic parsers and are extensively used for evaluating all types of parsers for English. The standard method for training and evaluating a probabilistic parser is to use sections 02-21 of the *Wall Street Journal* (WSJ) dataset of the PTB for training the parsing model, and sections 23 and 24 as the test and development sets, respectively (Ringger *et al*. 2004). This will enable an English parser to be evaluated with approximately 2,400 gold standard sentences from a prestigious financial newspaper written in American English.

The SUSANNE corpus (Sampson 1995) is another PS treebank that is frequently used in parser evaluation. It consists of a subset (64 out of the total of 500 texts) of the BC. The texts represent four different genres: press reportage, biography/memoirs, technical/scholarly and adventure/fiction. The size of the SUSANNE, approximately 7,000 sentences, means that it cannot be reliably used to train probabilistic parsers. It is nevertheless a useful resource for parser evaluation.

### *6.1.2 Test suites*

Test suites consist of artificially constructed *test items* (Balkan *et al.* 1994). They are collections of examples with both syntactic annotation and additional information that can be utilized for controlled testing and evaluation. Ungrammatical items are usually also included. This enables researchers to test whether or not parsers have the ability to handle ill-formed input. The test items are usually artificially constructed so that they contain either a single morphological or syntactic phenomenon or a combination of phenomena. The systematicity of a test suite is designed to obviate any kind of uncontrolled interaction between phenomena. Some of the best-known test suites are the *Test Suites for Natural Language Processing* (TSNLP) (Lehmann *et al.* 1996) and the *Hewlett-Packard Test Suite* (Flickinger *et al*. 1987).

The purpose of TSNLP is to evaluate parsers and grammar checkers. It consists of test items for three languages: English, French and German (Balkan *et al.* 1994).

The test suite itself is stored in a relational database from where test items can be retrieved by making use of several criteria. Each test item is marked with the information about the phenomenon for which it tests and other phenomena in the item, that are not designed for testing. Some items are nevertheless specifically designed to contain co-occurrences of phenomena. Ill-formed test items are also included. In the TSNLP framework, the ability to handle specific grammatical phenomena is measured in terms of classes of linguistic phenomena. Summary reports can be created, and the progress of a system monitored by comparing successive reports over a period of time.

A summary report is a table in which each row contains information about a particular grammatical phenomenon (Oepen & Flickinger 1998). The results are reported in terms of the percentage of items that have been covered, and they are recorded separately for grammatical and ungrammatical test items. Thus, for example, if the coverage of grammatical and ungrammatical test items in the *agreement* phenomenon is 59% and 14% respectively, this means (1) that the parser is not covering this particular phenomenon adequately (it is rejecting 41% of the grammatical items), and (2) that it is also overgenerating (it is accepting 14% of the ungrammatical test items).

### *6.1.3 A corpus of ungrammatical sentences*

Foster and Vogel (2004) reported a 20,000-word (approx. 1,000 sentences) corpus of naturally occurring, ungrammatical English sentences. Such a resource is useful for evaluating a parser’s ability to analyze noisy input. The corpus was collected from several resources that included newspapers, e-mails and student writings. The error types in the corpus include incorrect word forms, extraneous words, omitted words, and composite errors (errors that can be fixed by applying more than one correction operation). For each sentence, the corpus offers parallel correct and ungrammatical versions that are identical in meaning. The assumption is that the parse for an ungrammatical sentence should be as close as possible to the parse for its grammatical counterpart so that its true meaning is expressed. In a case where a particular erroneous sentence can be corrected in more than one way, all possible corrected versions are included in the corpus of grammatical sentences. The corpus is not annotated with morphological or syntactic information.

### *6.1.4 Analysis*

In this section, I shall compare the different types of linguistic resources (treebanks and test suites) and analyze their suitability for parser evaluation. The discussion about treebanks and test suites in Sections 6.1.4.1 and 6.1.4.2 will be limited to the two most commonly used PS treebanks and test suites for English respectively.

#### 6.1.4.1 Treebanks

It may be argued that the standard method of using the PTB in evaluation makes the current parsing practices appear in a rather-too-rosy light. It does this because there are certain properties of the PTB that make parsing easy (Manning & Carpenter 1997). Even non-lexicalized and CF approaches (such as non-lexicalized PCFG parsers) work well when it comes to parsing the PTB. The main reason why this happens is that the trees in the PTB are quite flat. The less detail there is in the structures, and the fewer brackets there are, the easier it is to assign structures correctly. As Manning and Carpenter also point out, the analyses assigned to certain kinds of sentence structures[55] do not give enough advantage to a parser that can analyze them correctly. Since the PTB is often used as the only evaluation resource in the development of a parser, there is a concern about the extent to which the parser will adapt to other data. Diversity of data is the major advantage that the SUSANNE corpus has over the PTB. While section 23 of the PTB that is often employed in evaluation is composed solely of WSJ sentences,[56] the SUSANNE contains texts from several genres.

#### 6.1.4.2 Test suites

The main difference between the older generation of test suites, such as the Hewlett Packard test suite (Flickinger *et al.* 1987) and the TSNLP introduced above, is that, in the latter, the test items are grouped into sets that define classes of linguistic phenomena. This enables evaluators to carry out a more controlled kind of testing because they can adjust the granularity to their needs (Oepen & Flickinger 1998). In more recent test suites special attention has also been paid to the systematicity of the phenomena covered. One of the deficiencies in TSNLP is that it does not have a mechanism that permits the automated replacement of lexical items (Oepen & Flickinger 1998). If therefore a test item contains words that the evaluated parser cannot recognize, the evaluator will either have to modify the lexicon of the parser or skip the test item.

---

[55] For example, the adjunction structures of the form [NP [NP *the ball*] [PP *in* [NP *the field*]]].
[56] Lin (2003) mistakenly claims that PTB consists only of WSJ data.

The main difficulty with test suites is that they are extremely complex to construct (King 1996). It is also difficult to define test items that will test *only* the desired phenomena – and nothing else. Furthermore, a test suite has to be very large (in the order of thousands of test items) if it is to cover all the major syntactic and morphological phenomena that occur in a language. Another problem that arises in the application of test suites is that they are often designed to test a specific system.

#### 6.1.4.3 Treebanks vs. test suites

Test suite evaluation is especially suitable for checking the consistency of a grammar and parsing model. After modifications have been made to the grammar or the parsing algorithm, it is possible to check the system for undesirable side effects. Test suites also provide the means to test phenomena that occur rarely in free text. But one disadvantage associated with test suites is the lack of variation in the lexical items. Another disadvantage is that the test items usually contain a single grammatical phenomenon that leaves interactions between phenomena untested.

Since treebanks are usually designed with diverse types of uses in mind, they are usually more general than test suites. The main distinction between treebanks and test suites, with regard to parser evaluation, is in their focus (Srinivas *et al.* 1998, Carroll *et al.* 1998). Parser developers usually use test suite-based evaluation to monitor the development of a system and to identify its strengths and weaknesses in controlled circumstances. Their aim as they do this is to measure competence, namely, how successful the system is in covering phenomena and whether or not it is consistently successful in treating phenomena in the same way in different contexts. One may contrast competence with performance, which is how the system behaves when it parses running texts or a treebank (measured as the number and type of errors in the output of the system). Table 6-1 compares some of the properties of the best-known test suites and treebanks.

**Table 6-1.** A comparison of the treebanks and test suites described above. Column "TB/TS" indicates whether or not the evaluation resource represented in the row is a treebank (TB) or a test suite (TS).

| Test suite / treebank | TB/ TS | Size | Size of vocabu-lary |
|---|---|---|---|
| PTB (Marcus *et al*. 1993) | TB | ~42,000 sentences | ?* |
| SUSANNE (Sampson 1995) | TB | ~7,100 sentences | ?* |
| TSNLP (Lehmann *et al*. 1996) | TS | ~5,000 test items/language | ~200 |
| Hewlett-Packard suite (Flickinger *et al*. 1988) | TS | ~1,230 test items | ~250 |

*The authors do not give the size of the vocabulary in the treebank. I was, moreover, unable to find this information from any other source.

Because treebanks and test suites are very different in their design, they perform different roles in evaluation and should be regarded as being complementary rather than competing evaluation resources (Balkan *et al.* 1995). Prasad and Sarkar (2000), for example, observed a small degree of duplication of the error types they found when they applied treebank evaluations and test suite-based evaluations respectively. Table 6-2 summarizes the discussion above.

**Table 6-2.** A comparison of treebank and test suite-based evaluation

| | Treebanks | Test suites |
|---|---|---|
| **Coverage of linguistic phenomena** | They contain naturally occurring sentence structures and sentences with several syntactic phenomena. They lack systematic variations. | They are restricted to structures that are taken into account by the creators of the test suite. They usually test only one phenomenon per sentence. |
| **Lexicon** | They are rich in lexical variety, even though this variety is usually restricted to one text genre. | This is usually restricted. It is necessary to provide a lexical replacement tool. |
| **Generality** | They are usually constructed to serve a number of purposes. | They are often designed for system-specific purposes. |

#### 6.1.4.4 Collections of ungrammatical sentences

The resource constructed by Foster and Vogel (2004) is the only one of its kind for English. It consists only of sentences with grammatical mistakes; no misspelled words are included. Such corpora with ungrammatical and grammatical sentences as well as tests suites with negative test items can both be applied to evaluate overgeneration in parsers. Another possibility would be to generate ungrammatical sentences by using an automatic sentence-generation method. This approach has, however, not yet been tried out.

## 6.2 Annotation of Evaluation Resources

The annotation scheme defines the tagsets, the inventory of tags, for describing the linguistic content and the principles of annotation. The annotation scheme of a treebank usually consists of word and syntactic levels. While word-level annotation typically consists of at least POS tags, it may also include lemma information and morphological descriptions. Some treebanks have an additional annotation level for semantics. The *Prague Dependency Treebank* for Czech, for example, is annotated on three levels: morphological, syntactical and semantic (Böhmová *et al*. 2003). There are usually only two choices when it comes to the type of syntactic annotation in a treebank: the linguistic resource is annotated either in terms of a PS or D structure scheme.

While there are some treebanks that have been annotated completely by hand, the manual construction of syntactic trees remains a slow and error-prone process. Since taggers and parsers are readily available to automate some of the work, such a method is rarely employed in state-of-the-art treebanking. The most common practice is to construct a treebank *semi-automatically* by combining automatic processing with human checking.

The differences in the annotation schemes of the linguistic resources and the output schemes of parsers are a hindrance to parser evaluation. There is no widespread agreement about which POS and morphological tagging schemes are best for the linguistic resources – let alone about which tagsets might be best for syntactic description. Apart from the fact that there are considerable differences in the size of the tagset, word-level annotation often assumes a different segmentation of text into lexical units and handles punctuation in different ways.[57] Such differences often result in many-to-many mappings between schemes. The problems caused by differences in word-level annotation can often, however, be solved – if not perfectly, then at least in a satisfactory way. This is effected by automatic mapping and alignment algorithms (see, for example, Leech *et al.* 1996, Déjean 2000, and Chiarcos 2006). The differences between the schemes are much greater in sentence-level annotation.

---

[57] Atwell *et al.* (2000) have pointed out a problem that arises when one compares schemes. This problem is how to identify precisely where a scheme is defined: whether in the annotation guidelines (if one exists), or in the annotations observed in the resource/output, or in the intuitions generated by the linguists who are in charge of the project. None of these sources is error-free. The annotations in the resource may contain inconsistencies, the guidelines may contain omissions, and the experts may have made mistakes.

There are several approaches that have been devised to overcome the problems caused by differences in annotation and output schemes. These are *mapping* algorithms between schemes (Section 6.2.1), *abstract* annotation models that can be transformed into resource-specific annotation schemes (Section 6.2.2), and *parallel annotations* that are undertaken in accordance with several parser-specific schemes (Section 6.2.3). One possibility, described in Section 6.2.4, is to organize the POS and syntactic tagsets hierarchically. This enables one to compare the annotated resource to a variety of parsers because the hierarchy can be used to allow inexact matches between the parser output and the annotation.

One may distinguish classes of annotation schemes on the basis of how closely they conform to a particular theory of syntax (Nivre 2003). A *theory-specific* annotation scheme that is constructed in accordance with a particular linguistic theory runs the risk of being useful only to those researchers and system developers who are applying the same framework. While a *theory-neutral* annotation scheme might in theory be serviceable to a wider range of users, it would be necessary to make so many compromises in the course of its design that there is a risk that it would generate far too little information to be useful to any possible group of users. A *theory-supporting* scheme is one that disavows the extremes of the two mentioned above: it supports an annotation scheme that can be mapped to theory-specific target annotations.

### *6.2.1 Mapping between annotation schemes*

The most important aim of the mapping approach is to base evaluation on a given annotation scheme and then to use mapping algorithms to automatically convert parsers' outputs to that scheme. The function of a mapping algorithm is to map from a source annotation scheme to a target scheme without changing the information content of the annotation. One may formally define a mapping in the following way:

**Definition 6-1**. Mapping from a source annotation scheme to a target scheme.

Let $S$ and $T$ be the source and target annotation schemes respectively. Let $A_S$ and $A_T$ be the set of all the annotations that arise from the annotation schemes $S$ and $T$ respectively. Let $S(txt)$ denote the annotation of text *txt* in scheme $S$; $S(txt) \in A_S$. Similarly, let $T(txt)$ denote the annotation of *txt* in $T$, $T(txt) \in A_T$.

1. Mapping $M: A_S \rightarrow A_T$ is a function for each text *txt*, $M(S(txt)) = T(txt) \in A_T$.

Nivre's (2003) notion of theory-supporting treebanks is based on the use of a set of mapping algorithms to create the possibility of converting a treebank to theory-specific formats. Figure 6-1 illustrates how a theory-supporting treebank might be utilized.

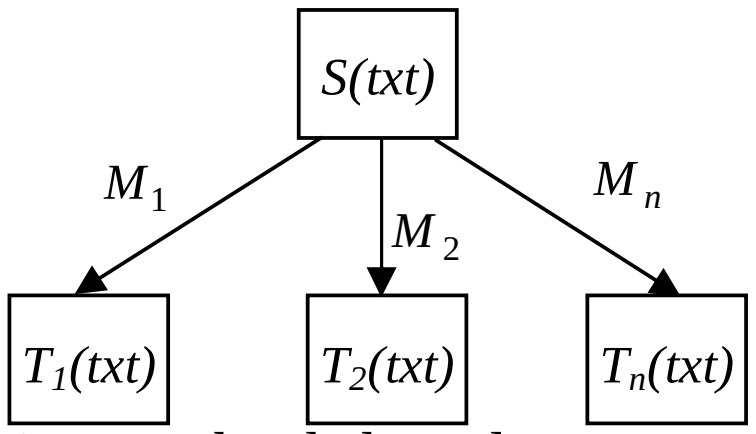

**Figure 6-1.** A theory-supporting treebank based on mapping algorithms. A text annotated according to the scheme *S* of the source treebank is defined in a way that it can be accurately converted into target schemes $T_{1..n}$ by using mapping algorithms $M_{1..n}$.

### *6.2.2 Abstract annotation models*

The aim of an abstract annotation model is to provide a general theory- and tagset-independent framework for linguistic annotation. XML-based *exchange formats*, such as *TIGER-XML* (Mengel & Lezius 2000), *ATLAS* (Bird *et al.* 2000) and *XCES* (Ide & Romary 2003), have the following goal in common: they each offer an intermediate level between the annotated data and the tools for browsing and for manipulating data. An advantage of such an approach (which is illustrated in Figure 6-2) is that it enables a common set of tools to be used for creating and manipulating treebanks that use different annotation schemes.

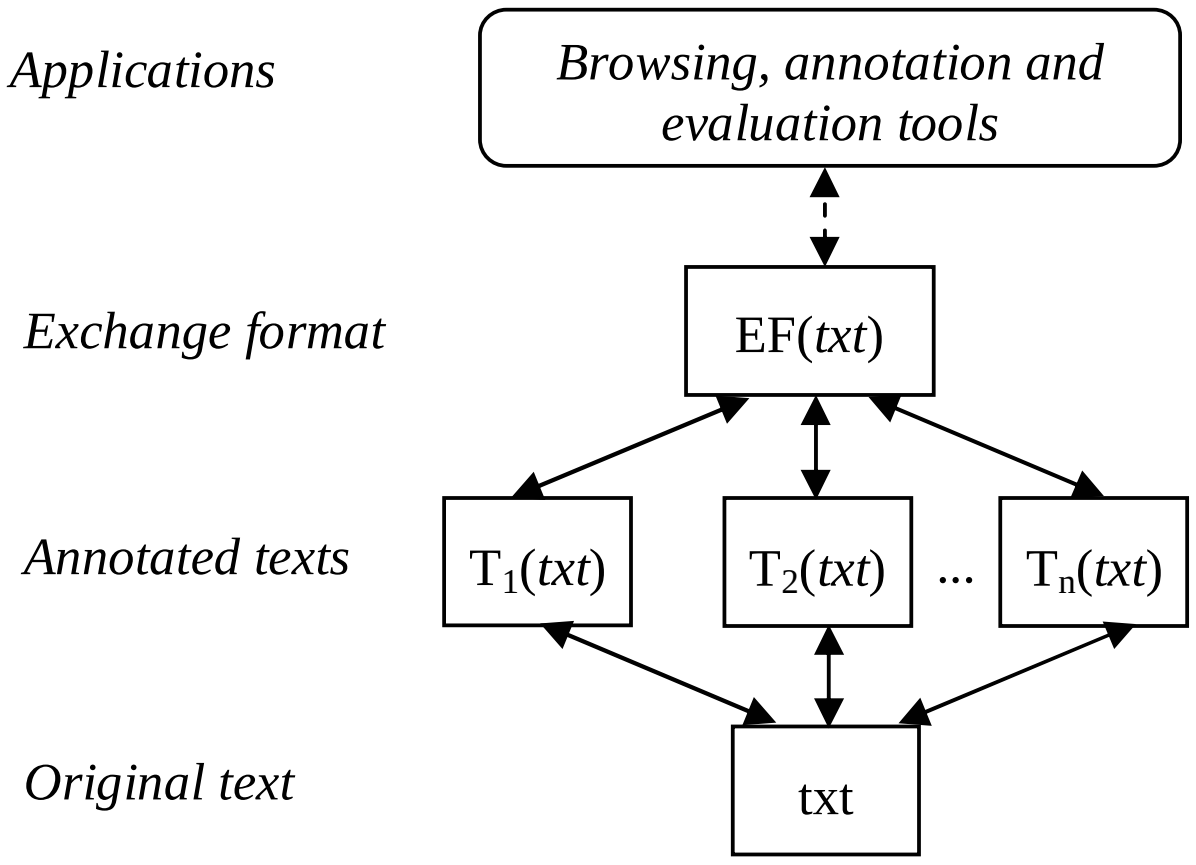

**Figure 6-2.** Using an abstract annotation format for treebanks. The exchange format *EF* works as an interface between the different annotations stored in treebanks that use schemes $T_{1..n}$. Instead of treebank-specific software, one only needs a set of tools that is able to understand the exchange format for manipulating, browsing and searching any of the treebanks.

TIGER-XML is an exchange format that is sufficiently general for existing treebank-specific annotations, both D and PS based, to be exported into its XML-based representations (Mengel & Lezius 2000). The model is based on the encoding of DAGs, each of which represents a sentence using four XML element types: sentence *<s>*, nonterminal *<nt>*, terminal *<t>*, and edge *<edge>*. Syntactic, POS, and other kinds of information are represented as attributes in *<t>* and *<nt>* elements. Edges encode labeled links between terminals and nonterminals. TIGER-XML also allows *secondary edges* to be encoded. These can, for example, be used for encoding semantic information and LDDs. Figure 6-3 shows examples of a TIGER-XML-encoded sentence.

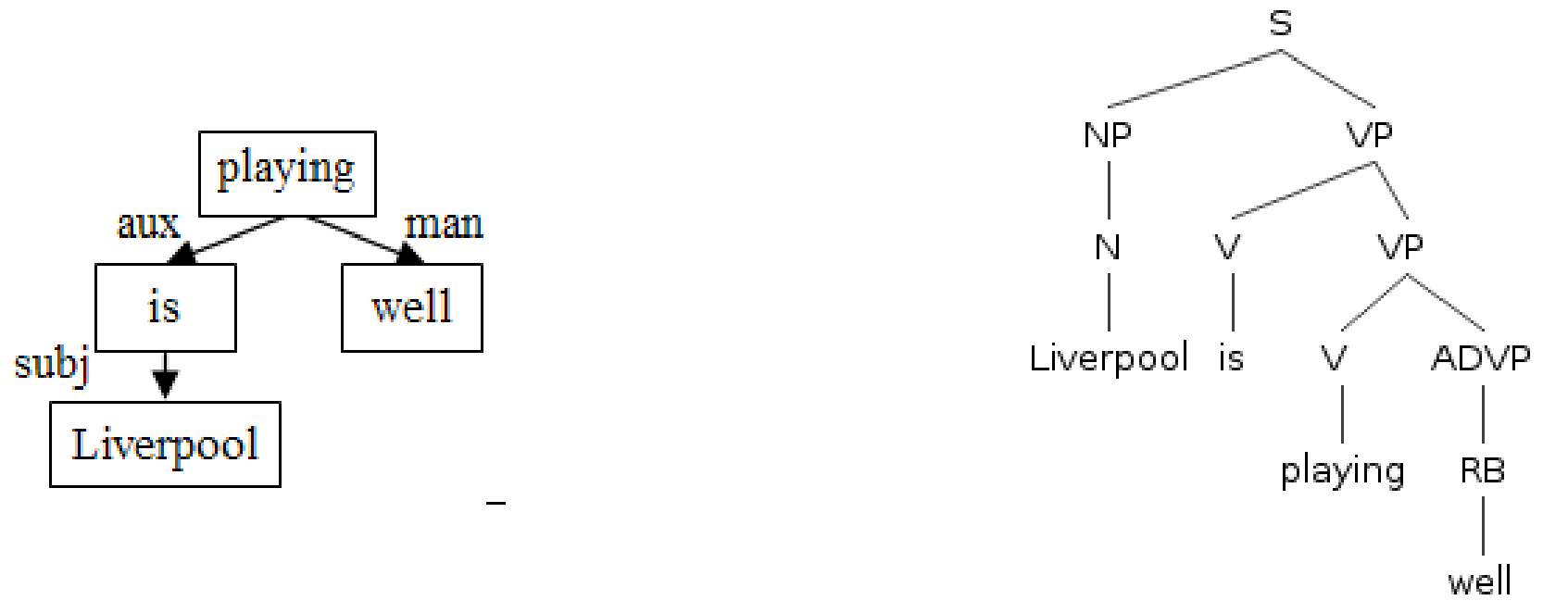

```
<s id="s1">
 <graph root="nt3">
  <terminals>
    <t id="t1" word="Liverpool" pos="N"/>
    <t id="t2" word="is" pos="V"/>
    <t id="t3" word="playing" pos="V"/>
    <t id="t4" word="well" pos="RB"/>
  </terminals>
  <nonterminals>
    <nt id="nt1" word="Liverpool" cat="N">
      <edge idref="t1" label="--" />
    </nt>
    <nt id="nt2" word="is" cat="V">
      <edge idref="t2" label="--" />
      <edge idref="nt1" label="subj" />
    </nt>
    <nt id="nt3" word="playing" cat="V">
      <edge idref="t3" label="--" />
      <edge idref="nt2" label="aux" />
      <edge idref="nt4" label="man" />
    </nt>
    <nt id="nt4" word="well" cat="RB">
      <edge idref="t4" label="--" />
    </nt>
  </nonterminals>
```

```
<s id="s1">
 <graph root="nt5">
  <terminals>
    <t id="t1" word="Liverpool" pos="N"/>
    <t id="t2" word="is" pos="V"/>
    <t id="t3" word="playing" pos="V"/>
    <t id="t4" word="well" pos="RB"/>
  </terminals>
  <nonterminals>
    <nt id="nt1" cat="NP">
      <edge idref="t1" label="--" />
    </nt>
    <nt id="nt2" cat="VP">
      <edge idref="t2" label="--" />
      <edge idref="nt3" label="--" />
    </nt>
    <nt id="nt3" cat="VP">
      <edge idref="t3" label="--" />
      <edge idref="nt4" label="--" />
    </nt>
    <nt id="nt4" cat="ADVP">
      <edge idref="t4" label="--" />
    </nt>
    <nt id="nt5" cat="S">
      <edge idref="nt1" label="--" />
      <edge idref="nt2" label="--" />
    </nt>
  </nonterminals>
  </graph>
```

**Figure 6-3.** A D and PS structure for the sentence “Liverpool is playing well” encoded in TIGER-XML. The POS tags are omitted from the D tree.

In the ATLAS architecture, the abstraction level between the physical and application level is called the *logical level*, with *Annotation Graphs* (AGs) as the central notion (Bird *et al*. 2000). It is possible to add time-stamps to AGs; this makes them suitable for representing, in addition to syntactic structures, videos and multi-modal interactions. The functions for manipulating ATLAS structures from applications are implemented in the *ATLAS Application Protocol Interface* (API). The *ATLAS Interchange Format* (AIF) serves as the common XML-based representation for specific annotations. The physical storage can be relegated to a database. Figure 6-4 illustrates ATLAS architecture.

**Figure 6-4.** The architecture of ATLAS (Bird *et al*. 2000).

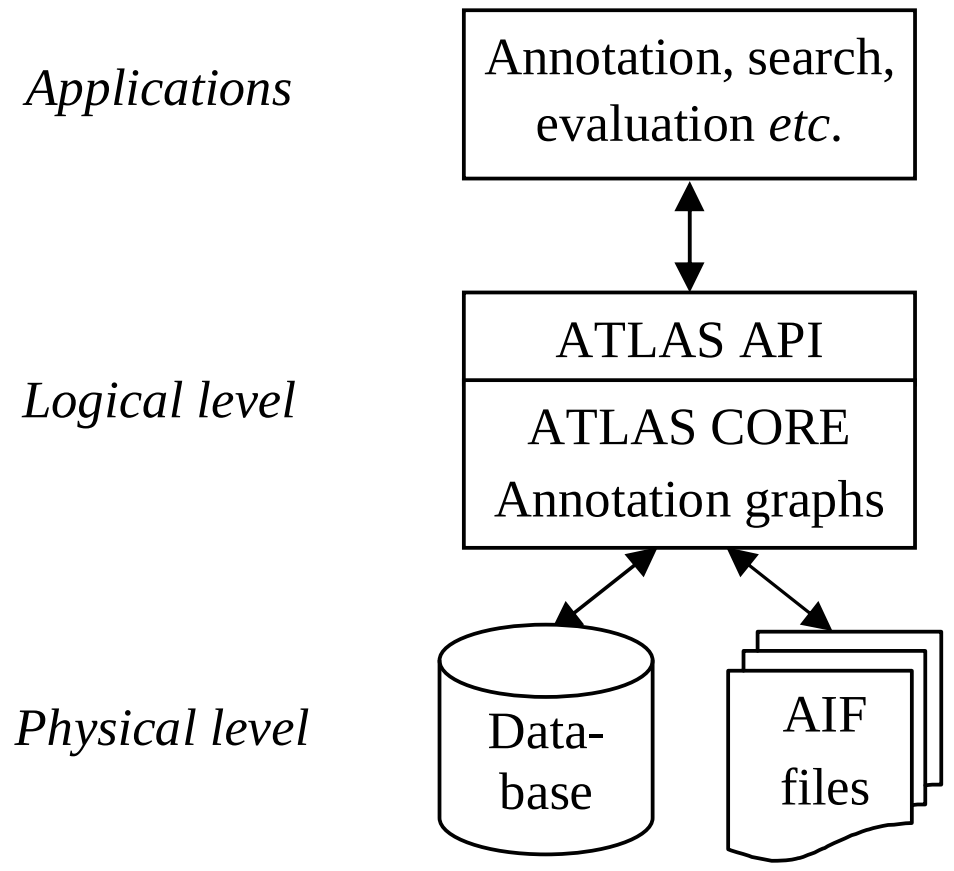

The XCES format has been designed for several different types of annotation such as morphosyntactic, syntactic and coreference annotation (Ide & Romary 2001, 2003, Ide *et al*. 2001). The XCES framework is divided into two levels: the *universal resources*, that are shared by all annotation models, and the *project specific resources*.

### *6.2.3 Parallel annotations*

In an approach that relies on parallel annotations, a linguistic resource is annotated in terms of several treebank- or parser-specific schemes. The only example of such a resource is the *MultiTreebank* of English which consists of 60 sentences annotated automatically according to nine different schemes (Atwell *et al.* 2000). Figure 6-5 illustrates this idea.

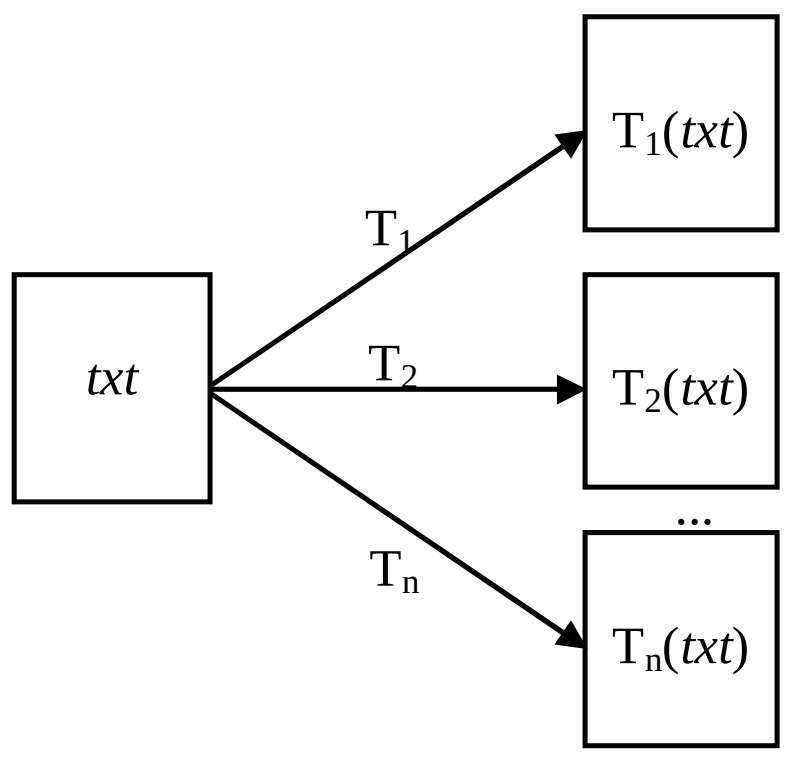

**Figure 6-5.** A parallel treebank. The source texts are annotated in formats $T_{1..n}$.

### *6.2.4 Hierarchical organization of tagsets*

The *Grammatical Relations Scheme* by Carroll *et al*. (1998, 2003) (referred to hereinafter as the GR scheme) uses an annotation scheme that is based on GRs between heads and dependents. In order to facilitate inter-system comparisons, the GRs are divided according to the hierarchy illustrated in Figure 6-6.

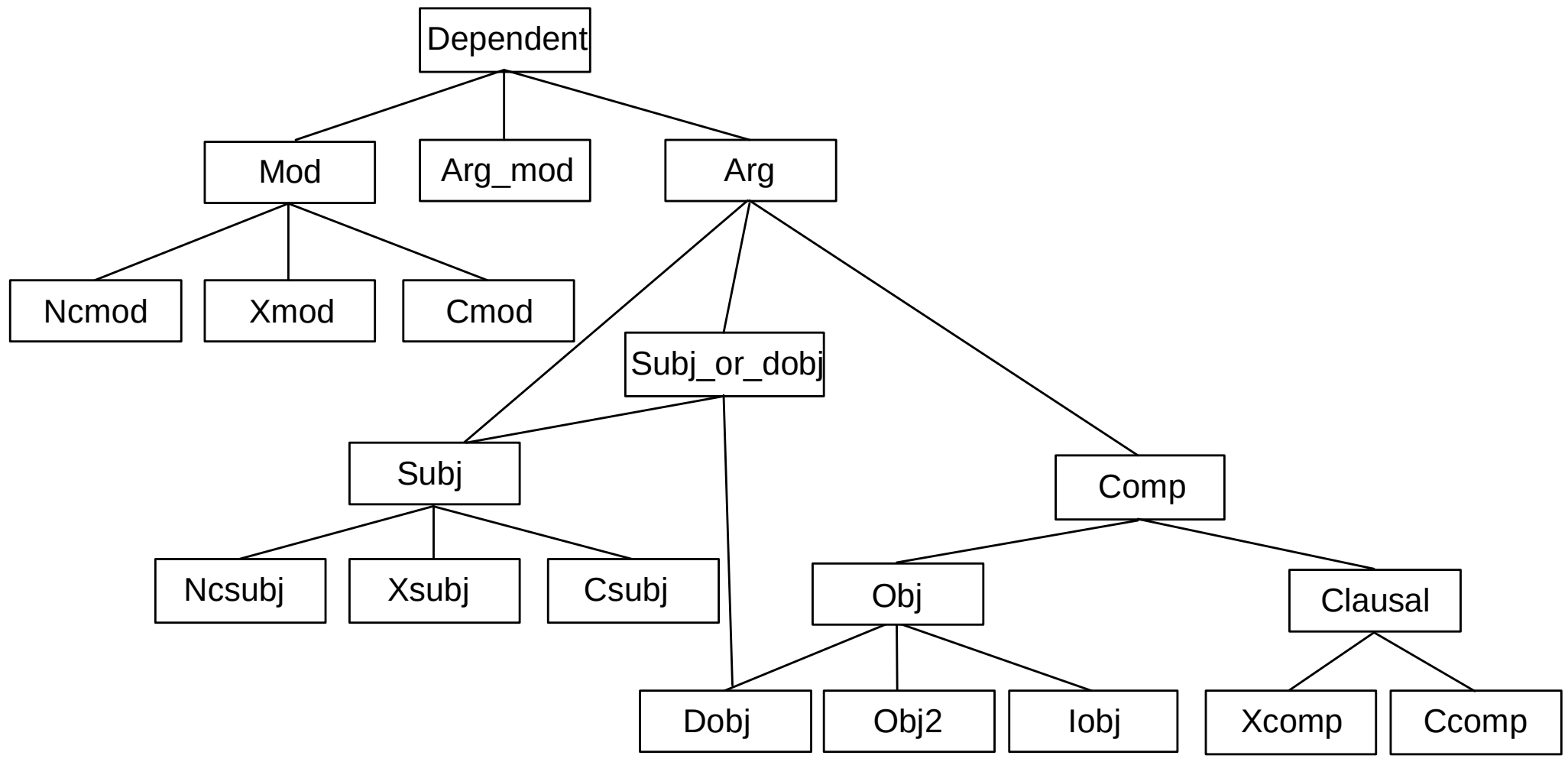

**Figure 6-6.** The hierarchy of grammatical relations in the GR scheme (Carroll *et al*. 2003).

Figure 6-6 shows that *dependent* is the most generic of relation type (Carroll *et al.* 1998). Dependents are further divided into three categories. The relation between a head and its modifier is *mod* (denoted as *mod(type, head, dependent)*), and the relation between a head and its argument *arg*. The *arg_mod* relation holds between a head and a semantic argument, which is syntactically realized as a modifier.

A test set of 500 sentences, consisting of texts from the BC, has been developed on the basis of the GR scheme (which, in this thesis, is called the GR corpus) (Briscoe *et al*. 2002). There are in total 4,690 relations in the test set, of which roughly 60% belong to *mod*s and 40% to *arg*s.

**Example 6-1.** The GR scheme annotation for the sentence “Liverpool is playing well”. The first item, for example, indicates that the word “playing” is in *dobj* relation to the word “well”.

(*dobj* playing well)
(*aux* _ playing is)
(*ncsubj* playing Liverpool _)

### *6.2.5 Analysis*

Table 6-3 summarizes the differences between the annotation schemes of four treebanks that are often used in evaluation and output schemes of five parsers that are based on different grammar formalisms.

**Table 6-3.** A comparison of the annotation and output schemes of some well-known treebanks and parsers for English.[58] In the column "Type", TB and P indicate whether the item is a treebank or a parser. "Grammar" column indicates the grammar formalisms on which the parser in question is based. The PS/D column indicates whether the scheme is based on PS or on D-style representations. The column labeled "Functional labels" indicates whether the scheme includes functional labels (subject/object, etc.). The last two columns show the number of POS tags and syntactic tags in the tagset.

| Parser / treebank | Type | Grammar | PS/D | Functional labels | No. of POS tags | No. of syntactic tags |
|---|---|---|---|---|---|---|
| *PTB* (Marcus *et al. 1993*) | TB | | PS | No | 36 + 12 | 17 |
| *PTB-II* (Marcus *et al. 1994*) | TB | | PS(D*) | Yes | 36 + 12 | 17 |
| *SUSANNE* (Sampson 1995) | TB | | PS** | Yes | 353 | 8 main, 53 sub |
| *PARC 700* (King *et al.* 2003) | TB | | D | Yes | - | 18 |
| *ENGCG* (Lingsoft Ltd 2006) | P | CG | D | Yes | 16 | 32 |
| *LG Parser* (Sleator *et al.* 1991) | P | LG | D/PS | Yes | 8 | 107 |
| *RADISP 2* (Carroll *et al.* 2006) | P | UG | D | Yes | 50 | 23 |
| *Stanford Parsed* (Klein *et al.* 2003) | P | PCFG | PS/D | Yes | 36+12 | 48 (D) |
| *StatCCG (Hockenmaier 2003)* | P | CCG | D | Yes | >1200 lex. cat. types | 4 atomic types |

*PAS in the PTB-II.

**Lin (2003) introduced an algorithm for transforming the SUSANNE structures into D format.

The tendency in treebank building has been to move away from theory-neutral annotation schemes toward theory-specific ones (Nivre 2002). This is a retrograde development for evaluation. When one considers the high cost of building linguistic resources, it would surely be more practical to develop reusable formats.

Each of the approaches discussed above has its own disadvantages. Since mappings are often complicated and usually have to be performed in several steps, there are a number of problems that may arise during the mapping process (Sasaki *et al.* 2003, Wang *et al.* 1994). The first of these is that the tagsets may not be identical: the number of tags may be different and the mapping is not necessarily one-to-one.[59] The second problem is that some constructions in one scheme may not be able to be represented by the other scheme.

[58] Some of these issues were discussed by Atwell (1996) and Atwell *et al.* (2000) on the basis of the classification of annotation levels by Leech *et al.* (1996a).

[59] As I have already pointed out in Chapter 3, a PS can be converted into D structure if each nonterminal has a *head child* (either a lexical head or a phrase containing a lexical head). The problem is that most PS treebanks do not provide information for unambiguously identifying the heads. Apart from this, the notion of head might not be compatible in the source and target annotations.

In those cases where mappings can be devised, they constitute a useful and rather simple method for facilitating a certain linguistic resource for evaluating parsers. But because of innumerable potential problems, it might never be possible to map between most of the existing parser-specific output schemes and resource-specific annotation schemes. While this approach might work reasonably well for evaluation taggers and morphological analyzers, it is not realistic for comparative parser evaluation.

Nivre (2003) has pointed out that there are no theory-supporting treebanks available. The most serious hindrance to practical implementation is in the construction of mapping algorithms. The problems that arise in mapping could be avoided, at least to some extent, by setting conditions for the source annotation. The source annotation, for example, should be required to contain sufficient information to identify the heads. This would enable successful PS/D mappings to take place. It is, however, undesirable to allow the annotation scheme to set conditions on the annotations that are allowed.

Each of the three abstract annotation formats discussed above has a different focus. TIGER-XML is designed for corpus and treebank annotation and offers an exchange format rather than a whole abstraction framework. ATLAS is similar to TIGER-XML in the sense that it uses DAGs for representation. ATLAS has a broader scope than TIGER-XML because one can apply the framework in the annotation of several different types of data including images and video. XCES is the most ambitious of the formats. It offers a complete framework and takes the abstraction a step further than either of the other two models. The problem with general models of linguistic categories, however, is that they often lead to the loss of theory-specific information (Sasaki *et al*. 2003).

The main advantage that the abstract annotation models as well as the mapping-based approaches have over parallel annotations with regard to parser evaluation, is that they offer more grounds for comparing diverse parsing systems. The main problem of the multi-treebank approach lies in guaranteeing the consistency of the annotations. Because of the size and complexity of annotation schemes, it would be too much to expect a single annotator to master several of them (Atwell 1996). The creation of a multi-treebank therefore requires the cooperation of several research teams, and this might introduce a possible source of inconsistencies. Because it is expensive to construct even a “normal” treebank, one would expect that the cost of building a multi-treebank would be prohibitive.

Organizing the tagset into a hierarchy, as in the GR scheme, allows for differences between the annotation of the evaluation resource and a parser output, and it still remains reasonably easy to annotate. In fact, once the tag hierarchy has been defined, no extra work is required from the annotators. The GR scheme has been applied only to the annotation of the rather small GR corpus in English.

## 6.3 Analysis of Dependency Treebanks

In this section, I will introduce the most important results of the D treebank analysis that I have already discussed in paper [1].

There has been a great deal of interest in recent years in the functional annotation of treebanks. Several D treebanks in particular have been constructed. In addition to this, grammatical function annotation has been added to some PS-type treebanks. The most commonly used argument for selecting the D format for building a treebank is that the treebank concerned is being created for a language with a relatively free word order. Such treebanks already exist for Basque, Czech, German and Turkish. D treebanks have been also developed for languages such as English, which are usually regarded as languages that can be better represented with PS formalisms. The reasons for using D annotation vary from the fact that the type of structure concerned is the one that is needed by many, if not most, applications – to the fact that it offers a rational interface between syntactic and semantic representation (Lombardo & Lesmo 1998). D trees can, moreover, be automatically converted into PS trees (Xia & Palmer 2001), and vice versa (Daum *et al*. 2004) – although not always with 100 % accuracy. The *TIGER Treebank* for German is an example of a treebank with both PS and D annotations (Brants *et al*. 2000). Such hybrid treebanks are also available for Danish (Bick 2003) and Dutch (Van der Beek *et al*. 2002).

The following treebanks are discussed in this section:

1. *Prague Dependency Treebank* (PDT) for Czech (Böhmová *et al.* 2003). It contains text from the *Czech National Corpus* (Czech National Corpus 2005). It is encoded in two treebank-specific formats, *feature structure* (FS) and *Czech Sentence Tree Strucutre* (CSTS).
2. The *TIGER Treebank* for German (Brants *et al*. 2000). This treebank was developed on the basis of the *NEGRA Corpus* (Skut *et al.* 1998), and consists of a set of articles on diverse topics sourced from a German newspaper.

3. *Arboretum* for Danish (Bick 2003), *L'Arboratoire* for French, *Floresta Sintá(c)tica* for Portuguese (Afonso *et al.* 2002), and *Arborest* of Estonian (Bick *et al.* 2005). These are sibling treebanks (Arboretum is the eldest sibling). The treebanks are hybrids with both PS and D annotation organized into two separate levels.
4. The *Dependency Treebank* for Russian (Boguslavsky *et al.* 2000, 2002). This treebank is based on the *Uppsala University Corpus* (Lönngren 1993). The texts were collected from contemporary Russian prose, newspapers and magazines.
5. The *Alpino Treebank* for Dutch (Van der Beek *et al.* 2002). This treebank, which comprises newspaper articles, was designed mainly for parser evaluation. The annotation scheme was taken from the *CGN Corpus* of spoken Dutch (Oostdijk 2000) and the annotation guidelines were based on the TIGER Treebank's guidelines.
6. The *Danish Dependency Treebank* (Kromann 2002, 2003). The annotation scheme of the treebank is based on *Discountinuous Grammar* and covers a wide range of topics. The morphosyntactic annotation was obtained from the *PAROLE Corpus* (Keson & Norling-Christensen 2005).
7. The *Turkish Treebank* (Atalay *et al.* 2003). The texts in this treebank are morphologically and syntactically annotated and were obtained from the *METU Turkish Corpus*, which covers 16 main genres of contemporary written Turkish (Oflazer *et al.* 2003).
8. The *Basque Dependency Treebank* (Aduriz *et al.* 2003). This treebank consists of manually annotated sentences from newspaper articles.
9. The *Turin University Treebank* (TUT) for Italian Treebank (Bosco 2000, Bosco & Lomardo 2003, Lesmo *et al.* 2002). This treebank is divided into four sub-corpora. The majority of these texts are from the civil law code and newspaper articles.
10. The *Dependency Treebank* for English (Rambow *et al.* 2002). This treebank consists of dialogues between a travel agent and customers, and is the only D treebank with annotation of spoken language transcripts.
11. The *PARC 700 Dependency Bank* (DepBank) (King *et al.* 2003). This treebank consists of 700 annotated sentences from the WSJ dataset of the PTB.

Table 6-4 summarizes some key properties of existing D treebanks. The size of the treebanks is usually quite limited, and they range from a few hundred sentences to 90,000 sentences. This is partly due to the fact that even the oldest of the D treebank projects, the PDT, was started less than ten years ago. The treebank producers have in most cases aimed at creating a multipurpose resource for evaluating and developing NLP systems and for studies in theoretical linguistics. Some have been built for specific purposes. The Alpino Treebank of Dutch, for example, was designed mainly for parser evaluation. Most of the D treebanks consist of written texts, and there is only one (The Dependency Treebank for English) that was based on a collection of spoken utterances. The written texts are usually obtained from newspaper articles. In other cases such as the Czech, German, Russian, Turkish, Danish and Dutch treebanks, they were sourced from an existing corpus. Annotation usually consists of POS and morphological levels accompanied by D-based syntactic annotation. In the case of the PDT, a higher, semantic layer of annotation is also included.

**Table 6-4.** A comparison of dependency treebanks. KEY: M = manual, SA = semi-automatic, TB = treebank.

| Name | Lang. | Genre | Size (sent.) | Annotation methods | Parser | Encoding schemes |
|---|---|---|---|---|---|---|
| PDT | Czech | Newspaper (general, economic), science mag. | 90,000 | M/SA | Lexicalized PCFG parser (Collins) | FS, CSTS SGML, Annotation Graphs XML |
| TIGER TB | German | Newspaper | 50,000 | SA, post-editing & interactive | Probabilistic parser / LFG parser | TIGER-XML & NEGRA export |
| Arbore-tum and others* | 4 lang. | Mostly newspaper | 21,600 (Arb.) 9,500 (Flor.) | D to PS mapping, M checking | A CG parser for each language | TIGER-XML & PENN export (Arb.) |
| Depend-ency TB for Russian | Russian | Fiction, newspaper & scientific | 12,000 | SA | Morphological analyzer and a parser | XML-based TEI-compatible |
| Alpino | Dutch | Newspaper. For parser evaluation | 6,000 | SA, partially M disambiguation aided by a parse selection tool | HPSG-based Alpino parser | Own XML-based |
| Danish Depend-ency TB | Danish | Range of topics & genres | ~5,500 | M. Morphosyntactic annotation obtained from a corpus | - | PAROLE-DK format with additions, TIGER-XML |
| METU-Sabanci TB | Turkish | 16 genres | 5,000 | M disambiguation & M dependency marking | Morph. analyzer based on XEROX FST | XML-based, XCES-compatible |
| Basque TB | Basque | Newspaper | 3,000 | M, automatic checking | - | XML-based, TEI-compatible |
| TUT | Italian | Mainly newspaper & civil law | 1,500 | M checking of parser & morph. analyzer output | Morph. analyzer, rule-based tagger and a parser | Own ASCII-based |
| Depend-ency TB of English | English | Spoken, travel agent dial. | 13,000 words** | SA, M correction of parser output & automatic checking of inconsistencies | Supertagger and Lightweight Dependency Analyzer | FS |
| Dep-Bank | English | Financial newspaper | 700 | 700 M checking & correction, autom. consistency checking | LFG parser, checking tool | Own ASCII-based |

*Not all the treebanks in the Arboretum "family" are included in the table. **Information of number of utterances was not available.

The definition of an annotation scheme always involves a trade-off between the accuracy of the representation, the coverage of the data and the costs of treebank development (Bosco 2000, Bosco & Lombardo 2003). The selection of the tagsets for annotation is critical. While the use of a large variety of tags guarantees a high degree of preciseness and specialization in the description, it makes the work of the annotators even more time-consuming. In addition to that, highly informative annotation in some applications (such as training of probabilistic parsers) will frequently cause problems with data sparseness. The opposite of this is that if the annotation is highly generalized, the annotation process will be faster but a lot of information will be lost. The TUT and Basque treebanks attempt to deal with this problem by organizing the set of GRs into a hierarchical taxonomy.

The choice of the type of application for the treebank also often affects the design of the annotation scheme. A treebank for evaluation allows for some remaining ambiguities but no errors, while the opposite may be true for a treebank used for training (Abeillé 2003). In an annotation scheme that consists of multiple levels, a definite separation between the levels is a source of concern. The format of the annotation is also directed by the specific language for which the treebank is being developed. The format needs to be suitable for representing the structures of the language. In the METU-Sabanci Treebank, for example, a special type of morphological annotation scheme was introduced because of the complexity of Turkish morphology.

Semi-automatic annotation that combines parsing and manual checking is the method most commonly applied in constructing treebanks. None of the D treebanks has been created completely manually; at least an annotation tool capable of visualizing the structures is used by each of the projects. There are no fully automatically created D treebanks simply because there are no parsers of free text that are capable of producing error-free analyses.

The most common way of dividing tasks between the human developers and the machine is to let the annotator work as a post-checker of the parser's output. Even though such a method is quite straightforward to implement, the method itself may generate difficulties. Firstly, if one begins annotation with parsing, this can lead to a high number of unresolved ambiguities and these ambiguities can make the selection of the correct parse a time-consuming task. A parser that is applied for the purpose of building a treebank should therefore perform disambiguation to ease the workload of the annotators. Secondly, because the work of a post-checker is purely mechanical, there might be a tendency for a checker simply to accept the parser's suggestions without any kind of rigorous inspection. The solution that was applied in the case of, for example, both the treebanks for English and the Basque treebank, was the application of a post-checking tool to the created structures before they were accepted.

Other variants of semi-automatic annotation do exist: the TIGER, TUT, Alpino, and Russian Treebanks apply a method which allows the parser and the annotator to interact. The advantage of this method is that since the errors made by the parser are already corrected by the human operator at the earlier stages of parsing, they have no opportunity to multiply during the later stages. This procedure makes it more probable that the parser will produce a correct analysis. In certain annotation tools, such as those of the Russian and the English D treebanks, the annotator is given the option of adding comments to annotation – a process that makes it easier to inspect doubtful structures. In the annotation tool of the TUT, a special type of relation can be assigned to mark doubtful annotations.

Recent years have seen an increase in collaboration between treebank projects. Thus, for example, the framework developed for the PDT is used in the *Prague Arabic Dependency Treebank* (Hajič *et al*. 2004) and the *Slovene Dependency Treebank* (Erjavec 2005, 2006). Nevertheless, the main problem with regard to current treebanks – insofar as their use and distribution is concerned – is that, instead of reusing existing annotation and encoding schemes, developers have created new ones. Another problem is that those schemes that have already been developed have usually been designed from theory- and even application-specific viewpoints, and are consequently of little use for recycling. When one considers the high costs involved in developing a treebank,[60] it seems obvious that the reusability of tools and formats should be given high priority. Apart from the difficulties that it creates for reuse, the creation of a treebank-specific representation scheme requires the development of a new set of tools for creating, maintaining and searching the treebank. But the existence of exchange formats, such as XCES (Ide & Romary 2003) and TIGER-XML (Mengel & Lezius 2000), already allow multipurpose tools to be implemented and used.

## 6.4 Conclusion

I would summarize the findings of Section 6.1 in the following way: A linguistic resource for comprehensive, full-scale parser evaluation should include the following four kinds of materials: 1) unannotated sentences, 2) annotated sentences, 3) test items for checking grammatical and morphological coverage, and 4) pairs of grammatical and ungrammatical sentences. To my knowledge, no such single unified resource exists for any language. While all these materials are indeed available in English, they are scattered over several resources.

[60] The estimated costs of the PDT, for example, are around USD 600,000 (Böhmová *et al.* 2003).

The genre-dependency of parsers is an established fact (see, for example, Sekine (1997)). Most of these evaluation resources consist of texts from a single genre. This constitutes a deficiency in most of the existing parser evaluations reported in the literature. This deficiency is caused, in turn, by a lack of treebanks divided into genre-specific sub-treebanks. A key issue in available evaluation materials is therefore genre homogeneity. The BC part of the PTB is the only sizeable evaluation resource to reflect such variations in a systematic way. Most English parser evaluations are usually performed on newspaper texts (namely, on section 23 of the PTB).

A further complication is that many parsing models are trained on the same treebank on which they are tested. Parsers therefore come to be applied to texts from numerous other genres without being tested. The obvious question that confronts us in these circumstances is: How well will a parser that performs well on financial texts from the WSJ generalize to other text types? An evaluation resource tailored for parser evaluation should include texts from several genres such as law, biomedicine and prose. The *EASY corpus* of French, with its 4,200 sentences (Paroubek *et al.* 2006), is the only purpose-built parser evaluation resource to reflect such variations in a systematic way.

Based on the findings of Section 6.2, the most promising approach from the point of view of practical implementability for annotating parser evaluation resources appears to be the approach in which the tagsets are organized into a hierarchy. This approach allows for more flexibility than other approaches. Moreover, this type of scheme can be more easily mapped into different types of schemes. The GR scheme is an example of such an approach. It allows for differences in the schemes while still remaining reasonably easy to annotate. In fact, once the tag hierarchy has been defined, no extra work is required from the annotators. But this kind of approach cannot be generally applied to all possible schemes. A hierarchical annotation scheme of this kind could be encoded with one of the abstract annotation schemes. The XML-based models have the advantage of their inbuilt validation (checking documents against the XML schema) and transformation (e.g. *XSL Transformations*) mechanisms.

It is interesting to note that none of the annotation schemes takes into account the possibility of inherently ambiguous sentences that cannot be disambiguated without contextual information that spans over a single sentence. An annotation and encoding scheme that is tailored for syntactic parser evaluation should allow for more than one analysis to be stored for such sentences.

I have used the results of the analysis of the existing linguistic resources and their annotation schemes as the basis for the design of the evaluation treebank and two unannotated evaluation resources represented in the next chapter. One could summarize the important findings as follows: Firstly, a comprehensive parser evaluation has to be based on several types of linguistic resources. Secondly, the linguistic resources must contain texts from several genres. Thirdly, it is desirable to organize the tagset of an annotated linguistic resource for evaluation in the form of a hierarchy. This facilitates the use of the resource in evaluating and comparing different kinds of parsers. An existing XML-based exchange format should be utilized in annotation and encoding in order to allow for a reuse of software tools for browsing and manipulating the resources. If the outputs of the parser to be evaluated are transformed into the same exchange format, the implementation of the evaluation tools is simpler. These issues are discussed in more detail in the next chapter.

# 7 New Evaluation Resources

This chapter introduces three new evaluation resources that were developed on the basis of the findings that I reported in Chapter 6. Section 7.1 deals with the design of the parser evaluation D treebank for Finnish called *FiEval*. Section 7.2 introduces the design and construction of the two new evaluation resources for English called *RobSet* and *Multi-Genre Test Set*.

## 7.1 Developing a Parser Evaluation Treebank for Finnish

FiEval is a treebank for Finnish that is currently under construction. It has been designed especially for the evaluation of syntactic parsers, taggers and morphological analyzers. This treebank consists of both naturally occurring and manually constructed sentences from several genres. The design of the treebank and its annotation and encoding schemes are introduced in Section 7.1.1. Section 7.1.2 describes the purpose-built annotation tool called *DepAnn*.

### *7.1.1 Designing FiEval - A parser evaluation treebank for Finnish*

This section describes the annotation and encoding schemes of the treebank and justifies the choices that were made in its design. It is essential to link the design of a treebank to its intended usage. Certain major design decisions, such as choosing whether or not a treebank will focus on a specific application, will affect other collateral choices. There are a number of major decisions that one has to make when designing a treebank. These decisions affect the content, the type of annotation and the methods and tools that will be used in the construction of the treebank. The *annotation guidelines* articulate the conventions that guide the annotators throughout this process. Figure 7-1 illustrates the process of creating the FiEval treebank.

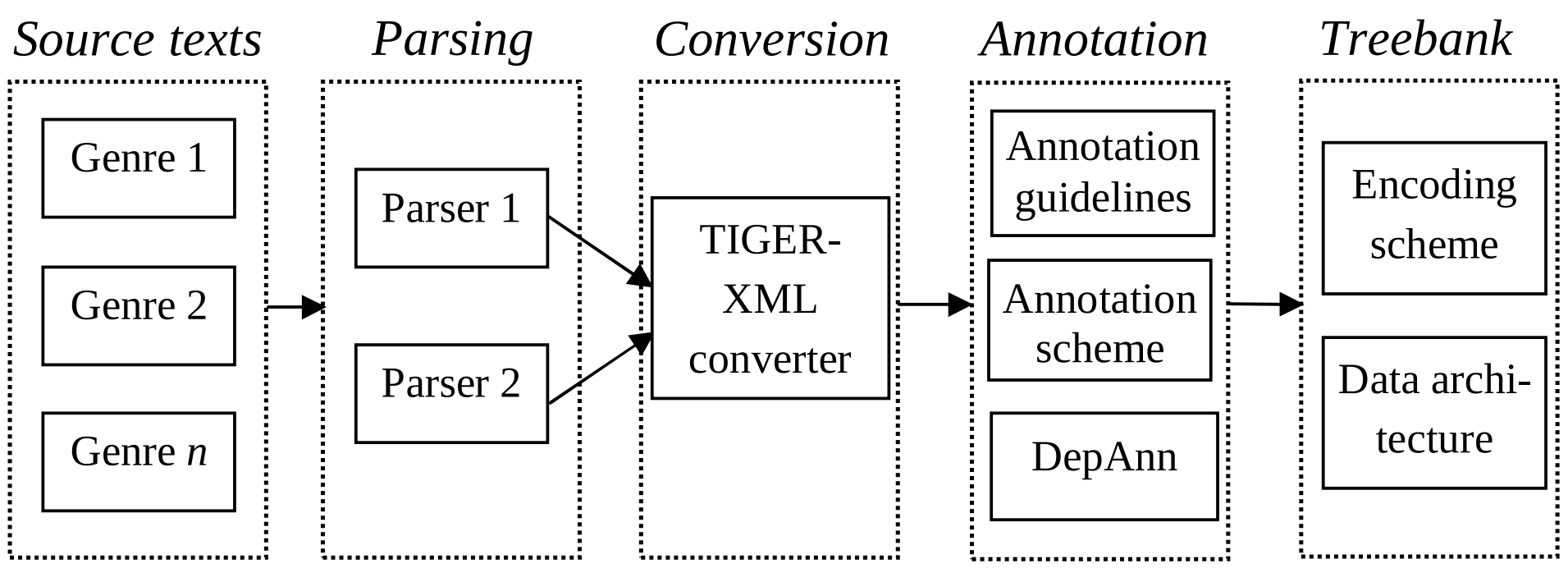

**Figure 7-1**. The process of creating FiEval.

Making a treebank especially suitable for parser and tagger evaluation has implications for the *selection of the content* (naturally occurring text vs. artificially constructed test sentences) and the *design of the annotation and encoding schemes* (so that one might better match the output of the tools that need to be evaluated). Decisions about the *methods and tools for creating the treebank* are important for making the treebank consistent and easy to construct. I will now analyze each of these issues in detail in the subsections that follow (some of the treebank design criteria have been described in Ide and Romary (2003)).

*Selection of the content*

- transcriptions of spoken language or written texts
- specific genre or balanced over multiple genres
- size

All the source materials used in FiEval are written texts. No transcriptions of spoken language have been included. I made this choice because most of the work performed by syntactic parsing focuses on the processing of written texts. It therefore seems logical that a treebank for parser evaluation should, in the first place, focus on the same modality.[61]

A key feature in a treebank which is intended to be used for parser evaluation is the inclusion of texts from several different genres. FiEval currently consists of texts from a work of fiction (a novel) (Gaarder 1994), a newspaper (Karjalainen 1999), and law code texts (European Union 2005, Finlex). Because we in the EdTech group of the University of Joensuu are currently carrying out research into automatic and semi-automatic essay assessment (see for example, Kakkonen & Sutinen 2004), we are particularly interested in the ability of parsers to analyze the kind of text that one will encounter in such a context. These texts consist mainly of student texts such as short answers to assignments, longer free-text responses (essays), and extracts from student theses. Examples of all such texts were sourced from undergraduate and graduate students from the University of Joensuu. Also included is a sub-corpus (referred to as TS) which consists of manually constructed sentences that reflect the main syntactic and morphological phenomena of Finnish.

At the time of writing, the FiEval treebank consists of 3,160 parsed sentences. While a larger treebank is always better, increasing the size of a treebank obviously also increases the cost of its development. While the current size of the treebank is acceptably large for parser evaluation, is not yet adequate for training probabilistic parsing models.

[61] As the parsing of spoken language is one of the future application areas of parsers, it might be worthwhile later on to add spoken language transcripts to the treebank.

*Design of the annotation and encoding schemes*

- phrase structure or dependency
- type of theory support: theory-neutral, theory-specific, theory-supporting
- using existing annotation and encoding schemes or designing new ones
- designing the tagsets: the set of tags for POS, morphological and syntactic annotation
- the type of the encoding scheme: markup-language, ASCII/text file
- data architecture: the files/database and annotations are interspersed throughout the document containing the primary text or are stored in one or more additional documents linked to the primary text

Because the Finnish language is characterized by a relatively free word order, and because existing parsers for Finnish are implemented in DG frameworks, D representation becomes an obvious choice for a parser/tagger evaluation treebank for Finnish. Moreover, by many (such as Yamada and Matsumoto (2003)) the D annotation is regarded as being more intuitive and easier to understand than PS annotation. This feature of course makes for faster and less error-prone treebank construction. Because semantic dependencies are embedded in the syntactic dependencies, the D description also offers a more straightforward interface between syntactic and semantic representations.

Instead of creating a theory-specific treebank, I designed the annotation scheme in such a way that it will allow the treebank to be used for evaluating several different types of parsers. This was partly achieved by organizing the POS tagset into a hierarchy. In that sense, the treebank is theory-supporting. The two grammar formalisms that were used as the basis for the design, CG and FDG – the only two frameworks in which parsers have been implemented thus far for Finnish– are closely related. From this point of view, one might argue that the treebank is theory-specific.

Designing an annotation format involves several steps. Firstly, the annotation scheme consisting of the morphosyntactic labels and the syntactic tags together with the general structural principles for the annotation needs to be designed. Then it is necessary to select an encoding scheme – the physical representation of the annotation information in a physical document with tags. Finally, a *data* *architecture* has to be chosen for the primary text and its annotations. This will dictate whether the treebank is stored in a database or files and whether annotations will be interspersed throughout the document containing the primary text or stored in one or more additional documents linked to the primary text. The annotation scheme of FiEval consists of word-level information (the word form, the lemma, and POS and morphological tags) and syntactic description. The encoding is based on TIGER-XML and has been designed in terms of the guidelines proposed by the Nordic Treebank Network (Kromann 2005). The annotation and encoding schemes and the data architecture of FiEval will be discussed in more detail in Sections 7.1.1.3. to 7.1.1.5.

*Construction method and tools*

- annotation method: manual, semi-automatic, automatic
- checking the annotation: checking the levels of annotation (POS and syntax, for example) separately or simultaneously
- annotation tool: the use of an existing tool or the development of a new one
- the morphological analyzer and parser: the selection of appropriate ones

I already pointed out above that the manual construction of syntactic trees is both slow and error-prone. It is unfortunate because there are no tools available for creating high-quality treebanks fully automatically. Treebanks are therefore usually created by means of a combination of automatic and manual processing. In addition to parsers/taggers for creating the initial structures, several types of resources are needed for semi-automatic treebank construction. These include the annotation guidelines on which the work of the annotators is predicated, and an annotation tool that one applies for checking and correcting automatically created structures as well the validity of structures that have already been created.

FiEval currently consist of sentences that have been automatically tagged by two morphological analyzers/parsers: *CG parser for Finnish*, (FINCG) (Karlsson 1990) and *FDG parser for Finnish* (FI-FDG) (Tapanainen 1997). A few sentences have been manually checked as part of the process of implementing the purpose-built annotation tool *DepAnn* (which will be introduced in Section 7.1.2) and defining the annotation format.

**7.1.1.1 Related work**

The Alpino Treebank of Dutch is similar to FiEval because it too focuses on parser evaluation (Van der Beek *et al.* 2002). Apart from having this feature in common,

FiEval and Alpino both utilize the TIGER Treebank of German (Brants *et al.* 2002) as the main inspiration for the annotation format. While FiEval only applies the TIGER-XML (Mengel and Lezius 2000) encoding scheme, both the annotation and encoding schemes of Alpino are similar to those of the TIGER Treebank. In addition to the TIGER Treebank, there are two Danish treebanks (Kromann 2003, Bick 2003) that also support the TIGER-XML format. The morphosyntactic tagset of FiEval has been organized in such a way that inexact tag matches can be accounted for in the process of parser/tagger evaluation. The idea was originally used by Carroll *et al.* (2003) in the syntactic annotation scheme based on GRs.

#### 7.1.1.2 The Finnish language

Finnish has a rich morphology. Nouns, for example, have fourteen cases. Finnish is an agglutinative language in which grammatical markers and endings are joined to a word root. For example, the word "autoissanikinko?" (which literally means "In my cars, too?"), can be broken down into the following structural components: root ("auto") + plural marker "i" + inessive "ssa" + 1st person sing. possessive "ni" + enclitic particle "kin" + question marker "ko".

As in English, the subject-verb-object sequence is the word order that occurs most frequently in Finnish (Karttunen & Kay 1985). Because it possesses such a rich inflectional system, word order in Finnish is relatively free. Because of its inflection, the function that a word plays in a sequence of words can be often understood without any additional reference to its specific position in a sentence. For instance, all the six permutations of the three words *hän* "he" (sg nom), *söi* "eat" (past sg 3rd), and *kalan* "fish" (sg acc) are grammatical sentences in Finnish. This looseness of ordering constraints is not, however, found in all syntactic categories in Finnish. For example, the order of constituents in NPs in Finnish is almost as fixed as it is in English. Furthermore, although all the possible word orderings in the example above are grammatical, they are not identical. While all six sentences express the same proposition, they are used in different discourse modalities to place emphasis on different aspects of the proposition.

#### 7.1.1.3 Morphosyntactic annotation scheme

Because of the complexity of Finnish morphology, the morphosyntactic annotation scheme of FiEval is rather extensive. The morphosyntactic tagsets of FINCG and FI-FDG are similar, and the tagset of FiEval closely resembles both of these. The scheme is defined as follows:

**Definition 7-1.** Word-level annotation in FiEval.

1. The word form WORD.
2. The base form LEMMA.
3. The part-of-speech tag POS ∈ {V, N, A, AD-A, ADV, PRON, PRE, PSP, NUM, CS, CC, CC>, INTJ, FW}.
4. Morphological tags MORPH ∈ {SG, PL, NOM, GEN, PTV, …}.

The word-level tagset is designed in a way that allows for inexact matches between the two parsers and the treebank. The morphosyntactic annotation scheme consists of 14 POS and 58 morphological tags. The POS tags and their descriptions are given in Table 7-1 (below). Figure 7-2 illustrates the organization of the POS tagset. FW is the only tag not present in either parser's tagset. All FI-FDG tags and all except two FINCG tags (ABBR for abbreviations and Q for quantifiers) have a matching tag in the tag hierarchy. Table 7-3 gives all the morphological tags for nouns.

**Table 7-1**. The POS tags in FiEval. KEY: coord. conj. = coordinating conjunction.

| Tag | Name | Example | Transl. | Tag | Name | Example | Transl. |
|---|---|---|---|---|---|---|---|
| *V* | Verb | mene | go | *N* | Noun | pallo | a ball |
| *A* | Adjective | suuri | big | *AD-A* | Adjective adjunct | melkein | almost |
| *ADV* | Adverb | nopeasti | quickly | *PRON* | Pronoun | minä | I |
| *PRE* | Preposition | ennen | before | *PSP* | Postposition | takana | behind |
| *NUM* | Numeral | yksi | one | *CS* | Subordinate conjunction | kun | when |
| *CC* | Coord. conj. | ja | and | *CC>* | Multipart coord. conj. | sekä..että | both...and |
| *INTJ* | Interjection | Hei! | Hey! | *FW* | Foreign word | word | - |

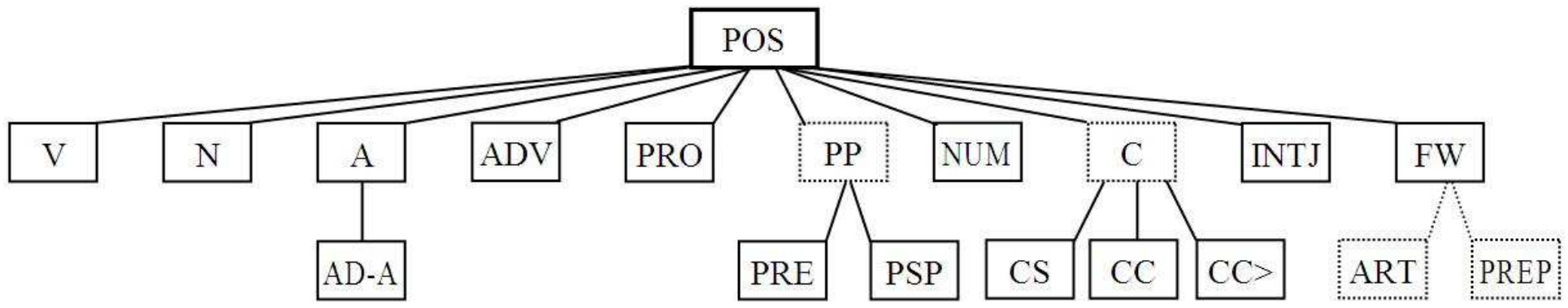

**Figure 7-2**. The part-of-speech tagset. The boxes with solid lines denote tags used in the treebank. The boxes with broken lines represent tags used by either of the parsers although they are not included in the treebank annotation. KEY: PP = post- or preposition, C = conjunction, ART = foreign article, PREP = foreign preposition.

**Table 7-2**. Morphological tags for nouns (N) in FiEval.

| Tag | Name | Example | Translation |
|---|---|---|---|
| *SG* | Singular | pallo | a/the ball |
| *PL* | Plural | pallot | (the) balls |
| | | *CASES* | |
| | | GRAMMATICAL CASES | |
| *NOM* | Nominative | pallo | a ball |
| *GEN* | Genitive | pallon | ball's |
| *PTV* | Partitive | palloa | some/(without a) ball |
| | | LOCATIVE CASES | |
| *INE* | Inessive | pallossa | in a ball |
| *ELA* | Elative | pallosta | out of a ball |
| *ILL* | Illative | palloon | into a ball |
| *ADE* | Adessive | pallolla | on a/by means of a ball |
| *ABL* | Ablative | pallolta | from a/the ball |
| | | ABSTRACT LOCATIVE CASES | |
| *ESS* | Essive | pallona | as a ball |
| *TRA* | Translative | palloksi | transformed into a ball |
| | | OTHER CASES | |
| *ABE* | Abessive | pallotta | without a ball |
| *CMT* | Comitative | palloineen | with a ball/with the balls |
| *INS* | Instructive | laivoin | by means of a boat |
| *PRO* | Prolative | meritse | by sea |
| | | POSSESIVE SUFFIXES | |
| *1SG* | 1st pers. sing. | palloni | my ball |
| *2SG* | 2nd pers. sing. | pallosi | your ball |
| *3* | 3rd pers. s./plu. | pallonsa | his/her/their ball |
| *1PL* | 1st pers. plur. | pallomme | our ball(s) |
| *2PL* | 2ns pers. plur. | pallonne | your ball |
| | | CLITICS | |
| *HAN* | -han/hän | pallohan | the ball (adds politeness) |
| *KAAN* | -kaan/-kään | ei pallokaan | not even the ball |
| *KIN* | -kin | pallokin | also the ball |
| *KO* | -ko | palloko | [do you mean] the ball? |
| *PA* | -pa/pä | pallopa | the ball (adds emphasis) |
| *S** | -s | pallopas | the ball (adds emphasis) |

*Often implies an explicit reversal of the other discussant's presupposition.

#### 7.1.1.4 Syntactic annotation scheme

The syntactic annotation scheme of FiEval follows the D description employed by the FI-FDG parser. The dependencies are marked in edges connecting the words. The formal definition of the syntactic annotation scheme is given in Definition 7-2 below. The D relations are labeled with 31 D types described in Table 7-3.

**Definition 7-2.** Sentence-level annotation in FiEval. $T_d$={MAIN, SUBJ, OBJ,…} is the set of dependency relation labels (tags). Let *S* be an annotated sentence, consisting of:

1. Sequence $\delta = (\alpha_1, \alpha_2, \ldots, \alpha_n)$ of words $\alpha_1 \ldots \alpha_n$, where each $\alpha_i$ is annotated with word-level information.
2. A partial function $D_\delta:\{1,2,\ldots,|\delta|\} \rightarrow T_d$, where $D(i, j) = l \in T_d$ if and only if there is a dependency link from the word *i* to the word *j*, tagged with the label $l \in \{MAIN, SUBJ, OBJ \ldots, ORD\}$. The root of the sentence is denoted with 0.

**Table 7-3**. The dependency tags in FiEval. The head of the relation and all the elements dominated by it are denoted by italics. KEY: comp. = complement; postmod. nom. = postmodifying nominal; attrib. adv. = attributive adverbial.

| Tag | Name | Example | Translation |
|---|---|---|---|
| *MAIN* | Main verb | *Potkaise* palloa. | *Kick* the ball. |
| *SUBJ* | Subject | *Hän* potkaisi palloa. | *He* kicked the ball. |
| *OBJ* | Object | Ostin *pallon*. | I bought *a ball*. |
| *DAT* | Indirect obj. | Syötätkö sen *minulle*? | Are you going to pass it *to me*? |
| *COMP* | Subject comp. | Pallo on *pyöreä*. | The ball is *round*. |
| *OC* | Object comp. | Hänet palkattiin *valmentajaksi*. | He was hired *as the coach*. |
| *PM* | Preposed marker | Syötin, *että* tekisit maalin. | I passed *so that* you can score a goal. |
| *PHR* | Phrase | Häntä *pidettiin kiinni*. | He *was held*. |
| *COPR* | Copredicative | Hän oli *tuomarina*. | He was acting *as the referee*. |
| *VOC* | Vocative | *Pele*, syötä pallo! | *Pele*, pass the ball! |
| *TMP* | Time | Pelasimme *eilen*. | We played *yesterday*. |
| *DUR* | Duration | Pelasimme *kaksi tuntia*. | We played for *two hours*. |
| *FRQ* | Frequency | Voitimme *kolmesti*. | We won *three times*. |
| *QUA* | Quantity | Se nousi yli kolme *prosenttia*. | It went up over three *percent*. |
| *MAN* | Manner | Hän taklaa *rajusti*. | He tackles *hard*. |
| *LOC* | Location | Pele asuu *Brasiliassa*. | Pele lives *in Brazil*. |
| *SOU* | Source | Liverpool FC on *Englannista*. | Liverpool FC is *from England*. |
| *GOA* | Goal | Potkaisin pallon *maaliin*. | I kicked the ball *into the goal*. |
| *PUR* | Purpose | Pelaamme *voittaaksemme*. | We play *to win*. |
| *PTH* | Path | Pallo tuli *postitse*. | The ball came *by mail*. |
| *RSN* | Reason | *Miksi* et syöttänyt? | *Why* didn't you pass? |
| *CND* | Condition | Hävetkää, *jos ette voita*! | Shame on you, *if you don't win*! |
| *META* | Clause adv. | Tein *melkein* maalin. | I *almost* scored a goal. |
| *QN* | Quantifier | Pystytkö tekemään *viisi* maalia? | Can you make *five* goals? |
| *ATTR* | Premodifying nominal | *Joukkueen* kannattajat huutavat. | The spectators *of the team* are shouting. |
| *MOD* | Postm. nom. | Se, *joka pystyy*, tekee maalin. | The one *who can*, will score. |
| *AD* | Attrib. adv. | Peli oli *todella* rankka. | The game was *really* tough. |
| *CC* | Coordination | Taklasin puolustajan *ja kaksi hyökkääjää*. | I tackled a defender *and two strikers*. |
| *INS* | Instrument | Löin palloa *mailalla*. | I hit the ball *with a bat*. |
| *COM* | Comitative | Pelaa *hänen* kanssaan! | Play *with him*! |
| *ORD* | Ordinance | *Sitten* potkaisin palloa. | *Then* I kicked the ball. |

**Example 7-1**. The parse tree and annotation of the sentence: "Liverpool pelaa hyvin" (Liverpool is playing well).

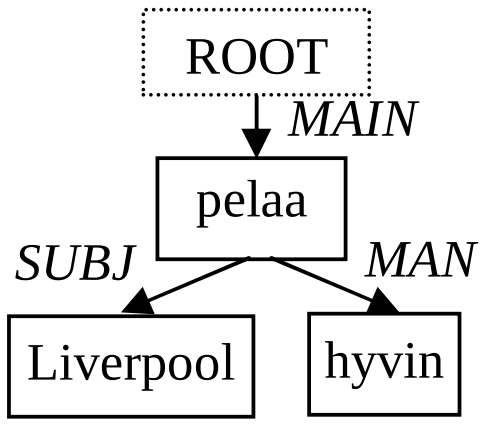

δ={
{WORD=Liverpool, LEMMA=Liverpool, POS=N, MORPH=SG NOM}
{WORD=pelaa, LEMMA=pelata, POS=V, MORPH=ACT PRES SG3}
{WORD= hyvin, LEMMA=hyvä, POS=ADV, MORPH=-}
}

$D_{\delta}(0, 2)$=MAIN, $D_{\delta}(2, 1)$=SUBJ, $D_{\delta}(2, 3)$=MAN

### 7.1.1.5 Encoding scheme and data architecture

In order to avoid having to implement browsing and searching tools for the treebank, I made a decision to base the encoding on an existing encoding format. After a detailed analysis of XML-based encoding schemes (which are described in Section 6.2.2 above and in paper [1]) had been made, TIGER-XML was selected. TIGER-XML offers an XML-based representation that is capable of encoding different kinds of corpus and treebank annotations (Mengel & Lezius 2000). Syntactic categories, POS, lemma and other word information, are described as attributes in the terminal nodes in DAGs. The nonterminals encode D links.

I selected TIGER-XML for the following reasons. Firstly, it is flexible and extensible enough to accommodate different treebank annotation types, both D- and PS-based. Secondly, several well-implemented tools such as the *TIGERSearch* viewing/query tool and *TIGERRegistry* indexing tool (König *et al.* 2003) already exist for TIGER-XML. These are capable of transforming many well-known corpus and treebank formats such as the SUSANNE and the PTB into TIGER-XM. Thirdly, TIGER-XML has already been successfully used for D annotation by other treebank projects such as the TIGER Treebank (Brants *et al.* 2002) and the Danish treebank (Kromann 2003). Fourthly, there are explicit specifications available on how to encode D structures TIGER-XML (Kromann 2005). Finally, the architecture of the annotation and encoding scheme of a treebank should allow for refining the encoded information at a later stage, both in width and depth (Skut *et al.* 1998). Adding depth refines the existing representation. TIGER-XML allows the user to define the attributes for nodes in a syntactic tree, thus enabling one to create a flexible depth of representations. "Increasing width" refers to adding new levels of annotation. An extension called SALSA/TIGER-XML (Erk & Padó 2004) has been devised that allows for incorporating semantic role annotation into a TIGER-XML encoded treebank. This same mechanism could be used for adding levels other than semantics.

In TIGER-XML the annotations are interspersed throughout the documents. But it is easy to extract the original content of a document if one needs to do so. The data architecture of the treebank is organized in the following way. Texts are divided into sub-corpora on the basis of genre. Each sub-corpus is then stored in a separate TIGER-XML file. One of the features of the treebank design is that it allows for the inclusion of parallel annotation for a part of the sentences. This feature is useful for annotating ambiguous sentences. When one evaluates a syntactic parser, it is fair to accept any of the ambiguous analyses as correct for sentences that cannot be disambiguated solely on the basis of syntax.

#### *7.1.2 DepAnn – An annotation tool for dependency treebanks*

Constructing a treebank – even with a semi-automatic method – is a labor-intensive undertaking. Efficient tools play a key role in keeping down the costs of treebank development and in allowing developers to create larger and better quality treebanks. A crucial component in semi-automatic treebank creation is the annotation tool. A well-designed and well-implemented tool can play an important role in making the work of annotators easier. A user can use an annotation tool to browse, check, and correct the parser’s output, and create structures from scratch. In some existing tools, the annotations are automatically checked against inconsistencies before they are saved to the treebank. The user is also able to add comments to the structures or mark them as doubtful where necessary.

After carrying out an investigation into existing annotation methods and tools, such as *GRAPH* (Böhmová *et al*. 2003), *Abar-Hitz* (Díaz de Ilarraza *et al*. 2004), *Annotate* (Plaehen & Brants 2000), *DTAG* (Kromann 2003), *CDG SENtence annotaTOR* (SENATOR) (White 2000), I came to the conclusion that none of the available tools were able to satisfy all the needs and requirements. The tools were either not suitable for D annotation or were incompatible with any common XML-based encoding schemes, the user interface was unsuitable, or the tool did not offer all the necessary functions. In addition to that, there are no annotation tools that are capable of showing outputs from several parsers for the purpose of assisting the annotator to make choices. A decision was therefore taken to design and implement an annotation tool that *did* contain all the desired characteristics.

#### 7.1.2.1 Design principles

It was necessary to conduct a prior analysis of existing annotation tools in order to get a clear idea of what I would need in the system that I intended to develop. On the basis of that analysis, I decided to include the following key features:

1. *Support for an existing XML encoding scheme*

   The use of an existing encoding scheme would make the system reusable. Existing tools that support the same scheme could be used for browsing, manipulating and searching the annotated treebanks.
2. *Both text and graphic display and manipulation of parse trees*

   It is necessary for any annotation tool to be able to show sentence structures in a visual display. Any such graphic display should also preferably be interactive so that the user can manipulate the structures wherever necessary. There are, however, certain annotation tasks for which a text view of the structure would be more suitable.
3. *An interface between morphological analyzers and parsers for constructing the initial trees*

   In order to generate the trees for human inspection and modification, the annotation tool must have an interface for a morphological parser, a POS tagger and a syntactic parser. This tool should be capable of simultaneously using outputs from several tools to guide the annotator's decisions.
4. *An inconsistency checker for both structures and encoding*

   The annotated sentences that will be saved to the treebank need to be checked for tagging inconsistencies. Apart from an XML-based validation of encoding, the inconsistency checker needs to be able to alert the annotator to several other kinds of mistakes such as mismatching combinations of POS and morphological tags, a missing main verb, and fragmented, incomplete parses.
5. *Menu-based tagging*

   In order to speed up the annotation process, tags need to be chosen from a pre-defined tag list rather than by the annotator typing the tag labels manually. Menu-based tagging is not only efficient: it also diminishes the error rate by eliminating the errors generated by typos in the labels. Keyboard shortcuts for selecting appropriate tags should also be provided for more advanced users.
6. *A commenting tool*

   For facilitating later revisions that might need to be performed by other annotators, a user should be in a position to add comments to the annotated structures. A user should also be able to mark a sentence as either *ready* or *unfinished* so as to make it easier to locate sentences needing further revision.

The predominant design principles, apart from making the annotation process faster and less error-prone, were that the tool should be reusable and modifiable. The system was therefore designed in such a way that the modules for processing the treebank output and input were kept separate from the structure viewing and manipulation modules. This makes it easier to modify the tool. Support for an existing encoding scheme is a crucial reusability feature of any treebanking software. The selection of the format was initially narrowed down by the decision that the format should be XML-based – because XML offers a set of validation capabilities – so that it could automatically check for encoding inconsistencies.[62]

### 7.1.2.2 Main functionality

Figure 7-3 illustrates the main frame of DepAnn's user interface.

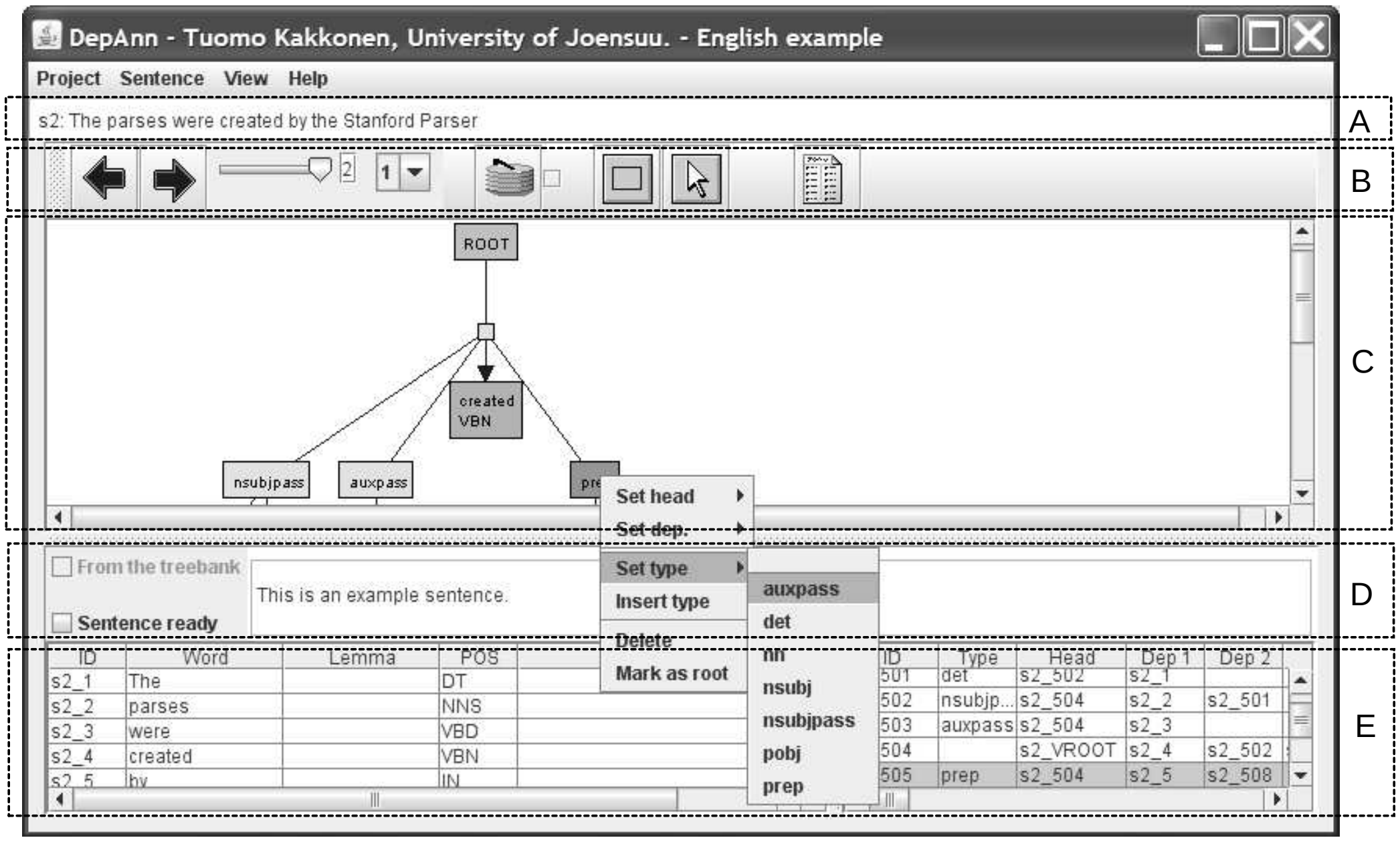

**Figure 7-3.** The main frame of the DepAnn tool.

The main groups of functions are indicated in Figure 7-3 by boxes A...E. The text field in the area bordered with box A shows the sentence being annotated in the raw text format. Area B is a toolbar with controls for treebank browsing (buttons for showing the next and the previous sentence and a slidebar for browsing), checking and saving the sentence, and modifying the tagsets. In area C, the user can graphically manipulate the structure by changing the values on nodes representing the words and D links, and by removing, adding and rerouting the links between the nodes. Area D consists of the revision functions. The user can also mark the sentence as *ready*: this indicates that further revision is not needed. In addition, the user can utilize the comment field to write notes about the sentence structure. Box E frames the tables for text-based structure manipulation and viewing. The parser and tagger outputs for aiding the annotation decisions are shown in a separate resizable and customizable dialog. In a computer system with multiple monitors, for example, the dialog can be placed into a separate desktop. In the current version, the user can select which parser's output is used as the initial tree for correction and modification.

[62] As described in Section 7.1.1.5, TIGER-XML was selected as the encoding scheme because of the several advantages that it offers. Since TIGER-XML is a general model of treebank encoding, it would be possible to show and manipulate PS structures with DepAnn. A decision was nevertheless made not to include both PS and D structures in the design of the tool because it was felt that too general a design would hamper the efficiency of D annotation. The visualization functions and user interface are therefore tuned for manipulating D structures.

In the DepAnn tool, the structure to be annotated is represented to the user in text and graphic formats in order to offer the use of the best possible option for his or her needs. Because the text and graphic views are fully integrated, changes applied in the graphic view immediately affect the text view, and vice versa. The user interface is also customizable to suit the task and the annotator's preferences. A user can add comments on annotations such as reminders about problematic parts in the sentence structures. Completed trees can be marked as ready, thereby indicating that no further inspection or modifications are needed. Outputs of several parsers and POS taggers can also be applied in parallel. These allow the annotator to compare the outputs before making any annotation decisions. To be able to use the output of a parser in DepAnn, a converter is needed to transform the output from the parser- or tagger-specific format to the format used by DepAnn. TIGER-XML (Mengel & Lezius 2000) is used as the input format for the structures obtained from the automatic tools, as well as the output format for the annotated treebank. For internal data representation, the TIGER-XML structures are transformed into Java objects.

The annotation process that uses DepAnn begins with processing the treebank texts with one or more parsers and taggers. It then applies a converter to the outputs in order to transform the tool-specific output into TIGER-XML. After the conversion, the annotator can view the parsed structures. The annotator can select the parser output to be used for creating the initial trees and modify these.

Once the user stops editing a sentence, the program performs an automatic consistency check to validate sentence structure, annotation, and encoding. In the first place, it runs a series of checks to verify that the sentence has a main verb and a root, that all the words have word form and lemma information and morphosyntactic tags, that the sentence is not fragmented, and so on. Secondly, if the first series of checks is passed, the program transforms the sentence into TIGER-XML and validates it against the XML schema to identify any possible errors in the encoding. If any such errors appear, the user is alerted. The user can select which checks should be run by modifying the system set-up.

#### 7.1.2.3 Implementation details

The annotation tool is implemented in Java. Because Java is platform-independent, the system can be used in any environment in which Java is available. The system consists of three main components: the interface to parsers and taggers, the annotation tool itself, and the output module. I used two freely available open-source packages, *OpenJGraph* (Salvo 2006) and *TIGER API* (Demir *et al*. 2006), for developing the system – although both had to be modified before they could be used as a part of DepAnn. TIGER API, a Java API for TIGER-XML, is used for input and output processing. The graphic annotation manipulation functionality was built on top of OpenJGraph. The annotation tool uses *Java Database Connectivity* (JDBC) both for storing the outputs from the parsing and tagging tools, user comments and information on completed sentences. Because of this, the *MySQL* database currently being used can be replaced by any other JDBC-compatible database.

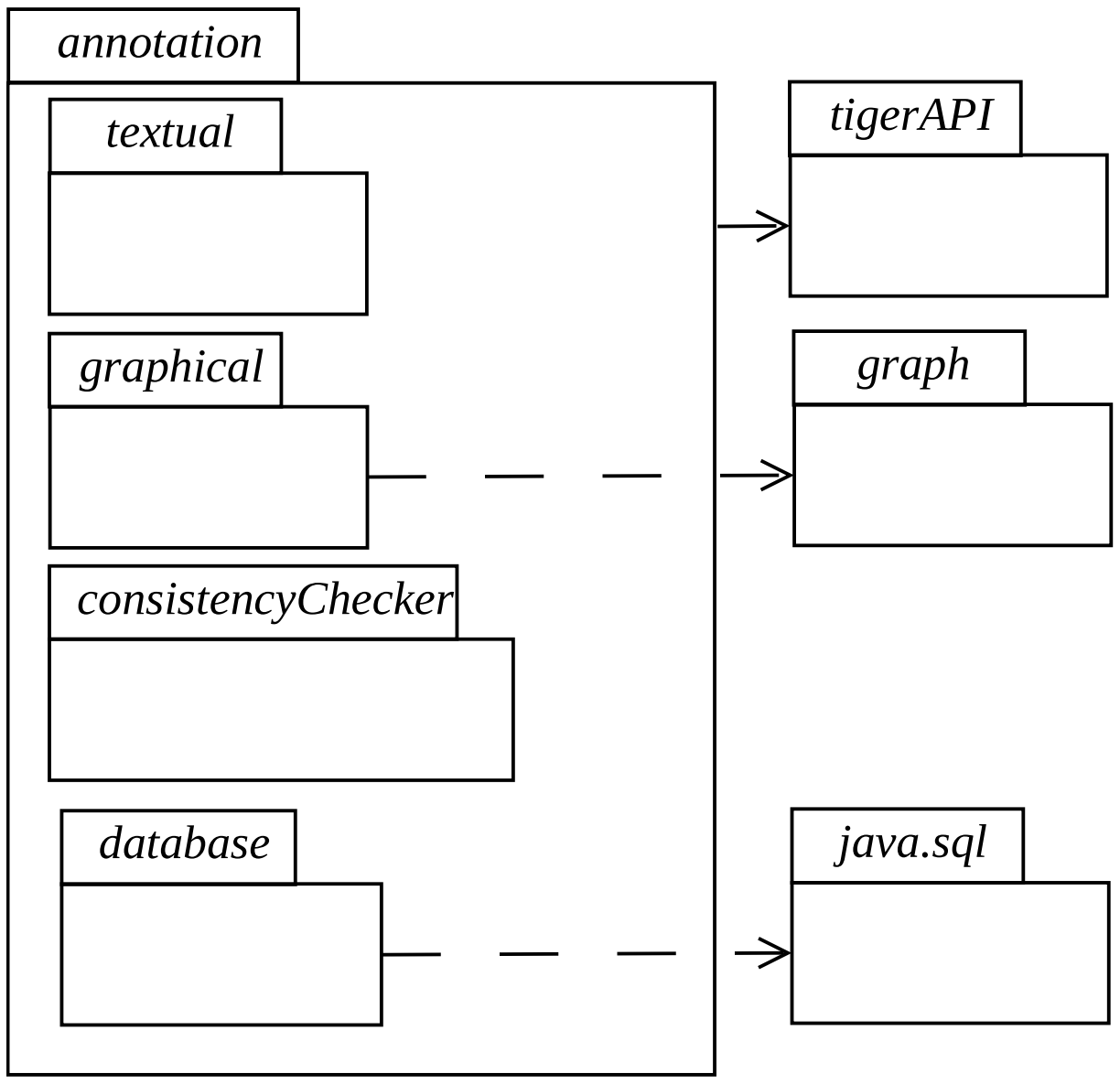

**Figure 7-4**. The main Java packages of DepAnn.

## 7.2 Evaluation Resources for English

In this section I will describe the two linguistic resources for English that I developed as part of this research. Sections 7.2.1 and 7.2.2 respectively describe the design and construction of the Multi-Genre Test Set and RobSet.

### *7.2.1 Multi-Genre Test Set*

The *Multi-Genre Test Set* (MGTS) is a test set that comprises 826,485 sentences in English. It contains six subsets that cover the following genres: newspaper, legislation, fiction, non-fiction, religion and biomedicine.

#### 7.2.1.1 Main design principles

MGTS aims to provide a sizeable corpus of texts for evaluating parsers' performance on diverse text genres. Because the corpus is divided into genre-specific subsets, it allows a user to measure the effects of genre variance on the performance of parsers. The texts differ not only with regard to genre but also in terms of discourse function and intended audience. While most of the texts address a general readership, the biomedical texts and some of the texts in the non-fiction data set are written for experts.

#### 7.2.1.2 Developing the resource

The text materials were obtained from the following sources: the *Leipzig Corpora Collection* (LCC) (Quasthoff *et al*. 2006), the *Oxford Text Archive* (OTA) (Oxford Text Archive 2006), the *Project Gutenberg* (PG) (Project Gutenberg 2006), *GENIA* (GEN) (GENIA Project 2006), the *Yapex Corpus* (YAP) (Yapex Project 2006) and the *Aligned Hansards of the 36$^{th}$ Parliament of Canada* corpus (HAN) (Natural Language Group of the USC Information Sciences Institute 2001). All the text sources were available for use without charge.

The test set was constructed in the following way:

- Deletions were made to the texts. These included the removal of some poems and song lyrics as well as sentences in languages other than English.
- Text normalizations were carried out. These included the removal of underscores (_), markup (e.g. XML tags) and other codes.
- The texts were divided into sentences by using the sentence tokenizer of the *Maximum Entropy Part of Speech Tagger* (MXPOST) tool

(Ratnaparkhi 1996, Ratnaparkhi 2007). A Java-based post-processor was implemented for correcting some of the systematic problems observed in MXPOST's output.

- Automatically tokenized texts were checked manually for inconsistencies. A considerable number of errors left by the automatic tools were corrected.[63]
- The texts were collected into sub-corpora of between 6 and 12 MBs size.
- Two separate versions were made of each file to be parsed: one for parsers that take raw text as input and another for those that take the input text in the PTB format. A small Java tool was implemented to produce the two versions automatically.
- Finally, the texts were tagged with MXPOS for use with the parsers that make use of a POS tagged input.

Table 7-4 shows the subsets of MGTS. There are 15,385,855 tokens in total (with an average length of 18.6 tokens per sentence). The average number of tokens per sentence in the sub-corpora varies between 15.9 and 27.1.

**Table 7-4**. The materials in the Multi-Genre Test Set.

| Genre | Description | No. of sent. | Avg. length | Sources |
|---|---|---|---|---|
| *Legislation* | Proceedings from the Canadian Parliament (Hansard) | 390,042 | 17.2 | HAN |
| *Newspaper* | Newspaper texts from Financial Times, Wall Street Journal & Associated Press | 217,262 | 19.5 | LCC |
| *Fiction* | Novels from the 20th and 21st centuries | 97,156 | 15.9 | OTA, PG |
| *Non-fiction* | Non-fiction books from the 20th and 21st centuries | 61,911 | 21.9 | OTA, PG |
| *Religion* | The Bible, the Koran, the Book of Mormon | 45,459 | 27.1 | OTA, PG |
| *Biomedicine* | Abstracts from biomedical journals | 14,655 | 21.6 | GEN, YAP |
| *TOTAL* | | 826,485 | 18.6 | |

### *7.2.2 RobSet*

*RobSet* consists of 443 test items, each of which contains a pair of erroneous and correct sentences. Each of the erroneous sentences contains one to three misspelled words.

[63] The manual checking and correction took approximately one week of working time. Checking the whole test set sentence-by-sentence would take up to several months. I estimated the error frequency to be 0.4% by sampling a total of 1450 randomly selected sentences. I considered the error rate to be low enough for the purpose to which the test set was to be used in this research.

#### 7.2.2.1 Main design principles

It is clear that as the level of noise in the inputs increases, the performance of a parser degrades. The effect of the noise on the performance can be measured by parsing sentences with successively more spelling mistakes and observing the consequent effect on the performance of a parser. RobSet enables one to make these kinds of evaluations.

#### 7.2.2.2 Developing the resource

I began the test set construction by selecting 19 sentences from a public domain web page. I then altered one, two or three words per test sentence, and this gave a total of 443 test sentences – 255 with one error and 94 with two and three errors respectively. The length of each of these sentences was between 5 and 36 words, and the average length was 16.3 words per sentence. I then manually introduced misspellings into the sentences by deleting, adding and transposing characters, permitting only one edit operation per word. The character additions were based on the keyboard proximity of letters in order to simulate errors in naturally-occurring texts. Since the purpose was not to introduce structurally distorted sentences, only alterations that did not create an acceptable word were permitted. Table 7-5 shows examples of sentences from the test set.

**Table 7-5**. Example sentences from RobSet. The errors that were introduced are indicated by in italics.

| Correct sentence | Noisy sentence |
|---|---|
| Your username is not logged. | *Yoru* username is not logged. |
| Please e-mail suggestions for improvements. | Please *e-mil* suggestions for improvements. |
| A text-only browser such as Lynx is great for viewing almost all the pages at this site. | A text-only browser *usch* as Lynx *si* great for *viewin* almost all the pages at this site |
| Once you have files loaded into the cache, reaccess is speedy. | Once you have files *loadsd itno* the cache, reaccess is *speedu*. |

## 7.3 Conclusion

I have introduced three resources for the evaluation of natural language parsers, and have discussed the design principles and the construction methods as well as the contents of the resources involved. The two resources for English are applied in Section 10 for evaluating parsers for English. As the treebank for Finnish currently consists of automatically parsed sentences, it can not yet be used reliably for parser evaluation. Manual checking of a treebank containing thousands of sentences is beyond the scope of this research.

# III EVALUATION METHODS AND TOOLS

## 8 Analysis of Existing Methods and Tools

It is an extremely laborious task for a human being to evaluate individual parser outputs manually, and it is also one that is likely to give rise to errors. There are some methods such as *TextTrees*, that have been developed to assist the work of inspection (Newman 2005).[64] Experiments have shown that formats such as TextFree can speed up the manual evaluation process. In spite of this, it is necessary to use automatic evaluation methods for systematic parser evaluation because of the major drawbacks inherent in manual evaluation practices. The first of these drawbacks is that manual inspection is both slow and error-prone. It would be extremely cumbersome in practice, if not impossible, to use manual inspection for comparing parsing systems that use different output formats. The second drawback is the possibility of introducing bias along with human evaluators. Whenever an evaluator decides whether or not a parse is appropriate or acceptable, he or she introduces bias by making such a judgment (Bod 1998). In those cases where an evaluator might judge an analysis to be appropriate, he or she might have assigned a completely different analysis to the sentence if he or she had not already seen the parser's analysis.

In addition to the linguistic resource that is used as a comparative material, metrics and measures are also needed for automatic evaluation. An *evaluation metric* allows one to compare a system's performance after improvements, as well as to compare the performance of different systems. Mappings often need to be applied to parser outputs and evaluation resources in order to make them sufficiently similar for metrics to be applied. *Measures* define the way in which the results are reported.

The quality of a parser can be approached from several perspectives. The most commonly used perspective in parser evaluation is the preciseness of the parses produced by the systems. Most of the work that has been done on parser evaluation has in fact concentrated solely on this particular point of view. Preciseness evaluation methods are described in Section 8.1 below. *Coverage* (Section 8.2) refers to the proportion of the structures in the gold standard to which the parser assigns one or more parses. A parser's ability to produce an error-free or only a slightly altered output from input sentences containing errors is referred to as *robustness*. Robustness evaluation methods are discussed in Section 8.3. *Efficiency* (Section 8.4) refers to the speed with which a parser performs analyses. It can be extended to cover the use of memory. Existing evaluation tools are discussed in Section 8.5 of this chapter.

[64] In TextTrees, parser output trees are converted into unlabeled indented strings. These trees contain a minimal amount of bracketing to ease inspection work.

### 8.1 Preciseness Evaluation

Preciseness refers to a parser's ability to correctly analyze grammatical structures. These structures, depending on the type of evaluation, may be either constituents, dependencies or sentences. Preciseness is measured by comparing the analyses in the gold standard to the parses in the parser output.

**Definition 8-1**. Gold standard.

*Gold standard* is a 2-tuple GS=(S,A) where

1. $S=(s_1,s_2,\ldots,s_n)$ is a finite sequence of grammatical structures, i.e. constituents, dependency links or sentences.
2. $A=(a_1,a_2,\ldots,a_n)$ is a finite sequence of analyses. For each $i$, $1 \le i \le n$, $a_i \in A$, is the analysis of $s_i \in S$.

**Definition 8-2**. Parser output in evaluation.

Let GS=(S,A) be a gold standard. Let *P* be a parser.

1. *Parser output* $O(P,GS)=(P(s_1), P(s_2),\ldots P(s_{|S|}))$ is a sequence of analyses such that $P(s_i)$ for each $i$, $1 \le i \le n$ is the analysis assigned by parser *P* for sentence $s_i \in S$.

Let GS=(S,A) be a gold standard and O(P,GS) a parser output for sentences *S* in the gold standard. Preciseness evaluation is carried out by comparing each element in O(P, GS) to each element in *A*. *GS* consist of two sets: *TP* (true positives) and FN (false negatives). *TP* is the subset of *GS* whose analyses in O(P,GS) match with the corresponding analysis in *A*; the parser produced a correct analysis for these sentences. The members of set *FP* (false positives) do not match in O(P, GS) and *A*. This means that the parser was not able to produce a correct analysis for these sentences. Set *FN* consists of the elements in *GS* that do not have an analysis in O(P, GS), i.e. they comprise the set of sentences for which the parser was unable to produce an analysis. $TP \cup FN$ constitutes the gold standard. The parser output is formed of *TP* and *FP*. Figure 8-1 illustrates these concepts.

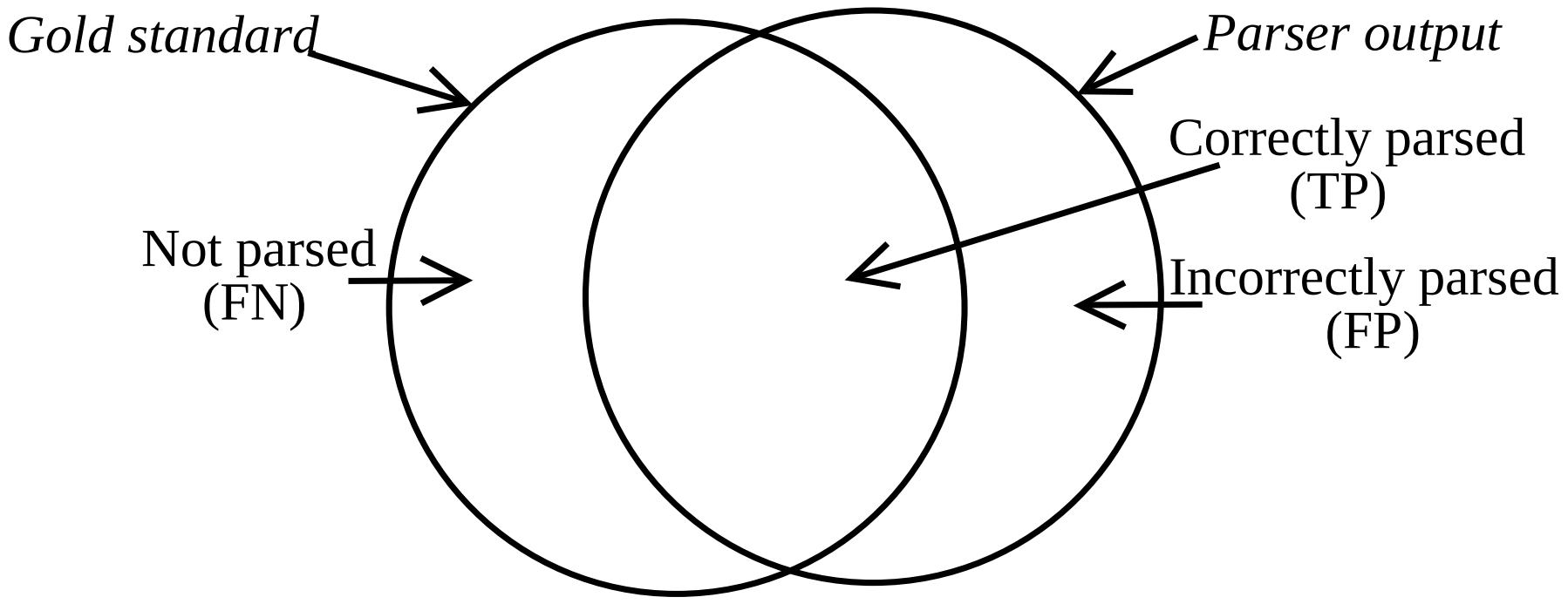

**Figure 8-1**. The sets of analyses in parser evaluation. TN is the set of all the sentences in the language L not included in the gold standard or parser output, called the true negatives.

**Example 8-1**. Let GS=(S, A) be a gold standard. Let O(P,GS) be a parser output for *S*. The following table shows for each sentence $s_1 \ldots s_5$ the gold standard analysis, the parser output and the group to which (TP, FP or FN) the parser analysis belongs.

| S | A | O(P,GS) | Set |
|---|---|---|---|
| s1 | a1 | a1 | TP |
| s2 | a2 | NULL | FN |
| s3 | a3 | a3 | TP |
| s4 | a4 | a4 | TP |
| s5 | a5 | p5≠ a5 | FP |

The most common measures used in preciseness evaluation are *precision* (P) and *recall* (R), which are defined as follows:

**Definition 8-3**. Precision and recall.

1. $\mathrm{P} = \frac{|TP|}{|TP \cup FP|}$
2. $\mathrm{R} = \frac{|TP|}{|TP \cup FN|}$

*F*-score is a combined measure of precision and recall that facilitates the comparison of evaluation results. *F*-score can be defined as in Definition 8-4. Figure 8-2 illustrates examples of *F*-scores for certain precision-recall pairs.

**Definition 8-4**. *F*-score.

$$F\text{-score} = \frac{2PR}{P+R}$$

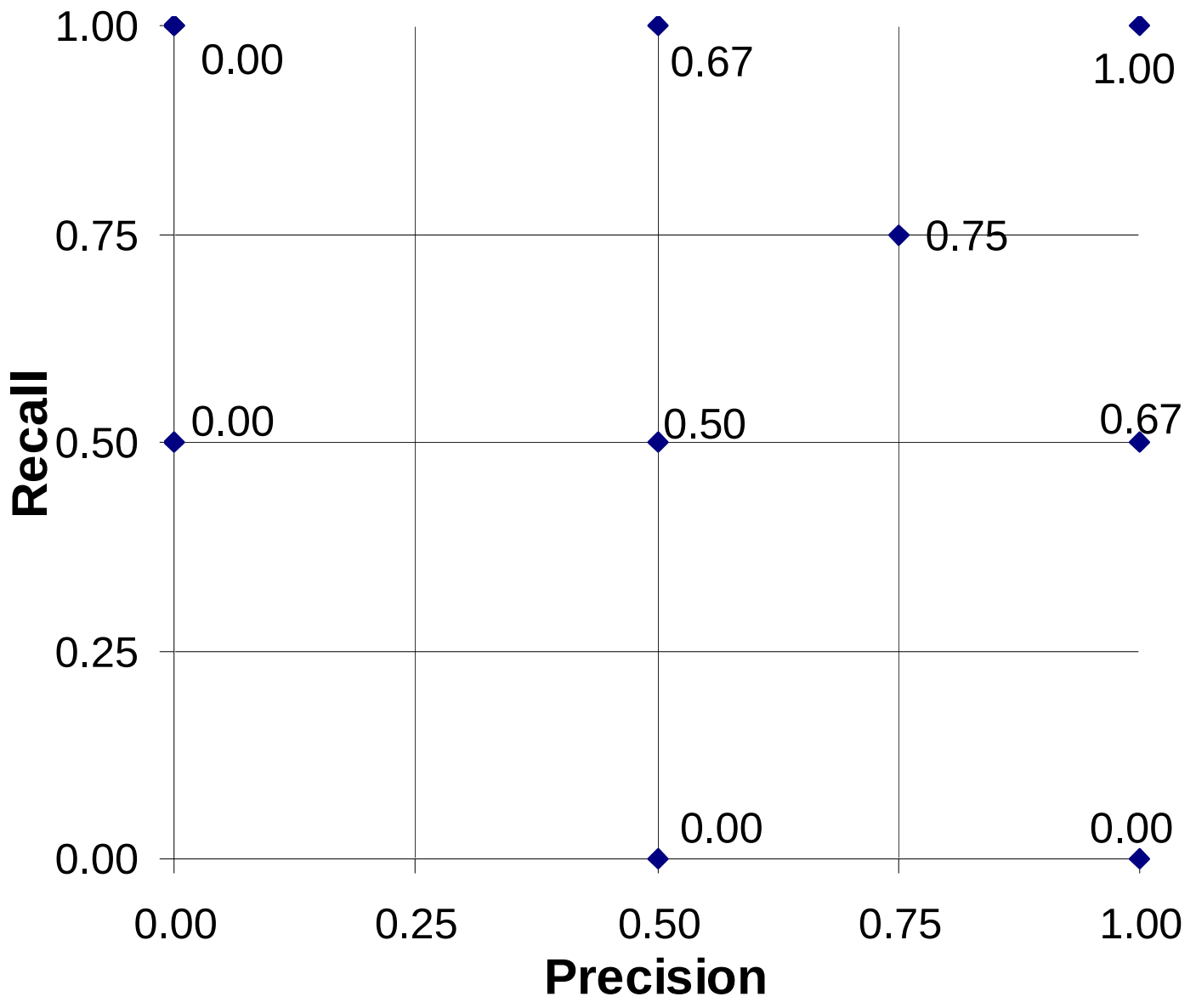

**Figure 8-2**. Precision-recall pairs with their corresponding F-scores.

### *8.1.1 Phrase structure based metrics*

#### 8.1.1.1 PARSEVAL

Perhaps the most widely used parser evaluation metric is known as *PARSEVAL* (Black *et al*. 1991). It uses PS bracketings to compare the structures in a parser output and a treebank. For the evaluation, parser output ($TP \cup FP$) and the gold standard ($\mathrm{TP} \cup \mathrm{FN}$) are represented in bracketed string format.

**Example 8-2**. Bracketing representation of the PS parse tree shown in Figure 3-2. Phrase boundaries are represented with brackets spanning an interval [*i*, *j*], where *i* indicates the index of the first and *j* the last word in the phrase. There are therefore five phrase boundaries in the sentence used as the example: [0,5], [0,1], [2,5], [3,5] and [4,5].

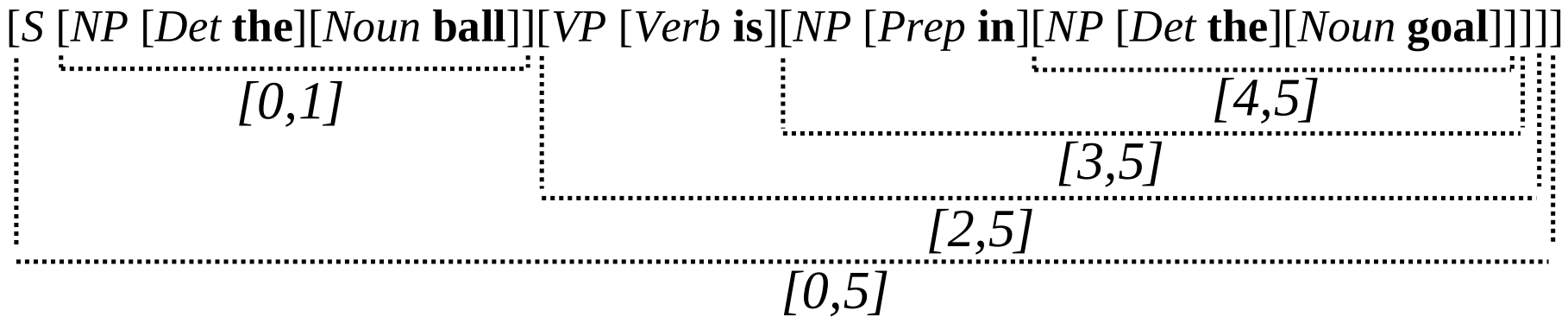

In PARSEVAL, precision and recall measures are operationalized on the basis of the brackets: |TP| is the number of bracket pairs for which the parser output and gold standard match. |FP| is the number of bracket pairs in which the analyses do not match. |FN| is defined as the number of the brackets in the gold standard for which the parser was unable to produce an analysis. While *unlabeled PARSEVAL* compares only the brackets, in *labeled PARSEVAL* two analyses match if and only if both the brackets and the labels (POS and syntactic tags) match. In addition to precision, recall and F-score, the *number of crossing brackets* is used as an evaluation measure. The number of crossing brackets is defined as the mean number of bracketed sequences in which the parser output overlaps with the gold standard structure. Figure 8-3 illustrates the idea of the crossing brackets measure. Example 8-2 shows an unlabeled PARSEVAL evaluation.

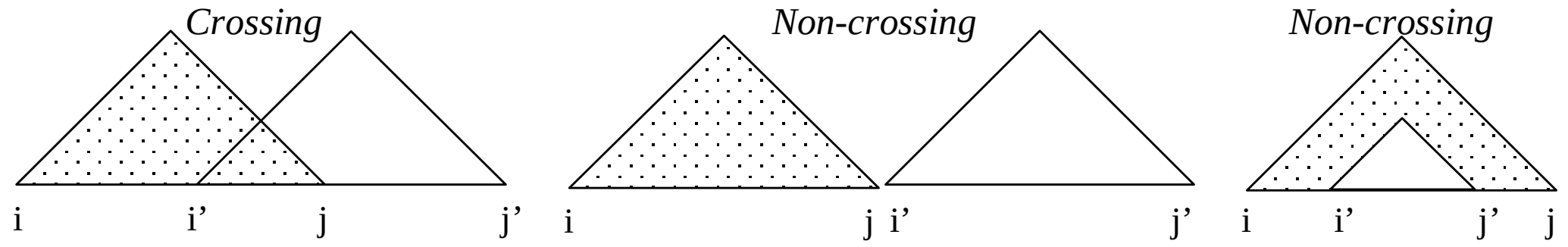

**Figure 8-3**. Non-crossing and crossing brackets. The phrase boundaries [*i*, *j*] and [*i’*, *j’*] are boundaries in the gold standard and the parser output respectively. Pair [*i*, *j*] [*i’*, *j’*] is defined as a pair of crossing brackets if they overlap, that is, if $i < i' \leq j < j'$.

**Example 8-2**. An example of PARSEVAL measures. A is the gold standard structure and B the parser output (adapted from (Lin 1998)).

*A) [[He [hit [the post]]] [while [[the all-star goalkeeper] [was [out [of [the goal]]]]]]]*

*B) [He [[hit [the post]] [while [[the [[all-star] goalkeeper]] [was [out of [the goal]]]]]]]*

| | |
|---|---|
| precision | = 75.0% (9/12) |
| recall | = 81.8% (9/11) |
| F-score | = 78.3% |
| crossing brackets | = 1 pair |

### 8.1.1.2 Maximal projections of heads

To eliminate some of the problems in PARSEVAL bracketing-based evaluation, Ringger *et al*. (2004) have proposed a method that they claim is suitable for evaluating and comparing both probabilistic (treebank) and rule-based (non-treebank) parsers. Their idea is to concentrate on the aspects that are common to different types of parsers. The method is therefore directed at comparative evaluation. Instead of using bracketed structures, the authors apply *maximal* *projections of heads* (MPHs)[65] to obtain a more suitable format for comparing parsers across frameworks.

#### 8.1.1.3 Leaf-ancestor metric

*Leaf-ancestor* (LA) metric is based on evaluating the similarity of *lineages* of individual terminal elements in the parser output and the gold standard (Sampson (2000), Sampson & Babarczy (2002)). A *lineage* is the sequence of node labels for the nodes on the path from a terminal to the root. A left bracket is inserted in the lineage of a terminal element immediately before the label of the highest nonterminal beginning with that element. A right bracket is inserted in the lineage of a terminal element immediately after the label of the highest nonterminal ending with that element.

The evaluation is carried out by comparing each word's lineage by using the *edit distance* measure (Levenshtein 1966). *Edit distance* is defined as the minimum number of insert, delete and replace operations needed to transform one string into the other. While deletions and additions have the cost one, LA metric uses a modified edit distance measure in which the cost of replacing a node label is defined by a function that sets a value between 0…2 to each replacement. This reflects the fact that when two grammatical categories are entirely dissimilar, they actually count as two separate errors (namely, failing to recognize the right tag and falsely assigning another tag). However, when the categories are more or less similar, the lineages should be recognized as less far apart. They are then assigned value 1. A sentence-level LA measure is defined as the average of the LA values for the words in the sentence.

[65] Maximal projections are projections of a head that unite with the start symbol or with a non-head daughter of a rule For example, in the structure [*X*'' [specifier] [*X*' [*X*] [complement]]] *X* is the head of the phrase and *X*' and *X*'' projections of *X*. More specifically, the top node *X*'' is referred as the maximal projection and *X*' as the intermediate projection of *X*. Unlabelled precision and recall are used as evaluation measures to further abstract away from differences between parser output formats.

**Example 8-3**. Leaf-ancestor metric. The analyses A and B are from the gold standard and the parser output respectively. The column "LA" gives the LA score for each word. The LA score for the whole sentence is 0.88.

*A) [S [NP [N Liverpool]] [VP [V is] [VP [V playing] [ADVP [ADV well]]]]]*

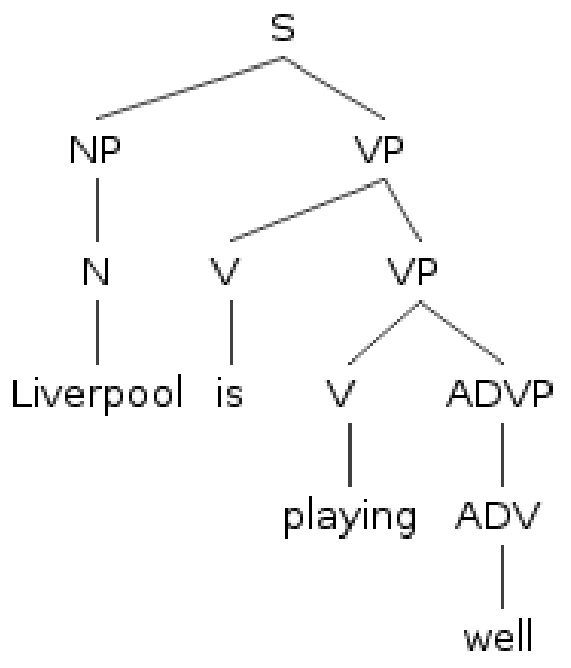

*B) [S [NP [N Liverpool]] [VP [V is] [VP playing well]]]*

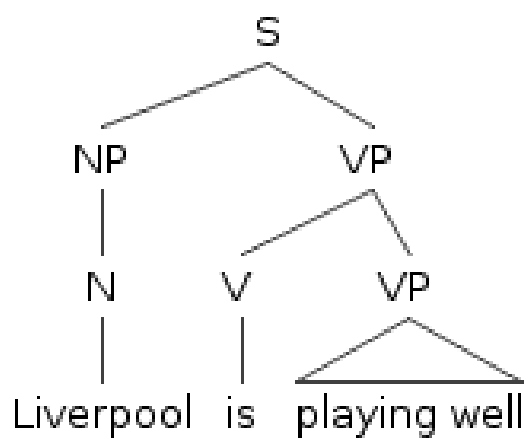

| A | LA | Word | B |
|---|---|---|---|
| N NP ] [ S | 1.0 | Liverpool | N NP ] [ S |
| V ] [ VP S | 1.0 | is | V ] [ VP S |
| V ] [ VP VP S | 0.8 | playing | [ VP VP S |
| ADV [ ADVP VP VP S ] | 0.73 | well | VP VP S ] |

### *8.1.2 Dependency-based metrics*

Several methods have been proposed that use D structures as the basis of evaluation instead of PS bracketings. In addition to the developers of D parsers, D-based evaluation methods have been widely applied within the PCFG parsing research community (by Collins (1999), for example), and by developers of CCG parsers (such as Clark and Hockenmaier (2002a, 2002b)). In addition to the evaluation methods based on D treebanks and D output, several methods that are based on mapping or mixed D/PS representation have been proposed. Lin's (1996, 1998, 2003) model maps the treebank and parser output into D structures. The *Relation Model* of Srinivas *et al*. (1996, 1998) – referred to hereinafter as RM – aims at combining PS and D representations by adding D relations between phrasal constituent chunks. The GR evaluation scheme (Carroll *et al*. (1998, 1999)) uses a corpus annotation scheme with GRs between heads and dependents.

#### 8.1.2.1 Pure dependency measures

Table 8-1 summarizes the pure D evaluation measures[66] that will be described in this section.

[66] The term "pure D evaluation" is used in this research to refer to the methods that are used for comparing D trees with one another – without using any mappings from other types of structures

**Table 8-1**. Evaluation measures based on D structures (adapted from Nivre & Scholz (2004)). All the measures are reported as percentages of the nodes or sentences that are correct.

| Name | Measure | Origins |
|---|---|---|
| *Unlabeled attachment score* (UAS) | Nodes that are assigned the correct head or no head if it is the root | Eisner (1996), Collins *et al*. (1999) |
| *Labeled attachment score* (LAS) | Nodes that are assigned the correct head and D label or no head if it is the root | Nivre *et al*. (2004) |
| *Dependency accuracy* (DA) | Non-root nodes that are assigned the correct head | Yamada & Matsumoto (2003) |
| *Root accuracy* (RA) | Sentences in which the root is recognized correctly | |
| *Complete match* (CM) | Sentences whose unlabeled D structure is completely correct | |

**Example 8-4**. Pure dependency evaluation measures. *A* is the gold standard parse and *B* the parser output. The roots are denoted with the bold face font. For example, UAS is calculated as 6/7 because out of the total of seven word nodes, six of which have been assigned the same head (no head for the root) in both analyses. Only the word "him" has been assigned a wrong head. Only three of the labels match in the D links. LAS is therefore calculated as 3/7. In DA score the root is excluded and it is defined as 5/6. Although the root matches (RA = 1/1), the whole sentence has not been correctly analyzed (CM = 0/1).

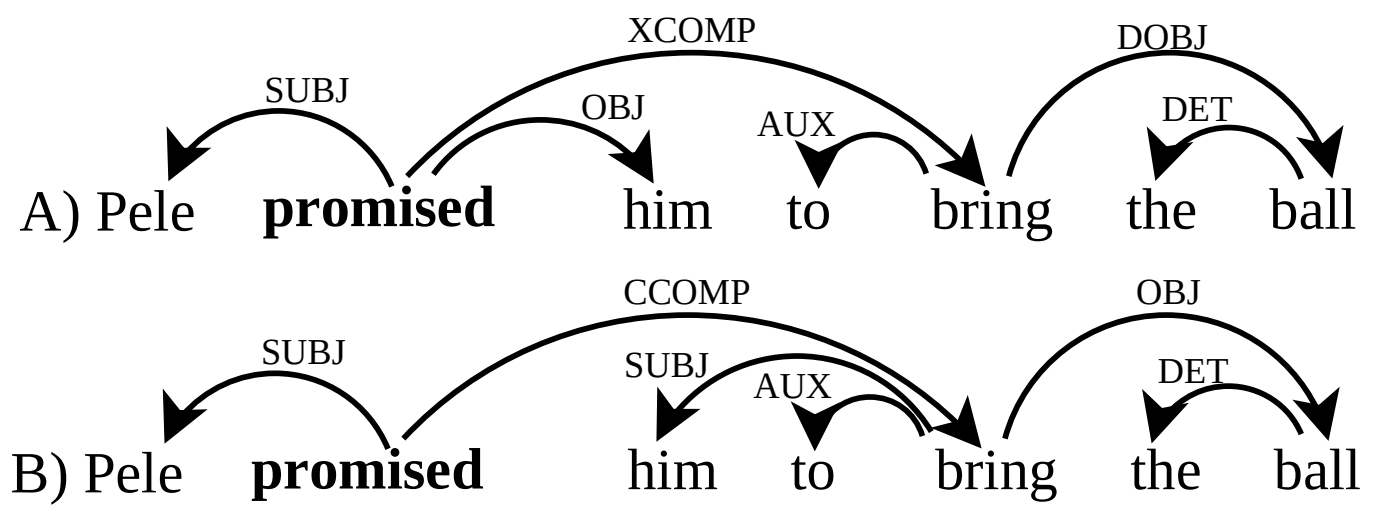

UAS = 85.7% (6/7)
LAS = 42.9% (3/7)
DA = 83.3% (5/6)
RA = 100% (1/1)
CM = 0% (0/1)

*The Shared Task on Multilingual D Parsing* at the 10th Conference on Computational Natural Language Learning (STMDP-CoNLL-X) (Buchholz & Marsi 2006, CoNLL-X Shared Task 2006) was the first evaluation to use standard

(such as Lin's model, which will be described in Section 8.1.2.2) or specific annotation format for the evaluation resources (such as in GR model in Section 8.1.2.4).

metrics and test data for a wide range of D parsers, including parsers for languages other than English. [67] LAS was the official measure used for ranking the parsers. UAS was used for an additional measure for system comparison and error analysis.

#### 8.1.2.2 Lin's mapping model

As I have already noted in Section 3.1.3, PS trees can be mapped, with certain restrictions, into D trees. Lin (1996, 1998, 2003) has proposed an evaluation method based on PS/D mappings. The mapping algorithm is based on the one proposed by Magerman (1994). Lin (1998) applied the method to evaluate the D-based MINIPAR parser on the SUSANNE corpus by first transforming the corpus from PS to D structures. A D relation in Lin's model consists of a modifier, a head and an optional label indicating the type of relationship between these two.

A D output or a mapped PS output is scored on the basis of either labeled or unlabeled D relations that classify each into one of the following four categories: (1) A "correct" word modifies the same word in the evaluation resource and parser output or else it does not modify any other words in either structure. (2) "Incorrect" words modify a different word in the parser output than in the gold standard. (3) If the word does not modify any word in the output, but does so in the gold standard, it is referred to as "missing". (4) In the opposite case of a missing link in the gold standard parse, the modifiee is referred to as being "spurious".

Precision and recall are calculated over the D links. Precision measures the percentage of D links in the parser output that match the D link in the gold standard parse. Recall is defined as the proportion of the D links in the gold standard that match the parser output. Lin (1996, 1998, 2003) does not report F-scores.

[67] The shared task in the 10th CoNLL provided test sets for D parsing in 13 languages.

**Example 8-5.** Lin's D-based evaluation method. This example represents a labeled evaluation. The correct words are "Pele", "promised", "to", "bring", and "ball". There is one incorrect word, "him", and a missing word, "the".

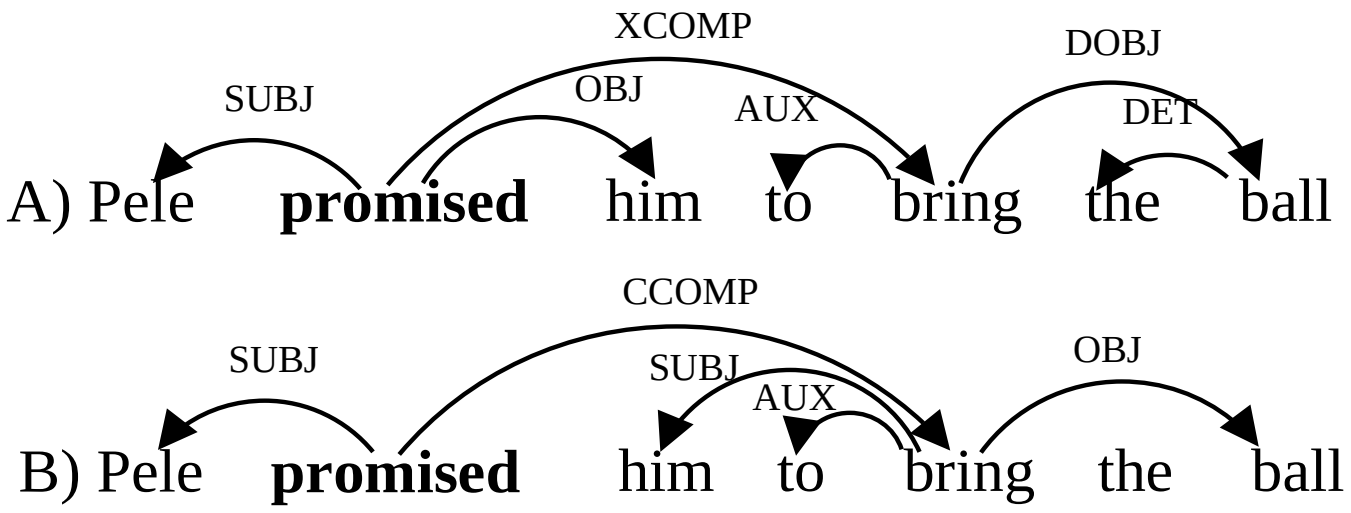

Precision = 60.0% (3/5)
Recall = 50.0% (3/6)

### 8.1.2.3 The relation model

The RM model (Srinivas *et al.* (1996, 1998)) is based on the idea that evaluation can be defined as measuring how well a parser can express a certain relation. Srinivas *et al.* (1998) propose two derivatives from this general model. In chunk-based evaluation, the relation is the one between the words starting and ending the chunk. In the D-based evaluation, the relation is defined as "word *x* depends on word *y*".

These two models can be applied in parser evaluation in the following way. The parser outputs are flattened into chucks and the evaluation is first carried out by using these chunks. D-based evaluation is next used to check the correctness of the internal structure of the chunks. Evaluation is carried out on the basis of the links between these. The results are reported by using either labeled or unlabeled precision, recall and F-score measures. Srinivas *et al.* (1998) represent an evaluation in which the output from a TAG parser and PTB structures used as the gold standard are compared firstly on the basis of chunks, and, secondly, on the basis of dependencies between words.

### 8.1.2.4 The grammatical relations metric

The GR metric is based on the GRs annotation scheme that was introduced in Section 6.2.4 (Carroll *et al.* 1998, Briscoe & Carroll 2006). Two GRs are considered to match if the relation assigned by the parser is on the same level or – in the case of *mod*, *sub* and *clausal* relations – one level apart in the hierarchy. A *clausal* relation, for example, matches both *xcomp* and *ccomp* relations. Furthermore, the *type* slot in the *mod*, *iobj* and *clausal* relations can be unspecified in the parser output. Recall and precision over GR structures are used as evaluation measures (Carroll *et al*. 1998).

**Example 8-6**. An example of GR measures. A is the gold standard structure and B the parser output (adapted from Lin (1998)). Precision and recall are both 1.0 in this case, since the *dobj* and *obj* reltations and *ncsbuj* and *subj* relations are on adjacent levels in the GR hierarchy (see Figure 6-6).

| ***A*** | **B** |
|---|---|
| (*dobj* playing well) | (*obj* playing well) |
| (*aux* _ playing is) | (*aux* _ playing is) |
| (*ncsubj* playing Liverpool _) | (*subj* playing Liverpool _) |

### *8.1.3 Analysis*

In this section I will offer an analysis of existing preciseness evaluation metrics on the basis of the methods and measures applied, the type of linguistic resources needed, and suitability for evaluating specific types of parsers. Table 8-2 summarizes the properties of the preciseness evaluation methods mentioned above.

**Table 8-2**. Comparison of preciseness evaluation metrics described. The column headed "Basis" indicates whether the metric is based on either PS or D structures. The two columns below the heading "Aim" indicate whether the metric is aimed at evaluating PS or D parsers, or both. Column "P/R" indicates whether the metric uses recall and precision as the evaluation measures. Column "Other" lists the other evaluation measures used by the metric. The last column of the table indicates whether the method relies on some existing annotation scheme (*e*), and whether it uses its own scheme (*o*) or mappings between annotation and output schemes (*m*).

| **Metric** | **Basis** | **Aim** | | **Measures** | | **Scheme** |
|---|---|---|---|---|---|---|
| | | *PS* | *D* | *P/R* | *Other* | |
| *Unlabeled PARSEVAL* | PS | Yes | No | Yes | F-score, no. of crossing brackets | e |
| *Labeled PARSEVAL* | PS | Yes | No | Yes | | e |
| *MPH* | PS | Yes | (Yes)* | Yes | No. of crossing brackets, no. of matching sentences | m |
| *LA metric* | PS | Yes | (Yes)** | No | Edit distance | e |
| *UAS, LAS, DR, CM, RA* | D | No | Yes | No | Percentage of correct words/D links | e |
| *Lin's model* | D | Yes | Yes | Yes | - | m |
| *Relation Model* | D&PS | Yes | Yes | Yes | F-score | o |
| *GR metric* | D | Yes | Yes | Yes | - | o |

*Ringger *et al.* (2004) claim that the approach generalizes to D structures – a claim remains as yet untested in practice. **While Sampson and Babarczy (2002) claim that this metric could be used for evaluating D trees, no such evaluations have as yet been reported in the literature.

### 8.1.3.1 PS-based evaluation

One of the advantages of PARSEVAL is that the evaluation can be based on a relatively undetailed treebank. PARSEVAL also provides one with a means to compare parsers by using rather different types of PS output schemes. Several criticisms of PARSEVAL have nevertheless been documented. Srinivas *et al.* (1996, 1998) point out, firstly, that parsers generating detailed analyses are penalized by the PARSEVAL precision measure if the comparison is based on an undetailed treebank, and, secondly, that PARSEVAL is unsuitable for evaluating partial or D parsers. Lin (1998) also argues that a crossing brackets measure counts a single bracketing error more than once in some cases. PARSEVAL is most commonly combined with the PTB as the gold standard, and this may cause serious problems. Because the PTB trees are flat and have few brackets in them, the number of crossing brackets is likely to be low; the less structure is assigned to a sentence, the fewer are the possibilities for error. This makes it easier to obtain high precision and recall scores. One can predict that attempt to generalize PARSEVAL to D structures would face serious problems.

While the method proposed by Ringger *et al*. (2004) makes PARSEVAL-style evaluation suitable for a wider range of parsers, such an approach would not be without drawbacks. Firstly, in order to perform MPH evaluation, both the parsers to be evaluated and the evaluation resource need to have an annotation scheme that marks the heads. This is not true with, for example, the PTB. This means that the heads have to be automatically marked – a process that might introduce errors into the resultant structures. Secondly, as Ringger *et al*. themselves point out, this approach cannot be generalized to all kinds of structures, for example *small clauses*. Thirdly, although Ringger *et al*. claim that the approach generalizes to D structures, this claim remains as yet untested in practice.

Sampson and Babarczy (2002) claim that the LA metric is better than PARSEVAL because it allows imperfect matches to be made between the parser output and the gold standard. Furthermore, LA accounts for the severity of the mismatch by applying the function that determines the replacement costs of each label pair. The LA metric is moreover better suited to locating parsing errors. While the LA metric is based on comparisons on the terminal level, PARSEVAL deals with global scores (bracket matches). LA assigns every terminal with a path to the root: this allows detailed error analyses to be carried out. Sampson and Babarczy suggest that LA could be used to identify configurations (particular words and structures) that are regularly associated with low scores. This would enable parser development to be focused on problematic areas.

#### 8.1.3.2 D-based evaluation

One may justify the use of D-based evaluation schemes by pointing to the fact that semantic dependencies are embedded in the syntactic dependencies. One can therefore argue that the results of a D-based evaluation are more meaningful than those that rely on PSs. Buchholz and Marsi (2006) made the following observation on pure D evaluation measures based on the experiences of the STMDP-CoNLL-X. Firstly, there is little difference in the ranking between the parsers that occur when one uses UAS or DA instead of LAS. Secondly, one can observe very little difference in scores or rankings when scoring is performed on all tokens (including punctuation) instead of only on word tokens.

However, one needs to be cautious when one uses pure D evaluation methods. When evaluating her probabilistic CCG parser using LAS, for example,

Hockenmaier (2003) observed that the results were not directly comparable to the evaluation results of Collins's PCFG; CCG categories encode subcategorization information and are more specific than those of Collin's (Hockenmaier 2003).[68] In addition to this, parsers use different sets of D types. Hockenmaier therefore argues that the scores based on unlabeled D relations are the only ones that can be compared across parsers that are based on different grammar formalisms.

Lin's precision measure is equivalent to the DA measure. The main advantage of Lin's metric is his use of PS/D mappings: this makes it applicable to a wide range of parsers. The problem with Lin's approach resembles the central problem of PARSEVAL: part of syntactic information is lost in the mapping, and some detail in the evaluation is also lost. Moreover, as I have already noted in Chapter 3, mapping from dependencies to PS structures is one-to-many: there are many possible trees that can be generated for a given D structure. Non-projective D structures can also not be mapped into single PS trees.

Srinivas *et al.* (1996, 1998) claim that the RM model is suitable for inter-system comparisons. This model, however, remains untested in comparative evaluations. In addition to this, the details of the metric are rather vaguely represented by Srinivas *et al.* (1996, 1998). Why this approach should be preferred to other models based on D/PS mappings, is not entirely clear. The two most problematic parts of the evaluation scheme appear to be the definition and automatic identification of chunks – as well as the fact that both the evaluation resource and parser output in most cases need to be mapped onto the RM scheme. Because parsers do not produce a chunked output, these need to be introduced automatically. On the other hand, if a treebank used in evaluation is of the PS type, mappings need to be done so that the second, word-word based evaluation phase can be carried out.

Evaluation using the GR scheme is in many ways similar to Lin's method. The main difference is that the GR scheme defines a specific inventory of D relations. Because the relations are organized into a hierarchy which enables parsers using varying levels of detail in their output to be compared to a GR-annotated treebank. Using GR metric requires the parser to be evaluated to identify heads and dependents. The main disadvantage of the GR scheme is that it requires the construction of a custom-built test set. GRs also lack uniformity across languages

[68] In order to get a D link correct, a CCG parser needs to identify whether the dependent is an adjunct or a complement. A PCFG treebank parser does not need to make this distinction.

(Van Valin & LaPolla 1997). The relations and the hierarchy would consequently need to be modified for each language or group of languages.

Both GR and Lin's models compare favorably with PARSEVAL in their ability to provide detailed information about the preciseness of parsing. In the GR scheme, for example, precision and recall scores can be provided on the word level by using either groups of relation types or single relation types.

#### 8.1.3.3 Conclusion

In conclusion, I make some recommendations about which of the evaluation metrics and measures would be best used for evaluating various types of parsers and suggest ways in which existing methods and measures might be improved.

The LA metric is undoubtedly best for PS evaluation. Firstly, it allows for imperfect matches between the parser output and the gold standard annotation. This facilitates system comparisons and makes it possible to evaluate a wider range of PS parsers against a single gold standard resource. Secondly, because it is based on paths between the nodes in the parse tree, the LA metric offers error analysis and error locating capabilities. In addition therefore to providing information about the performance of a parser, it also provides directions for improving the system.

The most important D-based word-level evaluation measures are UAS and LAS. LAS provides information about which D links are assigned completely correctly, and this, in the end, is the ultimate purpose of every D parser. While UAS is only concerned with the structure and not with the labels, it is better for comparing parsers, even when they use dissimilar tagsets. When this type of evaluation is accompanied by the use of an evaluation resource with a hierarchically organized tagset (as in the GR corpus and FiEval), it facilitates comparability between different systems. One should also report evaluation results separately for each relation/D type in order to allow for a detailed analysis of differences in the performance of various parsers.

When a direct comparison of the preciseness figures of several parsers is not a concern and when a linguistic resource that is annotated with a scheme similar to the parser's output scheme is available, it would be preferable to use that resource along with corresponding (PS or D) evaluation metrics. It is better to use D-based evaluation metrics for comparative preciseness evaluation when possible. This is because the results of a D-based evaluation are more meaningful than those that rely on PSs because of the fact that semantic dependencies are embedded in the syntactic dependencies. Moreover, at least one of the output schemes in the majority of parsers contains a type of word-word dependency-based scheme (i.e. it is either a D output, or a GR output or a PAS output).

It is interesting to note that many current evaluation metrics lack a sentence-level measure. Manning and Carpenter (1997) point out that what makes NLP hard is the fact that an NLP system has to make consecutive decisions correctly if it is to be successful. In order to parse a sentence correctly, a parser therefore needs to make a correct decision about each word as well as the sentence structure. The overall success rate in parsing a full sentence is thus the *n*th power of the individual decision success rates. In PARSEVAL, for example, precision, recall and crossing brackets measure success at the level of individual decisions – and not at the level of the sentence. This means that they are rather easy measures on which to do well. It may thus be argued that these measures miss those essential qualities that define a high-quality parser.

Hence, in addition to word and PS nonterminal/D link -level analysis, which is much more useful than sentence-level measures for error-analysis, it is crucial to report the percentage of sentences that have been correctly parsed. With D-based evaluation, this can be done by using the CM measure proposed by Yamada and Matsumoto (2003), which was discussed in Section 8.1.2.1 above. In addition, it is reasonable to report labeled CM figures because this measure indicates the number of sentences for which there are exactly correct analyses. This, in the end, should be the main goal of preciseness in parsing. The LA metric defines a sentence level measure for PS evaluation, that calculates the proportion of sentences that have matching structures and labels in the gold standard and parser output.

## 8.2 Coverage Evaluation and Error Mining

The most straightforward way of evaluating the coverage of a parser is to parse a set of sentences and measure for which proportion of the sentences the parser is able to produce a parse. Coverage is defined as:

$$\frac{|TP \cup FP|}{|TP \cup FP \cup FN|} \qquad (8\text{-}1)$$

Van Noord (2004), and Sagot & de la Clergerie (2006), have proposed *error mining* methods in which the results of coverage evaluations are used for further analysis. The aim of this is specifically to locate deficiencies in the lexicon and the grammar. The methods compare the sentences that the parser is able to analyze with those that it is unable to cover. Both methods are based on observing the frequencies of word *n*-grams. A *word n-gram* is either an individual word (*n*=1) or a sequence of words (*n*>1). The purpose of using word *n*-grams is to detect the words and sequences of words that make a parser fail to analyze a sentence.

**Example 8-7**. The sets $G_n$, $n=\{1, 2, 3\}$ of word *n*-grams of the sentence "Liverpool is playing well."

"Liverpool is playing well."
$G_1$ = {"Liverpool", "is", "playing", "well"}
$G_2$ = {"Liverpool is", "is playing", "playing well"}
$G_3$ = {"Liverpool is playing", "is playing well"}

### *8.2.1 Van Noord's method*

The error-mining method suggested by Van Noord (2004) works in the following way. Let GS(S,A) be a gold standard and O(P,GS) output of a parser for the sentences in the gold standard. The *parsability* of *n*-gram *q* is calculated as follows:

**Definition 8-5**. Parsability of *n*-gram *q* in Van Noord's error mining method.

Let $T = (\delta_1, \delta_2, ..., \delta_m)$ be a sequence of sentences. Let *q* be an *n*-gram. C(*T*,*q*) is the set of sentences in *T* that contain *q*:

$$\mathrm{C}(T,\, q) = \left|\left\{ i \,\middle|\, \delta_i[j ... j+n-1] = q, \text{ where } 1 \le j \le |\delta_i| - n + 1 \right\}\right|.$$

The parsability of *q* is defined as

$$\text{parsability}(q) = \frac{|C(TP \cup FN, q)|}{|C(TP \cup FP, q)|},$$

where TP, FN and FP are sets of sentences as defined in Section 8.1.

If an *n*-gram is to be considered normal, its parsability should be close to the overall coverage of the parser. A considerably lower parsability score indicates a problem in the lexicon or the grammar. Error mining is performed by using increasing *n* values, starting from 1.[69] The result of the mining is reported as an *n*-gram table, and is sorted according to the parsability scores and frequencies.

[69] When *n* is increased, only the *n*-grams that have lower parsability values than its sub *n*-grams (obtained with lower values of *n*) are considered (Van Noord (2004)). The purpose of this is to identify the most important *n*-gram and thus to reduce the number of redundant *n*-grams that are within each other.

### *8.2.2 Sagot and de la Clergerie's method*

In contrast to Van Noord, Sagot and de la Clergerie (2006) base their error mining method (hereinafter referred to as the SC method) on observing the sentences that were not covered. Their aim is to use statistical modeling for finding the cause for the parsing failures. The SC method uses the term *suspicious forms* to refer to the kind of results that Van Noord's method can detect. In addition to this, one of the capabilities of SC is to identify the cause of the failure in each uncovered sentence. This word is then called *the main suspect*.

Sagot and de la Clergerie have devised several ways of extending this basic model. For example, they have developed a method for providing an estimate of the benefit that might accrue from the correction of the corresponding error in the parser for each suspicious form. This process allows parser developers the option of identifying the most critical directions for further development.

### *8.2.3 Analysis*

I shall now analyze the two error mining methods by comparing their similarities and differences, their respective degrees of usability, and the information that they are able to provide. Both these methods are similar in their aim: they have been developed to detect errors and deficiencies in the lexicon and the grammar of a parser. The main advantage of both approaches is that no annotated resource is needed for carrying out error detection. Since both these methods are based on using unannotated texts, one can apply huge collections of text for mining.

The main deficiency of the SC method is that it is applied only to single words and word bi-grams. This limits the errors that are detectable by using this method mostly to problems in the lexicon. Van Noord's method has been applied with *n* values from 1 to 5, and this allows for the observation of a wider array of error types. While Van Noord's method does not analyze uncovered sentences directly at all, the SC method bases its analysis solely on these sentences. In addition to providing information about the items that most frequently cause problems, the SC method is able to provide estimates of the main causes of parse failure for each of the sentences that have not been covered. But Van Noord's method can only provide the former kind of information. Another advantage of the SC method is that it offers a graphic tool for viewing mining results.

## 8.3 Robustness Evaluation

A parser's ability to produce an error-free or only a slightly altered output from input sentences containing errors is referred to as robustness. A robust parser is able to provide as complete and correct an analysis of the input sentence as is possible under the circumstances. Foster (2004) and Biggert *et al*. (2003, 2005) have proposed metrics for evaluating the extent to which the parser is able to parse noisy inputs.

### *8.3.1 Evaluation based on manually constructed examples*

Foster (2004) has proposed a robustness evaluation method based on the corpus of ungrammatical sentences discussed in Section 6.2.4. Her idea is that the highest similarity score between the parse for an ungrammatical sentence and any of its grammatical counterparts is chosen for each sentence.[70] This may be formally expressed as follows. Let *U* and *G* be the set of analyses for the ungrammatical and grammatical sentences respectively. It is possible that there may be more than one grammatical (corrected) sentence corresponding to each ungrammatical sentence. Thus, each element $g \in G$ is the set $A=\{1,\ldots,n\}$ of analyses for the grammatical sentences. The sentences are compared as follows:

```
GetSentenceScores(U, G)
    sentScores←∅
    FOR i ← 1 TO |U| DO
        sim←∅
        u←U_i
        FOR j ← 1 TO |G_i| DO
            g←G_ij
            sim←sim∪CalculateLabeledPARSEVAL(u,g)
        sentScores←sentScores∪GetMax(sim)
RETURN sentScores
```

**Figure 8-4**. Pseudocode of the Foster's robustness evaluation method.

The robustness of a parser can then be defined as the precision and recall it displays over all of the test sentences. Foster (2004) reports on an evaluation of a state-of-the-art probabilistic parser by Charniak (2000). This parser was able to produce an exactly correct parse for one third of the ungrammatical sentences. For agreement errors and errors caused by the use of a wrong preposition, the accuracy rate was over 70%. The most problematic type of error, which reduced the accuracy rate to only 20%, was one erroneous word with a POS category than was different from the original correct one. It is not surprising that similar behavior was observed on sentences with more than one error.

[70] If, for example, an ungrammatical sentence has two possible corrections, and if its parse is 90% similar to the first correction's parse and 80% similar to the second correction's parse, it is allocated a score of 90%.

### *8.3.2 An unsupervised robustness evaluation method*

Bigert *et al*. (2003, 2005) deliberately constructed sentences that contained errors for the purpose of evaluating robustness by using an automatic tool that simulates naturally occurring typing errors and so introduces spelling errors into input sentences. This automated introduction of errors enabled the researchers to undertake controlled testing of the effect of increased error rates on outputs. The evaluation was conducted in the following way. Firstly, the parser to be evaluated was given an error-free text to parse. Secondly, the parser was given ill-formed input texts to parse. Finally, the results obtained from the first and second stages were compared. The degradation of a parser's output was then measured by comparing the parser's preciseness on error-free texts to its preciseness on ill-formed inputs.

The procedure described above was repeated for several levels of distortion. The lower the level of degradation is, the more robust is the parser in the face of ill-formed input. Bigert *et al*. conducted these experiments at error levels of 0%, 1%, 2%, 5%, 10%, and 20% respectively. They iterated the procedure several times and calculated the results as the average of the test runs. The preciseness of a robust parser was expected to deteriorate as much or less than the level of errors introduced.

### *8.3.3 Analysis*

I analyzed the two robustness evaluation methods on the basis of the type of information they can provide and on the type of resources that are needed for carrying out evaluations. While Bigert *et al*.'s measure is based on the overall preciseness on a certain error-level, Foster's measure provides sentence-level figures on the similarity of the parses returned for the grammatical and ungrammatical sentences. This information can be used for error analysis.

Both these methods base their evaluations on unannotated texts. This renders them applicable to a wide range of parsers without any further need for any modifications to the evaluation resource or for devising mapping algorithms between parser output and annotation schemes. Another advantage of the method of Bigert *et al*. is that it does not require any human intervention for constructing the evaluation resource. The only prerequisite is a set of unannotated sentences and an estimate of a parser's preciseness on error-free text. Foster's method, by contrast, relies on manually constructed test sentences. Bigert *et al*.'s tools, however, have only been applied to Swedish. While Foster's method has been applied to a wide variety of grammatical errors, Bigert *et al*. have only reported experiments on spelling errors. It nevertheless appears to be possible to generalize the method to other types of errors.

## 8.4 Efficiency Evaluation

Few studies have been undertaken to compare the efficiency of parsing algorithms on the basis of parse times and the effects of different grammars. I have already described one of the rare works that uses common grammars and test data, that of Van Noord (1997), in Section 4.5.3. The problem with experiments of this kind is, however, that when they are carried out with different machines and when the algorithms are implemented for different languages, comparisons between studies become problematic. In the following section, I describe several methods suggested for resolving these difficulties.

### *8.4.1 Heap- and event-based measures*

Carballo and Charniak (1998) have used the number of edges that have popped off the agenda of a chart parser[71] to measure efficiency in probabilistic parsing. Roark and Charniak (2000) propose a related measure, based on events considered, that is applicable to a wider range of parsing approaches. It measures the number of events for which a probability must be calculated. A search or pruning technique is more efficient than others if it reduces the number of events that must be considered.

Carballo and Charniak (1998) argue that their efficiency score cannot be artificially reduced through optimization. They claim, however, that the measure is general enough to cover different search and pruning techniques, and that it is independent of the execution environment and – to a certain extent – independent also of the implementation language. Carballo and Charniak (1998), and Roark and Charniak (2000), argue that such a measure enables parsing efficiency to be compared at the algorithmic level without any attention to the low-level optimization that has been performed. However, they also point out that some significant part of a parser's function may be disconnected from the heap operations.

[71] In addition to the chart, a chart parser has another data structure. It is the *agenda* that contains the items yet to be recognized. The data is recorded by using *edges* that contain information such as the head of the current item and the position of the word in the sentence. When an edge is placed onto the chart, it is said to be "popped off" the agenda.

### *8.4.2 Moore's method*

Moore's (2000) method is based on using common grammars, test data and standard implementations of reference algorithms in all programming languages of interest. The logic behind this is that the efficiency of an algorithm can be reported relative to the speed of this reference parser. This factors out the influences of different programming languages and computing platforms.

In contrast to Charniak *et al.*'s (Carballo & Charniak 1998, Roark & Charniak 2000) experiments, Moore evaluated non-probabilistic algorithms. His method works in the following way. Firstly, a set of grammars, a test set and reference parser implementations[72] in the most commonly used programming languages, are provided. Secondly, the evaluator parses the test sentences on one machine with his/her parser and the reference parser is implemented in the same programming language. Thirdly, efficiency is measured as the percentage of the execution time of the parser being evaluated over the time recorded for the reference parser.

### *8.4.3 Evaluation based on strongly equivalent grammars*

Yoshinaga *et al.* (2003) describe a method based on the use of strongly equivalent grammars obtained by grammar conversion. They represent an algorithm for converting LTAGs to strongly equivalent HPSG grammars and demonstrate an evaluation on manually constructed and automatically induced TAGs and an HPSG. The authors claim that grammar conversion abstracts away from the surface differences between grammar formalisms, and this means that one can gain a deeper insight into generic parsing techniques and share techniques that have been developed for different grammar formalisms.

[72] The CYK algorithm could, for example, be implemented with maximum efficiency in C and Lisp. The source code should be made available so that researchers will be able to examine and improve the implementation without changing the basic algorithm (Moore 2000).

### *8.4.4 Analysis*

While the observation of the running times of parsers is well-suited to monitoring the progress of a single parser, such a method does not allow for making reliable comparisons with other systems unless the same grammars and test sentences are used and unless it is possible to relate the parse times obtained across different implementation languages and platforms (such as Linux, Windows and SunOS).

Moore (2000) points out that Roark and Charniak (2000) have applied the events considered measure to a best-first parser and to a beam-search-based parser. Although these parsers differ in some respects, they also have a number of attributes in common. This makes it easier for researchers to identify a common measure for comparing their efficiency. Moore further claims that there are several ways in which the number of events that are being considered can fail to correlate with the parse time. The metric, for example, does not take into account the effort needed to compute the probabilistic models. The measure does not take into consideration the pruning phase, which may be one of the most time-consuming tasks.

Moore's (2000) experiments showed that the number of chart edges do in fact often fail to predict the running time of parsers correctly. This led him to conclude that it is necessary rather to measure the actual parse times. Moore's method does not address Roark and Charniak's search for a metric that is insensitive to different degrees of optimization. In fact, when one wants to compare different parsing algorithms, one needs to restrict oneself to a comparison of implementations that are as similar as possible in that regard.

Since the outputs of strongly equivalent grammars are equivalent, the method proposed by Yoshinaga *et al*. (2003) allows meaningful comparisons to be carried out among parsers for different grammar formalisms regardless of their surface differences. This method therefore seems to offer the ultimate solution to the problem of comparing the efficiencies of parsing systems. Where mappings between grammars exist, that is clearly the case. The problem is similar to the difficulties that one encounters when creating mappings between treebanks: such algorithms are difficult, and in many cases impossible, to devise. Where mappings do exist, this method is valuable for comparing the effects of specific parsing approaches, filtering mechanisms, and so on.

I argue that a clear separation should be maintained between the following two notions of efficiency. The first one focuses on efficiency in parsing algorithms and the parsability of grammar formalisms. This type of evaluation is of the greatest use to developers of parsing algorithms. The second one focuses on parsing times. NLP practitioners are interested in practical differences in the efficiency of parsers, i.e. in parsing times. An orientation towards either of the types of efficiency should also affect the choice of metrics and the measures applied. The methods of the type that Carballo and Charniak (1998), and Roark and Charniak (2000), have proposed are useful for comparing the efficiency of parsing algorithms in those cases where the algorithms are sufficiently similar. The method proposed by Yoshinaga *et al*. (2003) is especially useful for comparing the parsability of grammar formalisms. But such methods are only of limited use to NLP practitioners who want to compare actual parsing times. In order to do this, it is best to have the actual parsing times measured on standardized machine configurations.

## 8.5 Evaluation Tools

There are few freely available parser evaluation tools. Most of the tools are constructed for preciseness evaluation, and they use a single evaluation metric. I shall discuss these tools in this section.

*Evalb* is a freely available bracket-scoring program that reports the PARSEVAL measures precision, recall, and the number of crossing brackets (Sekine & Collins 2006). It also reports tagging accuracy as the percentage of the POS tags correctly assigned. This program takes the gold standard and parser output files as the inputs, and reports the scores for each sentence separately by printing them to the standard output. *Randomized Parsing Evaluation Comparator* by Bikel (2006) is a statistical significance tester for Evalb outputs. When parser evaluation is performed repeatedly on the same data and materials, it is important to know if the improvements in the preciseness are statistically significant. Bikel’s Perl script takes outputs from two separate Evalb evaluations as input, and calculates whether the differences in recall and precision are statistically significant or not.

The LA metric has been implemented in C (Sampson & Higgins 2006). The program takes two input files, one of which consists of parser output and the other of which consists of gold standard analyses. The output consists of a preciseness figure for each word and the whole parse tree. If one wants to make a change to the partial match function used in node label replacements, one needs to modify the source code.

The creators of STMDP-CoNNL-X released an evaluation script for the task (Buchholz & Marsi 2006). It is a Perl implementation of the three pure D evaluation measures (UAS, LAS and LA) used in the task, and is available on-line (CoNLL-X Shared Task 2006). The user can choose whether to use punctuation in evaluation or whether to base it only on the word tokens. This tool also reports error analysis statistics: the precision and recall figures for D relations, distance figures for heads and dependents and a list of *frame confusions*.[73]

*Trnstree* is Lin's C++ implementation of his preciseness evaluation metric, and it can translate SUSANNE structures into D trees and perform evaluations on those (Lin 1999). Since the tool's scripting language allows one to define mapping rules between annotation and output schemes, this makes it usable for treebanks other than SUSANNE. In spite of this, no such use has been reported in the literature. The tool reports precision and recall figures over the D structures.

Carroll's (2006) *GRAMRELEVAL* is a Lisp-implemented evaluation tool for the GR metric. The tool takes as input the GR corpus and the parser output and calculates precision and recall measures. The input consists of four files: a lemmatized word file, a file with the input sentences numbered, and files for gold standard and parser output GRs respectively. The system outputs precision, recall and an F-score for each GR and an overall figure over all the relations. It also gives a confusion matrix over the GR types for further error analysis.

The HPSG community uses the term *competence and performance profiling* for a structured snapshot of the parser status at a certain development point (Oepen & Flickinger 1998, Oepen & Callmeier 2000, Oepen & Carroll 2000). *[incr tsdb()]* is a software package for producing, maintaining, and inspecting such profiles. The tool has a graphic user interface and functions for profile analysis and comparison. It is also able to store profiles in a database for later use. This tool is suitable for comparing a parser's performance with earlier results and between various parameter settings. Approximately 100 attributes are recorded in a profile that consists of information about the system setup and parameters, the coverage (parse trees per reading, the number of analyses, etc.), ambiguity and resource consumption (the use of memory and time).

*AutoEval* and *Missplel* (Bigert 2006) are robustness evaluation tools that were used in the evaluations reported by Bigert *et al.* (2003, 2005). AutoEval has been designed to simplify the gathering, processing and counting of the kind of data often involved in NLP evaluation tasks. It includes a script language for describing the evaluation task that will be carried out. It is also able to store the data from test runs. While Missplel is a tool that introduces human-like spelling errors into text, it is trained only for Swedish data.

[73] A frame is a list of the D relations of a token and the relations of all its children. Frame confusion occurs when the gold standard and the parser output frame are not identical.

Table 8-3 summarizes the main characteristics of the evaluation tools introduced above.

**Table 8-3**. Evaluation tools. KEY: P = precision; R = recall.

| Tool | Type | Metrics | Measures |
|---|---|---|---|
| *Evalb* | Preciseness | PARSEVAL | P, R, no. of crossing brackets |
| *Randomized Parsing Evaluation Comparator* | Significance tester for Evalb | Comparison between R and P of two Evalb runs | p-value |
| *LA C implementation* | Preciseness | LA | LA word and sentence level |
| *STMDP-CoNNL-X evaluation script* | Preciseness | Pure D | UAS, LAS, LA |
| *Trnstree* | Preciseness | Lin's mapping based | P, R, F-score |
| *GRAMRELEVAL* | Preciseness | GR scheme | P, R, F-score |
| *[incr tsdb()]* | Profiling tool for HPSG parsers | Coverage, efficiency | Items covered, time and memory consumption, the number of chart edges + several others |
| *AutoEval* | General NLP evaluation | Definable by the script language | P, R, F-score |
| *Missplel* | Error induction | - | - |

Most of the existing evaluation tools are for preciseness evaluation and they implement a single evaluation method. AutoEval and [incr tsdb()] are the only tools that take a more broad perspective on evaluation. The advantage of AutoEval is its flexibility: the user can define an evaluation task by using the script language. However, AutoEval has only been used in the robustness evaluations reported by Biggert *et al*. (2003, 2005). While [incr tsdb()] implements a set of evaluation methods and measures, it is designed for evaluating only UG parsers and has been applied to parsers using HPSG. The tool has a graphic user interface for making browsing and the interpretation of results easier. The system, however, offers only a limited ability to make inter-system comparisons.

## 8.6 Conclusion

In conclusion I would like to make the following observations. Firstly, there is a need for a framework that defines the criteria for evaluating parsers and that establishes a coherent set of evaluation metrics and measures for each criterion. The evaluation methods should be based either on current best practices or their improved versions. New methods and measures should be devised for those criteria for which no suitable methods and measures are currently available. Secondly, it is necessary to define the interaction between the different aspects of parsers' performance. Since current evaluation methods focus on a single aspect, namely the preciseness, they are deficient in interactions of this kind. It is nevertheless clear that undergeneration, overgeneration and robustness, for example, are tightly connected (this point was discussed in Section 5.1.2). Thirdly, the greatest challenge in comparative evaluation is the integration of the level of detail in parser outputs. It is necessary to include a measure of this kind in the evaluation framework that is devised. Fourthly, what are needed are software tools for carrying out evaluations – especially the kind of tools that offer several types of evaluation metrics and measures and that also support several annotation and output schemes.

# 9 FEPa – A Framework for Evaluating Parsers

As described in Chapter 8, existing evaluation methods focus on one single aspect of parser performance. While each approach to parsing has distinctive strengths and weaknesses and the structure and detail of parser outputs varies, it becomes obvious that a single scalar value cannot fully or comprehensively reflect the quality of a particular parser. It is for this reason that more sophisticated and fine-grained methods are needed.

The *Framework for Evaluating Parsers* (FEPa) that I describe in this chapter focuses on intrinsic evaluation and provides useful information for parser developers. My purpose is to provide a fuller picture of a parser than other existing methods are able to do. The goal of FEPa is to provide a framework for practical evaluations of parser performance and to define a set of measures for evaluating parsers. FEPa can be used, moreover, for comparing parsers based on several criteria.

This chapter is organized in the following way. Section 9.1 describes the metrics and measures applied in the FEPa framework. Section 9.2 is concerned with the types of linguistic resources that are needed for carrying out evaluations with such a framework. Section 9.3 concludes the chapter with some closing remarks.

## 9.1 The Framework

An evaluation framework needs to be able to address the following five questions:

1. *Purpose*: What is the purpose of the evaluation? Purpose in this context would typically be defined by what the user intends to do with the evaluation results. One might ask, for example, whether the results will be used by parser developers or by NLP system developers who are trying to select a suitable parser.
2. *Criteria*: What is being measured? Evaluation criteria define the set of characteristics of the system that is being evaluated. PARSEVAL, for example, focuses on the single criterion of preciseness.
3. *Metrics*: What are the means that one uses to observe the performance of a system in terms of each of the criteria? A metric is a system of related measures that facilitates the quantification of assessing a criterion. The PARSEVAL metric, for example, is based on a comparison of bracketed tree structures.

4. *Measures*: How are the results reported? A measure is the way in which the results of an evaluation are quantified when one uses a specific metric. Measures can be used for monitoring the progress of a particular parser and for comparing system performances. Precision and recall, for example, are used as the measures of preciseness in PARSEVAL.
5. *Materials*: What kinds of resources (i.e. software tools, linguistic resources) are used for evaluation? PARSEVAL evaluations, for example, are based on a treebank – often the PTB. The EVALB tool is usually used for PARSEVAL evaluations.

### *9.1.1 Purpose and criteria for parser evaluation*

There are two main reasons why parsers need to be evaluated. The first reason is that evaluations give parser developers the information they need to guide their work. There are two separate kinds of developers: grammar writers and parsing algorithm developers, and they both have particular evaluation needs. The second reason is that evaluation provides NLP system developers with information about the relative performances of different parsing systems.

The evaluation criteria in FEPa are *preciseness*, *coverage*, *robustness*, *efficiency*, and *subtlety*. The evaluation process for each of these criteria consists of selecting the resources for evaluation, parsing selected texts with the parsers to be evaluated, and performing the calculations that are needed to measure their performances in terms of a given criterion. Distinctive metrics and measures are needed for each criterion. The evaluation criterion used will obviously also affect the choice of the linguistic resources that will be utilized. When the results of the evaluations undertaken in terms of the five criteria are combined into performance profiles for parsers, those parsers can then be compared from several different points of view. In the following five sections I will introduce each of these criteria in detail and describe the metrics and measures for observing parsers' performance in terms of each of them.

### *9.1.2 Preciseness*

Preciseness is the most important evaluation criteria for parsers: a parser that cannot produce correct analyses is not much of a use. Firstly, a preciseness evaluation method should measure the parser's preciseness in assigning syntactic tags. Secondly, if a parser performs a morphological analysis, the preciseness of morphological tagging should also be measured. Thirdly, since the tags cannot be correctly assigned if segmentation has failed, the preciseness of word and sentence segmentation will also obviously be indirectly assessed. Apart from its ability to define the preciseness of a parser, this kind of evaluation is also useful for producing error analyses. Detailed information on the preciseness of a parser in assigning certain tags can, for example, be provided.

I concluded on the basis of the results of the analysis described in Section 8.1.3.3 that FEPa should apply the following evaluation methods. In D-based evaluation, LAS and labeled CM should be applied for intrinsic evaluation and for a comparison of parsers whose tagsets match one another. UAS and CM need to be used for comparing systems in those cases where output schemes do not match. For PS-based evaluation FEPa uses the LA metric, because of its many desirable characteristics described in Chapter 8. D-based evaluation is the preferred type for comparative evaluation where applicable.

### *9.1.3 Coverage*

The concept of *coverage* has two meanings (Prasad & Sarkar 2000). In the first place, *grammatical coverage* refers to a parser's ability to cope with different linguistic phenomena. *Parsing coverage*, by contrast, measures the proportion of naturally occurring, free-text for which a parser is able to produce a full, unfragmented parse. I further divide parsing coverage into categories of genre coverage based on different types of text such as prose, newspaper, law, financial, religious, and so on. This allows one, for example, to measure the generalizability of a parsing approach over text genres. I use the term *generalizability* to refer to a parser's ability to analyze texts from diverse genres.

The results of grammatical coverage evaluation can be reported by listing the types of grammatical phenomena covered and not covered. This kind of information can be especially useful for grammar developers. Parsing coverage can be measured as the percentage of input sentences that a parser is able to assign a complete, unfragmented parse.[74] No annotated text is needed for performing this type of evaluation. While results obtained in this way are comparable across parsers, the detail in each parser output should be taken into consideration in order to assure the fairness in this type of evaluation.

---

[74] Note that a more strict definition of parsing coverage (namely, the proportion of sentences covered correctly) is equivalent to sentence-level preciseness evaluation.

On the one hand, one can argue that coverage alone constitutes a rather weak measure of parser performance. An obvious problem that arises from measuring coverage alone is that a parser that returns undetailed and flat analyses for every sentence will easily obtain high coverage scores, whereas a parser that outputs detailed analyses will suffer in covering all the input sentences. Another consideration is that one may regard preciseness and coverage as conflicting requirements for a parser. While increasing the preciseness of the grammar often causes its coverage to decrease, the addition of more constraints to the grammar will cause some of the sentences to be rejected even if they are acceptable to users of the language. In contrast, a loosening of the constraints will allow more sentences be parsed. While this will increase coverage, it can simultaneously cause overgeneration, problems with disambiguation and decreased preciseness (see Sections 5.1.2 and 5.1.3).

On the other hand, the points that I have raised above confirm that there is a strong relationship between coverage and preciseness. This implies that coverage can be used as an indirect measure of preciseness and generalizability. The aim of syntactic parsers is to analyze whole sentences (and not just fragments such as constituents/D links) precisely. The connection between coverage and preciseness is clear in the case of sentence-level preciseness evaluations.[75] One may define this connection by saying that a sentence that cannot be fully analyzed cannot have a complete match with the correct structure. Coverage can consequently be used an indirect measure of preciseness and can also, for example, be utilized for measuring the generalizability of a parser. Preciseness and subtlety, however, have to be taken into consideration while performing such an evaluation.

### *9.1.4 Robustness*

A robust parser is one that is able to recover from various kinds of exceptional inputs while parsing – and not crash in the process. A robust parser is able to provide a complete and correct analysis for a noisy input sentence. This is the underlying assumption of the two existing robustness evaluation methods introduced in Chapter 8. A robust parser, moreover, should perform in a coherent way when faced with increasing levels of noise. I accept the terminology suggested by Biggert *et al*. (2003, 2005) in this matter, and refer to this property as *degradation.* Secondly, robustness is connected to *stability*. Stability means that a system will not crash while attempting to parse given inputs.

[75] The CM measure, for example, uses the percentage of sentences whose unlabeled D structure is completely correct to evaluate the sentence-level preciseness.

FEPa makes the simplifying assumption, as do Bigert *et al*. (2005), that a parser is robust if it is able to produce a similar analysis for a correct sentence and a noisy version of the same sentence.[76] The assumption behind this is that if a parser is able to do this, it will also be able to perform in a robust way when it is confronted by noisy inputs. In order to measure the degradation of a parser, one needs a corpus of parallel correct and erroneous sentences. One can measure the stability of a parser by observing the number of times that a parser fails to parse or crashes while attempting to parse input sentences.

Robustness evaluations are carried out in FEPa in the following way. Firstly, the correct sentences ($CS_e$) are parsed. Secondly, the noisy sentences ($NS_e$), each representing an error-level *e*, are parsed. Thirdly, the analyses produced for each sentence in $CS_e$ and its corresponding noisy sentence in $NS_e$, are compared. Finally, the performance is measured by using two distinct evaluation measures. The first evaluation measure is the percentage of sentences for which a parser produced exactly the same structure for both the correct and noisy input sentence. I refer to this measure as an *unlabeled robustness score* (UR score). The second measure, *labeled robustness score* (LR score), is stricter: it accepts an analysis only if the two structures are the same and if, in addition, the labels on syntactic categories (GRs, dependencies) match.[77] The UR and LR scores are defined in the following way:

**Definition 9-1.** The structure and labels of a syntactic analysis.

Let $X$ be an analysis for sentence $s$ that consists of a sequence of grammatical structures $x \in X$ (i.e. bracket pairs in a PS tree or D links in a D tree) and a label (i.e. tag) for each $x$.

1. *Structure*($X$) denotes the sequence of grammatical structures in $X$.
2. *Labels*(X) denotes the sequence of labels assigned to each $x \in X$.

[76] Since some parsers are designed to check grammaticality of input sentences, their returning of a “failure to parse” in response to an ungrammatical sentence is (for them) a correct result. This metric is not applicable to such systems. In such cases, one could use the proportion of ill-formed sentences that the parser accepts as a measure of robustness. If one defines *grammar checker* to be a parser that locates errors in input sentences and corrects them, robustness can be defined as the proportion of noisy sentences that the system is able to correct so that it matches the original sentence, and in addition, is able to produce a parse to the corrected sentence.

[77] For example, the introduction of a single misspelling into a sentence often results in the type of the D link associated with the misspelled word to be altered.

**Definition 9-2.** Unlabeled and labeled similarity of two analyses.

Let *X* and *Y* be syntactic analyses.

1. *Unlabeled similarity* is defined as

$$\text{ULSim(X,Y)}=\begin{cases}0, structure(X) \neq structure(Y)\\ 1, structure(X) = structure(Y)\end{cases}.$$

I.e. the two analyses match in unlabeled similarity if every structure in the analyses is similar.

2. *Labeled similarity* is defined as

$$\text{LSim(X,Y)}=\begin{cases}1, structure(X) = structure(Y) \text{ and } labels(x) = labels(y)\\ 0, otherwise\end{cases}.$$

I.e. the two analyses match in labeled similarity if their structures are similar and if, in addition, the labels assigned to each structure match.

**Definition 9-3.** Unlabeled and labeled robustness score.

Let $CS=(s_1,s_2,\ldots,s_n)$ be the sequence of correct sentences. Then P(*CS*) is the sequence of analyses assigned to these sentences by parser *P*. Let $NS=(r_1,r_2,\ldots,r_n)$ denote the sequence of noisy sentences that correspond to the correct sentences *CS*. P(*NS*) is the set of analyses assigned by a parser to sentences *NS*.

The UR score is defined as $\text{UR} = \frac{\sum_{i=1}^{n} \text{ULSim}(\text{P}(s_i), \text{P}(r_i))}{n}$

The LR score is defined as $\text{LR} = \frac{\sum_{i=1}^{n} \text{LSim}(\text{P}(s_i), \text{P}(r_i))}{n}$

It is quite clear that as the level of noise in the inputs increases, the performance of a system degrades correspondingly. The extent to which this occurs, referred to as degradation, can be measured by increasing the number of errors in the noisy input sentences and then observing the effect that this will have on its performance.

### *9.1.5 Efficiency*

The FEPa framework is restricted to the type of efficiency defined in Section 8.4.4, namely, the one based on the measurement of a parser's practical efficiency in terms of the time and space spent in parsing a specific test set. This kind of efficiency is the most easily measurable and comparable of the five criteria in FEPa. In practical terms, the efficiency of a parser can be measured by observing the time and space it takes for a parser to analyze a sentence.

The most straightforward method for empirically evaluating the efficiency of a parser (described in detail in Chapter 8) is to record the time that the parser takes to parse a set of test sentences. This is exactly what FEPa does. Efficiency evaluations are carried out in the following way. Firstly, the same corpus of text is parsed with the parsers that are being evaluated in the same computer environment. Secondly, one records the parse times for each of the parsers. When one uses the same input texts and identical machines and environments, one can compare the parsers' performances. The efficiency figures thus obtained can be analyzed in terms of sentence length to provide insight into the time and space complexity of the parser. A parser's efficiency in analyzing ill-formed input can also offer insight into the way in which the robustness mechanisms of the system are implemented.

### *9.1.6 Subtlety*

I use the term *subtlety*[78] in reference to the level of detail in a parser's output. Subtlety refers to both the levels of description in the output (syntax, semantics, etc.) as well as to the complexity of the description for each level. Some parsers leave some of the ambiguities in the output unresolved because they either return multiple analyses for an input sentence or because they leave some words underspecified. While the amount of desired detail depends on the application in which a parser is to be applied, ambiguity and underspecification are regarded as negative properties in a parser.[79]

Subtlety has two uses in evaluation. In earlier discussion (Chapter 3), I noted that varying levels of detail in parser output are needed for different NLP tasks. NLP developers who are searching for a suitable parser for their application need information about the type and richness of syntactic description they will get in parser outputs. Subtlety can be applied, moreover, as a factor for the fair measurement and therefore comparison of systems that produce different levels of richness in their outputs. It is obvious that the level of detail in the analysis is proportionate to the number of decisions that are made during parsing. This makes it more difficult and time-consuming to assign a correct analysis which is rich in information. One should therefore use the subtlety of the outputs as a factor in the comparison of parsers in order to make such comparisons fairer. The lack of sensitivity to subtlety is one of the main deficiencies in existing methods of comparative evaluation.

---

[78] I have avoided using the term "delicacy" – coined by Atwell (1996) – because it might suggest fragility and therefore be misunderstood as a *negative* property of a parser.

[79] If one were to restrict oneself to considering those parsers that use only syntactic information (as most of the existing parsers do), one could assert that a parser *should* leave pending the types of ambiguity (for example, PP attachment (see Section 5.1.1)) that it cannot resolve reliably and that it should leave such decisions to later processing stages.

The subtlety of an output scheme is measured in FEPa in the following way. Firstly, the evaluator needs to consider the amount of detail in the output scheme. The amount of detail can either be defined manually after an examination of the parser's documentation or automatically by scrutinizing the complexity of the tagset. The subtlety measures for describing the detail in parser output in FEPa are: the number of POS tags, the types of output produced (PS, D, semantics, etc.) and the size of the syntactic tagset. Secondly, the level of ambiguity remaining in the output is measured by observing the average number of analyses per sentence and the proportion of sentences that are left with more than one analysis in the output. Thirdly, the underspecification remaining in the outputs is measured from the output. Underspecification is measured by the proportion of words that are left underspecified. This information is finally combined into a subtlety profile of the parser.

### *9.1.7 Conclusion*

While preciseness and coverage are essential criteria for grammar developers, robustness and efficiency metrics are needed mostly by the developers of parsing algorithms. Subtlety is useful both for those who are making inter-system comparisons and for NLP system developers who are looking for a parser that will suit their needs. The criteria, metrics and measures in FEPa and their prospective users are summarized in Table 9-1 below. Figure 9-1 illustrates the interactions between the criteria.

**Table 9-1.** FEPa evaluation criteria, metrics and measures. The column "Users" specifies the groups of users to which each criterion might be most useful. KEY: G = grammar developers; A = parsing algorithm developers; P = NLP practitioners

| Criterion | Sub-criterion | | Metrics | Measures | Users |
|---|---|---|---|---|---|
| *Preciseness* | Morphological tagging | | Comparing parser output and correctly tagged text | The percentage of correct tags | G, A |
| | Syntactic parsing | | D-based | UAS, LAS, CM, labeled CM | All |
| | | | PS-based (LA metric) | Precision, recall, f-score. sentence-level measure | |
| *Coverage* | Grammatical coverage | | Comparing parser output and a test suite | Grammatical construction types covered | G |
| | Parsing coverage | | Observing the proportion of analyzed sentences on unannotated texts | The percentage of sentences covered | All |
| | Genre coverage | | Observing the proportion of analyzed sentences on unannotated texts from several genres | The percentage of sentences covered | All |
| *Robustness* | Stability | | Observing the number of crashes | The percentage of failures | A |
| | Degradation* | | Comparing the outputs for correct and erroneous sentences | Unlabeled and labeled percentage of similar analyses | A, G |
| *Efficiency* | Time | | Parse time on a single machine & environment | Minutes/seconds | A, P |
| | Space | | Memory consumption | KBs/MBs | A, P |
| *Subtlety* | Level of detail | POS | Observing the output / consulting the manual | The number of tags | P |
| | | Syn-tax | Observing the output / consulting the manual | The number of tags | P |
| | Underspecification | | Observing the output | The number of underspecified words per sentence | All |
| | Ambiguity | | Observing the output | The number of analyses per sentence | All |

*Degradation can be further divided on the basis of the type of noise in the inputs (grammatical mistakes, misspellings).

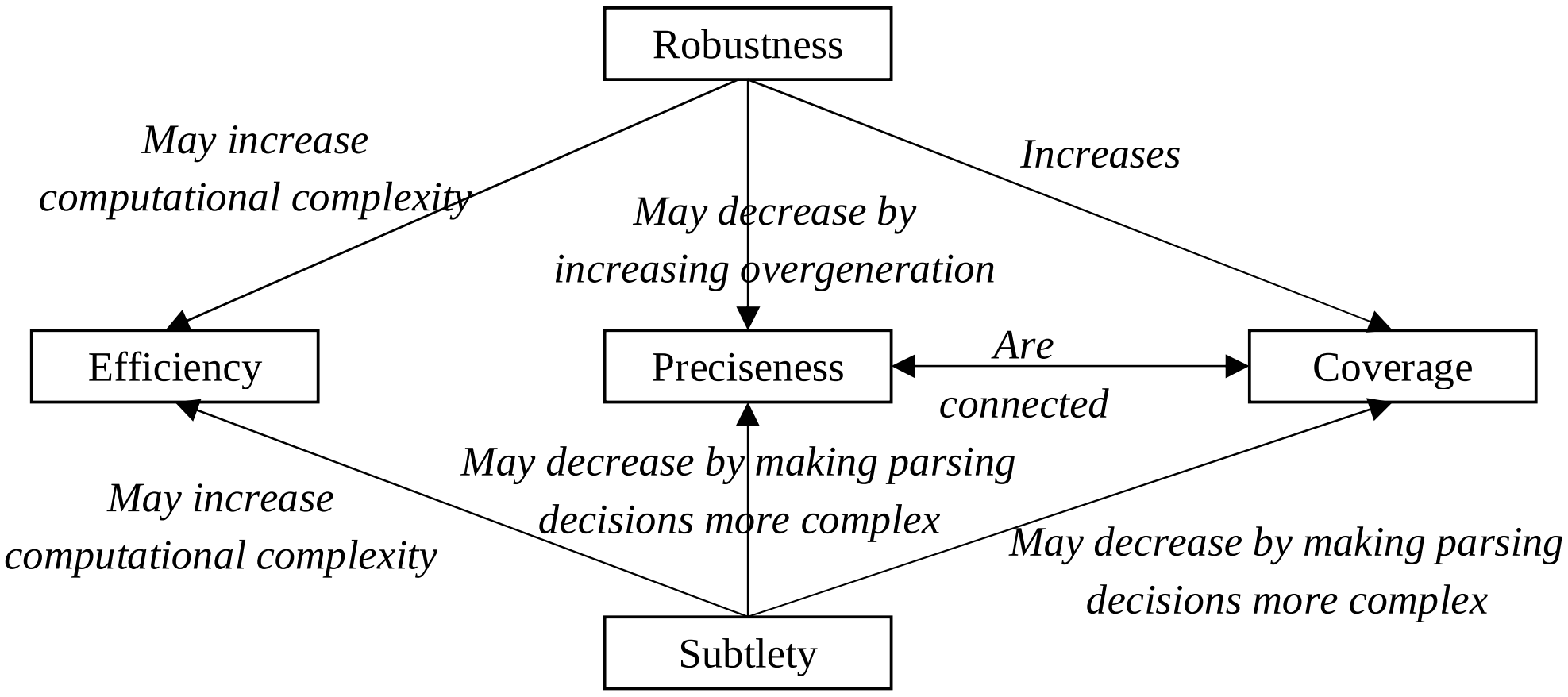

**Figure 9-1.** Connection between the properties of parsers.

Some of the connecting lines in Figure 9-1 need further explanation. The discussion above (see Section 9.1.3) emphasizes that there is a strong connection between the coverage and preciseness of a parser. The purpose of syntactic parsers is to analyze whole sentences rather than just fragments (constituents/D links) precisely. The connection between coverage and preciseness is clear in the case of sentence level evaluation measures such as CM: a sentence that cannot be fully analyzed cannot make a complete match with the correct structure in the evaluation resource.

The connection between preciseness and coverage is two-way. Preciseness and coverage can also be seen as conflicting requirements for a parser. Increasing the preciseness of the grammar often causes its coverage to decrease; adding more constraints to the grammar causes sentences to be rejected even when they are acceptable to users of the language. While the loosening of constraints allows more sentences to be parsed (and therefore increases coverage), it can at the same time easily cause overgeneration, problems with disambiguation and decreased preciseness.

The effect of subtlety on preciseness and coverage is two-fold. Firstly, the provision of detailed analyses makes it more difficult to cover all input sentences. Conversely, it is easier to achieve a high coverage and preciseness if only shallow analyses are provided. Secondly, a high performance is easier to achieve when parses contain underspecification and when ambiguities are left unresolved.

There is an important connection between robustness, preciseness and coverage. If robustness is achieved by adding new rules to the grammar and/or relaxing the constraints, the coverage of the parser increases. But, it is more than likely that such a parser will suffer from overgeneration and produce large numbers of candidate parses for every sentence, including ungrammatical sentences. This, in turn, would diminish the preciseness of the system. Robustness may also compromise the efficiency of a parser because robustness mechanisms often generate more processing and so overload a system's computational capacity.

## 9.2 Linguistic Resources for FEPa

It is best for an evaluation to be performed by using different kinds of linguistic resource: a treebank, a test suite, and a corpus of ungrammatical sentences. The resources should also preferably consist of texts from diverse text genres. Such resources, however, do not currently exist for any language or parser, and the evaluations have to be carried out with resources that lack exhaustiveness.

Along with evaluation of grammatical coverage, preciseness evaluation makes the greatest demands on the annotation of the evaluation resources. As for the resources used for preciseness evaluation, it is debatable whether the resources should be tailored towards linguistically interesting sentences, which are often rare in running text, or more commonly occurring cases.

All the above-mentioned types of resources are equally useful for measuring efficiency, stability and subtlety. While test-suite-based evaluation is more suitable for measuring grammatical coverage, evaluation that is based on a treebank or unannotated texts is better suited for evaluating parsing coverage. Measuring grammatical coverage calls for a test suite in which items are marked with the grammatical phenomena that they contain. Prasad and Sarkar (2000) point out that the difference between the two concepts of coverage can also be approached from the points of view of competence and performance that I described in Section 6.1.4.3. This observation can be understood in terms of the type of evaluation resources that are needed. Grammatical coverage, or the proportion of linguistic phenomena that a parser can handle correctly, depends on competence and is better measured by a test-suite-based evaluation. Parsing coverage, the proportion of sentences from a naturally occurring free text a parser is capable of parsing correctly, depends on the performance of the system, and is better captured by treebank-based methods or evaluations carried out with the use of unannotated texts.

An annotated evaluation resource would be needed to measure the robustness of a parser against human judgments. However, this would require the annotation of noisy sentences and their correct counterparts in a format that is compatible with the output formats of all the parsers that need to be evaluated. In such circumstances it is probable that the resource would have to be annotated in terms of more than one annotation scheme. This would be an enormous task. Assuring consistency between the annotations would be another difficulty inherent in this approach. Because of such complications, robustness evaluations in FEPa are performed on unannotated texts. In addition to robustness evaluation, a collection of ungrammatical sentences can be applied for measuring the performance of a grammar-checking parser. Table 9-2 summarizes the discussion on linguistic resources for evaluation.

**Table 9-2.** Types of linguistic resources most suitable for evaluating each FEPa criteria.

| Criterion | Sub-criterion | Material |
|---|---|---|
| *Preciseness* | Morphological tagging | Corpus, treebank |
| | Syntactic parsing | Treebank |
| *Coverage* | Grammatical coverage | Test suite |
| | Parsing coverage | Unannotated texts |
| | Genre coverage | Unannotated texts, several text genres, treebank |
| *Robustness* | Stability | All |
| | Degradation* | Corpus of ungrammatical sentences |
| *Efficiency* | Time | All |
| | Space | All |
| *Subtlety* | POS | All |
| | Syntax | All |
| | Underspecification | All |
| | Ambiguity | All |

*Can be further divided on the basis of the type of noise in the inputs (viz. grammatical mistakes, misspellings).

## 9.3 Comparative Evaluation in FEPa

The FEPa framework does not provide direct measures of relative preciseness of parsing systems working in different formalisms. As I will point out below, it remains debatable whether such an approach is feasible at all. FEPa nevertheless offers a common ground for measuring and representing the performance of parsers according to several dimensions and thus provides a way for the comparison of their strengths and weaknesses.

I concluded, on the basis of my analysis and discussion in Chapter 6 and 8 of the problems generated by comparative parser evaluation that it is unrealistic to expect to be able completely to harmonize the outputs of all parsers by mapping, an hierarchically organized tagset or by any other means or resource available for comparative evaluation. Only the dimensions common to all parsers (such as POS tagsets, nonterminal/D labels) can be compared directly, and the variety of detail in the output of a parser cannot be taken into consideration when one is making direct comparisons. Many parsers do not agree about even such low-level tasks as segmentation or basic word classes, let alone syntactic description. For this reason, metrics such as PARSEVAL that try to directly compare the preciseness of different types of parsers always need to abstract away from parser-specific information.

Santos (2003), among others, came to the same conclusion. Black (1998) is of the same opinion when he states that it may never be possible to compare all parsers of any given language in a uniform way. He suggests that, instead of comparing parsers across grammar formalisms and parsing approaches using coarse-grained scores based on dubious technical compromises, evaluation could be carried out by using highly accurate methods within the framework of the parser to be evaluated.

Because of these considerations, FEPa does not aim at directly comparing parsers that are based on different output formats, but rather at facilitating comparative evaluations by applying the subtlety measures. Asserting this is not, however, to claim (as has already been discussed in Chapters 6 and 7) that annotation schemes that support comparative evaluation should *not* be developed. Linguistic resources with well-defined annotation schemes that can be mapped into different parser output schemes are one of the most important components of successful evaluation practices. For parsers for which more direct methods of comparing relative performances are available (because they use the same or highly similar output formats or have a parallel treebank), such measures can be incorporated in evaluation to provide more direct measures of relative parser performance.

The evaluation scores can be made more comparable in FEPa by using the subtlety measures. This method takes into account the level of detail in a parser’s output and the possible ambiguity and underspecification that remains in the parser’s output. Thus, for example, the overall preciseness score ($PR_p$) of parser $p$ can be formulated according to Equation (9-1).

$$PR_p = \frac{preciseness_p * detail_p}{ambiguity_p * underspecification_p} \quad (9\text{-}1)$$

Parsers that produce detailed analyses should be accorded an advantage over those that produce undetailed parses. For example, the level of detail in the syntactic description of a D-based parser could be defined on the basis of the number of D link types in its tagset. The factors of $ambiguity_p$ and $underpecification_p$ are used respectively to account for remaining ambiguity and underspecification in the output. If the parser returns, say 1.05 syntactic tags per word, it should be penalized when it is compared to a parser that produces only a single tag per word.[80] Similarly, a parser that leaves part of the words or structures underspecified should be penalized in a comparison to a parser that assigns a tag for each word.

Another mechanism that FEPa uses to facilitate comparative evaluations and adaptability to evaluators' needs is the use of factors for emphasizing specific criteria. If a particular evaluator regards efficiency as a crucial component, she can emphasize it when making comparisons between parsing systems. For example, if preciseness is seen as the most important criterion, an evaluator can give it extra weight in parser comparisons.

## 9.4 Conclusion

In this chapter, I described in detail a framework for carrying out empirical parser evaluations and explained how such evaluation could be performed. The purpose of the method thus offered is to provide a basis for characterizing how well and efficiently a parser is able to analyze syntax. I also considered the ways in which this model can be utilized to compare the performance of different parsing systems – without the need for using compromising direct comparison metrics. I also outlined a method in Section 9.3 for using FEPa to make comparative evaluations. I utilize this method in Chapter 10 to compare two parsers that use a grammar based on the same formalism and also to compare parsers based on different grammar formalisms.

[80] Determining the level of ambiguity for a syntactic analysis is a rather complicated issue. In order to distinguish between overgeneration and real, inherent ambiguity, the treebank that is used for evaluation should include several parses for sentences that cannot be disambiguated purely on the basis of syntactic information.

The main advantage of this framework over existing evaluation methods is that FEPa adopts a broad approach to parser evaluation and provides a set of criteria for defining the quality of a parser. FEPa also provides a set of measures for carrying out evaluations according to each of the criteria.

The subtlety factors for comparative evaluation in Equation 9-1 are admittedly difficult to define and are therefore open to question. If, however, one wants to compare diverse parsing systems, one has to take into account the difference in the complexity of the decisions that a parser has to make in order to succeed. Although one has to make compromises when defining subtlety factors, one can avoid thereby the main problem associated with attempts to compare parser outputs directly: no compromises needs to be made in the evaluation itself. A parser output can be compared to an evaluation resource with a similar annotation scheme, and there is no need to abstract away from the differences in the outputs of the parsers under consideration. I will describe my practical experiments in which I compare parsers by using FEPa in Section 10.7 of the next chapter.

# IV EVALUATIONS

## 10 FEPa in Use

This chapter describes a series of experiments that use the FEPa. The parsers that were used in the experiments are described in Section 10.1. Section 10.2 analyzes the preciseness results reported in the literature for the six selected parsers. The experiments themselves are described in the following four sections: coverage (Section 10.3), robustness (Section 10.4), efficiency (Section 10.5), and subtlety (Section 10.6). The parsers are then compared in Section 10.7 on the basis of the data obtained from these experiments. Section 10.8 concludes with the findings and analyzes the amendments that will be made to the FEPa framework as a result of the information obtained from these experiments.

### 10.1 The parsers

The six parsers listed in Table 10-1 below were selected for the evaluation. The parsers are based on five different grammar formalisms: the four state-of-the-art parsers are based on CCG, PCFG and LG; one of the older parsers uses a DG formalism, and the remaining one uses its own parser-specific grammar formalism.

**Table 10-1**. The parsers in this experiment. The two "grammar" rows indicate the grammar formalism on which the grammar used by the parser is based, and the type of the grammar. The two "output" rows indicate the type of output used for evaluation in this experiment (Type 1), and the other scheme for those parsers that support more than one output scheme (Type 2). KEY: A= automatically induced; M= manually constructed.

| | | **Apple Pie Parser (APP)** | **C&C Parser** | **Link Grammar Parser (LGP)** | **MINI-PAR** | **Stanford Parser (SP)** | **StatCCG***** |
|---|---|---|---|---|---|---|---|
| *Version* | | v. 5.9, 4 April 1997 | v. 0.96, 23 November 2006 | 4.1b, January 2005 | unknown version, 1998 | v. 1.5.1, 30 May 2006 | preliminary release, 14 January 2004 |
| *Grammar* | *Formalism* | semi context-sensitive*, probabilistic | CCG | LG | DG | Unlexicalized PCFG** | CCG |
| | *Type* | A (PTB) | A (PTB) | M | M | A (PTB) | A (PTB) |
| *Algorithm* | | probabilistic, bottom-up, chart | probabilistic, log-liner, packed chart, supertagger | dynamic programming | principle-based, distributed chart | probabilistic, CYK | probabilistic, generative, CYK |
| *Output* | *Type 1* | PS, 20 | GR, 48 | LG, 107 | D, 27+20 | GR, 38 | CCG derivations**** |
| | *Type 2* | - | PAS | PS | PS | PS | PAS |
| *Reference* | | Sekine (1998) | Clark & Curran (2004) | Sleator & Temperley (1991) | Lin (1998) | Klein & Manning (2003), de Marneffe *et al*. (2006) | Hockenmaier (2003) |

*Sekine (1998) states that the grammar is semi context-sensitive, but leaves this grammar class undefined. **SP supports both unlexicalized and lexicalized grammars. I used the unlexicalized grammar bundled with the parser in the experiments. ***Because StatCCG does not perform POS tagging, I used the MXPOST tagger (Ratnaparki 1996) to preprocess the texts before inputting them to StatCCG. ****StatCCG's lexical category set contains approximately 1,200 types, and there are four atomic types in the syntactic description.

While I wanted to include parsers that use different formalisms, I also wanted to include two CCG parsers in order to find out how well FEPa is capable of comparing parsers that use the same formalism. I chose two parsers, APP and MINIPAR, that were developed in the 1990s, and four state-of-the-art parsers, for this evaluation. My inclusion of older generation systems was motivated by my desire to evaluate the level of development in parsing systems over the last decade. One of these older parsers is a PS parser (APP) and the other a D parser (MINIPAR).

Language processing applications that involve parsing must set practical limits on resource consumption. In order to create similar and equal conditions for all parsers throughout the evaluation, I limited the use of memory to the same value for all the parsers and experiments. I accordingly selected 650 MB as the limit because it is a realistic setting for free working memory in an average personal computer with 1GB of memory.[81] In addition, parsing in the order of hundreds of thousands of sentences with six parsers consumes thousands of hours of processor time. I was therefore obliged to limit memory consumption so that I could run the experiments in parallel on a server and complete them all within a reasonable period of time.

## 10.2 Preciseness

For preciseness evaluation, I used the results reported in the literature. There are several reasons for this. Firstly, these results were available for all the evaluated parsers. Secondly, I did not have access to the PTB, and I would have needed such access for carrying out further experiments. Table 10-2 summarizes the preciseness figures for the six parsers.

[81] I used several methods (depending on the parser concerned) for limiting memory usage. In the Java-based parsers, I set the limit according to the size of the Java heap. With the C&C parser, for example, setting values at 1,250,000 to 1,300,000 for the maxsupercats parameter limited the memory usage to approximately 650 MB. In APP, the limit for the chart size was set at compile time.

**Table 10-2**. Preciseness results reported in the literature. KEY: P = precision, R = recall; F = F-score.

| Parser | Source | Gold standard | Metric | Results | | |
|---|---|---|---|---|---|---|
| | | | | *P* | *R* | *F* |
| *APP* | Sekine (1998) | PTB | Unlabeled PARSEVAL | 71.1 | 70.3 | 70.7 |
| *C&C* | Clark & Curran (2004a) | PTB | Labeled PARSEVAL | 84.8 | 84.5 | 84.6* |
| *LGP* | Molla & Hutchinson (2003) | SUSANNE | GR-based** | 54.6 | 43.7 | 48.5 |
| *MINIPAR* | Sampson (1995) | SUSANNE | Lin's D-based | 88.0 | 80.0 | 83.8 |
| *SP* | Klein & Manning (2003) | PTB | Labeled PARSEVAL | 86.3 | 85.1 | 85.7 |
| *StatCCG* | Hockenmaier (2003) | PTB | Labeled PARSEVAL | 83.7 | 84.2 | 84.0*** |

*90.7 for unlabeled F-score. **Only four GR relations (subj, obj, xcomb, mod) were considered in the evaluation. ***91.2 for unlabeled F-score (Clark & Curran 2004a). F-score 91.3 and 83.3 for unlabeled and labeled word-word dependencies in the PAS output (Hockenmaier 2003).

A direct comparison of the results reported in Table 10-2 would be unreliable. While four of the parser were evaluated on the PTB, the SUSANNE corpus was used for evaluating the two others. Although the same measures were used in the SUSANNE-based evaluations, the results were calculated with different metrics. Even if the parsers had been evaluated on the same data and the same metrics had been applied, the figures would not take into the account the differences in the subtlety of the parsers' outputs. A CCG parser, for example, needs to distinguish between complements and adjuncts and to identify LDDs in order to get a derivation right. Most PCFG parsers, such as SP, do not need to predict these in order to do well on labeled PARSEVAL measures because the LDDs are ignored by these parsers.

One may draw the following conclusions about performance of the parsers on the basis of the preciseness results.[82]

1. SP performs best. Its labeled PARSEVAL F-score on the PTB data is 85.7. This is slightly higher than those of the two CCG parsers.[83]

[82] This ranking does not address the differences in the subtlety of the parser's outputs. See Section 10.6 for preciseness evaluation that considers subtlety.

[83] Hockenmaier (2003) points out two sources of difficulty when comparing CCG preciseness figures to parsers using other grammar formalisms. The first is that since CCG trees use a fine-grained category set, PARSEVAL scores cannot be compared. The second is that the grammar in StatCCG has over 1,200 lexical category types. FEPa does, however, address both kinds of difference by using subtlety measures.

2. C&C, StatCCG and MINIPAR form the second group. No reliable ranking can be made between these three parsers on the basis of the reported scores. Depending on the test settings, either C&C or StatCCG is more precise than the other.[84] While MINIPAR achieved an F-score of 83.8 in D evaluation on the SUSANNE, StatCCG scored 83.3 on labeled PAS evaluation. This indicates a comparable preciseness.
3. APP is the second-worst performer. If one compares it to the unlabeled PARSEVAL evaluations of C&C and StatCCG, the parser scored an F-score which is over 20 percentage points lower than the F-scores of these two parsers. Although the evaluations were performed with different measures on different test sets, it is reasonable to assume that APP is more precise than LGP because it achieved an F-score that is higher by 22.2 percentage points than that achieved by LGP.
4. LGP performs worst. Although the reported evaluation used only four GR types, the F-score is more than 30 percentage points lower than the two parsers for which there are comparable evaluation results available: MINIPAR's F-score on labeled D-based evaluation was 83.8 and CCG's F-score on labeled PAS dependencies was 83.3.

## 10.3 Coverage

In this section I describe a set of parsing coverage evaluations. In addition to examining overall parsing coverage, I will also consider the genre coverage of the evaluated parsers. This will allow me to make judgments about the generalizability of the parsers. The three questions for which I sought answers were as follows:

- What is the parsing coverage of state-of-the-art parsers?
- How does the text genre affect the parsing coverage?
- How much progress has been made on parsing coverage in the last decade?

### *10.3.1 Previous work*

This experiment is the only one reported in the literature that compares the coverage of a set of parsers for English. The studies that critically examine the genre dependency of parsers have all come to the same unsurprising conclusion that the text genre has an effect on parser performance. Clegg and Shepherd (2005) conducted experiments on biomedical data by using the *GENIA treebank*

[84] In the experiments reported by Clark and Curran (2004a), the model was slightly better than Hockenmaier's. But Hockenmaier (2003) compared the performance of the two parsers and found

(GENIA Project 2006). They point out that biomedical English is distinctly different from newspaper English, and that it might indeed be characterized as a sub-language (Friedman *et al*. 2002). Laakso (2005) reports experiments on the CHILDES corpus that consists of transcriptions of conversations between parents and their children. Mazzei and Lombardo (2004) report cross-training experiments in Italian on newspaper and civil law texts, in which an LTAG, CFG and DG were trained on one data set and tested on the other. When they did this, they observed a dramatic drop of usually around 10 to 30 percentage points in the parsing coverage of all the three grammars.

It has been found out that the performance of a probabilistic parser degrades – often to a considerable extent – when it parses texts from genres other than those that were used for training the model. Sekine (1997) observed the following order of performance from best to worse between the training and testing data: the same genre, the same class,[85] all genres, another class, and other genres. Sekine also found out that even a significant increase in the size of the training data from a genre other than the one being parsed often has no effect on the performance. Gildea (2001) reports similar findings.

One of the rare studies that reflects upon the possible reasons for the drop in performance when a parser is applied to a genre other than the one intended when developing the system, is reported by Baldwin *et al*. (2004). The authors evaluated the performance of the manually constructed ERG HPSG grammar (Copestake & Flickinger 2000) on 20,000 sentences from the *British National Corpus*. The grammar was created on the basis of corpus of data extracted from informal genres such as conversations about schedules and e-mails about e-commerce. Baldwin *et al*. (2004) restricted the experiment to sentences with a *full lexical span* (i.e. sentences that only contain words included in the lexicon). ERG had a full lexical span for 32% of the sentences in the test data. The parser was able to generate a parse for 57% of these. A total of 83% of the analyses were correct. The parser was thus able correctly to parse 47.3% of the sentences with full lexical span. This represented 15.1% of all the sentences. These results indicated that an extension of the grammatical coverage of the grammar increases the coverage on the unparsable sentences with a full lexical span (18% of all the sentences). However, extending the lexical coverage raises the coverage on the sentences without a full lexical span (68% of the whole test set). They therefore came to the conclusion

---

out that when they were trained on the same data, her parser slightly outperformed the parser of Clark *et al*. (2002) with 90.5% unlabeled precision and 91.1% recall.

[85] In Sekine's experiment, class refers to the distinction between fiction and non-fiction texts.

that lexical expansion should be the first step in the process of parser enhancement.

### *10.3.2 Test settings*

I performed a parsing coverage evaluation by using unannotated texts drawn from MGTS. The design of the test settings was guided by the three questions enumerated above. I answered the first question by parsing a document collection that contained hundreds of thousands of sentences and by then measuring the coverage of the parsers on the data. Because MGTS is divided into genre-specific subsets, this allowed to measure the effects of genre variance and so provide an answer to the second research question.

One might argue on the one hand that coverage alone is a rather weak measure of a parser's performance and therefore of its generalizability. An obvious problem inherent in the measurement of coverage alone is that a parser, such as APP, that returns undetailed and flat analyses will easily produce high coverage, whereas a parser (such as C&C and StatCCG) that outputs detailed analyses will be unable to cover all the input sentences.

On the other hand, as I pointed out in Section 9.2.6, it is clear that coverage and preciseness are connected. The connection between coverage and preciseness is clear in the case of sentence-level evaluation measures: a sentence that cannot be fully analyzed cannot have a complete match with the correct structure in the evaluation resource. I consequently argue that coverage can be used an indirect measure of generalizability. It sets the upper bound for performance on sentence-level measures such as the CM and LA sentence measure. An evaluation should nevertheless always be accompanied by data about the preciseness of the parser and the level of detail in its output.

The most important decision about parsing coverage evaluation is how to make the distinction between a covered and an uncovered sentence. Since it was also my intention to collect data about the proportion of sentences for which the parsers generated a fragmented analysis, I had to define other criteria for this purpose. These criteria have to be defined separately for each parser. I set out the criteria that I used in this experiment in Table 10-3.

**Table 10-3**. The criteria for defining whether a sentence was covered.

| Parser | Criteria | |
|---|---|---|
| | ***Covered sentences*** | ***Fragmented analyses*** |
| *APP* | A single S non-terminal is found which dominates the whole sentence. | More than one S non-terminal is found. |
| *C&C* | The parser marks the sentence as fully parsed. | After projecting each GR to a graph that allows cycles, more than one connected set is found. |
| *LGP* | At least one linkage without null links is found. | No linkage is found that does not contain null links. |
| *MINIPAR* | A single root is found for the sentence that is connected to all the words in the sentence through a path. The root is, in addition, assigned with a phrase/sentence type marker. | The analysis contains U tags that indicate unrecognized structures. |
| *SP* | After projecting each GR to a graph that allows cycles, only one connected set is found. | More than one connected set is found. |
| *StatCCG* | StatCCG does not mark the sentence as "failed" or "too long" in its output. | No sentence-level non-terminal is found in the CCG derivations output. |

I implemented a set of tools in Java to record the statistics from the parsers' outputs, and also devised experiment runner tools for some of the parsers. Whenever a parser crashed, these tools restarted the evaluation process from the following sentence. The results are reported by determining an overall percentage of the sentences covered over all the text genres and also by determining separate results for each genre.

### *10.3.3 Results*

Table 10-4 summarizes the results of the experiment. The parsing coverage of the parsers for each of the sub-corpora in MGTS is reported separately. Total figures are given on parser and sub-corpus level. The generalizability of the parsers was measured by comparing their coverage on the newspaper genre to their coverage on the lowest-scoring genre:

$$Generalizability = 1 - \frac{MIN(Coverage(all\quad genres)) - Coverage(newspaper)}{Coverage(newspaper)}$$

**Table 10-4.** Comparison of the parsing results for each sub-corpus and parser. The column labeled "Average" gives the average of the coverage figures for the six genres weighted according to the number of sentences in each genre. The column labeled "Generalizability" shows the percentage of the coverage in the lowest-scoring genre compared to the coverage in the newspaper genre.

| **Parser** | **Newspaper** | **Legislation** | **Fiction** | **Nonfiction** | **Religion** | **Biomedicine** | **Average** | **Gener-alizability** |
|---|---|---|---|---|---|---|---|---|
| *APP* | 99.8 | 98.9 | 97.5 | 96.4 | 93.1 | 98.9 | 98.5 | 93.3 |
| *C&C* | 87.8 | 84.9 | 86.0 | 81.2 | 75.5 | 84.8 | 85.0 | 86.0 |
| *LGP* | 74.1 | 38.7 | 38.4 | 42.1 | 15.0 | 49.4 | 50.2 | 20.2 |
| *MINIPAR* | 88.0 | 68.8 | 68.0 | 71.5 | 34.4 | 70.1 | 72.1 | 39.1 |
| *SP** | 99.8 | 99.5 | 98.0 | 98.3 | 98.9 | 98.5 | 99.2 | 98.2 |
| *StatCCG* | 96.7 | 85.2 | 87.7 | 86.7 | 94.0 | 83.3 | 89.1 | 86.1 |
| ***Average*** | 91.0 | 79.3 | 79.3 | 79.4 | 68.5 | 80.8 | 82.4 | 70.5 |

*SP experienced a coverage drop of tens of percentage points in comparison to other genres on the Hansard dataset. This was caused mainly by a single issue: the dataset contained a number of sentences that contained only a single word – sentences such as "Nay.", "Agreed.", "No." and so on. Because no root node is assigned to D analysis by SP, the parser did not return any analysis for such sentences. These sentences were omitted from the evaluation. When the sentences were included, the coverage on legislation data was 59.5%.

Table 10-5 breaks down the coverage figures to indicate the percentage of the analyses that failed or were incomplete, and the number of occasions on which the parser crashed or terminated during the process.

**Table 10-5.** Breakdown of the failures. All the results are reported as a percentage of the total number of sentences (826,485). The column labeled "Incomplete" reports the proportion of sentences that were parsed but the analysis was not full. The column labeled "Failed" indicates those cases in which the parser was not able to return a parse. The column labeled "Terminated" shows the proportion of the cases in which the parser crashed or terminated during the process of parsing a sentence.

| **Parser** | **Incomplete** | **Failed** | **Terminated** |
|---|---|---|---|
| *APP* | 1.5 | 0.0 | 0.001 |
| *C&C* | 12.8 | 2.2 | 0.006 |
| *LGP* | 42.2 | 7.4 | 0.206 |
| *MINIPAR* | 27.9 | 0.0 | 0.009 |
| *SP* | 0.5 | 0.4 | 0.002 |
| *StatCCG* | 9.6 | 1.4 | 0.000 |
| ***Average*** | 15.8 | 1.9 | 0.037 |

APP and SP performed at the highest rate of coverage in this experiment. APP produces shallow parses, which enables it to obtain a high coverage. When it is unable to recover a full parse for a sentence, this parser enters a fitted parsing mode in which a fragmented parse is recovered from the partial trees in the chart. This typically causes the analysis to have two root (S) constituents. Klein and Manning (2003) claim that SP is able to analyze all the sentences in section 23 of the PTB in a machine with 1GB of memory. If one excludes the one-word sentences from the legislation dataset, SP gave the best coverage and best generalizability rate. The most common reason for not fully covering a sentence was a fragmented analysis caused by segmentation errors in which a delimiter was considered to be a node in the D tree.

The overall performance of C&C (85.0%) was slightly worse than the performance of the other CCG-based system that scored an average coverage of 89.1%. Compared to StatCCG, C&C's coverage was, however, more consistent over the genres. Although StatCCG skipped extremely long sentences, it only did this on 353 occasions.

LGP (average coverage 50.2%, generalizability rate 20.2%) and MINIPAR (72.1%, 39.1%) gave the worst coverage and lowest generalizability in the experiment. While MINIPAR achieved an 88.0% coverage on the newspaper corpus, its performance dropped over 15 percentage points on other corpora. Its coverage was only 34.4% with the religion corpus. The most commonly occurring problem with this data was a fragmented analysis occasioned by sentences beginning with an "And" or "Or" that was not connected to any other words in the parse tree. LGP coverage on the religion dataset was the lowest in the whole experiment, only 15.0%.

### *10.3.4 Conclusion*

The six parsers were able to cover, on average, 82.4% of the sentences. The coverage was, unsurprisingly, highest on the newspaper genre. The lowest average coverage was achieved on the religion and legislation genres. The difficulties in parsing the religious texts are attributable at least in part to the length of the sentences in the sub-corpus (on average 27.1 words per sentence), which was the highest over all the genres. The legislation genre consists of transcribed speech, which may be the main reason for the lower-than-average performance on that data. Contrary to my expectation, the biomedical genre, with its specialist terminology, was not the most difficult genre for the parsers.

If one compares these results to those obtained by MINIPAR, it is clear that the coverage of the newer parsers has improved. APP produces a shallow analysis that enables it to achieve a high coverage. The good performance of the APP may be partly explained by its rather poor preciseness: its rate of just over 70% is much lower than that of other parsers. The poor performance of the two parsers that are based on manually constructed grammars, MINIPAR and LGP, supports what was said in Section 5.2 about how probabilistic parsers typically have an advantage over rule-based ones with regard to coverage.

## 10.4 Robustness

The robustness comparison of parsers for English that I describe in this section are the only evaluations of their kind reported in the literature. In making this evaluation I sought answers to the following questions:

- What is the overall robustness of the evaluated parsers?
- What effect does an increasing error level have on the parsing results?
- How stable are the parsers?

### *10.4.1 Previous work*

Not much work has been done on methods for empirically evaluating the robustness of parsers. Foster (2004) is the only researcher to have reported a robustness evaluation of a parser for English in the literature. She evaluated Charniak's (2000) PCFG parser by using the corpus described in Section 6.1.3. The parser returned the same analysis for correct and erroneous versions of the same sentence in 32% of the cases. The highest score was achieved on agreement errors and the use of the wrong preposition. Of these, over 70% of cases obtained a complete match.

The research most closely resembling my own is reported by Bigert *et al*. (2005) (see Section 8.3.2). The automatic introduction of errors enabled the researchers to undertake a controlled testing of degradation which is the effect of an increased error rate on a parser's output. But they only applied this method for Swedish.

### *10.4.2 Test settings*

In order to evaluate the degree to which a parser can handle noisy input, and spelling errors in particular, the following experiment was set up. A set of test sentences, both correct and erroneous, was obtained from the RobSet. I then used the metric and measures defined in Section 9.1.4, namely UR score and LR score, to compare the overall robustness of the six evaluated parsers and considered the degradation in the parsers' performances in relation to increasing error levels. In order to evaluate the stability of the parsers, I recorded the number of crashes while parsing the MGTS. The final step was to consider the preciseness figures reported in the literature for the parsers and to compare them to these findings.

While some might claim that input with misspelled words tests the accuracy of POS tagging rather than the preciseness of syntactic description, it is my opinion that such an assertion is inadmissible. Firstly, if a parser is able to distinguish grammatical sentences from ungrammatical ones, it should be able to rule out a considerable number of analyses generated by erroneous POS sequences which the misspelled words might have caused. Secondly, if a parser's POS tagger is well designed, it should not try to disambiguate the POS tags of the words that it cannot recognize, but rather leave disambiguation to the syntactic analysis component.

### *10.4.3 Results*

Table 10-6 summarizes the results of the experiment on noisy input and indicates the overall robustness scores as well as separate scores for each error level (1,2,3).

**Table 10-6**. The results of the experiment are reported separately for the two evaluation metrics. Separate scores for each error level are also given. The column labeled "D" gives the degradation rates. These are defined by comparing the robustness scores on levels 1 and 3 and calculating the drop in performance in percentages.

| **Parser** | **UR** | | | | | **LR** | | | | |
|---|---|---|---|---|---|---|---|---|---|---|
| | *Avg.* | *1* | *2* | *3* | *D* | *Avg.* | *1* | *2* | *3* | *D* |
| *APP* | 43.3 | 59.2 | 28.7 | 14.9 | 74.9 | 37.0 | 54.5 | 19.2 | 7.5 | 86.3 |
| *C&C* | 63.8 | 72.9* | 62.8 | 40.4 | 44.6 | 45.4 | 60.8 | 34.0 | 14.9 | 75.5 |
| *LGP*** | 29.8 | 40.4 | 22.3 | 8.5 | 78.9 | 17.6 | 22.0 | 20.2 | 3.2 | 85.5 |
| *MINIPAR* | 37.9 | 57.4 | 22.1 | 1.1 | 98.2 | 20.6 | 33.2 | 6.3 | 1.1 | 96.8 |
| *SP* | 55.3 | 71.0 | 42.6 | 25.5 | 64.0 | 19.2 | 29.4 | 9.6 | 1.1 | 96.4 |
| *StatCCG* | 57.1 | 72.6 | 41.5 | 30.9 | 57.5 | 44.0 | 58.8 | 27.7 | 20.2 | 65.6 |
| ***Average*** | 47.8 | 62.3 | 36.7 | 20.2 | 69.7 | 30.6 | 43.1 | 19.5 | 8.0 | 84.4 |

*C&C failed to parse 23 correct sentences in this sub-corpus. Because I considered the inability of a parser to cover some sentences to be a serious robustness flaw, I deliberately included these sentences in the calculations. This brought the scores down from 80.2 to 72.9 and 66.8 to 60.8 for UR and LR respectively. **LGP leaves some of the ambiguities unresolved and returns several parses for such sentences. This occurred with most of our test sentences. But in the interest of ensuring an entirely fair comparison among the parsers, I only considered the first highest-ranking linkage in the evaluation.

The results show, as was expected, that the performance of the parsers degrades as the level of distortion in the input sentences increases. The results indicate that performance usually declines by tens of percentage points when parsers are presented with texts that contain misspellings. While the parsers produced the same analyses for correct and erroneous sentences for 43.1% of the sentences in labeled evaluation on error level 1, the average score on error level 3 was only 8.0%.

When tested on the purpose-built test set of 443 sentences, the best parser in the experiment (C&C parser) was able to return exactly the same parse tree for the grammatical and ungrammatical sentences for 60.8%, 34.0% and 14.9% of the sentences with one, two or three misspelled words, respectively.

The overall performance of StatCCG and SP are similar on the unlabeled evaluation in which the LGP performs considerably worse than the other five parsers. Error level 3 on labeled evaluation was the only category in which StatCCG outperformed C&C in this experiment. Figure 10-1 shows the performance of the parser at each error level for unlabeled evaluation.

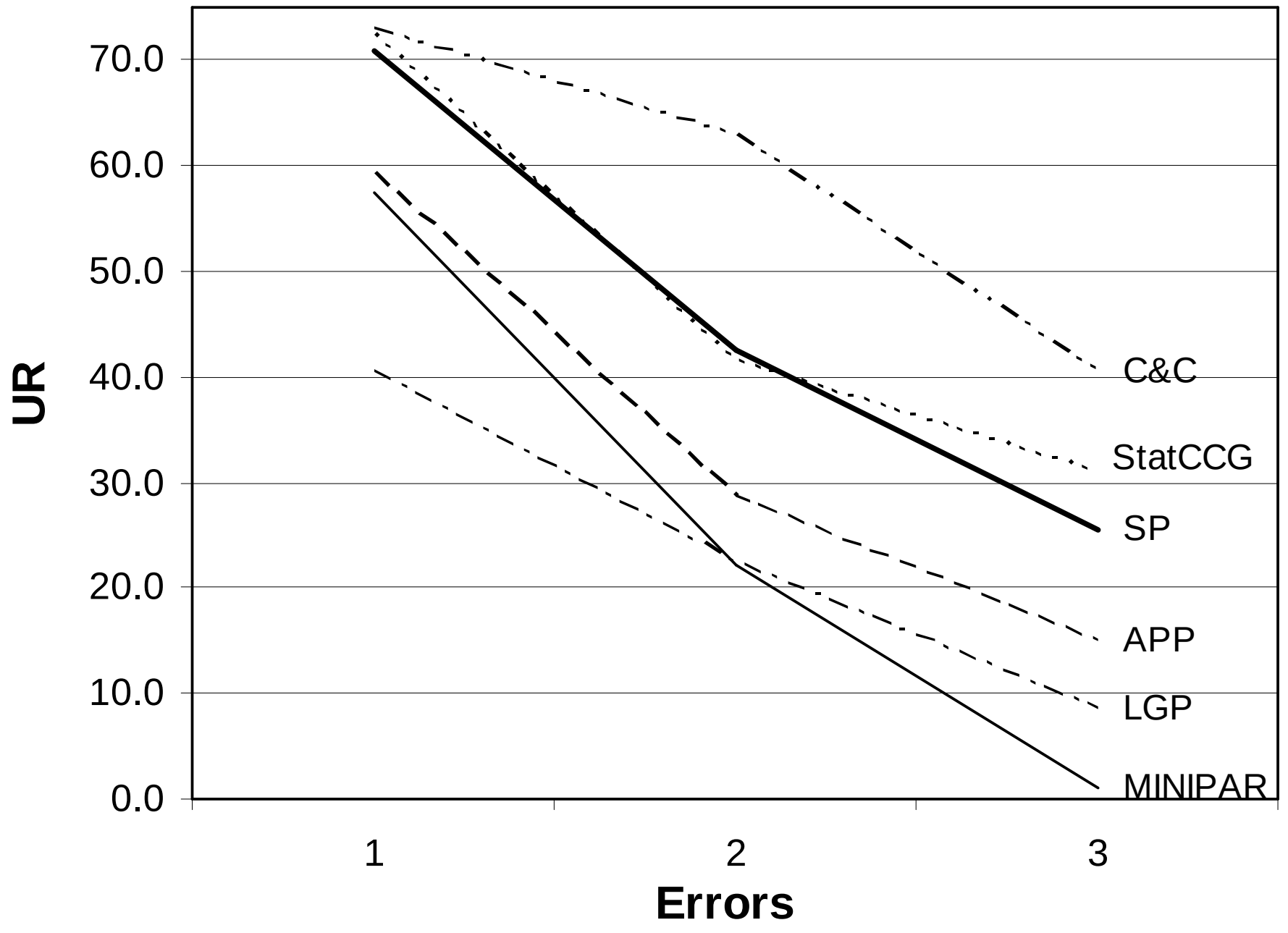

**Figure 10-1**. Unlabelled robustness scores on the three error levels. The UR-axis indicates the robustness score on the three error levels given on the x-axis.

Figure 10-1 shows that while C&C is the best performer, LGP and MINIPAR are the worst performers on all three error levels on structural similarity evaluation.

StatCCG beats SP on error levels 1 and 3. The same pattern of performance degradation is repeated as the error level rises. While the robustness score of C&C decreases from 72.9% on error level 1 down to 40.4% on level 3, equaling to 44.6% drop, the figure for LGP is 78.9% and 98.1% for MINIPAR. StatCCG and SP are again in the upper middle-class, at around 60%, while APP positions itself between these two and C&C with 74.8% degradation.

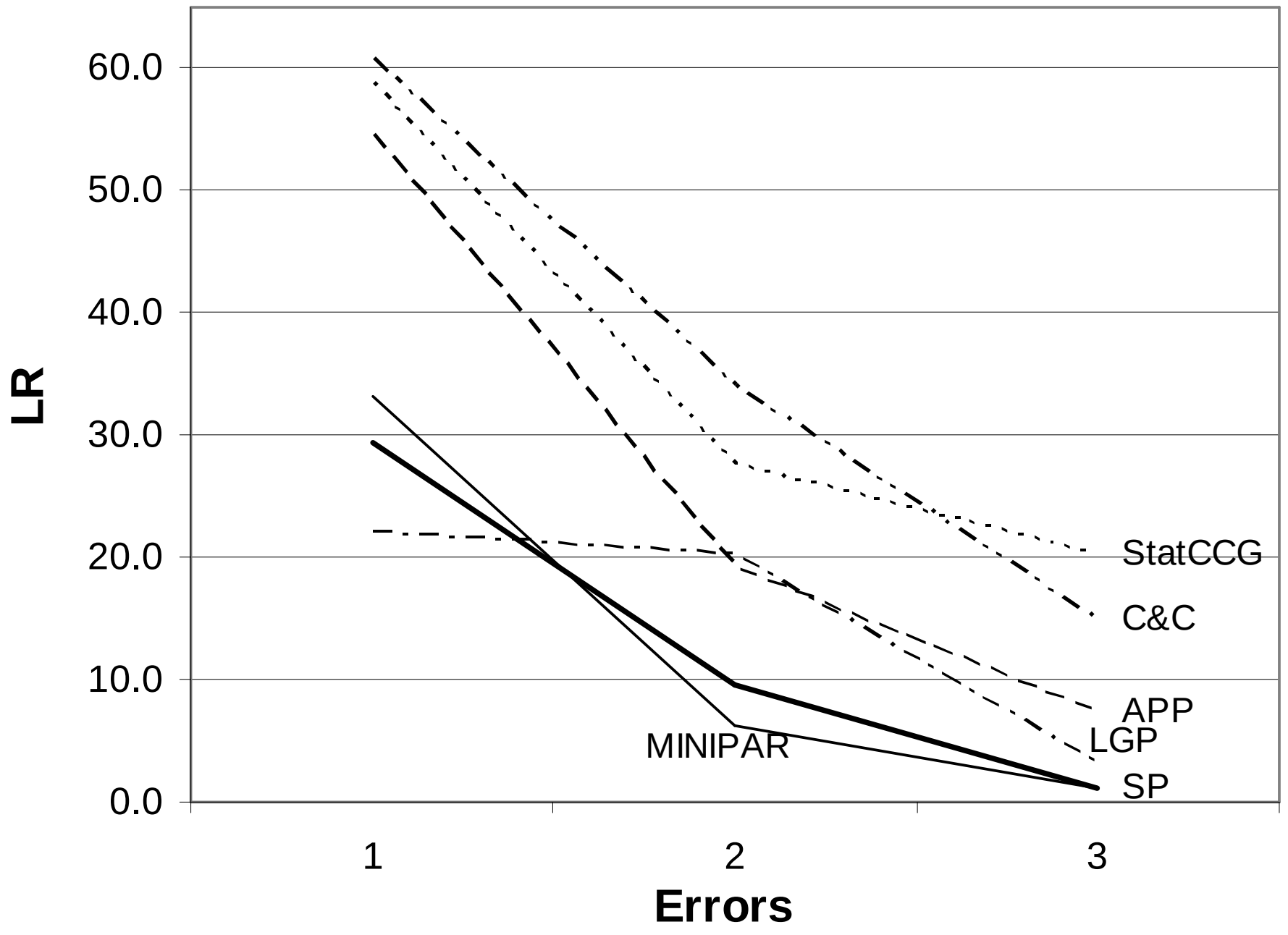

**Figure 10-2**. Labeled robustness scores.

The results of the labeled evaluation (see Table 10-6 and Figure 10-2 above) indicate that while structural similarity can be preserved for close to 73% of the sentences on error level 1 by the two best parsers (StatCCG and C&C), the performance drops to around 60% when the labels are also required to match. The degradation figures in labeled evaluation are: StatCCG (65.6%), C&C (75.5%), LGP (85.5%), APP (86.3%), SP (96.4%), and MINIPAR (96.8%).

The large tagset of 107 tags makes it difficult for LGP to obtain a good performance on labeled evaluation. LGP was able to rank third on degradation, largely due to the fact that its performance was poor (22.0%) even on the lowest error rate. It was rather surprising, however, to observe how dramatically SP's robustness scores decreased between the unlabeled and labeled evaluations. This indicates a flaw in the robustness mechanisms of the parser, a flaw that might be attributable to the poor ranking of candidate parse trees for noisy sentences or problems with the POS tagging model of unknown words.

In addition to being able to perform well when faced with just slightly distorted input, C&C is the most consistent parser when faced with increasing distortion: the unlabelled robustness score dropped by 44.6% from error level 1 to error level 3. LGP has the lowest reported preciseness and its performance also degrades drastically, especially on unlabelled evaluation (a 78.9% drop from error level 1 to error level 3). MINIPAR was the worst performer with regard to degradation, scoring 98.2 and 96.8 for unlabeled and labeled degradation respectively.

I carried out a stability evaluation by parsing the whole of the 826,485-sentence MGTS and recording the number of times each of the parsers terminated during the parsing. These figures are given in the column "Terminated" in Table 10-5 (above). LGP crashed in 0.21% of the cases, which makes it the most unstable parser in the experiment. StatCCG and APP proved to be the most stable parsers in the experiment. While StatCCG did not crash at all, APP terminated once. MINIPAR crashed 72 times altogether (i.e. for 0.009% of the sentences). SP and C&C were on the middle ground, terminating on 0.002 and 0.006 per cent of the cases.

### *10.4.4 Conclusion*

Table 10-7 summarizes the findings of the robustness experiments.

**Table 10-7.** The overall robustness of the parsers. The columns labeled "Noisy input" and "Degradation" show the ranking on the basis of the labeled evaluation. The column labeled "Rank" shows the overall ranking of the parser on the basis of the averages of the rankings in the three measures.

| Parser | Noisy input | Degra-dation | Stability | Rank |
|---|---|---|---|---|
| *APP* | 3 | 4 | 1 | 3 |
| *C&C* | 1 | 2 | 4 | 2 |
| *LGP* | 6 | 3 | 6 | 6* |
| *MINIPAR* | 4 | 6 | 5 | 5 |
| *SP* | 5 | 5 | 3 | 4 |
| *StatCCG* | 2 | 1 | 1 | 1 |

*LGP and MINIPAR scored the same average rank. But because of its instability, which was much greater than that of any other parser in the experiment, it ranks after MINIPAR.

C&C was the only parser in the experiment that produced coverage of less than 100% on our test set. In spite of this, it proved to be by far the best-performing parser. In comparison to StatCCG, C&C performed better – especially on the unlabeled measure. The performance of APP was on the middle ground between the three best and two worst parsers. It compared especially well on error levels 1 and 2 for the labeled evaluation. However, its performance degraded drastically on error level 3.

The performance of MINIPAR and LGP left a lot to be desired across all of the evaluation categories. They are both based on rule-based, manually constructed grammars. LGP also returned up to thousands of linkages per input sentence. All these observations, together with the low preciseness figures reported in the literature, indicate serious problems in the disambiguation model of the parsers. These results support the observation made in Section 5.2 about the lack of robustness in many rule-based parsers. APP was worse than the state-of-the-art parsers – with the exception of LGP. This indicates that considerable progress has been made in probabilistic parsing with regard to robustness.

## 10.5 Efficiency

This section reports a set of efficiency evaluations. These evaluations were carried out for the purpose of measuring practical efficiency of the six parsers. This kind of efficiency answers the question: How long does it take a parser to analyze a particular set of sentences on a given machine? As the experiments were run on a single machine configuration, a meaningful comparison of the parsers could be made.

### *10.5.1 Test settings*

The experiments were run under Linux Fedora Core 5 on a Pentium M 1.72 GHz machine with 1 GB of memory. The machine was dedicated only to the parsing task during the experiments. A 20,000-sentence subset of the MGTS newspaper subcorpus was parsed by each of the parsers and the total parse time recorded. For parsers that did not perform POS tagging, the processing time of the external tagger was added to this time.

### *10.5.2 Results and conclusion*

Table 10-8 and Figure 10-3 give the results of the efficiency evaluation performed on a 20,000-sentence test set consisting of newspaper texts.

**Table 10-8.** A comparison of running times of six parsers on newspaper data of 20,000 sentences. The column labeled "Time complexity" shows the complexity of the best known algorithm for the type of grammar the parser uses.

| **Parser** | **Time (min.)** | **Sec./sent.** | **Time complexity** |
|---|---|---|---|
| *APP* | 130.14 | 0.39 | N/A |
| *C&C* | 6.58 | 0.021 | $n^6$ |
| *LGP* | 214.18 | 0.64 | $n^3$ |
| *MINIPAR* | 4.40 | 0.014 | $n^3$ |
| *SP* | 230.31 | 0.69 | $n^3$ |
| *StatCCG* | 657.22 | 1.97 | $n^6$ |
| ***Average*** | 240.35 | 0.72 | - |

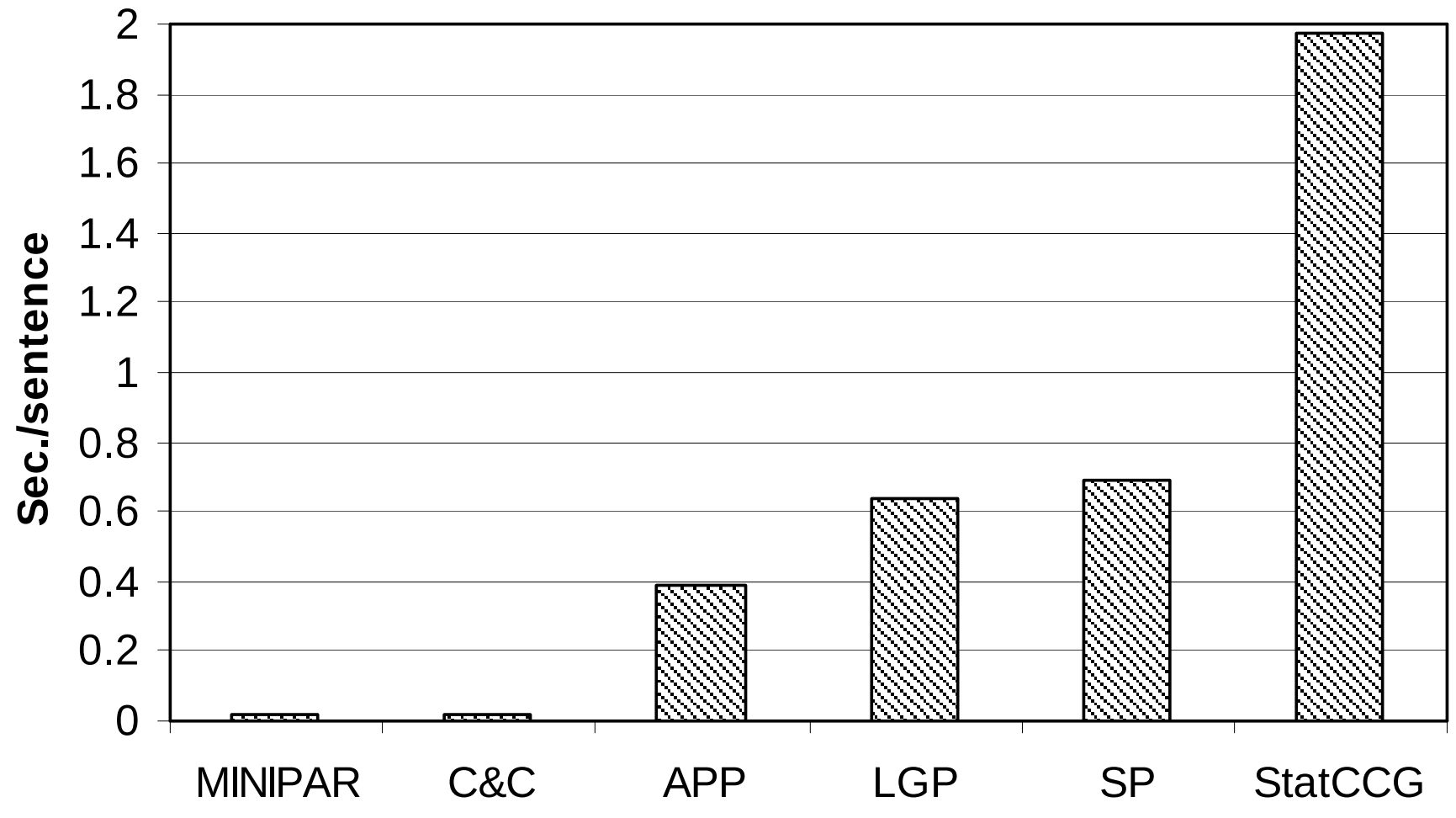

**Figure 10-3**. Average parsing times in seconds per sentence.

The parsers can be grouped into four categories in terms of their efficiency. MINIPAR (0.014 sec. per sentence) and C&C (0.021) are considerably faster than the other parsers. The speed of APP lies between that of MINIPAR and C&C and the group formed by LGP and SP, with an average parse time of 0.39 sec. per sentence. StatCCG is almost three times slower than SP. It took StatCCG almost two seconds on average to parse a sentence. This experiment confirms the observation that I made in Chapter 4, namely, that the theoretical upper bounds for the time complexity of parsing a certain grammar formalism have little to do with a parser's *practical* efficiency.

## 10.6 Subtlety

This section describes the evaluation of the subtlety of the six parsers. In this experiment, I addressed four notions of subtlety: 1) the levels of representation in the parser's output, 2) the detail on each level, 3) the level of ambiguity, and 4) the amount of underspecification remaining in the output.

### *10.6.1 Test settings*

I determined the levels of representation in each parser's output on the basis of the documentation of the system. I used both the documentation and observation of outputs to determine the size of the POS and syntactic tagsets, and measured the level of ambiguity and underspecification remaining in the parsers' outputs. For this purpose I implemented a set of software tools in Java that provide statistics about the richness, the level of ambiguity and the amount of underspecification in the outputs.

### *10.6.2 Results*

Table 10-9 summarizes the levels of representation and detail in the output of the six parsers. The amount of detail on level *x* (i.e. POS or syntax) of an output scheme *s* is calculated in the following way:

$$score_{s,x} = \log\left(\left|tagset_{s,x}\right|\right) \tag{10-1}$$

When considered intuitively, this equation appears to give a reasonable advantage to parsers that produce output that is more detailed. For example, out of the six parsers in this experiment APP receives the lowest score for its word-level annotation (0.0), StatCCG, C&C and SP achieve the highest scores (1.65). On a syntactic level, APP once again scores the lowest score (1.30), and StatCCG achieves the highest score, 3.02.

A scoring scheme for the level of detail in an output scheme should give more credit to the richness of the syntactic description because it is a more important feature of a syntactic parser's output than the richness of the word level description. The scheme therefore assigns a double weight to the syntactic output scheme in comparison to the POS scheme. The overall score of an output scheme *s* is defined in FEPa as:

$$socre_S = \frac{score_{S,WORD\ \ LEVEL}}{2} + score_{S,SYNTAX} \qquad (10\text{-}2)$$

**Table 10-9.** A comparison of the output schemes. The two columns included under the label "Word level" indicate whether the word level tags are included in the output, and the number of tags in the POS tagset. The four columns included under the label "Syntactic level" indicate whether the scheme includes PS or D representations respectively. The columns "Type" and "No. tags" indicate whether the parser supports PS and D-style output schemes, and the type and number of tags in the syntactic description used for evaluation in the experiments. The column labeled "Score" shows a score for the richness of the output scheme as defined in Equations 10-1 and 10-2.

| **Parser** | **Word level** | | **Syntactic level** | | | | **Score** | **Rank** |
|---|---|---|---|---|---|---|---|---|
| | | **No. tags** | **PS** | **D** | **Type** | **No. tags** | | |
| *APP* | No | 0 | Yes | No | PS | 20 | 1.3 | 6 |
| *C&C* | Yes | 36+9* | No | Yes | GR | 48 | 2.5 | 2 |
| *LGP* | Yes | 8 | Yes | Yes | LG linkage | 107 | 2.5 | 2 |
| *MINIPAR* | Yes | 18 | No | Yes | D | 27 | 2.1 | 5 |
| *SP* | Yes | 36+9* | Yes | Yes | GR | 48 | 2.5 | 2 |
| *StatCCG* | (Yes)** | 36+9* | Yes | Yes | PAS | 1044*** | 3.8 | 1 |

*The tagset of the PTB consist of 36 word categories and 9 tags for punctuation. *StatCCG does not produce POS tags, but expects a tagged input. ***As observed from the output of the parser for the MGTS legislation sub-corpus of over 390,000 sentences.

Table 10-10 is concerned with the level of ambiguity and underspecification in the outputs of the parsers.

**Table 10-10.** A comparison of level of ambiguity and underspecification. The results are given as the average number per covered sentence, calculated over the same 20,000-sentence test set that was used in efficiency evaluation.

| Parser | Underspecified words per sent. | No. of analyses per sentence | Rank |
|---|---|---|---|
| *APP** | 0.0006 | 1.0 | 5 |
| *C&C* | 0.0 | 1.0 | 1 |
| *LGP* | 0.0 | 1.259*** | 6 |
| *MINIPAR* | 0.0 | 1.0 | 1 |
| *SP** | 0.0** | 1.0 | 1 |
| *StatCCG* | 0.0 | 1.0 | 1 |

*The underspecification in APP and SP is marked by the tag X. **The average number of X tags per sentence in the PS output was 0.005. The GR output does not contain underspecified relations. The underspecification in the SP's output was excluded from the evaluation because the parser was evaluated using the GR-style output. ***The number of analyses returned after ranking the linkages based on their costs. On average the parser found 50182.0 linkages per sentence.

### *10.6.3 Conclusion*

Table 10-11 combines the results represented in Tables 10-9 and 10-10 to provide an overall subtlety ranking of the six parsers.

**Table 10-11.** Overall subtlety ranking of the parsers. The column labeled "A&U" gives the ranking on the basis of ambiguity and underspecification (see Table 10-10).

| Parser | Detail | A&U | Overall |
|---|---|---|---|
| *APP* | 6 | 5 | 6 |
| *C&C* | 2 | 1 | 2 |
| *LGP* | 2 | 6 | 5 |
| *MINIPAR* | 5 | 1 | 4 |
| *SP* | 2 | 1 | 2 |
| *StatCCG* | 1 | 1 | 1 |

In addition to creating the kind of subtlety profile of the parsers that is given in Table 10-11, subtlety can also be taken into consideration directly in the evaluation measures. Equation 9-1 suggested a way in which these factors can be applied in preciseness evaluation. I have already defined in Equations 10-1 and 10-2 how the level of detail can be measured. The factor for underspecification in the output of scheme *o* is formulated as follows:

$$factor_o = 1 + score_o \quad (10\text{-}3)$$

In Equation 10-3, $score_o$ refers to the underspecification rate for the output scheme *o* given in Table 10-10. The ambiguity factor is equal to the average number of analyses per sentence given in Table 10-10. Table 10-12 gives the preciseness evaluation of the parsers that takes the subtlety of the output schemes into consideration according to the above definitions.

**Table 10-12.** The preciseness evaluation of the parsers that takes the subtlety of the output schemes into consideration. F-scores were obtained from Table 10-2. The column labeled "Detail" indicates the score for the level of detail in the parser's output as defined in Table 10-10. The figures in the columns labeled "Ambiguity" and "Underspecification" were obtained based on Table 10-10 as defined above. The column labeled "Score" gives the preciseness score for the parser as defined in Equation 9-1.

| Parser | Preciseness | | Detail | Ambiguity | Under-specification | Score | Rank |
|---|---|---|---|---|---|---|---|
| | F-score | Rank | | | | | |
| *APP* | 70.7 | 5 | 1.3 | 1.0 | 1.0006 | 91.9 | 5* |
| *C&C* | 84.6 | 2 | 2.5 | 1.0 | 1.0 | 212.2 | 2* |
| *LGP* | 48.5 | 6 | 2.5 | 1.26 | 1.0 | 95.6 | 5* |
| *MINIPAR* | 83.8 | 2 | 2.1 | 1.0 | 1.0 | 172.5 | 4 |
| *SP* | 85.7 | 1 | 2.5 | 1.0 | 1.0 | 214.9 | 2* |
| *StatCCG* | 84.0 | 2 | 3.8 | 1.0 | 1.0 | 323.0 | 1 |

*The differences between the pairs APP/LGP and SP/C&C were considered too small to make a reliable ranking between the parsers.

## 10.7 Overall Results and Comparison

Table 10-13 shows an overall comparison of the parsers based on their average rankings according to the five criteria: preciseness, coverage, robustness, efficiency, and subtlety.

**Table 10-13.** An overall comparison of the parsers. The column labeled "Rank" indicates the final comparative ranking of the parsers.

| **Parser** | **Preciseness** | **Coverage** | **Robustness** | **Efficiency** | **Subtlety** | **Rank** |
|---|---|---|---|---|---|---|
| *APP* | 5 | 2 | 3 | 3 | 6 | *5* |
| *C&C* | 2 | 4 | 2 | 2 | 2 | *1* |
| *LGP* | 5 | 6 | 6 | 4 | 5 | *6* |
| *MINIPAR* | 4 | 5 | 5 | 1 | 4 | *4* |
| *SP* | 2 | 1 | 4 | 5 | 2 | *2* |
| *StatCCG* | 1 | 3 | 1 | 6 | 1 | *2* |

FEPa allows the evaluator to adjust the comparative evaluation scheme for particular purposes. What follows are a few examples of how this may be done. One may argue that preciseness is the most important evaluation criterion for a parser. Quickly produced analyses that cover a wide range of sentences structures are of limited use if they are erroneous. In example A in Table 10-13, preciseness is allocated a double weight in comparison to the other criteria when calculating the average rankings for the parsers. Because a detailed analysis was not regarded as crucial for the application for which the parser was being selected, subtlety was not used as a factor in the evaluation. This might occur, for example, when a parser is intended to be used in a named entity recognizer, in which its most important function is to recognize the NP chunks reliably.

In example B, the evaluator has determined that efficiency is not a key factor in the particular application for which the parser will be used. This might occur, for example, in a text mining system that is not used for real-time queries, but rather for mining information that will be used later. Instead of giving the weight *one* to all of the criteria, efficiency is allocated only one half of the weight.

In example C, the NLP application for which the parser was selected was such that coverage and efficiency were considered to be the most important evaluation criteria. These two criteria are therefore given a double weight. This might happen, for example, when one selects a parser for an MT system, in which the user inputs need to be handled in real-time and the parser needs to cover as high proportion of sentences as possible so that it will be possible to translate the sentences. Table 10-14 shows the ranking of the parsers according to each of the comparative schemes described above.

**Table 10-14.** Overall comparisons of the parsers that each emphasize different criteria.

| Parser | Preciseness | Coverage | Robustness | Efficiency | Subtlety | Rank | | |
|---|---|---|---|---|---|---|---|---|
| | | | | | | A | B | C |
| APP | 5 | 2 | 3 | 3 | 6 | *5* | *4* | *5* |
| C&C | 2 | 4 | 2 | 2 | 2 | *2* | *2* | *2* |
| LGP | 5 | 6 | 6 | 4 | 5 | *6* | *6* | *6* |
| MINIPAR | 4 | 5 | 5 | 1 | 4 | *4* | *5* | *4* |
| SP | 2 | 1 | 4 | 5 | 2 | *3* | *3* | *1* |
| StatCCG | 1 | 3 | 1 | 6 | 1 | *1* | *1* | *2* |

## 10.8. Conclusion and Future Work

This section concludes with the findings obtained from the experiments (Section 10.8.1). Section 10.8.2 considers possible directions for future work.

### *10.8.1 Experiments and FEPa*

The two CCG parsers were the best performers to emerge from these comparisons. One could rank either of the two as top performers depending on the comparison scheme. The only criteria for which neither StatCCG nor C&C emerged as top performers was coverage. StatCCG also failed to rank either first or second in efficiency evaluation. The only criteria in which SP performed much worse than C&C and StatCCG was robustness. While one could locate the performance of MINIPAR in the middle on the basis of the other criteria, it proved to be the most efficient of all the six parsers. APP was especially compromised by its lack of subtlety and preciseness in comparison to the performance of the top four parsers. LGP failed to reach a ranking higher than fourth on any of the criteria and thus was ranked last in the overall ranking.

In summary, one may conclude that the results of the present study indicate that probabilistic parsers – because of their design – *tend* to be more robust and to have a better coverage than rule-based ones as suggested by the theoretical evaluation in Part I of the thesis. However, the sample of parsers that I selected was not large enough to allow me to make a general claim. The original observation about the poor connection between the theoretical upper bounds of computational complexity of formalisms and practical efficiency was also supported by the empirical findings. If one excludes LGP, the evaluations also gave clear evidence that more recently developed parsers are superior in performance to those parsers that were developed in the 1990s. The findings moreover confirm the observation made in Section 3.4.1.6 that some sentence constructions in English seem unable to fit the LG framework.

One finding that should be made based on the experiments relates to the user friendliness and documentation of the parsing systems. Despite the fact that all the system requirements were met, two of the parsers that were intended to be included in this evaluation were unable to function on our machine configuration. While all the parsers have a set of parameters that can be adjusted, the accompanying documentation about their effects is in many cases insufficiently detailed. Some exceptions, however, exist. Such an exception in particular was the documentation provided for C&C. From the NLP practitioner's point of view, the process of selecting an appropriate parser for a given task is complicated by the fact that the output format of a parser is frequently described in insufficient detail. It would also be helpful in many NLP applications if the parser were to indicate whether it could parse a sentence completely or not. It would also be ideal if a confidence score that indicated the reliability of the returned analysis could be provided.

The process of parsing the test material of hundreds of thousands of sentences with several parsing systems was neither simple nor straightforward. To begin with, most of the parsers crashed at least once during the course of the experiments. This is obviously an unacceptable feature in any kind of computer software. The C&C parser, for example, terminates when it encounters a sentence with two spaces between words. It would be far more convenient for users if the parser were automatically to skip or normalize such sentences.

In its first application, FEPa proved to be a useful and practical tool for empirical parser evaluations. It can provide a rich and layered picture of a parser's performance and can compare the performances of different systems. While I was carrying out these experiments, I made several modifications to the framework proposed in paper [5].

### *10.8.2 Future work*

There are three general future directions for the type of experimentation described in this section. Firstly, the inclusion of more parsers would offer a broader view of the current state of the art in parsing. In particular, because of their wide use, it would be necessary to include an HPSG and an LFG parser in the experiments. Secondly, the evaluation materials could be enhanced both in terms of their size and quality. Thirdly, the findings of the experiments could be used for further developing the FEPa framework.

One of my most important plans includes the performance of preciseness evaluations. The lack of such empirical work is the biggest deficiency in the evaluation reported in this chapter. The first phase of such evaluations could be carried out by using the PTB and PTB-derived resources such as DepBank and CCGBank. *CCGBank* is a CCG-style treebank that is derived from the PTB-II (Hockenmaier & Steedman 2007). If one used this method, one could evaluate PS, D and CCG parsers on the same data.

The most obvious directions for work on coverage evaluation would include other text genres and even larger collections of texts. One could also pinpoint the most problematic types of sentence structures by applying error-mining techniques to the results of the experiments. Another extension of this line of research would be to include grammatical coverage evaluations. I am specifically interested in finding out how well the parsers would handle specific types of LDD constructions because they represent one of the greatest challenges in the way of extending the current capacity of the best contemporary state-of-the-art parsers.

Several directions for future work also suggest themselves in robustness evaluation. One could collect more data for more comprehensive system comparisons. One could in addition extend the research to include various kinds of noise other than only misspellings. A future researcher might, for example, use the corpus of Foster and Vogel (2004) as a source for syntactically distorted sentences. It would also be interesting to apply error mining techniques to the sentences that parsers fail to analyze.

The weakest part of the comparative evaluation scheme in FEPa are the factors represented in Equations 10-2 and 10-3; they are insufficiently motivated. More experiments, particularly ones carried out with the use of standard metrics, measures and test data, are needed for a more comprehensive definition of these aspects of the framework. I believe, however, that FEPa, even in its current state, is quite capable of providing a great deal of illuminating and helpful insight into the relative strengths and weaknesses of parsers.

# 11 Conclusions and Future Work

This research comprises an extensive analysis of natural language parsers and the relative merits and disadvantages of their respective evaluation methods. In order to test the validity of my theoretical assumptions, hypotheses and conclusions, I also developed a set of evaluation resources and tools and applied them under controlled experimental conditions in order to evaluate a select number of parsing systems.

This chapter summarizes the main contributions of this dissertation (Section 11.1) and suggests directions for further research (Section 11.2). Four main research tasks were identified in Chapter 1. These I grouped into four main categories in order to address the following topics and issues: 1) The theoretical analysis of parsers, parser evaluation resources and evaluation methodology; 2) The development of linguistic resources for parsers and 3) a framework for parser evaluation; 4) The application of the developed resources, methods and tools in practical parser evaluations. The concerns addressed in Sections 11.1 and 11.2 are based on these divisions.

## 11.1 Summary of Results

### *11.1.1 Theoretical evaluation of current state-of-the-art parsing*

In Chapter 2, 3 and 4, I described the theoretical foundations on which state-of-the-art parsers are being constructed. I noted in Chapter 2 that a high degree of preciseness can be achieved in the preprocessing stages. It appears that these figures (close to 100% in segmentation and 96-97% of correct tags in POS tagging) are close to the upper bound. 100% accuracy in POS tagging is not achievable with the current methods. Probabilistic taggers will always encounter training data that contain errors. The rules of a rule-based parser can rarely, if ever, be error-free. The only way therefore to improve the preciseness of preprocessing is by incorporating segmentation and tagging into the syntactic (and perhaps semantic) processing by permitting interaction to take place between the preprocessing and latter processing stages.

One can characterize grammar formalisms , as I did in Chapters 3 and 4, on the basis of their descriptive power, computational complexity and equivalence to other formalisms. On the basis of this analysis I concluded that grammar formalisms, which in many respects constitute the foundations of parsing systems, are difficult to evaluate on theoretical grounds alone. A grammar formalism is merely a language in which linguistic theories can be expressed. There are many different ways of encoding the same linguistic information, and every formalism contains its own advantages and disadvantages. It would be invidious to assert categorically that one particular grammar formalism is superior to others. It is, after all, the grammar theory that the grammar formalism expresses in combination with the properties of the formalism itself and, most importantly, with *the quality of the grammar*, that suggests how well a grammar may perform in practice.

It is only possible to predicate superiority on the quality of the grammar itself, whether it be automatically acquired or manually constructed. It is in this sense that the quality of available grammars, grammar development and induction tools becomes a decisive factor in the choice of a grammar formalism. One of the most prominent developments in parsing grammars is the development of automatically acquired deep grammars. It has often been demonstrated that such grammars can outperform their hand-crafted counterparts which have been developed over several years or even decades. It is, however, my personal conviction that the possible benefits that could be obtained from a combination of manually encoded linguistic knowledge and the use of probabilistic information have not as yet been fully exploited. This line of research work might well be a precursor to future improvements.

An NLP system developer might well find any of the available formalisms suitable for the task at hand. English parsers, for example, that are freely available for research purposes exist for PCFG and many DG formalisms (such as, for example, CG, XDG), LG, CCG, HPSG, and different versions of TAG. The selection should be guided by the intended usage of the parser.

After taking expressive power, maturity of theory and the availability of resources into consideration, I concluded that LTAG, CCG, LFG and HPSG are currently the most attractive and viable grammar formalisms for a parser developer who is looking for a suitable formalism for a parsing system. It is their generative capacity and ability to handle LDDs that enable to assert that HPSG, LFG, the MCSGs analyzed in this work are in fact the most attractive and viable formalisms currently available for parser development. I would also add non-projective DGs into this group. XDG will offer a number of advantages once the open issues in the formalism have been resolved. The lack of free grammar development tools and parsers disfavors FDG.

It is a more straightforward task to determine the necessary criteria for comparing parsing algorithms and computational properties of formalisms: time and space complexity are the main factors. Even though one should keep in mind the distinction between the theoretical upper bound for complexity and actual performance, the findings with regard to search algorithms in Chapters 4 and 5 could be distilled into the following observation: a practical parsing algorithm should be efficient (i.e. use minimum computational work) and robust (i.e. behave in a sensible way when parsing a sentence that it cannot analyze fully).

From the point of view of computational properties, CG, LG, MCSGs (CCG and LTAG) and PCFG are among the most attractive ones available because of their relatively low computational complexity. Much to my surprise, however, the efficiency of the LTAG parser that I ran experimentally was, in fact, poor. It will be necessary to carry out practical evaluations on HPSG, LFG and XDG and other non-projective D parsers in order to verify exactly how they will behave in practice. The recognition problems of all these grammars have been reported to be intractable.

The observations that I made in Chapter 5 lead me to make the following conclusions. Firstly, there appears to be a limit to the performance of parsers that use only probabilistic or rule-based approaches. How exactly to combine the two approaches is an attractive research proposition.

Secondly, a problem in the typical sequential organization of parsing is potential error-propagation. When the output of a component is used as input for the next processing level, it may generate errors on that level and might even cause a failure to produce a parse. What is needed in such circumstances are more flexible processing architectures that allow interactions to take place between the different stages of processing. Intelligent ways of combining predictions from several sources (each of which is unreliable alone) may result in improved performances. This is especially important when one combines information from rule-based and probabilistic sources.

Thirdly, apart from the fact that a successful parser relies both on syntactical information as well as lexical and contextual information, it is evident that researchers will need semantically richer lexical resources and parsing models if they are to improve the performance of contemporary state-of-the-art parsers. However, as intuitively viable as the idea of using semantic information to guide syntactic parsing decisions might appear to be, the incorporation of semantic information into syntactic parsing is anything but a straightforward task.

The conclusion that one might draw from these findings is that the interaction of the levels of linguistic knowledge and processing stages appears to offer great potential for improving the performance of the best contemporary parsers.

### *11.1.2 Designing and implementing linguistic resources for parser evaluation*

Linguistic resources play a crucial role in parser evaluation. I analyzed and reviewed existing linguistic resources in Chapter 6 and considered the various ways in which these resources might be designed and annotated in order to render them useful for purposes of parser evaluation. I can summarize the results of that analysis in the following way.

Firstly, a comprehensive evaluation experiment requires the use of a diversity of linguistic resources. Secondly, the annotation of evaluation resources is a demanding task that compels one to make a choice between a more parser-specific resource and the preservation of high adaptability in the comparison of parsers. Thirdly, the need to balance the requirements of generality and specificity create a tension during the process of constructing resources for parser evaluation. Most of the resources that one uses in parser evaluation are general in the sense that they are also used for other purposes. However, in order to be practically useful, an evaluation resource should be sufficiently specific to accomplish the main purpose of evaluation – which is to measure how well a parser is able syntactically to analyze sentences.

I utilized the results of Chapter 6 in Chapter 7 to design and implement a set of linguistic resources that I deliberately created for the specific purpose of parser evaluation. I adapted the design of a treebank for Finnish so that it would be effective for evaluating parsers. I described, moreover, the design and implementation of an annotation tool, DepAnn, for D treebanks. The outstanding features of this treebank are the inclusion of diverse text genres, an hierarchically organized tagset, the use of an XML-based exchange format for encoding, and the fact that the annotation scheme allows multiple analyses to be saved for each sentence.

I also described two new evaluation resources for English. These resources can be used for carrying out evaluations of coverage, efficiency, subtlety and robustness. RobSet will, moreover, be made freely available to users.

### *11.1.3 Deriving an evaluation framework*

In Section 9, I described, on the basis my analysis of the parser evaluation methods and tools described in the literature (Chapter 8), the FEPa framework for carrying out parser evaluations and comparisons. I initiated the experimental part of this research project by identifying the following five requirements for the design of such a framework: purpose, criteria, metrics, measures and resources. I then explained how each of these requirements are addressed within FEPa. Most importantly, I explained how parsers are evaluated by FEPa from the following five points of view: preciseness, coverage, efficiency, robustness and subtlety. I also considered ways of using the framework for comparing the strengths and weaknesses of parsers.

### *11.1.4 Empirical parser evaluations*

In Chapter 10, I reported a series of parser evaluations in which I experimented with FEPa in practice. I selected six parsers that use five different grammar formalisms as the parsers that I wanted to evaluate. The parser that performed best in these experiments was C&C, which achieved the rank 1 to 3 on all the evaluation criteria. After taking into account the findings from Chapters 3 and 4 about the properties of grammar formalisms, I concluded that C&C was the best parser in this experiment. The good performance of StatCCG confirmed my conclusion that CCG seems to be among the most attractive grammar formalisms for grammar and parser development – for English at least.

These experiments also confirmed that the performance of parsers has improved considerably over the last decade. The best parsers from the comparative experiment performed in most cases better than the older parsers of the 1990s, and they also produced outputs that were more informative.

FEPa proved its worth as a useful tool for carrying out empirical comparisons between various approaches to parsing. But these comparative experiments also revealed the number of shortcomings in the initial version of the framework. I accordingly implemented changes and amendments to the framework during the experiments – but left others for future research.

## 11.2 Future Work

Because this dissertation reports on such a wide variety of topics and experimental research undertakings, it has been necessary for me to strike a balance between the scope of presentation (the number and variety of topics and problems) and the effective range of depth and detail of my approach to such topics and problems. Because of this, the research reported in this dissertation constitutes in many ways but the first steps towards making a complete set of resources and tools available, and using these for empirical evaluations at some time in the future. This section outlines several directions for future work.

### *11.2.1 Theoretical evaluation of state-of-the-art parsing*

It appears to be the case that analysis of the theoretical upper bounds of the complexity of parsing grammar formalisms is of limited use in practice. The most interesting challenge presented by theoretical evaluation is to conduct more in-depth explorations of the details of the generative capacities of the formalisms. It is the ability of a grammar formalism to describe natural languages that remains its most important property in a parsing system.

The treatment of LDDs specifically represents itself as an interesting topic in this field. Insufficient attention has been devoted to this issue in parser evaluations reported in the literature. Apart from the need for theoretical evaluation, there is also a great need for practical evaluations that will be based on a set of test sentences containing different types of LDDs.

### *11.2.2 Designing and implementing linguistic resources for parser evaluation*

The most ambitious future plan involves the construction of a manually checked D treebank. The treebank that I designed for this thesis has not yet been manually checked. The manual checking of parser-created analyses is beyond the scope of a single dissertation. The treebank is as yet too small for a comprehensive testing of a parser's performance. Both the annotation tool and the treebank are sufficiently developed to offer adequate grounds for developing a full-scale treebank for Finnish. My hope is that it will be possible to establish a project that will undertake the construction of the treebank for Finnish on the basis of the data produced by this research. In the project, the annotations would be checked by hand and the size of the treebank extended. But because the construction of a manually checked D treebank requires huge resources (up to hundreds of thousands of euros by present standards), it remains to be seen how the work on the treebank can be continued.

Before a full-scale manual checking process can begin, it will be necessary to construct annotation guidelines that will guide the checking process and ensure consistent annotation. The annotation tool will be improved during the manual checking process by more stringent automatic consistency checks and a higher level of automation. In order to obtain fine-grained feedback from errors in parsing and morphological analysis, the sub-corpus used for testing grammatical coverage could be transformed into a test suite in which the sentences will be organized as test items and grouped into test sets. Information about the morphological and syntactic phenomena that occur in it would also be added to each test item.

An obvious direction for the work on MGTS would be to include other text genres and even larger collections of texts. More comprehensive data could be collected in order to make robustness evaluation with RobSet more reliable. This work could also be extended in the direction of permitting kinds of noise other than misspellings, for example grammatical errors.

### *11.2.3 Deriving an evaluation framework*

There are some details of the FEPa framework, most of them connected to the comparison of parsers, that need further revision. Such modifications are carried out on the basis of practical evaluation experiments with the framework. The only way to evaluate the usefulness and ease of use of the framework is by applying it in practice and then using the feedback that one gets to improve it.

A set of tools were implemented to carry out evaluations within the framework. Such tools would be more convenient to use if they were combined into a single full-scale evaluation environment for performing evaluations by using the proposed framework. Such a system should be able to make use of several existing linguistic resources. It should also be able to provide the user with detailed information about all of the five aspects of the evaluation framework. The key properties that such a tool would provide would be:

1. Support for several types of parsers,
2. support for diverse linguistic resources,
3. implementations of evaluation metrics and measures for preciseness, coverage, efficiency, robustness and subtlety,
4. tools for browsing, storing and comparing results of separate test runs, and

5. a graphic user interface that will permit parse tree viewing and graphs for suppressing evaluation results in graphic form.

### *11.1.4 Empirical parser evaluations*

FEPa should be used to carry out more practical parser evaluations so that a more comprehensive picture of the current state of the art in parsing will emerge and so that researchers will be able to identify the problems of particular parsers. More specifically, I see it as essential to include a HPSG, LFG, and FDG parser in such experiments in order to determine how they will stand up to a comparison with CCG and other parsers. Evaluating an FDG parser, for example, would provide a fuller picture of the state-of-the-art in rule-based parsing. Any attempt to include an LTAG parser in the experiments were hindered by the enormous computational complexity of the system which may reflect the currently immature state of LTAG parser development.

Apart from future uses of the FEPa framework to conduct evaluations of other parsers, it is also necessary to clarify the role of the subtlety measure in comparison of systems. For example, in those cases where the same test set is not used for carrying out evaluations, the subtlety measure should take into account the degree of difficulty in parsing the test sets.

# Appendix A: Glossary of Grammatical Terms

The descriptions are based on the following sources: Trask (1993), Crysmann (2006), Van Eijck (2003), and Kim (2000).

| Term | Description |
|---|---|
| Agreement | Agreement is a form of cross-reference between different parts of a sentence or phrase. Agreement occurs when one word changes its form in dependence on other words to which it is related. For example, “He kick” is incorrect, but “He kicks” is correct. |
| Argument | Predicate and argument are the two main parts of a sentence. Arguments can be realized as nouns, groups of nouns or sentences. See also *predicate* and *predicate-argument structure*. |
| Argument clustering | Also argument cluster coordination. In sentences such as “Sponsors gave [Harry three balls] and [Robbie some books]” the NPs denoted with brackets form *argument clusters* coordinated by a coordinating conjunction (“and”). |
| Center-embedding | A type of structure in which a clause is interrupted by a second clause. For example, in the sentence “The ball [that the attacker kicked] hit the post”, “that the attacker kicked” is embedded in the clause “The ball hit the post”. |
| Chunking | This is also called partial parsing. Instead of recognizing the internal structure of the phrases as one does in full parsing, only the base phrases (referred to as chunks), such as NPs, VPs and PPS, for example, are identified in chunking, even though their structure is left unidentified. |
| Coordination | Coordination refers to the combination of like or similar syntactic units into groups of the same category or status. The joined sentence elements are referred to as *conjuncts*. In a coordinate structure, such as “strikers and defenders”, conjunctions like “and” or “but” conjoin words or phrases. At first glance coordination might appear to be a relatively simple phenomenon. It is, however, notoriously difficult for linguistic theories to define. |
| Constituent | A part of a sentence that forms a distinct syntactic unit is referred to as a constituent. A phrase in a PS tree forms a constituent. |
| Conjunction | Lexical items belonging to this category are used for constructing coordinate structures. Examples are “but”, “and”, “or”. |
| Control | In control, a VP complement with no overt subject is interpreted as having an NP as subject. For example, in the sentence “The coach asked the attacker [PRO to score goals]”, PRO is controlled by the subject “The coach”. This is an example of a specific type of control called *subject-control*. |
| Dependent | This is also called a modifier or daughter. In dependency grammars, a |
| | dependency relation is defined as the relation between a head and a dependent. See also *head*. |
| Ellipsis | A construction from which an element is missing although it can be recovered from the context. For example, the sentence “Defenders can’t score goals, but attackers can”, contains an ellipsis, namely the VP “score goals”. |
| Extraction | Extraction occurs when a subconstituent of some constituent is missing, and some other constituent of the incomplete constituent represents that missing constituent in some way. For example, the sentence “The referee hated the man that$_i$ the goalkeeper sold the ball to $t_i$:” contains left extraction represented by the annotation that links the pronoun “that” to its trace $t_i$. |
| Extended domain of locality | The domain of locality specifies many of the properties of a grammar formalism because it defines the domain over which dependencies (syntactic and semantic) can be stated. For example, the domain of locality in CFPSG, corresponding to a single rule, is one level in a PS tree. The term extended domain of locality is used especially with TAGs and refers to the fact that the elementary trees are larger units than the rules in CFPSG, thus allowing an extended domain over which dependencies can be stated. |
| Gapping | Gapping is a type of ellipsis that occurs in conjoined sentences. In gapping, the main verb of a clause is missing and the coordinator is presented instead. For example, “John kicks the ball and George _ the attacker.” |
| Grammatical relation | This is also called *grammatical function*. It describes connections of a grammatical nature between parts of a sentence. The most widely accepted GRs include subject, predicate and object. |
| Head word | This is also called *head*. The syntactically central element of a constituent, D link or GR. Most grammar formalisms assume that there is one daughter which can be identified as the head among the daughters introduced by a rule. For example, the head of an NP is typically a noun. In DGs, a D relation is defined as the relation between a head and a dependent. See also *dependent*. |
| Immediate dominance | An immediate dominance relation holds between a mother node and its daughter in a parse tree if there is no other node between them. For example, the rule $X \rightarrow Y\ Z$ defines that $X$ immediately dominates $Y$ and $Z$. |
| Lexical item | This is also referred to as a *lexeme*. It is an abstract object that has a consistent meaning or function but which can vary in form. For instance, the lexical item BALL can be realized in the forms “ball” and “balls”. |
| Linear precedence | A relation that holds between two nodes in a parse tree in which one of the two is located before (i.e. precedes) the other in left to right order. For example, an immediate dominance rule with no linear precedence rules, $X \rightarrow Y\ Z$ can be expanded into a structure in which the two siblings of $X$, $Y$ |

and *Z*, may occur in either order. After adding the linear precedence rule *Y* < *X*, only the expansion *Y Z* is allowed.

**Local ambiguity**
A type of ambiguity that occurs only when a part of a sentence is considered in isolation.

**Long-distance dependency**
A dependency in which the two elements may be separated by an arbitrary distance. The connection between the elements has to be defined recursively. *LDD*s occur, for example, in wh-questions such as “Whom did you give to ball to?”. The word “whom” serves in a sense as the NP in the preposition phrase (PP) “to whom”. A constituent (the filler) appears to be “dislocated” from its usual place (the gap).

**Predicate**
A predicate must contain a verb and can contain objects or phrases governed by the verb. It gives information about the arguments.

**Predicate-argument structure**
The predicate and argument are the two main parts of a sentence. A predicate-argument structure represents the hierarchical relations between a predicate and its arguments. The set of the relations types depends on the theory. For example, the following types could be recognized: Agent / Experiencer / Theme / Location / Goal.

**Principle**
A statement in a grammatical theory that has universal validity. For example, the Head Feature Principle in HPSG projects the properties of a head word onto headed phrases.

**Relative clause**
This is a clause that modifies an NP. For example, “The coach is watching the defender [the scout told him about]”.

**Right-node raising**
A construction that consists of a coordination of two sentences that lacks its rightmost constituent when a single further constituent appears on the right filling the gaps in both sentences. For example, “The goalkeeper passed, and the attacker kicked, the ball.”

**Self-embedding**
A phrase is self-embedding if it is embedded into another phrase of the same type and is there surrounded by lexical material. For example, “The attacker the defender the goalkeeper helped blocked stumbled.”

**Simplex sentence**
A simplex sentence consists of a single clause, and thus has one predicate and one subject.

**Small clause**
A small clauses is a minimal predicate structure. It lacks a finite verb. Small clauses usually occur within full clauses and may act as the direct ovject of the verb. For example, “The attacker [wearing red shoes] passed the ball”.

**Topicalization**
In topicalization an element of a sentence is marked as the topic. For example, in English topicalization is done by preposing the element. For instance, “[This ball] is round.”

**Tough-movement**
Movement of a direct object of a verb to appear as the subject. For example, the sentence “It is hard to please the coach.” changes to the form “The coach is hard to please.” by the tough-movement of the direct object “the coach”.

| | |
|---|---|
| Well-formedness | A well-formed structure is consistent with all the requirements of the grammar. |
| Wh-item | A lexical item that serves to ask a question. For example, “who”, “why”, “where”. |
| Wh-movement | A construction that consists of a wh-item that appears in a sentence or clause-initial position |
| Wh-relative clause | A relative clause in which the relative pronoun is of wh-type. For example, “The attacker [who scored five goals] was suspended.” |

## Dissertations in Computer Science

**Rask, Raimo**. Automating Estimation of Software Size during the Requirements Specification Phase – Application of Albrecth's Function Point Analysis Within Structured Methods. Joensuun yliopiston luonnontieteellisiä julkaisuja, 28 – University of Joensuu. Publications in Sciences, 28. 128 pp. Joensuu, 1992.

**Ahonen, Jarmo**. Modeling Physical Domains for Knowledge Based Systems. Joensuun yliopiston luonnontieteellisiä julkaisuja, 33 – University of Joensuu. Publications in Sciences, 33. 127 pp. Joensuu, 1995.

**Kopponen, Marja**. CAI in CS. University of Joensuu, Computer Science, Dissertations 1. 97 pp. Joensuu, 1997.

**Forsell, Martti**. Implementation of Instruction-Level and Thread-Level Parallelism in Computers. University of Joensuu, Computer Science, Dissertations 2. 121 pp. Joensuu, 1997.

**Juvaste, Simo**. Modeling Parallel Shared Memory Computations. University of Joensuu, Computer Science, Dissertations 3. 190 pp. Joensuu, 1998.

**Ageenko, Eugene**. Context-based Compression of Binary Images. University of Joensuu, Computer Science, Dissertations 4. 111 pp. Joensuu, 2000.

**Tukiainen, Markku**. Developing a New Model of Spreadsheet Calculations: A Goals and Plans Approach. University of Joensuu, Computer Science, Dissertations 5. 151 pp. Joensuu, 2001.

**Eriksson-Bique, Stephen**. An Algebraic Theory of Multidimensional Arrays. University of Joensuu, Computer Science, Dissertations 6. 278 pp. Joensuu, 2002.

**Kolesnikov, Alexander**. Efficient Algorithms for Vectorization and Polygonal Approximation. University of Joensuu, Computer Science, Dissertations 7. 204 pp. Joensuu, 2003.

**Kopylov, Pavel**. Processing and Compression of Raster Map Images. University of Joensuu, Computer Science, Dissertations 8. 132 pp. Joensuu, 2004.

**Virmajoki, Olli**. Pairwise Nearest Neighbor Method Revisited. University of Joensuu, Computer Science, Dissertations 9. 164 pp. Joensuu, 2004.

**Suhonen, Jarkko**. A Formative Development Method for Digital Learning Environments in Sparse Learning Communities, University of Joensuu, Computer Science, Dissertations 10. 154 pp. Joensuu, 2005.

**Xu, Mantao**. K-means Based Clustering and Context Quantization, University of Joensuu, Computer Science, Dissertations 11. 162 pp. Joensuu, 2005.

**Kinnunen, Tomi**. Optimizing Spectral Feature Based Text-Independent Speaker Recognition. University of Joensuu, Computer Science, Dissertations 12. 156 pp. Joensuu, 2005.

**Kärkkäinen, Ismo**. Methods for Fast and Reliable Clustering. University of Joensuu, Computer Science, Dissertations 13. 108 pp. Joensuu, 2006.

**Tedre, Matti**. The Development of Computer Science: A Sociocultural Perspective. University of Joensuu, Computer Science, Dissertations 14. 502 pp. Joensuu, 2006.

**Akimov, Alexander**. Compression of Digital Maps. University of Joensuu, Computer Science, Dissertations 15. 116 pp. Joensuu, 2006.

**Vesisenaho, Mikko**. Developing University-level Introductory ICT Education in Tanzania: A Contextualized Approach. University of Joensuu, Computer Science, Dissertations 16. 200 pp. Joensuu 2007.

**Huang, Haibin**. Lossless Audio Coding for MPEG-4. University of Joensuu, Computer Science, Dissertations 17. 86 pp. Joensuu 2007.

**Mozgovoy, Maxim**. Enhancing Computer-Aided Plagiarism Detection. University of Joensuu, Computer Science, Dissertations 18. 131 pp. Joensuu 2007.

**Kakkonen, Tuomo**. Framework and Resources for Natural Language Parser Evaluation. University of Joensuu, Computer Science, Dissertations 19. 264 pp. Joensuu 2007.